Akademia Górniczo-Hutnicza im. Stanisława Staszica
Wydział Informatyki, Elektroniki i Telekomunikacji
Katedra Informatyki

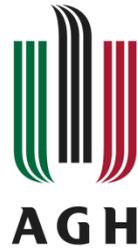

Rozprawa doktorska

# DETEKCJA UPADKU I WYBRANYCH AKCJI NA SEKWENCJACH OBRAZÓW CYFROWYCH

**mgr inż. Michał Kępski**

Promotor: dr hab. inż. Bogdan Kwolek, prof. nadzw. AGH

Kraków 2016

.

# Spis treści

# Wstęp

W ostatnich latach obserwuje się duży wzrost zainteresowania zagadnieniem rozpoznawania akcji, a w szczególności jedną z jego dziedzin jaką jest detekcja upadku. Znaczący wzrost długości życia o około trzydzieści lat w krajach rozwiniętych stanowi wyzwanie dla społeczeństwa do adaptacji pod kątem potrzeb ludzi starszych. Rozwój technologiczny pozwala na znaczne usprawnienie automatycznego lub zdalnego nadzoru samotnie mieszkających osób starszych. Jednak obecne rozwiązania komercyjne nie są pozbawione wad, przez co nie zostały w pełni zaakceptowane przez potencjalnych użytkowników. Trudności związane z wdrożeniem systemów detekcji upadku w środowisku seniorów wiążą się z niedoskonałością technologii: brakiem wystarczającej precyzji, dużą liczbą fałszywych alarmów, niewystarczającym poszanowaniem prywatności osoby podczas akwizycji i przetwarzania danych oraz z inwazyjnością rozwiązań, która wynika z tego, że użytkownik musi nosić sensory przytwierdzone do ciała bądź ubrania. Ponadto niewiele jest algorytmów zdolnych do detekcji upadku na energooszczędnych platformach w architekturze ARM. Zastąpienie komputera klasy PC systemem wbudowanym pozwoliłoby na istotną redukcję kosztów, zużycia energii oraz poziomu hałasu wynikającego z braku potrzeby aktywnego chłodzenia systemu, co z kolei miałoby pozytywny wpływ na potencjał aplikacyjny proponowanego rozwiązania.

Potrzeba usprawnienia systemów detekcji upadku zwróciła uwagę zespołów badawczych z całego świata. Ze względu na umiejscowienie urządzeń do akwizycji danych, proponowane w literaturze naukowej podejścia można podzielić na: noszone przez użytkownika (Bourke & Lyons, 2008; Bourke et al., 2010) oraz instalowane w jego otoczeniu (Stone & Skubic, 2011; Anderson et al., 2006; Stone & Skubic, 2013). Metody te nie są pozbawione wad, w szczególności nie pozwalają one na nieprzerwane monitorowanie osoby w trybie całodobowym.

Mając na względzie zapotrzebowanie na metody detekcji upadku, cele niniejszej pracy zdefiniowano w następujący sposób:

- opracowanie efektywnych algorytmów do detekcji upadku na podstawie sekwencji obrazów,
- opracowanie algorytmu umożliwiającego detekcję upadku w czasie rzeczywistym na urządzeniu wbudowanym,

- opracowanie efektywnych algorytmów detekcji (segmentacji) osoby na obrazach RGB-D dla systemu wbudowanego,
- opracowanie układu wnioskującego o upadku z uwzględnieniem kontekstu sytuacji,
- przebadanie opracowanych algorytmów pod kątem dokładności.

Tezy rozprawy doktorskiej sformułowano w następujący sposób:

**Zastosowanie obrazów głębi w systemach detekcji upadku prowadzi do znaczącego obniżenia liczby fałszywych alarmów**

oraz:

**Dzięki wykorzystaniu cech pochodzących z obrazów głębi oraz uwzględnieniu kontekstu sytuacji następuje wzrost skuteczności detekcji upadku oraz zmniejszenie liczby fałszywych alarmów w porównaniu do systemów operujących na sekwencjach obrazów RGB lub pomiarach z akcelerometru/żyroskopu.**

W ramach niniejszej pracy opracowano, przebadano oraz zaimplementowano trzy algorytmy umożliwiające detekcję upadku na podstawie sekwencji obrazów cyfrowych. Algorytmy te pozwalają na dużą dowolność wyboru umiejscowienia kamery. Opracowane algorytmy wykorzystują dane akcelerometryczne pochodzące z bezprzewodowego czujnika ruchu noszonego przez osobę monitorowaną do zamodelowania kontekstu sytuacji. Przebadano i wybrano zestaw cech pochodzących z obrazów głębi pozwalający na klasyfikację pozy w jakiej znajduje się osoba, jak i akcji, która jest przez nią wykonywana. Opracowano także algorytm segmentacji postaci na obrazach głębi mający zastosowanie zarówno w systemie z jednostką obliczeniową klasy komputera PC oraz w systemie wbudowanym z procesorem o architekturze ARM. Ponadto zaproponowano rozwiązanie, które dzięki zamontowaniu kamery na obrotowej głowicy aktywnej znacznie poszerza obszar monitorowany przez system. W tym celu wykonano głowicę oraz opracowano algorytmy śledzenia postaci i sterowania ruchem kamery. Na potrzeby oceny jakości opracowanych metod przygotowano zbiór danych składający się z ok. 13000 obrazów (70 sekwencji wideo), przedstawiający symulowane upadki i czynności życia codziennego wraz z danymi akcelerometrycznymi. Opracowany zbiór danych został udostępniony w sieci Web. Dzięki upublicznieniu zbioru danych inne zespoły mają możliwość rozwoju zaproponowanych rozwiązań lub porównania własnych metod z rozwiązaniem przedstawionym w niniejszej pracy.

Praca składa się z czterech rozdziałów. Pierwszy z nich omawia zagadnienie analizy i rozpoznawania akcji ze szczególnym uwzględnieniem problemu detekcji upadku. Rozdział ten zawiera uszeregowanie stosowanych rozwiązań oraz przegląd metod detekcji upadku. Przedstawiono w nim najczęściej stosowane miary do oceny jakości klasyfikacji oraz zbiory danych do ewaluacji skuteczności detekcji upadku.

Drugi rozdział poświęcono omówieniu systemu do detekcji upadku. Omówiono w nim dane wejściowe oraz scharakteryzowano urządzenia do akwizycji danych. W omawianym rozdziale przedstawiono algorytmy przetwarzania obrazów i ich modyfikacje wykonane na potrzeby systemu detekcji akcji. Ponadto omówione zostały metody detekcji i śledzenia postaci ludzkiej opracowane dla potrzeb detekcji upadku.

Kolejny rozdział opisuje założenia i zasadę działania systemu do detekcji upadku. Przedstawiono w nim algorytmy detekcji upadku opracowane i zaimplementowane w ramach zrealizowanych prac oraz wyniki badań eksperymentalnych nad zagadnieniem detekcji.

Rozdział czwarty omawia propozycję zastosowania rozmytego systemu detekcji upadku. Przedstawia podstawy logiki rozmytej i wnioskowania rozmytego oraz wykorzystanie ich do detekcji upadku. W rozdziale dokonano porównania efektywności systemu działającego w oparciu o wnioskowanie rozmyte z wcześniej opracowanymi algorytmami.

Pracę podsumowano omówieniem uzyskanych wyników oraz kierunków dalszych badań. Rozprawę zamyka wykaz pokrewnych prac, w szczególności związanych z problematyką detekcji upadku.

# 1. Analiza i rozpoznawanie zachowań ludzi

W niniejszym rozdziale omówiono problematykę analizy i rozpoznawania akcji ze szczególnym uwzględnieniem problemu detekcji upadku. Przedstawiono wyzwania i oczekiwania stawiane systemom w kontekście skuteczności i efektywności działania. Rozdział zawiera uszeregowanie stosowanych rozwiązań oraz przegląd metod detekcji upadku. Przedstawiono także najczęściej stosowane miary wykorzystywane w procesie oceny jakości klasyfikacji oraz zbiory danych do badań eksperymentalnych.

## 1.1. Problematyka rozpoznawania zachowań ludzi

Celem rozpoznawania zachowań jest sklasyfikowanie akcji wykonywanej przez człowieka w oparciu o analizę danych pochodzących z różnego rodzaju sensorów, najczęściej wizyjnych. Zagadnienie analizy zachowań jest powiązane z problemem estymacji pozy postaci ludzkiej, który można rozpatrywać jako problem regresji (Poppe, 2010), a omawiane zagadnienie jako problem klasyfikacji. Obie tematyki mają wiele cech wspólnych, szczególnie na poziomie reprezentacji obrazu. Duże zainteresowanie omawianym tematem jest umotywowane mnogością potencjalnych zastosowań, zarówno w systemach działających w czasie rzeczywistym jak i off-line. Do głównych obszarów potencjalnych zastosowań należą: systemy monitoringu i nadzoru, systemy umożliwiające interakcję człowiek-komputer oraz systemy przeszukiwania i indeksacji zbiorów multimedialnych.

Na przestrzeni kilku ostatnich lat zaobserwowano znaczny wzrost liczby kamer montowanych na potrzeby systemów monitoringu, zabezpieczeń i nadzoru (Brzoza-Woch et al., 2013). Zapewnienie stałego i efektywnego monitoringu realizowanego przez ludzi przy ciągle zwiększanej liczbie kamer jest zadaniem bardzo trudnym i kosztownym, gdyż wzrost liczby kamer wymaga zaangażowania większej liczby osób pracujących przy obsłudze danego systemu. W związku z tym coraz częściej opracowuje się rozwiązania mające na celu automatyzację pewnych zadań, a tym samym obsługę dużej liczby kamer przez pojedynczego operatora. Systemy monitoringu w oparciu o analizę obrazu mogą wykrywać ruch w monitorowanym obszarze (Tian & Hampapur, 2005), śledzić obiekty (Arsic et al., 2009), dokonywać detekcji osób (Dalal & Triggs, 2005; Dollar et al., 2010) lub innych obiektów, wykrywać podejrzane zachowania (Niu et al., 2004), interakcje z przedmiotami (Porikli et al., 2008) czy interakcje grup osób (Mehran et al., 2009). Osobną grupę systemów stanowią rozwiązania stosowane w telemedycynie, mające na celu opiekę nad osobami starszymi. Szczególnym obiektem zainteresowania twórców takich systemów jest wykrywanie nadzwyczajnych sytuacji, takich jak upadki (Kępski et al., 2012).

Rozpoznawanie akcji znajduje zastosowanie także w grupie systemów wspomagających interakcje człowiek-komputer. Wzrost liczby zastosowań takich systemów związany jest z rozwojem systemów multimedialnych, a w szczególności za sprawą gier wideo i specjalnych kontrolerów umożliwiających rozpoznawanie ruchów ciała. Szczególną popularnością cieszy się sensor *Kinect* firmy *Microsoft* oraz kamera *Sony PlayStation Eye*. We wspomnianych aplikacjach przeważa rozpoznawanie gestów przydatnych do sterowania przebiegiem akcji w grach komputerowych (np. machanie ręką, kopnięcie, rzut oraz gesty imitujące elementy gry w tenisa lub golfa) (Wang et al., 2012; W. Li et al., 2010). Prowadzone są także badania nad zastosowaniem rozpoznawania akcji do wydawania poleceń robotom (Yang et al., 2007; Kwolek, 2003) czy też rozpoznawaniem gestów reprezentujących litery alfabetu bądź słowa języka migowego (Kasprzak et al., 2012). W systemach wspierających interakcję człowiek-komputer najczęściej zastosowanie znajduje pojedyncza kamera RGB, kamera RGB-D lub kamera stereowizyjna.

Szybki rozwój technik rejestracji obrazu (szczególnie w urządzeniach mobilnych) oraz pojawienie się serwisów internetowych umożliwiających umieszczanie i odtwarzanie filmów (*YouTube, Vimeo*) spowodował zainteresowanie rozpoznawaniem akcji na potrzeby etykietowania dużych zbiorów multimedialnych (Jingen et al., 2009). Zagadnienie to jest trudne ze względu charakter danych, dlatego że ich akwizycja często odbywa się w warunkach odbiegających od środowiska laboratoryjnego, np. kamera nie jest statyczna, występują duże zmiany w skali obiektów, perspektywie i oświetleniu.

W zależności od sytuacji, zachowania ludzkie mogą mieć różną formę, od prostych ruchów do złożonych aktywności. Pod względem złożoności, można wyodrębnić trzy grupy zachowań (Moeslund et al., 2006): gesty (ang. *gestures*) określane również jako akcje podstawowe (ang. *action primitives*), akcje (ang. *actions*) i zachowania (ang. *activities*). Gesty są najmniej złożonymi rodzajami ruchu, które można opisać na poziomie zmiany położenia pojedynczych stawów i kończyn (np. wyprostowanie ręki). Akcje składają się z kilku podstawowych ruchów, często powtarzanych w pewnej kolejności (przykładowo bieganie jest sekwencją podstawowych ruchów kończyn dolnych i górnych). Zachowania składają się z kilku akcji i są najbardziej złożone (np. sportowiec uczestniczący w biegu płotkarskim wykonuje kilka akcji: start, bieganie, skakanie). Inne taksonomie (Aggarwal & Ryoo, 2011) dzielą zachowania na: gesty, akcje, interakcje człowiek-obiekt, interakcje człowiek-człowiek.

Rozpoznawanie akcji jest problemem trudnym ze względu na potrzebę rozwiązania szeregu złożonych zagadnień (Aggarwal & Xia, 2014) mieszczących się w zakresie dyscypliny jaką jest widzenie maszynowe. Pierwszą grupą zagadnień są problemy związane z reprezentacją osób na obrazie (zob. rysunek 1.1). Przeszkodą do poprawnego wydzielenia sylwetki mogą być przysłonięcia, występujące cienie i zmiany oświetlenia. Złożone tło, szczególnie o teksturze podobnej do postaci lub podlegające dynamicznym zmianom, wpływa negatywnie na skuteczność systemu wydzielającego sylwetkę człowieka. Podjęto prace (Weinland et al., 2010) nad krzepkością (ang. *robustness*) algorytmów i odpornością na wspomniane czynniki. Innym zagadnieniem, któremu poświęcono dużo uwagi w literaturze, jest zmienność wyglądu akcji w zależności od punktu umieszczenia kamery, a więc perspektywy (Parameswaran & Chellappa, 2006). Rozwiązanie tego problemu w systemach RGB polega między innymi na wykorzystaniu większej liczby zsynchronizowanych ze sobą kamer. W systemach pozwalających na akwizycję danych 3D, problem nie jest tak znaczący, między innymi ze względu na znajomość kształtu obiektu lub przynajmniej jego części, które określone są na podstawie danych o odległości między obiektami a kamerą (Shotton et al., 2011). Problemy związane z brakiem zadowalającego odwzorowania osób na obrazie, czy zmiennością w reprezentacji akcji w zależności od położenia obserwatora, nie są jedynymi wyzwaniami dla algorytmów rozpoznawania zachowań. Dla wielu grup akcji występują znaczne różnice między poszczególnymi jej instancjami, innymi słowy, inne lub nawet te same osoby mogą wykonywać tę samą czynność w różny sposób. Zmienność tempa czy zakresu ruchów wykonywanych przez człowieka, różnice antropometryczne między poszczególnymi osobami utrudniają klasyfikację akcji. Różnice między takimi samymi akcjami (duża wariancja wewnątrz-klasowa) oraz podobieństwo między różnymi akcjami (mała wariancja między-klasowa) są wyzwaniem dla twórców algorytmów rozpoznawania zachowań (Poppe, 2010).

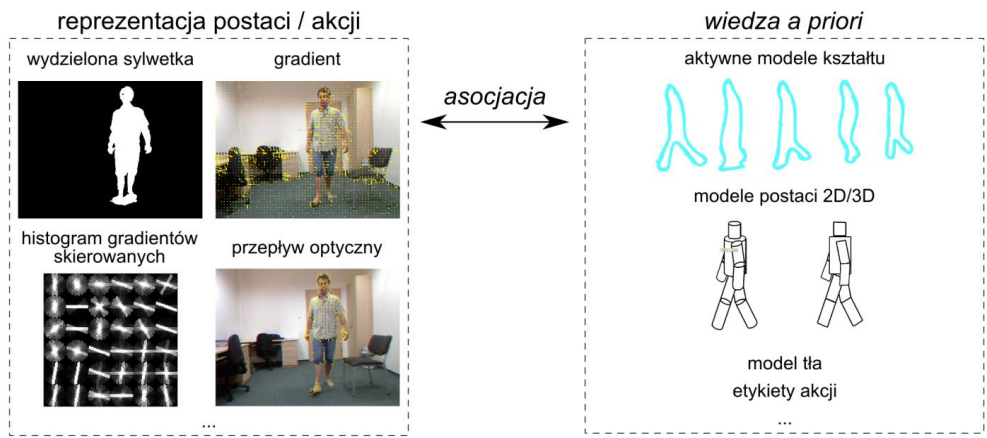

Rysunek 1.1. Rozpoznawanie akcji człowieka: kluczowe elementy składowe.

Na przestrzeni kilkunastu ostatnich lat opublikowano wiele prac podejmujących problem rozpoznawania akcji wykonywanych przez człowieka. Pod względem reprezentacji cech wydzielonych z sekwencji obrazów, podejście do rozpoznawania akcji na obrazach RGB można podzielić na dwie grupy: oparte o reprezentację globalną oraz lokalną (Poppe, 2010). Globalna reprezentacja cech jest uzyskiwana w metodach opartych na podejściu *top-down*, w którym dokonywana jest lokalizacja postaci na obrazie (metodą odjęcia tła, śledzenia lub rozpoznawania postaci), po czym cały obszar zainteresowania (ang. *ROI - Region of Interest*) jest opisywany przy pomocy wybranego deskryptora i poddawany klasyfikacji. Do takich metod zaliczyć można podejścia oparte o wolumeny obrazów (Bobick & Davis, 2001), aktywne modele kształtu (ang. *active shape models*) (Cootes et al., 1995), lokalnych wzorców binarnych (ang. *LBP – Local Binary Patterns*) (Kellokumpu et al., 2008), przepływ optyczny (ang. *optical flow*) (Black et al. 2007) lub histogramy gradientów (ang. *HOG - Histogram of Oriented Gradients*) (Thurau & Hlavac, 2008). Lokalna reprezentacja cech jest podejściem typu *bottom-up*, w którym dokonywana jest detekcja punktów zainteresowań na obrazach, a następnie wyznaczane są lokalne deskryptory w otoczeniu wybranych punktów. Grupa deskryptorów jest łączona w końcową reprezentację. Do grupy tych metod należą m.in. *Space-Time Interest Points* (Laptev, 2005), *Bag-of-Features* (Laptev et al., 2008). Metody działające w oparciu o mapy głębi pozwalają zazwyczaj na łatwiejsze i dokładniejsze wydzielenie postaci (Aggarwal & Xia, 2014). Ponadto informacja o odległości od kamery pozwala na wyznaczenie kształtu sylwetki postaci nie tylko w otoczeniu jej krawędzi, lecz dla całego jej fragmentu skierowanego w stronę kamery. Z tej własności korzystają algorytmy wykorzystujące sylwetki 3D do reprezentacji postaci, takie jak: *Bag-of-3D-Points* (Li et al. 2010), *Depth Motion Maps* (DMM) (Yang et al. 2012). Właściwości kinematyczne ciała człowieka można opisywać za pomocą modelu będącego kinematycznym drzewem, które składa się z segmentów (zwanych także kośćmi) połączonych stawami (ang. *joint*). Każdy staw ma określoną liczbę stopni swobody (ang. *degrees of freedom*). Definiują one sposób w jaki rozpatrywany staw może się poruszać. W 1975 roku Johansson w swoim eksperymencie (Johansson, 1975) wykazał, że ludzie potrafią rozpoznawać akcje wykonywane przez człowieka na podstawie obserwacji położenia i ruchu jego stawów. Metody rozpoznawania akcji w oparciu o model kinematyczny zostały z powodzeniem zastosowane w domenie obrazów kolorowych (Yao et al., 2011) jak i głębi (Ohn-Bar & Trivedi, 2013).

Rozwój w dziedzinie rozpoznawania akcji i zachowań ludzi oraz postępujący rozwój w dziedzinie widzenia maszynowego i sztucznej inteligencji spowodował wyłonienie nowych obszarów zainteresowań badawczych jakimi są rozpoznawanie zachowań grup ludzi, rozpoznawanie akcji z perspektywy pierwszej osoby oraz rozpoznawanie osób na podstawie chodu (Poppe, 2010).

Większość prac naukowych skupia się na analizie zachowań człowieka na podstawie sekwencji wideo zarejestrowanych z perspektywy trzeciej osoby, co oznacza, że kamera zazwyczaj znajduje się w pewnej odległości od osób i obiektów przedstawionych na obrazie. Metody rozpoznawania akcji oparte na tym założeniu mogą być niewystarczające w przypadku bezpośredniej interakcji kamery lub osoby ją trzymającej z osobami wykonującymi akcje. Rozpoznawanie akcji z perspektywy pierwszej osoby (ang. *first person activity recognition*) zakłada, że kamera znajduje się w centrum zdarzeń (przykładowo jest przymocowana do elementu ubioru człowieka), co oznacza, że w większości przypadków znajduje się w ruchu, często szybkim, chaotycznym lub niekontrolowanym. Motywacją do podjęcia badań w tym kierunku może być chęć rozpoznawania akcji wykonywanych przez człowieka noszącego kamerę. Algorytmy te docelowo mogą znaleźć zastosowanie w systemach indeksacji zbiorów multimedialnych (ze względu na popularność kamer sportowych, np. kamery *GoPro*) lub w urządzeniach mobilnych noszonych przez człowieka (ang. *wearable devices*) takich jak *Google Glass*.

Większość literatury naukowej dotyczącej rozpoznawania akcji skupia się na scenariuszu polegającym na rozpoznawaniu zachowania jednego człowieka. W rzeczywistych zastosowaniach, np. w systemach dozoru wizyjnego w miejscach, w których przebywają zbiorowiska osób, prawidłowe rozpoznanie akcji wykonywanej przez każdą osobę nie zawsze jest możliwe (Lan et al., 2012). Przyczyną tego może być fakt, że analiza zachowania konkretnej osoby nie niesie wystarczającej informacji o wykonywanej przez niego akcji, a do prawidłowego jej rozpoznania potrzeba zrozumienia kontekstu w jakim znajduje się ta osoba. Przez kontekst można rozumieć środowisko w jakim znajduje się osoba, jej wcześniejsze akcje czy też relacje z innymi osobami znajdującymi się w jej otoczeniu. Aktywności grupowe to takie, w których osoby uczestniczące należą do jednej lub większej liczby grup pojęciowych (Aggarwal & Ryoo, 2011). Przykładem prostej aktywności grupowej może być grupa osób niosących duży przedmiot. Do prawidłowego rozpoznania takiej aktywności potrzebna jest zarówno analiza zachowań jednostek jak i powiązań pomiędzy nimi. Istnieje kilka typów zachowań grupowych i w większości przypadków literatura poświęca uwagę każdemu z osobna. W pracach (Cupillard et al., 2002; Dai et al., 2008) uwagę skupiono na rozpoznawaniu aktywności grup, w których każdy z jej członków pełni inną, unikalną rolę. Dla tego typu aktywności system musi być w stanie rozpoznawać zachowania każdej jednostki i analizować ich strukturę. Następstwem tego jest często hierarchiczna konstrukcja modeli zachowań, gdyż rozróżniamy akcje na poziomie jednostki od akcji całej zbiorowości (Aggarwal & Ryoo, 2011). Inny typ zachowań zbiorowych jest charakteryzowany przez ogólny ruch całej grupy. W przeciwieństwie do wcześniej wspomnianego typu, w tym przypadku analiza poszczególnych osób nie jest tak ważna jak analiza na poziomie ich zbioru. Zazwyczaj w procesie analizy i rozpoznawania tego typu aktywności hierarchiczne modele nie znajdują zastosowania (Khan & Shah, 2005).

Analiza parametrów czasowo-przestrzennych chodu (ang. *gait analysis*) znajduje liczne zastosowania w medycynie i rehabilitacji, m. in. do oceny zmian organizmu w procesach chorobowych lub starzenia się. Analiza sposobu poruszania się człowieka może być użyteczna ze względu na możliwość zastosowań do identyfikacji osób na podstawie chodu. W oparciu o wielokamerowe, bezmarkerowe systemy śledzenia ruchu możliwe jest zbudowanie algorytmów uzyskujących wysoką dokładność rozpoznawania osób (Nixon & Carter, 2006; Switonski et al., 2011; Krzeszowski et al., 2013). Algorytmy te opierają się na założeniu, że sposób poruszania się jest indywidualną cechą osobową, która pozwala dokonać identyfikacji człowieka. Zadaniem rozpoznawania akcji jest natomiast generalizacja różnic między tymi samymi czynnościami poszczególnych osób i przydzielenie ich do wspólnych klas.

## 1.2. Miary jakości

Celem klasyfikacji jest znalezienie odwzorowania danych w zbiór predefiniowanych klas, a więc przydzielaniu danym $x_1, ..., x_n$ etykiet klas $C_1, ..., C_l$. Proces ten można podzielić na dwa etapy: budowa modelu w oparciu o znany zbiór danych i klas (zbiór uczący) oraz zastosowanie modelu do klasyfikacji nowych danych (testowy zbiór danych). Mając na względzie sposób odwzorowywania danych w zbiór klas, zagadnienie klasyfikacji jako problem uczenia maszynowego, może być podzielone na cztery grupy (Sokolova & Lapalme, 2009):

- klasyfikacja binarna – każda z danych wejściowych $x_1, ..., x_n$ jest przydzielana do jednej z dwóch nienakładających się na siebie klas $C_1, C_2$,
- klasyfikacja wieloklasowa – każda z danych wejściowych $x_1, ..., x_n$ jest przydzielana do jednej z $l$ nienakładających się na siebie klas $C_1, ..., C_l$,
- klasyfikacja wieloetykietowa – każda z danych wejściowych $x_1, ..., x_n$ jest przydzielana do jednej lub większej liczby nienakładających się na siebie klas $C_1, ..., C_l$,
- klasyfikacja hierarchiczna – każda z danych wejściowych $x_1, ..., x_n$ jest przydzielana do jednej z klas, która może być podzielona na podklasy lub tworzyć klasy nadrzędne. Hierarchia klas jest zdefiniowana przed klasyfikacją i nie może być zmieniana podczas jej trwania.

Ocenę jakości klasyfikatora dokonuje się najczęściej w sposób empiryczny. Można tego dokonać zliczając odpowiedzi klasyfikatora dla próbek, które mogą być:

- *TP* - prawdziwie dodatnie (ang. *true positive*) – poprawnie rozpoznana próbka, która należy do klasy.
- *TN* - prawdziwie ujemne (ang. *true negative*) – poprawnie rozpoznana próbka, która nie należy do klasy.
- *FP* - fałszywie dodatnie (ang. *false positive*) – niepoprawnie rozpoznana próbka, zakwalifikowana jako należąca do klasy, lecz w rzeczywistości do niej nie należy. Jest to błąd pierwszego rodzaju (ang. *type I error*).
- *FN* - fałszywie ujemne (ang. false *negative*) – niepoprawnie rozpoznana próbka, zakwalifikowana jako nienależąca do klasy, lecz w rzeczywistości do niej należy. Jest to błąd drugiego rodzaju (ang. *type II error*).

Zestawienie wyników przedstawia się często w postaci macierzy pomyłek (ang. *confusion matrix*) (zob. tabela 1.1.)

Tabela 1.1. Macierz pomyłek dla klasyfikacji binarnej.

|  |  | *Klasy rzeczywiste* | |
|---|---|---|---|
|  |  | *pozytywna* | *negatywna* |
| *Klasy uzyskane* | *pozytywna* | TP | FP |
|  | *negatywna* | FN | TN |

Na podstawie wyników klasyfikacji można określić miary klasyfikatora, które pozwalają porównywać i oceniać jakość klasyfikacji. Dla klasyfikacji binarnej są to (Sokolova & Lapalme, 2009):

- dokładność (ang. *accuracy*) – pozwala ocenić ogólną efektywność klasyfikacji:

$$accuracy = \frac{TP + TN}{TP + FP + FN + TN} \quad (1.1)$$

- precyzja (ang. *precision*) – pozwala określić, jaka część wyników wskazanych przez klasyfikator jako pozytywne, faktycznie należy do tej klasy:

$$precision = \frac{TP}{TP + FP} \quad (1.2)$$

- czułość (ang. *sensitivity*, *recall*) – zdolność klasyfikatora do identyfikowania wyników dodatnich:

$$sensitivity = \frac{TP}{TP + FN} \quad (1.3)$$

- swoistość (ang. *specificity*) – zdolność klasyfikatora do identyfikowania wyników ujemnych:

$$specificity = \frac{TN}{TN + FP} \quad (1.4)$$

- *F-score* – określa relację pomiędzy próbkami prawdziwie dodatnimi, a tymi uznanymi przez klasyfikator za dodatnie, przy $\beta$ określającej wagę precyzji względem czułości:

$$F_\beta = \frac{(1+\beta^2)TP}{(1+\beta^2)TP + \beta^2 FN + FP} \quad (1.5)$$

## 1.3. Problematyka detekcji upadku

Znaczący wzrost długości życia o około trzydzieści lat w krajach rozwiniętych, takich jak państwa Europy Zachodniej, Stany Zjednoczone, Kanada, Australia, Japonia, wydaje się być jednym z najważniejszych osiągnięć ludzkości w XX wieku (Christensen et al., 2009). W rozwiniętych krajach segment populacji powyżej 60. roku życia rośnie bardzo szybko i obecnie wynosi on około 600 milionów osób. Według Światowej Organizacji Zdrowia liczba ta podwoi się w ciągu najbliższych dziesięciu lat. Jest to efekt coraz skuteczniejszej ochrony zdrowia, rozwoju technologicznego i ekonomicznego, ale jednocześnie wyzwanie dla społeczeństwa do adaptacji pod kątem potrzeb ludzi starszych.

Jednym z poważniejszych zagrożeń dla zdrowia i życia ludzi starszych są upadki. Od 28 do 35 procent ludzi w wieku powyżej 65. roku życia upada przynajmniej raz w roku, zaś dla segmentu populacji powyżej 80. roku życia liczba ta zwiększa się do 50%. Od 20 do 30 procent upadających ludzi odnosi obrażenia, które mieszczą się w grupie od średnich do poważnych (Heinrich et al., 2010), a upadki są przyczyną około 90% złamań biodra (Fuller, 2000). Innymi częstymi obrażeniami powstałymi w wyniku upadków są urazy głowy oraz złamania kończyn górnych. Badania dotyczące ryzyka obrażeń przy upadkach wskazują także, że połowa osób, które upadły nie było w stanie podnieść się bez pomocy osoby trzeciej. We wspomnianych badaniach empirycznych uzyskano wyniki, które wskazują na to, że 14% osób leżało dłużej niż 5 minut bez uzyskania pomocy (3% powyżej 20 minut i 0,55% dłużej niż 8 godzin (Nevitt et al., 1991). Brak możliwości podniesienia się o własnych siłach wynika z powstałych obrażeń ciała lub malejącej z wiekiem aktywności mięśni, a długie przebywanie w pozycji leżącej po upadku może prowadzić do hipotermii, odwodnienia, zapalenia płuc czy odleżyn (Rubenstein & Josephson, 2002). Biorąc pod uwagę fakt, że około jedna trzecia ludzi powyżej 60. roku życia mieszka samotnie, negatywnym zjawiskiem jest także wytworzenie się strachu przed upadkiem, który zwykle skutkuje zmniejszeniem aktywności fizycznej, a przez co pogarsza się jakość życia osoby starszej.

Problem detekcji upadku sprowadza się do ciągłej analizy danych pochodzących, zależnie od metody, z sensorów noszonych przez osobę lub zewnętrznych urządzeń i odróżnieniu zdarzeń związanych z upadkiem od innych czynności życia codziennego (ang. *Activities of Daily Living – ADL*). Wraz z rozwojem badań nad metodami detekcji upadku ukształtowały się wymagania (Igual et al., 2013; Yu, 2008) stawiane tej technologii. Decyzja o wystąpieniu lub niewystąpieniu upadku podjęta przez system powinna być tak wiarygodna, jak jest to tylko możliwe. Wyniki osiągane w warunkach doświadczalnych, szczególnie w przypadku prostych metod, ulegają pogorszeniu po zastosowaniu systemu w rzeczywistym, niekontrolowanym środowisku (Noury et al., 2007). Innymi słowy, solidny system detekcji upadku powinien poprawnie klasyfikować upadki i czynności życia codziennego w naturalnym środowisku osoby starszej. Następną ważną cechą jest zdolność do ciągłego monitorowania użytkownika systemu i podejmowania decyzji w jak najkrótszym czasie. Dobry system detekcji upadku powinien zapewniać ciągły monitoring osoby starszej oraz działać nieprzerwanie przez całą dobę każdego dnia roku (ang. *year-round, continuous operation, 24/7/365*). Sukces w przystosowaniu technologii detekcji upadku do potrzeb użytkowników oraz późniejszym jej wdrożeniu możliwy jest po uwzględnieniu oczekiwań osób starszych. Badania wykazały, że coraz większa część starzejącej się populacji zaakceptowałaby nowe technologie w swoim otoczeniu celem poprawy jakości życia (Brownsell et al., 2000). Zaakceptowanie takiego systemu przez osobę starszą wiąże się z kilkoma wymaganiami. System powinien działać w pełni automatycznie lub wymagać minimum obsługi przez docelowego użytkownika, który nie zawsze radzi sobie z nowoczesną technologią (Kurniawan, 2008). Drugim ważnym wyzwaniem jest zapewnienie prywatności użytkownika. Problem ten dotyczy szczególnie systemów wizyjnych opartych na jednej lub wielu kamerach RGB. Użytkownik systemu nie powinien mieć poczucia inwigilacji czy też ingerencji w jego prywatne środowisko. Problem ten może się wydawać niezbyt istotny wtedy, gdy system instalowany jest w miejscu publicznym (np. korytarz czy biblioteka w domu pomocy społecznej). Jednak zastosowanie systemu niechroniącego prywatności w miejscach takich jak łazienka czy sypialnia osoby starszej jest niewskazane. Istnieje przekonanie, że korzyści płynące z detekcji upadku nie powinny być uzyskiwane kosztem ingerencji w prywatność użytkownika.

W literaturze brakuje jednoznacznego podziału technologii detekcji upadku. W pracy (Perry et al., 2009) zaproponowano podział metod na mierzące przyspieszenie, nie mierzące przyspieszenia oraz takie, które mierzą przyspieszenie w połączeniu z innymi metodami. Inny podział wprowadzony został w pracy (Noury et al., 2008) i dzieli metody detekcji na analizujące tylko moment uderzenia oraz takie, które uwzględniają jeszcze fazę po upadku (ang. *postfall phase*). W niniejszym podrozdziale przegląd metod został przedstawiony analogicznie do podziału zamieszczonego w pracy (Igual et al., 2013). We wspomnianej pracy systemy detekcji upadku podzielono na dwie grupy: urządzenia noszone przez użytkownika (ang. *wearable devices*) oraz instalowane w otoczeniu użytkownika (ang. *context aware systems*). Podział metod detekcji upadku według przyjętego kryterium przedstawia rysunek 1.2.

Większość proponowanych rozwiązań opiera się o analizę danych z urządzeń noszonych przez człowieka, które monitorują ruch, rozpoznają upadek i wyzwalają alarm. Szeroko stosowane są akcelerometry i żyroskopy. Ze względu na charakter danych ich gromadzenie jest stosunkowo łatwe oraz pozwala na pozyskanie ich w długim horyzoncie czasowym. Metody detekcji upadku działające w oparciu o dane z inercyjnych czujników ruchu można podzielić na trzy grupy: urządzenia wykorzystujące akcelerometr, żyroskop oraz akcelerometr wspólnie z żyroskopem. Decyzja o klasyfikacji akcji jako upadek podejmowana jest zazwyczaj w oparciu o predefiniowany próg lub polega na analizie pozycji w fazie po upadku. W pracy (Bourke et al., 2007) zaproponowano algorytm detekcji upadku działający w oparciu o maksymalną wartość przyspieszenia (ang. *Upper Peak Value – UPV*), natomiast w pracy (Bourke & Lyons, 2008) przedstawiony został algorytm oparty na prędkości kątowej – sygnału uzyskiwanego z żyroskopu. Podstawową wadą tych metod jest trudność odpowiedniego dopasowania progu wartości, przy którym wzbudzany jest alarm. Niektóre czynności życia codziennego, takie jak siadanie czy kucanie, w szczególności wykonywane szybko, mają zbliżone wartości przyspieszeń w porównaniu do upadków. Konsekwencją tego jest wysoki odsetek fałszywych alarmów w sytuacjach poza laboratorium, co zaprezentowano w pracy (Bagalà et al., 2012). Algorytm zaproponowany w pracy (Li et al., 2009) zakłada, że upadek zawsze kończy się w pozycji leżącej. Pozwala to na zniwelowanie części fałszywych alarmów poprzez odróżnienie upadków od czynności takich jak siadanie czy kucanie. Jednak takie założenie może prowadzić do spadku skuteczności wykrywania upadków, zwłaszcza gdy upadek kończy się w pozycji siedzącej.

Podjęto szereg prób wyeliminowania wymienionych wcześniej wad. Niektóre z nich sprowadzały się do wykorzystania dwóch lub więcej urządzeń noszonych przez osobę (Noury et al., 2007). Jednak takie podejście może być niekomfortowe w użytkowaniu przez osoby starsze i zostać uznane za inwazyjne. System działający jedynie w oparciu o sensory noszone przez osobę może być nieskuteczny, gdyż niektóre czynności, takie jak przebieranie się czy kąpiel, wymagają zdjęcia tych urządzeń.

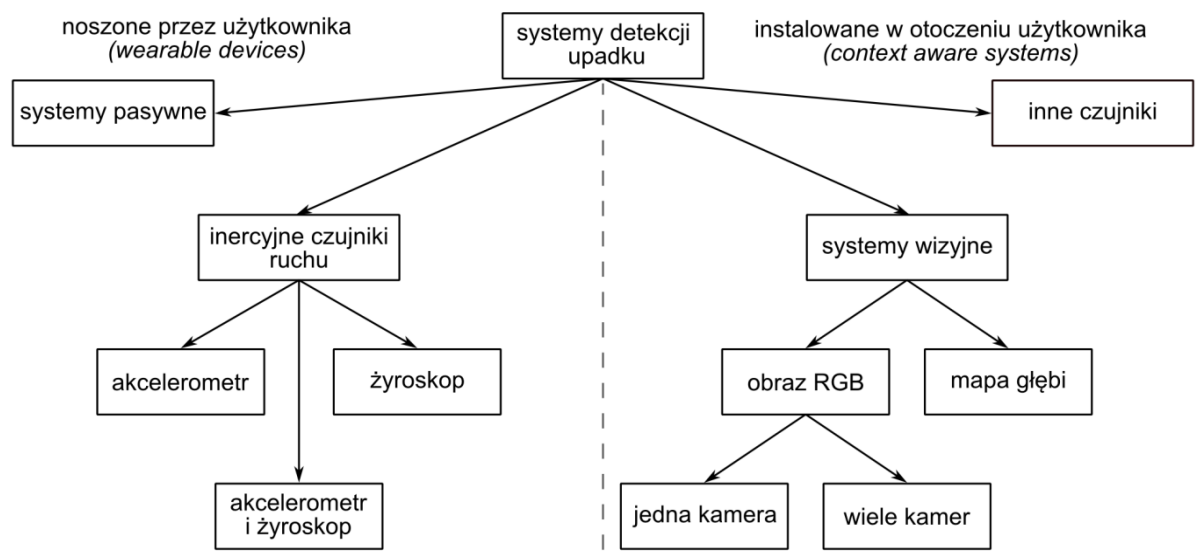

Rysunek 1.2. Podział systemów detekcji upadku ze względu na wykorzystaną technologię.

Urządzenia takie są niewygodne i zwykle muszą być zdejmowane na czas snu. Nie zapewniają więc właściwego monitoringu podczas wstawania i przebierania się, a więc czynności podczas wykonywania których często dochodzi do upadków. Często też osoby starsze zapominają o noszeniu ich czy konieczności ładowania baterii. Urządzenia pasywne w zasadzie nie dokonują automatycznej detekcji upadku, zaś fakt wystąpienia upadku i potrzeba udzielenia pomocy jest zgłaszany przez użytkownika. Systemy te mają najczęściej postać niewielkiego urządzenia wyposażonego w przycisk i możliwość komunikacji z systemem telemedycznym. Takie rozwiązania są niewystarczające ze względu na bezpieczeństwo użytkownika, gdyż statystyki pokazują, że duża część osób po upadku traci przytomność. Przykładem takiego urządzenia może być *Philips Lifeline HomeSafe Standard*[1].

Następnym podejściem jest zastosowanie urządzeń monitorujących otoczenie, instalowanych w miejscach przebywania użytkownika. Pierwsza grupa metod zakłada wykorzystanie sensorów nie będących urządzeniami wizyjnymi takich jak czujniki nacisku, wibracji (Zigel et al., 2009), pasywne czujniki podczerwieni (A. Sixsmith et al., 2005), mikrofony (Y. Li et al., 2010) czy radary (Piorek & Winiecki, 2015). Zastosowanie tego typu technologii zwalnia osoby starsze z konieczności noszenia przy sobie urządzeń, lecz wymusza czasochłonną instalację oraz częściową ingerencję w miejsce zamieszkania (ułożenie przewodów, montaż czujników i doprowadzenie zasilania), co może spotkać się z brakiem akceptacji przez użytkowników oraz ze znaczącymi kosztami użytkowania.

Druga grupa metod opiera się na wykorzystaniu technik widzenia komputerowego do śledzenia postaci i wykrywania upadku. Do tego celu wykorzystano systemy oparte na jednej kamerze CCD (Rougier et al., 2006), kilku kamerach (Cucchiara et al. 2007; Rougier et al. 2011), specjalnej kamerze dookólnej (Wang et al., 2006) lub kamerze głębi (Mastorakis & Makris, 2012; Kępski & Kwolek, 2012b; Kwolek & Kępski, 2013). Metody analizy video wydają się być adekwatne, gdyż dane z kamery są semantycznie bogatsze i niosą więcej informacji od zwykłych sensorów inercyjnych. Niestety do przetwarzania i analizy obrazów wymagane są często algorytmy o znacznych wymaganiach obliczeniowych. Najprostszym podejściem jest analiza sylwetki lub jej prostokątnego obrysu (ang. *bounding box*). Jednak dokładność tych metod zależy w dużym stopniu od pozycji człowieka względem kamery, umiejscowienia kamery oraz występujących przysłonięć. Aby wyeliminować te trudności niektórzy autorzy (Lee & Mihailidis, 2005) proponują umiejscowienie kamery w centralnym punkcie sufitu pomieszczenia lub kamerę dookólną. Przedstawiono kilka rozwiązań opartych o systemy wielokamerowe, których skuteczność jest wyższa niż w przypadku użycia pojedynczej kamery (Cucchiara et al., 2007). Jednak systemy te potrzebują znaczących mocy obliczeniowych, co sprawia, że w celu uzyskania przetwarzania w czasie rzeczywistym wymagane jest niekiedy przetwarzanie na GPU (Auvinet et al., 2011). Bardziej zaawansowane metody polegają na analizie konfiguracji modelu 3D reprezentującego pozę postaci. Ze względu na brak informacji o głębokości, precyzyjne dopasowanie modelu jest trudne do przeprowadzenia, szczególnie w tak podatnym na przysłonięcia środowisku jak mieszkanie czy też pojedynczy pokój. Warto przy tym wspomnieć, że kamery internetowe czy kamery działające w technologii *GigE* mogą posłużyć do zdalnej weryfikacji, czy nastąpił upadek. Główną wadą metod opartych na analizie obrazów RGB jest niemożliwość zapewnienia prywatności osobie, która jest monitorowana. Zastosowanie kamer w środowisku domowym osoby starszej może być odebrane jako zbyt inwazyjne. To, że system podłączony jest do sieci, np. w celu zdalnej weryfikacji upadku, może być krępujące, nie wspominając już o możliwości obserwacji czynności dnia codziennego takich jak mycie się czy przebieranie. Poza tym, czas działania systemów opartych na kamerach RGB jest ograniczony, gdyż mogą one działać jedynie przy odpowiednim oświetleniu, a więc do momentu zapadnięcia zmierzchu lub wyłączenia oświetlenia sztucznego. Nie zapewniają one odpowiedniej ochrony po wyłączeniu oświetlenia, w szczególności gdy osoba kładzie się spać lub wstaje w nocy z łóżka.

---

[1] http://www.lifelinesys.com/content/lifeline-products/confident-home-homesafe

Wraz ze wzrostem popularności urządzeń do obrazowania 3D zaprojektowanych na potrzeby interakcji człowiek - komputer (takich jak *Microsoft Kinect*), do detekcji upadku rozpoczęto wykorzystywać kamery udostępniające mapy głębokości (Rougier et al. 2011; Kępski & Kwolek, 2012a; Stone & Skubic 2013). Przetransformowanie mapy głębokości pozwala na uzyskanie trójwymiarowej reprezentacji sceny w postaci chmury punktów. Jest to istotne w zagadnieniach detekcji upadku, gdyż dzięki mapom głębi możliwe jest określenie rzeczywistych rozmiarów i położenia ludzi oraz przedmiotów. Ponadto, akwizycja obrazów pozbawionych informacji o kolorze i teksturze obiektów nie prowadzi do znacznego naruszenia prywatności użytkownika, co w przypadku kamer RGB zazwyczaj jest problemem. Dodatkowym atutem jest możliwość zastosowania takiego systemu do ciągłego monitoringu użytkownika, gdyż urządzenia do akwizycji map głębi, takie jak sensor *Kinect* potrafią działać w dowolnych warunkach oświetleniowych, za wyjątkiem bardzo silnego światła słonecznego (dotyczy to jedynie scenariuszy, gdy urządzenie znajduje się na zewnątrz). W pracy (Rougier et al. 2011) przedstawiono algorytm oparty na analizie odległości środka geometrycznego postaci od podłogi oraz prędkości środka ciężkości sylwetki. W pracy (Mastorakis & Makris, 2012) zaadaptowano metodę analizy prostokątnego obrysu sylwetki do danych 3D. Podjęto także kilka prób wykorzystania szkieletu 3D na potrzeby detekcji upadku (Parra-Dominguez et al., 2012; Planinc & Kampel, 2012).

W tabeli 1.2. przedstawiono zestawienie wybranych systemów detekcji upadku z podziałem na trzy grupy, którym literatura naukowa poświęca najwięcej uwagi: inercyjne czujniki ruchu, systemy wizyjne oraz systemy wizyjne oparte o kamery głębi. Ze względu na obszerność literatury przedmiotu w tabeli ujęto jedynie kilka prac dla każdej z grup, a bardziej szczegółowy przegląd można znaleźć w pracach (Igual et al., 2013; Webster & Celik, 2014). Celem umożliwienia analizy porównawczej wyników, przy wyborze prac przyjęto pewne kryteria. W omawianym zestawieniu zaprezentowano tylko te prace, które opisują jaki był scenariusz wykonywanych eksperymentów: liczba osób, liczba wykonanych akcji (z podziałem na upadki i czynności życia codziennego). W zastawieniu pominięto wyniki, dla których liczba eksperymentów lub osób wykonujących akcje była mała (1 - 3 osoby). W grupie systemów opartych o inercyjne czujniki ruchu przedstawiono rezultaty uzyskane w pracy porównawczej (Bagalà et al., 2012), co pozwala na rzetelną ocenę skuteczności metod detekcji upadku - eksperymenty odbywały się w oparciu o te same dane, uzyskane w środowisku niekontrolowanym, poza laboratorium. W grupie systemów wizyjnych, ze względu na brak odpowiednich eksperymentów porównawczych, przedstawiono wyniki uzyskane przez poszczególne zespoły badawcze.

Tabela 1.2. Porównanie wybranych systemów detekcji upadku.

| Autor | Rok | Użyte cechy | Liczba osób | Wyniki |
|---|---|---|---|---|
| Detekcja upadku w oparciu o inercyjne czujniki ruchu | | | | |
| Chen et al. | 2005 | wartość przyspieszenia, zmiana orientacji | | *accuracy:* 93,7% *sensitivity:* 76,0% *specificity:* 94,0% |
| (Bourke et al.) | 2007 | wartość przyspieszenia | 15 osób 32 upadki 1170 ADLs wiek: 66.4 ± 6.2 lat | *accuracy:* 21,3% *sensitivity:* 100% *specificity:* 19,0% |
| (Kangas et al.) | 2009 | wartość przyspieszenia, prędkość, położenie ciała człowieka po upadku | | *accuracy:* 96,7% *sensitivity:* 28,0% *specificity:* 98,0% |
| (Bourke et al.) | 2010 | wartość przyspieszenia, prędkość, położenie ciała człowieka po upadku | | *accuracy:* 96,3% *sensitivity:* 83,0% *specificity:* 97,0% |
| Detekcja upadku w oparciu o kamery RGB | | | | |
| Lee & Mihailidis | 2005 | cechy geometryczne sylwetki człowieka (obwód i oś główna), prędkość środka ciężkości sylwetki | 21 osób<br>126 upadków<br>189 ADLs<br>wiek: 20 - 40 lat | *sensitivity*: 93,9%<br>*specificity*: 80,5% |
| Miaou et al. | 2006 | cechy geometryczne sylwetki (wysokość, szerokość) | 20 osób<br>33 upadki<br>27 ADLs<br>wiek: brak danych | *accuracy*: 81,0%<br>*sensitivity*: 90,0%<br>*specificity*: 86,0% |
| Liu et al. | 2010 | szerokość i wysokość prostokątnego obrysu sylwetki | 15 osób<br>45 upadków<br>45 ADLs<br>wiek: 24 - 60 lat | *accuracy*: 84,4% |
| Shoaib et al. | 2011 | odległości między elementami sylwetki (głowa, nogi), kontekst sceny | 5 osób<br>43 upadki<br>30 ADLs<br>wiek: brak danych | *sensitivity*: 97,7%<br>*specificity*: 90,0% |
| Detekcja upadku w oparciu o kamery głębi (w nawiasach podano dokładność metody według pracy (Stone & Skubic, 2014)) | | | | |
| Zhang et al. | 2012 | wysokość postaci w 3D | 5 osób<br>200 sekwencji (upadki + ADLs)<br>wiek: brak danych | *accuracy*: 94,0% |
| Mastorakis & Makris | 2012 | obrys sylwetki w 3D | 8 osób<br>48 upadków<br>136 ADLs<br>wiek: brak danych | *accuracy*: 100% (*sensitivity:* < 80%) |
| Planinc & Kampel | 2012 | orientacja ciała w 3D na podstawie modelu szkieletu postaci | 4 osoby<br>40 upadków<br>32 ADLs<br>wiek: brak danych | *accuracy*: 95,8% (*sensitivity:* < 60%) |
| Stone & Skubic | 2014 | orientacja ciała w 3D oraz orientacja ciała | 16 osób<br>454 upadki<br>wiek: brak danych | *sensitivity*:<br>z pozycji:<br>  stojącej: 98,0%<br>  siedzącej: 70,0% |

Analizując rezultaty badań przedstawione w tabeli 2. zauważyć można, że uzyskanie wysokiej czułości detekcji przy pomocy inercyjnych czujników ruchu wiąże się z wysokim odsetkiem wyników fałszywie pozytywnych. Najprostsze algorytmy detekcji upadku, działające w oparciu o wartość przyspieszenia, poddane ciągłej, długotrwałej ewaluacji generują od 22 do 85 fałszywych alarmów na dobę (Bagalà et al., 2012), co jest wartością nie do zaakceptowania przez potencjalnych użytkowników tej technologii. Detekcja upadku w oparciu o obraz RGB charakteryzuje się mniejszą skutecznością niż w przypadku kamer głębi. Jednak opracowane dotychczas algorytmy nie są pozbawione wad i ich skuteczność maleje dla trudniejszych scenariuszy. Przykładowo w pracy (Stone & Skubic, 2014) uzyskano wysoką czułość dla upadków z pozycji stojącej, jednak detekcja zdarzeń rozpoczynających się z pozycji siedzącej okazała się być problematyczna.

## 1.4. Bazy danych

W dziedzinie uczenia maszynowego ocena skuteczności algorytmów realizowana jest najczęściej w sposób empiryczny poprzez analizę miar jakości dla zestawu danych testowych. Problem ewaluacji algorytmów opracowywanych przez różne zespoły badawcze jest istotny ze względu na możliwość analizy porównawczej otrzymanych wyników. Ze względu na charakter danych problem ten jest szczególnie istotny w dziedzinie widzenia maszynowego. Obiekty uwidocznione na obrazach mogą mieć różny wygląd, ponadto mogą być przedstawione z różnego położenia kamery. Na użyteczność danych wpływ mają także czynniki zewnętrzne, takie jak poziom naświetlenia czy stopień skomplikowania sceny. Rozpoznawanie akcji jest bardziej wymagającym zadaniem ze względu na mnogość akcji z jednej strony, z drugiej zaś strony dane mogą różnić się złożonością, np. dwie pozornie te same akcje różnią się znacząco pod względem trudności ich analizy i klasyfikacji. Zdecydowanie trudniej jest dokonać detekcji upadku jeżeli osoba upada z pozycji siedzącej w porównaniu do upadku z pozycji siedzącej, w szczególności gdy osoba uległa przysłonięciu lub dokonała interakcji z otoczeniem. Ponadto, różne osoby wykonują te same akcje w różny sposób. Różna liczba próbek danych oraz liczebność klas decyzyjnych może mieć wpływ na rezultaty klasyfikacji. Wszystkie te czynniki mogą mieć wpływ na uzyskiwane wyniki, co zostało zasygnalizowane w pracach (Bagalà et al., 2012; Stone & Skubic, 2014).

Rozwiązaniem tego problemu jest ewaluacja algorytmów w oparciu o wspólne, ogólnodostępne dane, zgodnie z jednym protokołem postępowania. Wraz z rozwojem metod przetwarzania i analizy obrazów powstało wiele publicznie dostępnych zbiorów sekwencji wideo. Repozytoria danych mogą zawierać obrazy RGB, pary obrazów RGB i głębi lub jedynie dane przestrzenne. Często, szczególnie w przypadku trudności praktycznych w analizie obrazów z wielu kamer, oprócz sekwencji obrazów udostępniane są dane pomocnicze, np. parametry kalibracji kamer wykorzystanych do akwizycji. Opracowano stosunkowo niewiele zbiorów danych umożliwiających ewaluację algorytmów do detekcji upadku. Istniejące repozytoria zawierają głównie obrazy RGB z jednej[2] lub wielu kamer (Auvinet et al., 2010; Anderson et al., 2009) oraz kamer głębi[3,4,5]. Przykładowe obrazy ze wspomnianych zbiorów danych przedstawiono na rysunku 1.3.

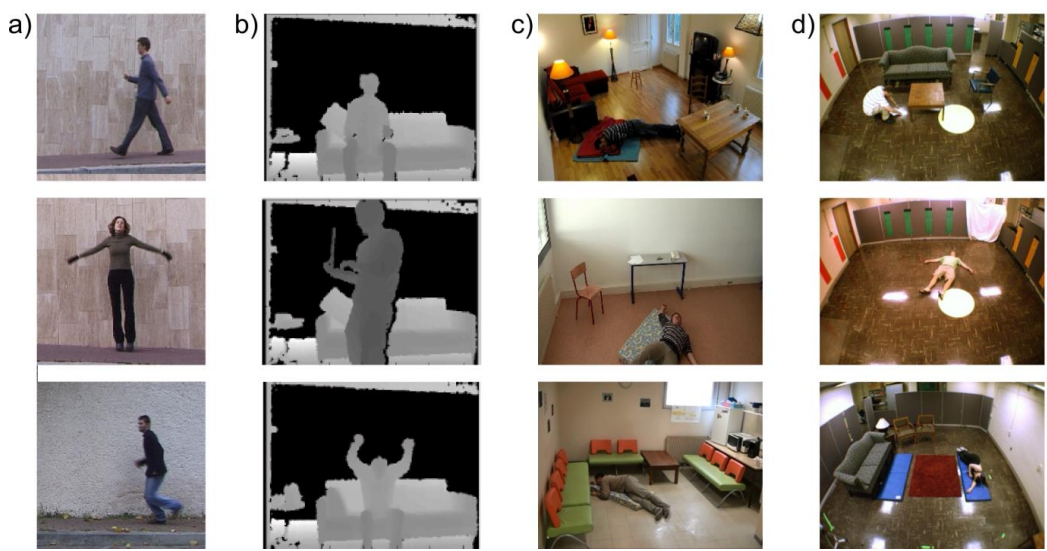

Rysunek 1.3. Przykładowe obrazy z różnych zbiorów danych do rozpoznawania akcji a) i b) oraz do detekcji upadku c) i d).

Na potrzeby oceny skuteczności detekcji metod prezentowanych w niniejszej pracy został przygotowany i opublikowany zbiór sekwencji UR Fall Detection Dataset (Kępski & Kwolek, 2014b). Przygotowany zbiór danych został wykorzystany przez kilka zespołów badawczych (Mazurek et al., 2015; Konstantinidis & Bamidis, 2015) oraz przytoczony w pracy przeglądowej o tematyce detekcji upadku (Zhang et al., 2015). Dane zostały zarejestrowane dwiema kamerami *Microsoft Kinect*, różniących się między sobą ustawieniem względem sceny oraz akcelerometrem x-IMU. Na wspomniany zbiór składa się 70 sekwencji prezentujących upadki (30 sekwencji) i czynności dnia codziennego (40 sekwencji) wykonywane przez 6 osób (5 w wieku 23-28 lat oraz jedna osoba w wieku 50+). Każda klatka sekwencji upadków reprezentowana jest w postaci 4 obrazów: dwóch par (kolorowego i głębi) dla każdej z kamer oraz zsynchronizowanego odczytu danych akcelerometrycznych. Dane te zostały zarejestrowane w typowym pokoju biurowym, o umiarkowanie skomplikowanej scenerii (jednak niepozbawionej innych obiektów sceny niż aktor). Każda klatka sekwencji ADLs jest reprezentowana w postaci jednej pary obrazów (kolorowego i głębi) oraz zsynchronizowanego odczytu danych akcelerometrycznych. Dane zostały zarejestrowane w pokoju biurowym (tym samym co upadki) oraz w trzech różnych pomieszczeniach w domu mieszkalnym (o typowej, niezmodyfikowanej na potrzeby nagrań scenerii). Wysokość umieszczenia kamery jest typowa dla pomieszczeń użyteczności publicznej. Przykładowe obrazy ze zbioru danych przedstawiono na rysunku 1.4. Część obrazów zawiera przysłonięcia, na niektórych fragmenty postaci są poza kadrem, co charakteryzuje dane pochodzące z systemów wizyjnych zastosowanych poza laboratorium. Warto wspomnieć, że opracowany zbiór jest pierwszym i do tej pory jedynym zbiorem, który zawiera zsynchronizowane obrazy RGBD oraz dane inercyjnego czujnika ruchu.

---

[2] http://le2i.cnrs.fr/Fall-detection-Dataset
[3] http://fenix.univ.rzeszow.pl/~mkepski/ds/uf.html
[4] https://sites.google.com/site/kinectfalldetection/
[5] http://www.sucro.org/homepage/wanghaibo/SDUFall.html

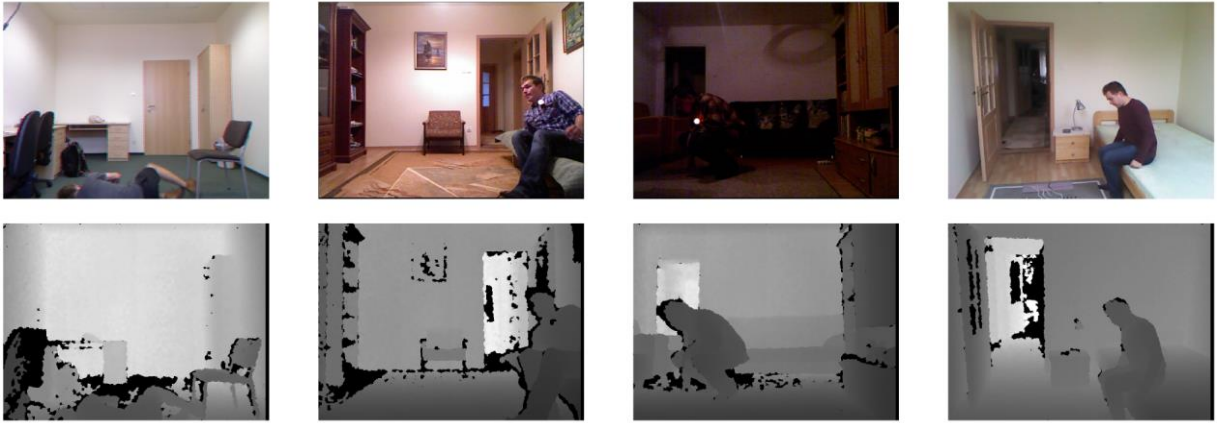

Rysunek 1.4. Przykładowe obrazy ze zbioru UR Fall Detection Dataset. Górny rząd: obrazy kolorowe, dolny rząd: korespodujące mapy głębi.

Sekwencje obrazów są zapisane w zbiorach plików PNG (ang. *Portable Network Graphics*) - rastrowym formacie plików wykorzystującym bezstratną metodę kompresji danych. Mapy głębi są zapisane w formacie PNG16, pozwalającym na przechowywanie obrazów w reprezentacji 16-bitowej. Zostały one przeskalowane tak, aby wykorzystać cały przedział możliwych wartości pikseli: <0, 65535>, dzięki czemu możliwa jest łatwa wizualizacja mapy głębi. Aby uzyskać wartość odległości od kamery należy dokonać przekształcenia:

$$d = \frac{C_i P(x,y)}{65535} \qquad (1.6)$$

gdzie $d$ - wartość odległości od kamery, $C_i$ współczynnik skalowania zależny od ustawienia kamery, a $P(x,y)$ to wartość piksela obrazu PNG16 o współrzędnych $(x,y)$. Dla każdej z sekwencji przygotowano dwa pliki CSV (ang. *Comma Separated Values*, wartości rozdzielone przecinkiem). Plik synchronizacyjny zawiera: numer klatki obrazu, czas w milisekundach jaki upłynął od pierwszej klatki sekwencji i korespondującą z daną klatką obrazu, interpolowaną wartość przyspieszenia:

$$SV_{total} = \sqrt{A_x^2 + A_y^2 + A_z^2} \qquad (1.7)$$

gdzie $A_x, A_y, A_z$ to wartość przyspieszenia względem danej osi układu współrzędnych. Plik z danymi akcelerometrycznymi zawiera znacznik czasowy momentu pomiaru przyspieszenia (ang. *timestamp*), liczony od pierwszej klatki sekwencji i wartości zmierzone w tym czasie: przyspieszenie $SV_{total}$ oraz przyspieszenia względem poszczególnych osi akcelerometru: $A_x, A_y, A_z$. Wszystkie wartości akcelerometryczne podane są w jednostkach przyspieszenia ziemskiego [$g$]. Ponadto, dla każdego z obrazów, zbiór zawiera zestawy wartości deskryptorów wykorzystanych w procesie klasyfikacji i opisanych w rozdziale 3.

## 1.5. Proponowane podejście do problemu

Punktem wyjścia do badań nad detekcją upadku były metody oparte o klasyfikatory i zestawy deskryptorów opisujące akcje wykonywane przez człowieka. Podejście takie wpisuje się w główny kierunek badań prowadzonych w dziedzinie detekcji upadku (Mastorakis & Makris 2012; Planinc & Kampel 2012). Zgodnie ze sformułowaną w niniejszej pracy tezą podjęto prace badawcze mające na celu wykorzystanie sensora głębi i inercyjnego czujnika ruchu, który dostarcza informacji o kontekście w jakim znajduje się osoba (Kępski & Kwolek, 2013). Te wstępne badania złożyły się na sformułowanie dodatkowych kryteriów, które są istotne w kontekście potencjalnych zastosowań. Mając na względzie powyższe czynniki, zdecydowano się na zaprojektowanie systemu wbudowanego o niskim zużyciu energii oraz niewymagającego kalibracji. Ze względu na ograniczoną moc obliczeniową platformy sprzętowej z procesorem o architekturze ARM zaproponowano hierarchiczną architekturę systemu, w której:

- wstępna hipoteza o wystąpieniu upadku jest podejmowana na podstawie analizy danych z inercyjnego sensora ruchu,
- obrazy przechowywane są w buforze cyklicznym w celu weryfikacji hipotezy o upadku postaci lub dokonania aktualizacji modelu tła w wypadku zmian monitorowanego otoczenia.

Idea hierarchicznej architektury systemu została przedstawiona na rysunku 1.5. Zaletą takiego podejścia jest brak konieczności ciągłego przetwarzania obrazów, szczególnie w sytuacjach dłuższej nieaktywności użytkownika. Okresy nieaktywności użytkownika, czyli takie, w których osoba pozostaje w bezruchu dłuższy czas, mogą zostać wykryte na podstawie analizy danych pochodzących z inercyjnego sensora ruchu. Biorąc pod uwagę przyjęte założenie, że docelowy system powinien nieustannie monitorować osobę, liczba obrazów, które nie są przetwarzane, jest znaczna.

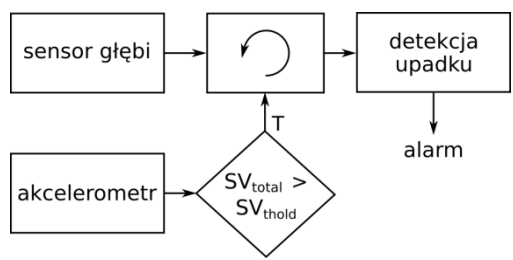

Rysunek 1.5. Schemat ideowy hierarchicznego systemu detekcji upadku.

Mając na uwadze parametry dostępnych na rynku sensorów głębi, w szczególności ich kąty widzenia oraz ich zasięg, które określają wielkość monitorowanego obszaru, zaproponowano i przebadano scenariusz, w którym kamera umieszczona jest na suficie (Kępski & Kwolek, 2014a). Zbyt mała powierzchnia monitorowania, a także brak możliwości jej zwiększenia w oparciu o różnego rodzaju nakładki na obiektyw, doprowadziły do umieszczenia kamery na obrotowej głowicy *pan-tilt,* którą wykonano na potrzeby niniejszej pracy.

Przygotowano także zestaw algorytmów wstępnego przetwarzania obrazów, do którego zaliczyć można:

- algorytm detekcji podłogi – przedstawiono propozycję połączenia metody *v-disparity* i algorytmu RANSAC do estymacji parametrów modelu płaszczyzny reprezentującej podłogę monitorowanego pomieszczenia,
- algorytm budowy referencyjnego modelu tła – dokonano modyfikacji klasycznego algorytmu opartego o medianę obrazów, pozwalającą na aktualizację tła w wybranych obszarach zainteresowania obejmujących te elementy sceny, które uległy zmianie,
- algorytm rozrostu obszarów – zaproponowano modyfikację klasycznego algorytmu rozrostu obszarów, w której dokonywany jest rozrost pojedynczego obszaru reprezentującego monitorowaną postać, co w konsekwencji prowadzi do zmniejszonych wymagań obliczeniowych algorytmu,
- algorytm śledzenia postaci – zaproponowano algorytm śledzenia osoby w oparciu o filtr cząsteczkowy, w którym kształt głowy postaci zamodelowany jest za pomocą elipsoidy,
- algorytm detekcji osoby – zaproponowano algorytm detekcji osoby w oparciu o histogramy gradientów i klasyfikator oparty o maszynę wektorów wspierających (ang. *support vector machine,* SVM).

Efektem badań i wynikiem doświadczenia nabytego przy pracy nad metodami detekcji upadku była propozycja zastosowania w rozpatrywanym problemie rozmytego systemu wnioskującego (Kwolek & Kępski, 2016).

# 2. System do detekcji i śledzenia osób

W niniejszym rozdziale zaprezentowano system do detekcji i śledzenia osób na podstawie map głębi. Omówiono dane wejściowe wraz z charakterystyką urządzeń do akwizycji danych, mobilną platformę obliczeniową oraz aktywną głowicę *pan-tilt*. Przedstawiono algorytmy przetwarzania obrazów i percepcji sceny, które składają się na system detekcji akcji. Następnie przedstawiono metody detekcji i śledzenia postaci ludzkiej opracowane dla potrzeb detekcji upadku.

## 2.1. Urządzenia i dane wejściowe

### 2.1.1. Sensor i mapy głębi

Do akwizycji sekwencji obrazów wykorzystano urządzenie RGB-D, posiadające kamerę kolorową i sensor głębi. Urządzenia tego typu (*Microsoft Kinect, Asus Xtion PRO Live*) umożliwiają użytkownikowi interakcję z konsolą *Microsoft Xbox 360* lub komputerem bez konieczności używania kontrolera, poprzez interfejs wykorzystujący gesty wykonywane przy pomocy kończyn i całego ciała. Stosunkowo niewielka cena oraz fakt, że sensor jest wyposażony w zwykłe złącze USB spowodowały, że *Kinect* nie tylko odniósł sukces komercyjny, ale również wzbudził zainteresowanie zespołów badawczych zajmujących się między innymi: rekonstrukcją 3D, interakcją człowiek-komputer, technikami SLAM i innymi dziedzinami multimediów.

Urządzenie posiada dwie kamery służące do akwizycji obrazu. Pierwsza z nich to zwykła kolorowa kamera RGB o rozdzielczości 640x480, która pobiera dane z częstotliwością 30 klatek na sekundę (ang. *frames per second – fps*) dostarcza obrazy w 8 bitowej głębi. Za pozyskiwanie informacji o głębokości odpowiedzialna jest para: emiter podczerwieni – kamera podczerwieni. System działa na zasadzie pomiaru odległości w oparciu o triangulację. Emiter podczerwieni wyświetla siatkę punktów (zob. rysunek 2.1.), których położenia rejestrowane są przez monochromatyczną kamerę CMOS. Układ punktów zarejestrowany przez kamerę, porównywany jest z referencyjnym obrazem siatki zapisanym w pamięci sensora. Jeśli wzorzec został wyświetlony na obiekcie, którego odległość jest inna niż odpowiadająca mu odległość na obrazie referencyjnym, wówczas pozycja wzorca na obrazie z kamery będzie przesunięta. Różnice w położeniu wyznaczane są dla każdego piksela, co skutkuje uzyskaniem obrazu dysparycji, nazywanej także rozbieżnością (ang. *disparity*). Zależność pomiędzy rozbieżnością zobrazowania a odległością przedstawiona jest na rysunku 2.2. Aby wyznaczyć współrzędne punktu w przestrzeni trójwymiarowej rozważmy układ współrzędnych o początku w środku kamery (C). Oś X jest przedłużeniem odcinka pomiędzy kamerą a projektorem (P), zaś oś Z jest prostopadła do płaszczyzny obrazu. Odległość pomiędzy kamerą a projektorem (odcinek CP) jest nazywana linią bazową (ang. *baseline*) i oznaczona jako $b$. Jak wspomniano wcześniej, na podstawie różnicy położenia wzorca z siatki odniesienia i położenia tego samego wzorca na obrazie uzyskać można informację o odległości obiektów od kamery.

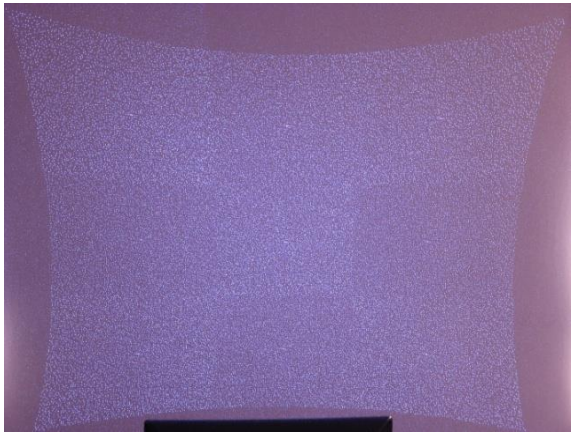
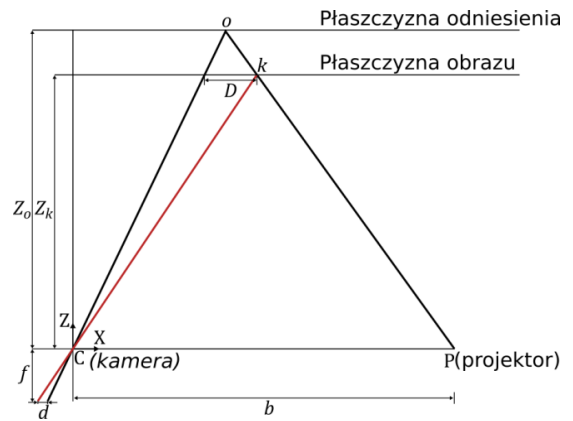

Rysunek 2.1. Siatka punktów projektora IR.  Rysunek 2.2. Zależność rozbieżność – głębokość.

Załóżmy, że punkt $k$ znajduje się na płaszczyźnie obrazu w odległości $Z_k$ od kamery, a na płaszczyźnie odniesienia ten sam punkt $o$ znajduje się w odległości $Z_o$ od kamery. Gdy obiekt będzie przesuwał się w stronę sensora lub od niego to będzie zmieniało się położenie punktu $k$ wzdłuż osi X. Dzięki temu, korzystając z podobieństwa trójkątów, można wyznaczyć wartość rozbieżności $d$:

$$\frac{D}{b} = \frac{Z_o - Z_k}{Z_o} \qquad (2.1)$$

oraz

$$\frac{d}{f} = \frac{D}{Z_k} \qquad (2.2)$$

gdzie, $Z_k$ i $Z_o$ to odległości od kamery, $f$ jest ogniskową kamery, $b$ oznacza linię bazową, zaś $D$ oznacza obserwowane przesunięcie punktu względem osi X. Przy połączeniu równań (2.1) i (2.2) otrzymujemy zależność na szukaną wartość $Z_k$:

$$Z_k = \frac{Z_o}{1 + \frac{Z_o}{fb}d} \qquad (2.3)$$

Dzięki znajomości parametrów $Z_o$, $f$ oraz $D$, które można uzyskać w procesie kalibracji kamery, w oparciu o równanie (2.3) dla danej mapy rozbieżności można zbudować mapę głębokości. Na rysunku 2.3. przedstawiono przykładową mapę głębokości i odpowiadający mu obraz RGB.

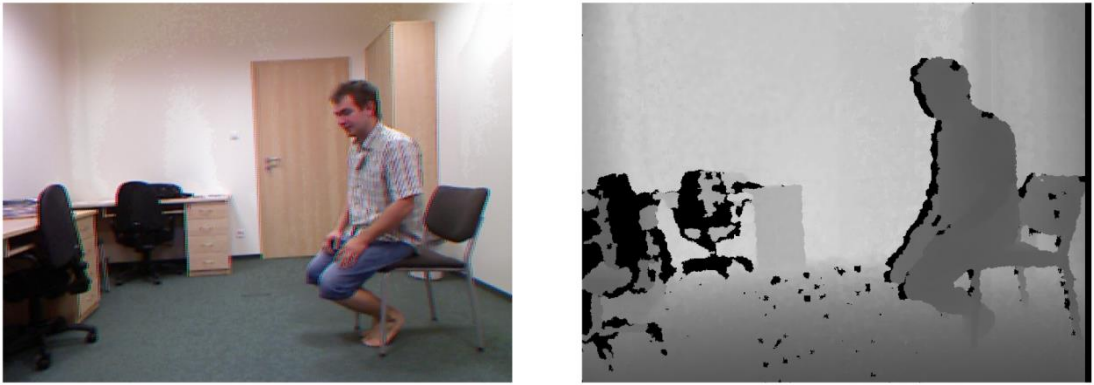

Rysunek 2.3. Obraz kolorowy i odpowiadający mu obraz głębi. Odległość na obrazie głębokości reprezentują odcienie szarości. Im większa jest jasność piksela tym odległość od kamery jest większa. Czarny kolor obrazuje piksele, dla których nie udało się określić głębokości (ang. *non-measured depth, nmd*).

Dysponując daną odległością od kamery oraz parametrami kalibracji, można łatwo wyznaczyć pozostałe współrzędne punktów w przestrzeni trójwymiarowej:

$$X_k = -\frac{Z_k}{f}(x_k - x_o + \delta x) \qquad (2.4)$$

$$Y_k = -\frac{Z_k}{f}(y_k - y_o + \delta y) \qquad (2.5)$$

gdzie $x_k$ i $y_k$ to współrzędne punktu na obrazie, $x_o$ i $y_o$ współrzędne punktu referencyjnego, a $\delta x$ i $\delta y$ oznaczają współczynniki dystorsji – wady układu optycznego polegającej na różnym powiększeniu obrazu w zależności od jego odległości od osi optycznej instrumentu. Dzięki znajomości wszystkich trzech współrzędnych zbudować można model 3D sceny reprezentowany przez chmurę punktów (ang. *point cloud*). Chmura punktów przedstawiająca geometrię skanowanych obiektów jest wygodnym sposobem prezentacji skanowanej sceny. Dodatkowym atutem jest możliwość przetwarzania danych w oparciu o dedykowane dla chmur punktów biblioteki o otwartym kodzie źródłowym, takie jak PCL (ang. *Point Cloud Library*) (Rusu & Cousins, 2011). Dzięki temu istnieje możliwość zastosowania gotowych algorytmów filtrowania, rekonstrukcji powierzchni, segmentacji i wielu innych.

W pracy (Khoshelham & Elberink, 2012) wyznaczono parametry kalibracji dla kamery podczerwieni oraz zbadano błędy pomiaru urządzenia *Kinect* dla obiektów znajdujących się w różnych odległościach od kamery. Jakość pomiaru określono poprzez porównanie z danymi referencyjnymi, za które uznano pomiary skanera laserowego FARO LS 880. Urządzenie to zostało przyjęte jako wzorzec, ponieważ jego dokładność zawiera się w zakresie 0,7-2,6 mm dla odległości 10 m, przy powierzchniach silnie odbijających światło. Aby zmniejszyć błędy wynikające z niedoskonałej korespondencji pomiędzy punktami, pobrane z obu urządzeń chmury punktów porównano ze sobą wykorzystując dwie metody: porównanie punktów z badanego urządzenia z referencyjnymi punktami przy wykorzystaniu algorytmu ICP (ang.

*Iterative Closest Point*) (Besl & McKay, 1992) oraz porównanie punktów z urządzenia z płaszczyznami wyznaczonymi za pomocą algorytmu RANSAC (Fischler & Bolles, 1981). W tabeli 2.1. przedstawiono wartości błędów pomiarowych urządzenia *Kinect* dla punktów po przekształceniu ich do współrzędnych rzeczywistych oraz wykorzystaniu wzorów (2.4) i (2.5).

Tabela 2.1. Statystyka błędów pomiarowych odległości dla urządzenia *Kinect* oraz urządzenia referencyjnego.

|  | **Kamera skalibrowana** | | | **Kamera nieskalibrowana** | | |
|---|---|---|---|---|---|---|
|  | dx | dy | dz | dx | dy | dz |
| Średnia [cm] | 0,1 | 0,0 | 0,1 | -0,5 | -0,6 | -0,1 |
| Mediana [cm] | 0,0 | 0,0 | -0,1 | -0,2 | -0,1 | -0,4 |
| Odchylenie standardowe [cm] | 1,0 | 1,1 | 1,8 | 1,4 | 1,5 | 1,8 |
| Odsetek pomiarów w przedziale [-0,5 cm; 0,5 cm] | 63,4 | 63,4 | 38,9 | 55,0 | 56,8 | 25,1 |
| Odsetek pomiarów w przedziale [-1,0 cm; 1,0 cm] | 83,4 | 80,7 | 61,6 | 74,3 | 72,7 | 51,6 |
| Odsetek pomiarów w przedziale [-2,0 cm; 2,0 cm] | 95,0 | 93,2 | 82,1 | 90,9 | 82,9 | 81,2 |

Uzyskane wyniki wskazują, że błędy zawierają się w przedziale 0-7,3 cm, przy średnim błędzie 1,8 cm i odchyleniu standardowym 1,2. Ponadto, można zaobserwować, że największy błąd występuje względem osi Z, czyli względem odległości od kamery. Wykonanie kalibracji pozwala zwiększyć dokładność pomiaru, jednak urządzenie nieskalibrowane (przeliczenie zmierzonych odległości do współrzędnych 3D zostaje wykonane przy wykorzystaniu fabrycznej kalibracji, czyli takich samych wartości dla każdego egzemplarza tego modelu urządzenia), nie charakteryzuje się znacząco większym błędem pomiaru. Z przeprowadzonych pomiarów i porównania z przyjętym w pracy (Khoshelham & Elberink, 2012) teoretycznym modelem błędu można wyciągnąć następujące wnioski:

- odpowiednio skalibrowane urządzenie nie jest źródłem dużego błędu systematycznego w porównaniu do skanera laserowego,
- błąd losowy zwiększa się z kwadratem odległości od sensora, osiągając maksymalną wartość około 4 cm,
- wraz ze wzrostem odległości maleje także gęstość punktów.

Dla potrzeb niniejszej pracy zbadano wpływ natężenia światła na jakość otrzymywanych obrazów. Pomiary przeprowadzono w trzech scenariuszach:

- kamera umieszczona wewnątrz budynku, obrazy rejestrowane w słabym świetle słonecznym, aż do jego całkowitego zaniku,
- kamera umieszczona wewnątrz budynku, obrazy rejestrowane przy silnym świetle słonecznym,
- kamera umieszczona na zewnątrz, obrazy rejestrowane w cieniu i przy silnym świetle słonecznym.

Na rysunku 2.4 przedstawiono przykładowe pary obrazów (obraz kolorowy – obraz głębi) dla sekwencji zarejestrowanych w przebadanych scenariuszach oświetlenia. Jak można zauważyć, pomiar głębi w pomieszczeniu nie jest znacząco zakłócony przez zmieniające się oświetlenie. Co więcej, można z powodzeniem dokonywać akwizycji obrazów głębi w warunkach kompletnego braku źródła światła, a silne promienie słoneczne oświetlające pomieszczenie wpływają nieznacznie na zwiększenie procentowego udziału pikseli *nmd* w całym obrazie (rysunek 2.4.b).

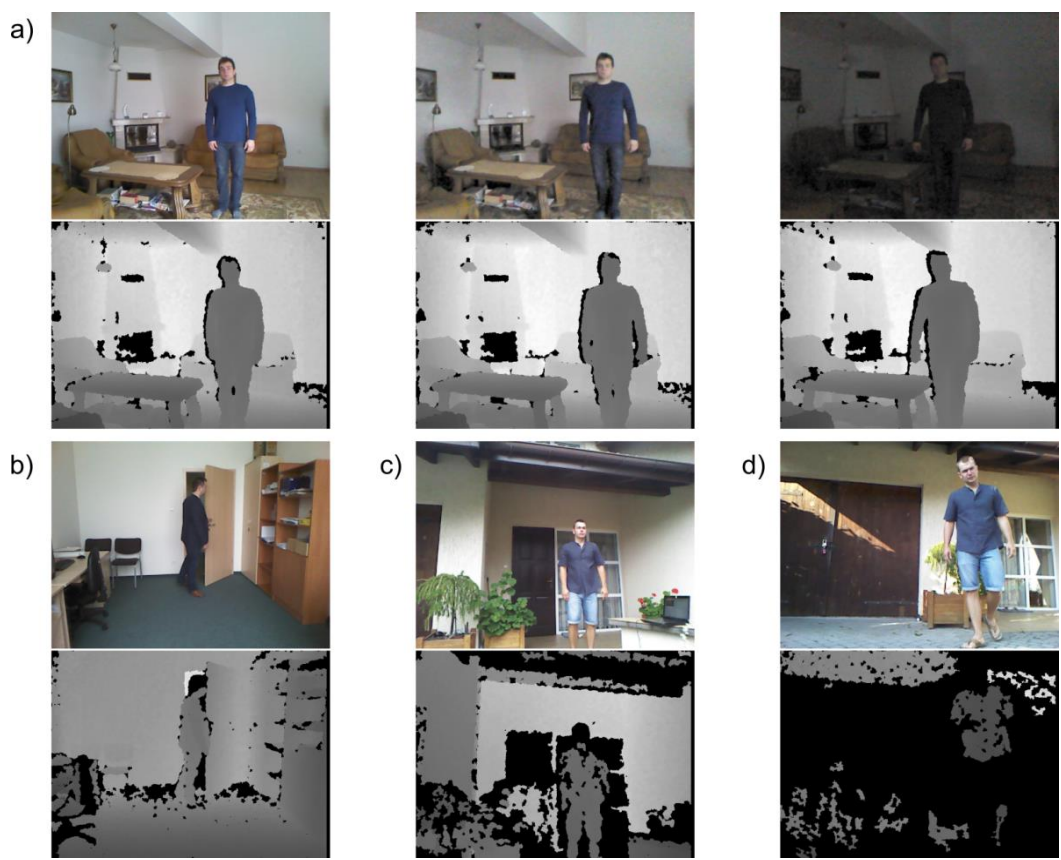

Rysunek 2.4. Obrazy kolorowe i mapy głębi dla różnych warunków oświetlenia: a) obrazy w warunkach słabnącego światła słonecznego, zarejestrowane wewnątrz pomieszczenia, b) obrazy przy silnym świetle słonecznym zarejestrowane wewnątrz pomieszczenia, obrazy zarejestrowane na zewnątrz: c) przy słabym i d) przy dużym natężeniu światła.

Umieszczenie sensora *Kinect* na zewnątrz znajduje uzasadnienie tylko w zacienionym miejscu, gdyż urządzenie to nie jest przystosowane do pracy przy silnym nasłonecznieniu, co zaobserwować można na rysunku 2.4d. Na większą liczbę pikseli o nieokreślonej głębokości wpływa także obecność na scenie przedmiotów, które dobrze odbijają światło np.: przedmioty szklane, okna, itp.

Mając na względzie dużą popularność sensora *Kinect*, opracowano kilka niezależnych pakietów sterowników oraz zestawów narzędzi programistycznych (SDK). Najpopularniejsze z nich to:

### *OpenNI*

*OpenNI* (*Open Natural Interaction*) jest organizacją non-profit, zajmującą się rozwojem, certyfikacją i kompatybilnością urządzeń NUI (ang. *natural user interface*), aplikacji i pakietów oprogramowania. Twórcami tej organizacji są m.in. *PrimeSense* (twórca technologii wykorzystywanej w *Kinect*) oraz firma *Asus*. Biblioteka *OpenNI* to zbiór API w językach C++ i C# o otwartym kodzie źródłowym, pozwalających na pobieranie obrazów z kamer RGB i kamer głębokości. Wspomniane biblioteki dostarczają warstwę abstrakcji do połączenia USB, implementują dedykowane typy danych, pozwalają na obsługę zdarzeń (ang. *events*), tworzenie logów oraz profilowanie. Dostępne są wersje dla Windows, Linux, Mac OS X. Dostępność kodu źródłowego pozwala na kompilację bibliotek także dla innych architektur, między innymi dla platformy ARM.

Bardzo przydatną z punktu widzenia programisty funkcją wprowadzoną przez *OpenNI* jest możliwość nagrywania sekwencji w postaci specjalnego formatu pliku z rozszerzeniem .oni. Dzięki temu dane mogą być poddane wielokrotnemu przetwarzaniu (*off-line*), gdyż traktowane są przez inne moduły bibliotek jak strumień danych z urządzenia.

NITE to oprogramowanie pośredniczące (ang. *middleware*) rozwijane przez *PrimeSense*. Zawiera ono zaimplementowane algorytmy detekcji użytkownika, kalibracji, śledzenia szkieletu i detekcji gestów. Kod źródłowy jest zamknięty i chroniony patentami. Warto wspomnieć, że nie istnieją wersje NITE dla innych architektur niż x86 i x64.

### *Microsoft Kinect for Windows SDK*

*Microsoft* udostępnia środowisko deweloperskie do tworzenia aplikacji wykorzystujących urządzenie *Kinect*. Posiada ono interfejsy w językach C++, C# oraz Visual Basic. Pozwala na uzyskanie dostępu do danych z kamer oraz urządzeń audio. Udostępnia algorytmy śledzenia szkieletu, umożliwia także na rozpoznawanie gestów oraz przetwarzanie danych dźwiękowych.

### *OpenKinect (libfreenect)*

*OpenKinect* jest biblioteką o otwartym kodzie źródłowym, zawierającą niezbędne elementy do komunikacji z urządzeniem *Kinect*. Omawiana biblioteka zawiera sterowniki i wieloplatformowe API działające w systemach Windows, Linux oraz Mac OS X. Posiada wsparcie dla wielu języków programowania: C/C++, C#, VB.NET, Java oraz Python. Dodatkowo rozwijana jest *OpenKinect Analysis Library* – biblioteka, która ma dostarczać funkcjonalności takich jak śledzenie dłoni, śledzenie szkieletu, generowanie chmury punktów, a nawet rekonstrukcję 3D.

Na potrzeby niniejszej pracy przeprowadzono eksperymenty celem ustalenia czy algorytmy detekcji i śledzenia całej postaci będą na tyle niezawodne, aby można było je wykorzystać w systemie detekcji upadku. Problemy z detekcją i śledzeniem występują szczególnie często, gdy osoba znajduje się w pozycji leżącej lub upada. Może to być spowodowane niedoskonałością algorytmów, które są przygotowane do działania w systemie interakcji człowiek-komputer, w szczególności gdy użytkownik znajduje się w pozycji stojącej. Ponadto, inne poruszające się obiekty na scenie także uznawane są za użytkownika. Niektóre prace (Parra-Dominguez et al., 2012; Planinc & Kampel, 2012) do detekcji upadku wykorzystują informacje o położeniu stawów (ang. *joints*) szkieletu postaci uzyskanego za pomocą *OpenNI NITE* lub *Microsot Kinect SDK*. Badania wykazały, że często niemożliwa jest niezawodna estymacja szkieletu postaci, w szczególności gdy człowiek znajduje się w pozycji leżącej. Co więcej, w takiej pozycji osoba zwykle jest niepoprawnie segmentowana i zazwyczaj podzielona na dwa lub więcej segmentów (zob. rysunek 2.5).

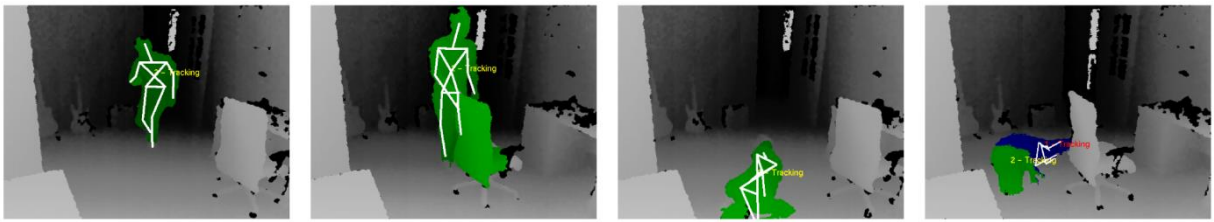

Rysunek 2.5. Przykładowe błędy w śledzeniu szkieletu w oparciu o bibliotekę NITE podczas upadku.

Uzyskane wyniki badań eksperymentalnych są zgodne z przedstawionymi w pracach (Stone & Skubic, 2014; Rojas et al., 2013). Ze 165 upadków przebadanych w pracy (Stone & Skubic, 2014), mimo braku znaczących przysłonięć, algorytm śledzenia zawiódł przynajmniej raz na jedną klatkę sekwencji podczas 46 upadków, a podczas 30 zawiódł całkowicie. W niektórych przypadkach, po zgubieniu postaci przez algorytm śledzący, po ponownej inicjalizacji osoba była rozpoznawana jako nowy użytkownik, co skutkowało brakiem ciągłości śledzenia osoby.

### 2.1.2. Inercyjny czujnik ruchu i dane akcelerometryczne

Informacje o ruchu obiektu lub zmianie pozy postaci można otrzymać nie tylko dzięki analizie obrazu z systemów wizyjnych. Stosując odpowiednie sensory uzyskać można informacje o charakterystyce ruchu i położeniu obiektu. Do rejestracji tych sygnałów służą akcelerometry – mierzące przyspieszenie, żyroskopy – mierzące położenie kątowe oraz magnetometry – mierzące pole magnetyczne. Omawiane sensory stosowane są dość powszechnie w wielu dziedzinach nauki i przemysłu: automatyce, biomedycynie, informatyce, mechanice czy lotnictwie.

Przyspieszenie jest jedną z wielkości fizycznych opisujących zmianę wektora prędkości w czasie. Zgodnie z drugą zasadą dynamiki Newtona występowanie przyspieszenia związane jest z istnieniem stałej, niezrównoważonej siły działającej na obiekt. Wielkości fizyczne, takie jak przyspieszenie, prędkość i położenie są ściśle powiązane, gdyż wektor prędkości jest pochodną wektora położenia, zaś przyspieszenie jest pochodną prędkości.

Zasadę działania akcelerometru można zilustrować posługując się jego uproszczonym modelem przedstawionym na rysunku 2.6. Model ten sprowadza się do kuli umieszczonej w sześcianie, którego każda para ścian jest prostopadła do jednej z osi układu współrzędnych.

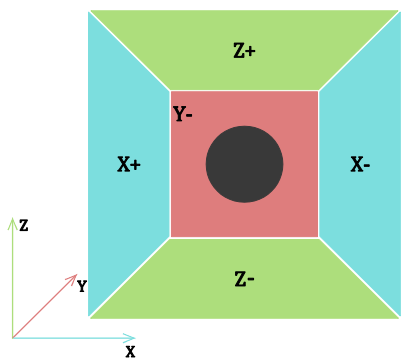

Rysunek 2.6. Uproszczony model akcelerometru z układem współrzędnych.

Załóżmy, że model akcelerometru został umieszczony w środowisku, w którym nie działają na niego siły grawitacji. W takim wypadku kula znajdowałaby się w środku sześcianu, nie powodując nacisku na żadną ze ścian. Przyłożenie do sześcianu siły $F$, a w konsekwencji przesunięcie go w stronę dodatnich wartości osi X z przyspieszeniem o wartości 9,81 m/s$^2$, spowodowałoby nacisk kuli na ścianę X- (zob. rysunek 2.7.). Dzięki temu zjawisku akcelerometr byłby w stanie zmierzyć wartość przyspieszenia jakie nadała mu siła $F$.

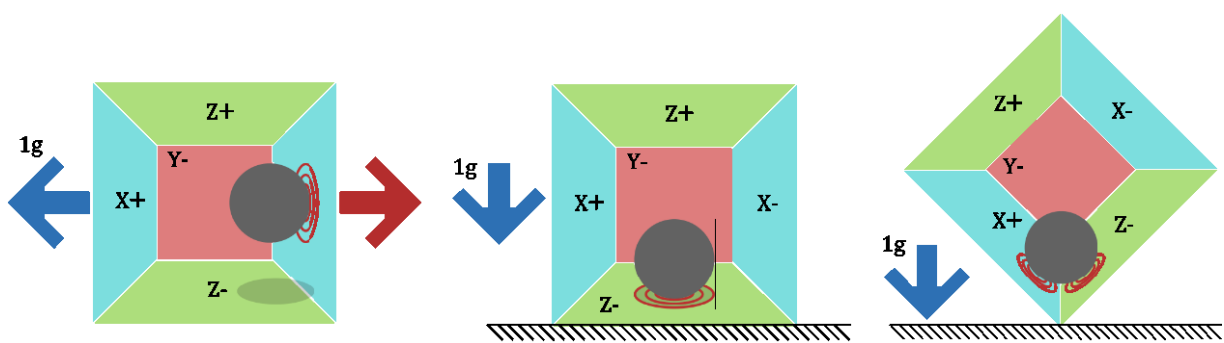

Rysunek 2.7. Rozkład sił w uproszczonym modelu akcelerometru: a) przyspieszającego względem osi X o wartości 1 [g] b) w stanie spoczynku, w polu grawitacyjnym Ziemi c) rozkład wektora przyspieszenia względem lokalnego układu współrzędnych.

Rozpatrując model akcelerometru w środowisku rzeczywistym, tj. w polu grawitacyjnym Ziemi warto zwrócić uwagę, że kula cały czas będzie źródłem nacisku na ścianę Z- akcelerometru, co powoduje wskazanie przyspieszenia o wartości przyspieszenia ziemskiego (-1g). Ważną własnością akcelerometru jest fakt, że mierzy on wartość przyspieszenia we własnym układzie współrzędnych, który jest niezależny od układu globalnego (zob. rysunek 2.7.c.). Każdy wektor przyspieszenia $R$, działającego na urządzenie, posiada składowe $Rx$, $Ry$, $Rz$, które są rzutami wektora na osie układu X, Y, Z. Wartości wektorów składowych są danymi uzyskiwanymi z akcelerometru. W praktyce, różne zjawiska fizyczne wykorzystywane są do mierzenia przyspieszenia, w zależności od typu akcelerometru:

pojemnościowe, piezoelektryczne, piezorezystywne, magnetorezystywne, czy działające na zasadzie efektu Halla.

Poza zastosowaniami w przemyśle, inercyjne sensory ruchu znalazły szerokie zastosowanie w medycynie. Są one wykorzystywane do rozpoznawania zachowań (Fahrenberg et al., 1997), w tym do detekcji upadków, diagnozowania i monitorowania chorób (np. układu ruchowego, lub choroby Parkinsona (Keijsers et al., 2003)), czy detekcji ataków epilepsji (Nijsen et al., 2005). Umieszczając akcelerometry na ciele człowieka, należy uwzględnić, że sygnał wyjściowy będzie wynikiem sumy kilku sygnałów składowych (Nijsen et al., 2010):

- szumów pochodzących z otoczenia człowieka:
    - przyspieszeń o źródle zewnętrznym, spowodowanych np. ruchem pojazdów,
    - przyspieszeń związanych z uderzaniem ciała lub sensora o inne obiekty,
- szumów o źródle w ciele człowieka:
    - skurczów mięśni,
    - rytmu serca,
    - oddechu,
- przyspieszenia ziemskiego,
- przyspieszenia związanego z ruchami człowieka,
- szumów sensora i systemu pomiarowego.

Gdy osoba nosząca akcelerometr pozostaje w bezruchu, wydzielenie składowych przyspieszenia, pochodzących z fizjologicznych zjawisk, takich jak oddech czy skurcz mięśni nie stanowi trudności. Wartości przyspieszenia ziemskiego zawierają się w granicach -1g do 1g w zależności od położenia sensora. Duży wpływ na wartości mierzonego przyspieszenia ma umiejscowienie sensora, co powoduje nie tylko wzrost niektórych składowych (np. umiejscowienie na klatce piersiowej spowoduje widoczny wzrost składowej sygnału pochodzącego od oddechu i rytmu serca), ale także inną charakterystykę sygnału związanego z ruchem ciała człowieka. Lokalizacja sensora ściśle powiązana jest z zastosowaniem. I tak na przykład w systemach detekcji upadku, akcelerometry zazwyczaj umiejscowione są przy biodrze (Bourke et al., 2010) lub mostku (Bourke et al., 2007), zaś w systemach detekcji epilepsji bardziej efektywnym miejscem są ręce i nogi (Nijsen et al., 2005). Przedstawione w pracy (Bourke et al., 2007) wyniki wskazują, że umiejscowienie akcelerometru na klatce piersiowej pozwala uzyskać największą dokładność detekcji upadku. W niniejszej pracy przyjęto za lokalizację akcelerometru okolice nieco powyżej biodra, z uwagi na to, że tam znajduje się środek masy człowieka, a także założono, że takie umiejscowienie czujnika będzie wygodniejsze i mniej inwazyjne (czujnik może być przypięty przykładowo do paska spodni użytkownika).

W niniejszej pracy wykorzystano inercyjny czujnik ruchu x-IMU firmy x-io Technologies (rysunek 2.8). Jest to niewielkich rozmiarów urządzenie, posiadające 3-osiowy akcelerometr, magnetometr oraz żyroskop. Sensor pozwala na akwizycję danych z częstotliwością 512 Hz, oraz przesyłanie ich przez *Bluetooth*, USB lub zapis na karcie microSD. Urządzenie pozwala na długotrwały zapis danych z wysoką dokładnością i częstotliwością, co z kolei stwarza szerokie możliwości zastosowań w systemie detekcji upadku.

### 2.1.3. Mobilna platforma obliczeniowa

Jako sprzętową platformę developerską wybrano płytę *PandaBoard ES*, która zdobyła popularność wśród twórców rozwiązań przeznaczonych na urządzenia mobilne. Płyta ta możliwościami obliczeniowymi odpowiada najwydajniejszym telefonom komórkowym (ang. *smarpthones*), jednocześnie charakteryzuje się niskimi kosztami produkcji oraz niewielkimi rozmiarami i zużyciem energii. Płyta wyposażona jest w platformę OMAP czwartej generacji, której sercem jest dwurdzeniowy procesor ARM Cortex-A9, pracujący z częstotliwością 1,2 GHz. Inne ważne parametry to:

- procesor graficzny POWERVR™ SGX540 wspierający technologie: OpenGL® ES v2.0, OpenGL ES v1.1, OpenVG v1.1,
- 1 GB DDR2 RAM,
- czytnik kart SD/MMC,
- łączność 10/100 Ethernet i 802.11 b/g/n oraz *Bluetooth*® v2.1,
- 3x USB 2.0 High-Speed pracujące również w trybie host,
- złącze różnego przeznaczenia (ang. *General Purpose Input/Output – GPIO*) pozwalające na obsługę protokołów: I2C, GPMC, USB, MMC, DSS, ETM,
- złącza: JTAG, RS-232,
- wymiary: 114,3 x 101,6 mm, waga: 81,5 gramów.

Duże zainteresowanie technologiami mobilnymi skutkowało powstaniem wielu projektów opartych o *PandaBoard*. Dzięki popularności platforma ta została wybrana przez fundację *Linaro* jako jedna z kilku platform sprzętowych, dla których co miesiąc przygotowywane są dedykowane obrazy systemów operacyjnych. *PandaBoard* może działać pod kontrolą systemów operacyjnych takich jak Linux, Android, Mozilla Firefox OS czy OpenBSD.

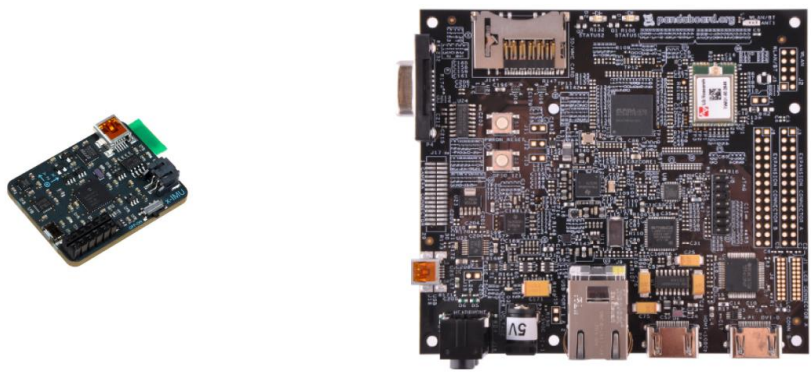

Rysunek 2.8. Urządzenia wykorzystane w pracy: po lewej sensor x-IMU, po prawej platforma mobilna *PandaBoard ES*.

Aby porównać możliwości obliczeniowe platformy *PandaBoard* ze współczesnym komputerem PC, przeprowadzono testy wydajnościowe przy wykorzystaniu dwóch testów wydajnościowych (ang. *benchmark*): *Dhrystone 2* oraz *Double-Precision Whetstone*. Uzyskane wyniki zostały zestawione w tabeli 2.2.

Tabela 2.2. Porównanie wyników testów wydajnościowych platformy *PandaBoard* i komputera klasy PC

|  | *PandaBoard ES* | *Intel* **i7-3610QM 2.30 GHz** |
|---|---|---|
| *Dhrystone 2*: | | |
| **wynik [lps]** | 4 214 871,3 | 37 423 845,1 |
| **indeks** | 361,2 | 3 206,8 |
| *Double-Precision Whetstone*: | | |
| **wynik [MWIPS]** | 836,3 | 4 373,7 |
| **indeks** | 152,1 | 795,2 |

### 2.1.4. Głowica aktywna

*Kinect* zawieszony na suficie na wysokości 2,6 metra i skierowany w dół, posiadający kąty widzenia 43 i 57 stopni, pozwala na obserwację obszaru o wielkości około 5,5 m$^2$. Aby zwiększyć powierzchnię monitorowania zaproponowano wykorzystanie aktywnej głowicy pozwalającej na rotację kamery w dwóch stopniach swobody (ang. *pan-tilt head*) (Kępski & Kwolek, 2014a). Podczas działania systemu kontroler, na podstawie danych o położeniu osoby, obraca kamerę tak, aby stale utrzymywać postać w środku obrazu. Pozwala to na monitorowanie obszaru o wielkości średniego pokoju o powierzchni 15 - 20 m$^2$.

Własnoręcznie wykonana głowica została przedstawiona na rysunku 2.9. Głowica została wykonana z aluminiowych kształtowników. Możliwe jest obracanie kamerą za pomocą dwóch serwomechanizmów sterowanych mikrokontrolerem Arduino Uno[6]. Mikrokontroler ten oparty jest o układ ATmega328, działający z częstotliwością 16 MHz i posiadający 2 KB pamięci RAM. Płytka mikrokontrolera wyposażona jest w 6 analogowych wejść oraz 14 złączy cyfrowych wejścia/wyjścia, z których sześć pozwala na sterowanie przy użyciu modulacji szerokością impulsu (ang. *pulse width modulation* – PWM). Wykorzystany mikrokontroler posiada kilka protokołów komunikacyjnych takich jak port szeregowy UART TTL, I2C oraz SPI. Do sterowania orientacją głowicy wykorzystano dwa regulatory PID, których zadaniem było utrzymanie obiektu zainteresowania w środku obrazu.

---

[6] http://arduino.cc/en/main/arduinoBoardUno

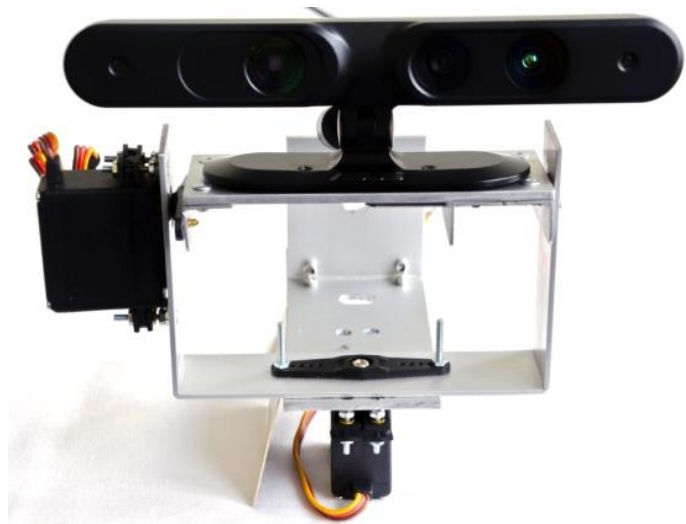

Rysunek 2.9. Głowica pan-tilt i umieszczony na niej sensor głębi.

## 2.2. Algorytmy przetwarzania obrazów i percepcji sceny

Skuteczne rozpoznawanie akcji w systemach wizyjnych wymaga zwykle wstępnego przetwarzania obrazów. Jakość rozpoznawania zależy nie tylko od poprawnego doboru cech oraz parametrów w oparciu o które działa klasyfikator, ale też wcześniejszego przygotowania danych (ang. *preprocessing*). Przykładowo, na pogorszenie wyników rozpoznawania akcji może wpływać niedokładna segmentacja postaci, czy też błędy w jej śledzeniu.

W opisywanym systemie detekcji upadku, mając na względzie etapy działania systemu, metody te można podzielić na dwie grupy:

- metody działające przy inicjalizacji systemu – wykonywane jednorazowo podczas uruchomienia systemu lub przy zmianie pozycji kamery: detekcja podłogi, budowa modelu tła,
- metody działające w czasie rzeczywistym – wykonywane ciągle podczas działania systemu: odjęcie tła, detekcja postaci, śledzenie, aktualizacja modelu tła.

### 2.2.1. *V-disparity* i transformata Hougha

Detekcja podłogi jest ważnym etapem wstępnym detekcji upadku, a jej wykorzystanie w systemie ma na celu:

- Redukcję rozmiaru danych w celu zmniejszenia czasu przetwarzania, gdyż dzięki znajomości równania płaszczyzny, po pobraniu obrazu z urządzenia, można usunąć te elementy chmury punktów, które należą do podłogi.
- Określenie cech opisujących postać, potrzebnych do detekcji upadku. Ponieważ przyjęte założenia projektowe systemu zakładają możliwość dowolnego umieszczenia kamery, występuje potrzeba odnalezienia pewnego referencyjnego elementu sceny, względem którego będzie badany ruch osoby.

Detekcja podłogi przebiega zgodnie ze schematem przedstawionym na rysunku 2.10.

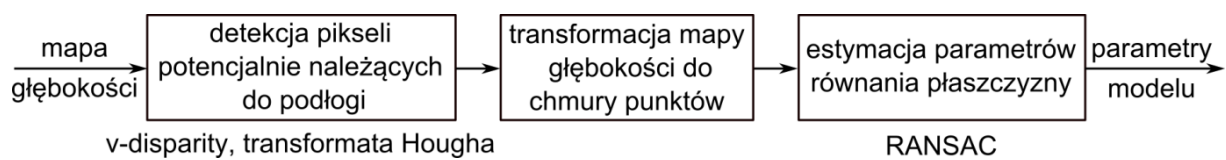

Rysunek 2.10. Schemat algorytmu do detekcji podłogi.

Mapy v-dysparycji (ang. *v-disparity*) zostały oryginalnie zaproponowane w (Labayrade et al., 2002) jako metoda detekcji przeszkód na podstawie obrazu dysparycji z kamery stereowizyjnej. Niech $H$ będzie funkcją dysparycji $d$, określoną jako $H(d) = I_d$, gdzie $I_d$ jest obrazem v-dysparycji. Funkcja $H$ akumuluje piksele $d$ o tej samej wartości różnicy zobrazowania z danej linii obrazu $I_d$. Innymi słowy, na obrazie *v-disparity* wartość $k$-tego punktu $i$-tej linii reprezentuje liczbę punktów na obrazie dysparycji o wartości $k$, znajdujących się w linii $i$. Rozważmy kamerę stereowizyjną, która umieszczona jest na wysokości $h$ nad płaszczyzną $G$ i nachylona do niej pod kątem $\theta$ (rysunek 2.11.). W takim ustawieniu można wyróżnić dwa układy współrzędnych: $R_w$ (globalny układ współrzędnych) oraz $R_c$ (układ współrzędnych kamery).

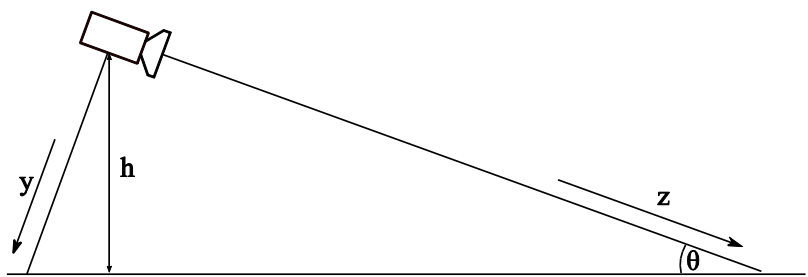

Rysunek 2.11. Ilustracja umieszczenia kamery na scenie, nad płaszczyzną podłogi.

Pozycja punktów na obrazie będzie oznaczona jako *(u, v)*, zaś projekcja optycznego środka jako *($u_0$, $v_0$)*, który jest jednocześnie środkiem obrazu. Korzystając z modelu kamery otworkowej, projekcja punktu o współrzędnych globalnych $X,Y,Z$ na płaszczyznę obrazu realizowana jest zgodnie z poniższym równaniem:

$$\begin{cases} u = \alpha_u \cdot \dfrac{X}{Z} + u_0 \\ v = \alpha_v \cdot \dfrac{Y}{Z} + v_0 \end{cases} \qquad (2.6)$$

gdzie $\alpha_u = f/t_u$, $\alpha_v = f/t_v$, $f$ – ogniskowa, $t_u, t_v$ – rozmiary piksela względem $u$ i $v$. Obiekty znajdujące się w przestrzeni 3D mogą być poddawane rozmaitym przekształceniom, np. przesunięciu, skalowaniu, obrotowi. Przekształcenia te nazywane są transformacjami geometrycznymi. Macierze jednorodne pozwalają na przedstawienie dowolnego przekształcenia geometrycznego w przestrzeni trójwymiarowej. Transformacje składające się z wielu przekształceń geometrycznych można reprezentować za pomocą operacji mnożenia macierzy tych przekształceń. Celem opisania położenia kamery względem środka układu współrzędnych można posłużyć się macierzami reprezentującymi translację (przesunięcie o wektor) i rotację (obrót obiektu). Dla zilustrowanego na rys. 2.11. przypadku można wyznaczyć macierze translacji i rotacji (odpowiednio $T$ i $R$) w oparciu o następujące zależności:

$$T = \begin{bmatrix} 1 & 0 & 0 & \frac{-b}{2} \\ 0 & 1 & 0 & h \\ 0 & 0 & 1 & 0 \\ 0 & 0 & 0 & 1 \end{bmatrix}, \quad R = \begin{bmatrix} 1 & 0 & 0 & 0 \\ 0 & \cos\theta & -\sin\theta & 0 \\ 0 & \sin\theta & \cos\theta & 0 \\ 0 & 0 & 0 & 1 \end{bmatrix} \tag{2.7}$$

Aby odpowiednio przedstawić współrzędne punktu na obrazie, oprócz transformacji, należy dokonać projekcji perspektywicznej, którą można przedstawić za pomocą macierzy $M_{proj}$:

$$M_{proj} = \begin{bmatrix} \alpha_u & 0 & u_0 & 0 \\ 0 & \alpha_v & v_0 & 0 \\ 0 & 0 & 1 & 0 \end{bmatrix} \tag{2.8}$$

Dzięki znajomości macierzy translacji, rotacji i projekcji, możemy dokonać przekształcenia z układu współrzędnych $R_w$ do współrzędnych obrazu:

$$D = RT \tag{2.9}$$

$$T_r = M_{proj}D \tag{2.10}$$

Punkt $P$ opisany współrzędnymi jednorodnymi $(X,Y,Z,1)$ można poddać przekształceniu z globalnego układu współrzędnych do współrzędnych jednorodnych obrazu:

$$p = T_r P = (x, y, z)^T \tag{2.11}$$

Na podstawie tego można uzyskać współrzędne $(u, v)$, dzieląc dwie pierwsze współrzędne przez współrzędną $z$. Rzędną $v$ projekcji punktu na obrazie można wyznaczyć w oparciu o następującą zależność:

$$v = \frac{(v_0 \sin\theta + \alpha\cos\theta)(Y + h) + Z(v_0\cos\theta - \alpha\sin\theta)}{(Y + h)\sin\theta + Z\cos\theta} \tag{2.12}$$

Dysparycję punktu $P$ można określić w oparciu o równanie:

$$\Delta = \frac{\alpha b}{(Y + h)\sin\theta + Z\cos\theta} \tag{2.13}$$

Na podstawie równań 2.13 i 2.14, płaszczyzna o równaniu $Z=aY+d$ może być przedstawiona jako linia prosta na obrazie v-dysparycji:

$$\Delta_M = \frac{b}{ah - d}(v - v_0)(a\cos\theta + \sin\theta) + \frac{b}{ah - d}\alpha(a\sin\theta - \cos\theta) \tag{2.14}$$

Okazuje się, że metodę v-dysparycji można zastosować także dla obrazów z sensora *Kinect* (Kępski & Kwolek, 2013). Biorąc pod uwagę zasadę działania tego urządzenia, na podstawie przekształcenia wzoru 2.2 oraz znajomość odległości od kamery, obliczyć można różnicę zobrazowania:

$$d = \frac{b \cdot f}{z} \tag{2.15}$$

gdzie $z$ – odległość od kamery (w metrach), $b$ – linia bazowa (w metrach), $f$ – ogniskowa (w pikselach). Dla sensora *Kinect* długość linii bazowej wynosi 7,5 cm, zaś ogniskowa równa jest 580 pikseli. Rysunek 2.12. przedstawia obraz v-dysparycji dla przykładowej mapy głębokości oraz korespondujący z nimi obraz RGB. Zakładając, że kamera umieszczona jest na wysokości powyżej 1 metra nad podłogą, początek odpowiadającego jej na obrazie *v-disparity* odcinka zawiera się w przedziale (20, 27) wartości dysparycji. Wyznaczenie położenia tego odcinka na obrazie, pozwoli dla każdej linii obrazu wejściowego (dysparycji, a w konsekwencji głębokości) znaleźć te piksele, które należą do podłogi.

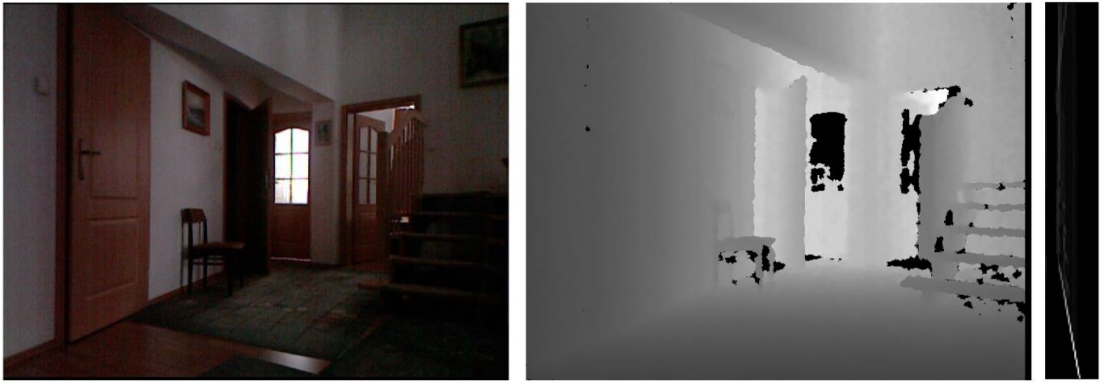

Rysunek 2.12. Metoda *v-disparity* zastosowana dla danych z sensora *Kinect*, od lewej: obraz RGB sceny, mapa głębi, obraz v-dysparycji.

Detekcja linii na obrazie może być zrealizowana w oparciu o transformatę Hougha (Szeliski, 2010), która została opracowana do wykrywania regularnych kształtów na obrazach. Pozwala ona na wykrywanie prostych kształtów takich jak linie, okręgi czy elipsy z dużą odpornością na zakłócenia obrazu. Najpowszechniej używanym jej wariantem jest tzw. uogólniona transformata Hougha (ang. *generalized Hough transform*) (Duda & Hart, 1972). W kartezjańskim układzie współrzędnych prostą można opisać następującym równaniem:

$$y = ax + b \tag{2.16}$$

gdzie $a$ to współczynnik kierunkowy, a $b$ tzw. wyraz wolny (rzędna punktu, w którym prosta przecina oś OY). Prostą można też opisać w biegunowym układzie współrzędnych, wyznaczonym przez pewien punkt $O$ (biegun) oraz półprostą $OS$ o początku w punkcie $O$ (oś biegunowa). W omawianym układzie współrzędnych, każdemu punktowi $P$ można przypisać dwie współrzędne:

- promień wodzący punktu $P$, czyli odległość $|OP|$ od bieguna oznaczony jako $\rho$,
- amplitudę punktu $P$, będącą wartością kąta skierowanego pomiędzy półprostą $OS$ a wektorem $\overrightarrow{OP}$ oznaczonym jako $\theta$.

Równanie prostej w biegunowym układzie współrzędnych ma postać:

$$\rho = x \cos\theta + y \sin\theta \tag{2.17}$$

W przestrzeni parametrów $(\rho, \theta)$ każdy punkt obrazu wejściowego jest reprezentowany jako sinusoida (przez dowolny punkt można przeprowadzić pęk prostych). Sinusoidy w przestrzeni parametrów, reprezentujące punkty leżące na jednej prostej, przecinają się w jednym punkcie (zob. rysunek 2.13). Proces wyznaczania wartości transformaty sprowadza się do głosowania (ang. *voting*) w przestrzeni parametrów, zwanej akumulatorem. Pary $(\rho, \theta)$, przyjmujące największe wartości akumulatora reprezentują proste znalezione na obrazie wejściowym. Algorytm transformaty Hougha można przedstawić następująco:

Algorytm 1 (Transformata Hougha):

1. Dokonaj kwantyzacji przestrzeni akumulatora, $\theta \in [0, \pi)$, $\rho \in [-R, R]$
2. Dla każdego piksela obrazu wejściowego $(x_i, y_i)$:
3. $\quad \rho = x_i \cos\theta + y_i \sin\theta$ , $\forall \theta \in [0, \pi)$
4. $\quad$ Zwiększ wartość akumulatora: $A(\rho, \theta) = A(\rho, \theta) + 1$
5. Dla przyjętego progu $A_{thold}$, każda wartość akumulatora spełniająca warunek $A(\rho, \theta) > A_{thold}$ reprezentuje linię na obrazie wejściowym o parametrach $(\rho, \theta)$

W praktyce obrazem wejściowym jest zazwyczaj obraz wydzielonych krawędzi uzyskanych przykładowo operatorem Sobela. Aby zmniejszyć wymagania obliczeniowe algorytmu, można zastąpić operację generowania pęku prostych dla każdego punktu obrazu wejściowego poprzez przeprowadzenie przez niego prostych o orientacji gradientu w tym punkcie (Vernon, 1991).

    Biorąc pod uwagę charakterystykę obrazu v-dysparycji, przy detekcji linii metodą Hougha można dodać do oryginalnego algorytmu pewne modyfikacje, które wpłyną na poprawę skuteczności detekcji oraz na zmniejszenie nakładów obliczeniowych. Przede wszystkim, znając specyfikę obrazu *v-disparity* można znacznie ograniczyć przestrzeń akumulatora zarówno w zakresie kąta $\theta$ jak i promienia wodzącego $\rho$. W trakcie badań eksperymentalnych, dla typowych położeń kamery, wyznaczono przedział kąta $\theta$ opisującego linię reprezentującą podłogę na obrazie v-dysparycji (Kępski & Kwolek, 2013). Ograniczenie takie pozwala na poszukiwanie kąta o wartościach $[0, 30)$ stopni, a nie całego przedziału $[0, 180)$, co wpływa na zmniejszenie czasu obliczeń. Ponadto biorąc pod uwagę fakt, że obraz v-dysparycji jest swego rodzaju histogramem, można przeprowadzić detekcję linii bezpośrednio na nim (a nie na obrazie krawędzi) i dzięki temu zwiększać odpowiednie elementy akumulatora o wartość piksela obrazu $I_d$. Powoduje to wzmocnienie w przestrzeni akumulatora tych linii z obrazu $I_d$, które odpowiadają bardziej licznym obszarom o wspólnych wartościach na obrazie głębi (zob. rysunek 2.14.).

    Dzięki znajomości parametrów wykrytej prostej, można dla każdej linii obrazu wejściowego wyznaczyć wartość dysparycji, którą powinny posiadać piksele należące do płaszczyzny. Na podstawie tego kryterium można dokonać detekcji obszarów sceny, należących do podłogi.

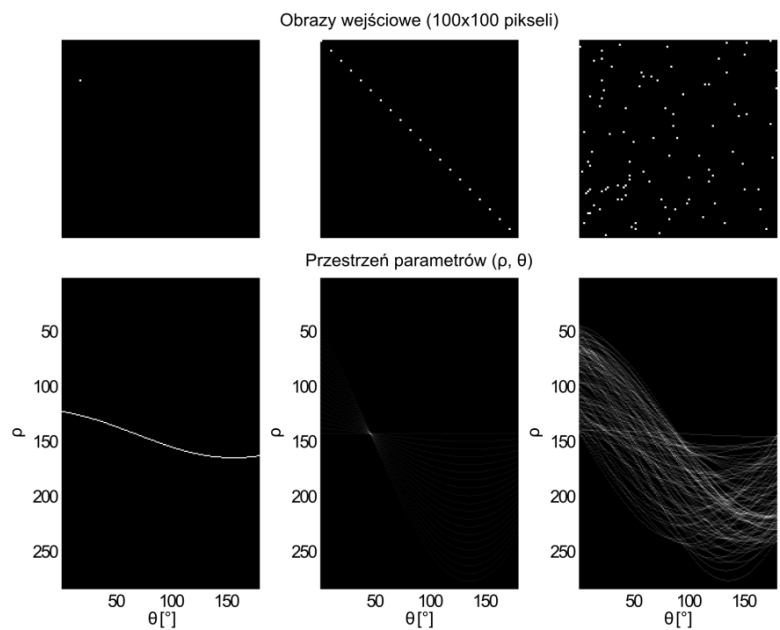

Rysunek 2.13 Wejściowe obrazy i reprezentacja przestrzeni Hougha dla: pojedynczego punktu, punktów współliniowych, losowego szumu.

Ze względu na pewną niedokładność pomiaru urządzenia, przyjęto, że piksele różniące się od wyznaczonej wartości dysparycji nie więcej niż o pewną wartość $d_t$, również należą do płaszczyzny. Jeśli $d_y$ jest wyznaczoną wartością dysparycji, która reprezentuje piksele podłogi w linii $y$ obrazu, to za piksele również do niej należące uważa się te, które posiadają wartość dysparycji $d \in (d_y - d_t, d_y + d_t)$.

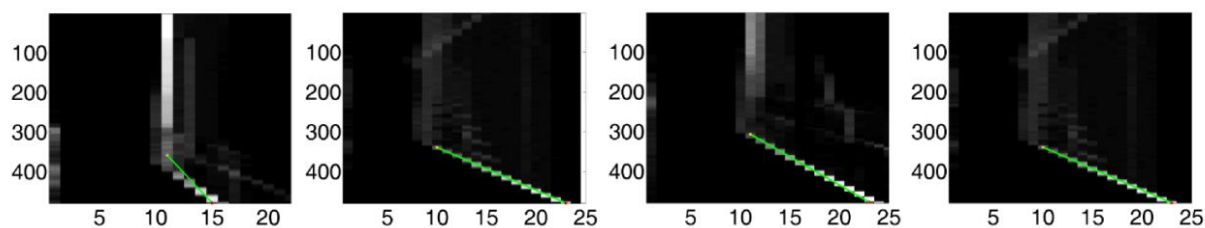

Rysunek 2.14. Detekcja linii na obrazie *v-disparity* w oparciu o zmodyfikowaną transformatę Hougha.

Ewaluację metody *v-disparity* przeprowadzono na obrazach z publicznie dostępnej bazy danych NYU Depth Dataset V2[7], która zawiera obrazy RGB-D wykorzystywane do oceny skuteczności algorytmów segmentacji sceny. Na rysunku 2.15 przedstawiono przykładowe rezultaty segmentacji.

---

[7] http://cs.nyu.edu/~silberman/datasets/nyu_depth_v2.html

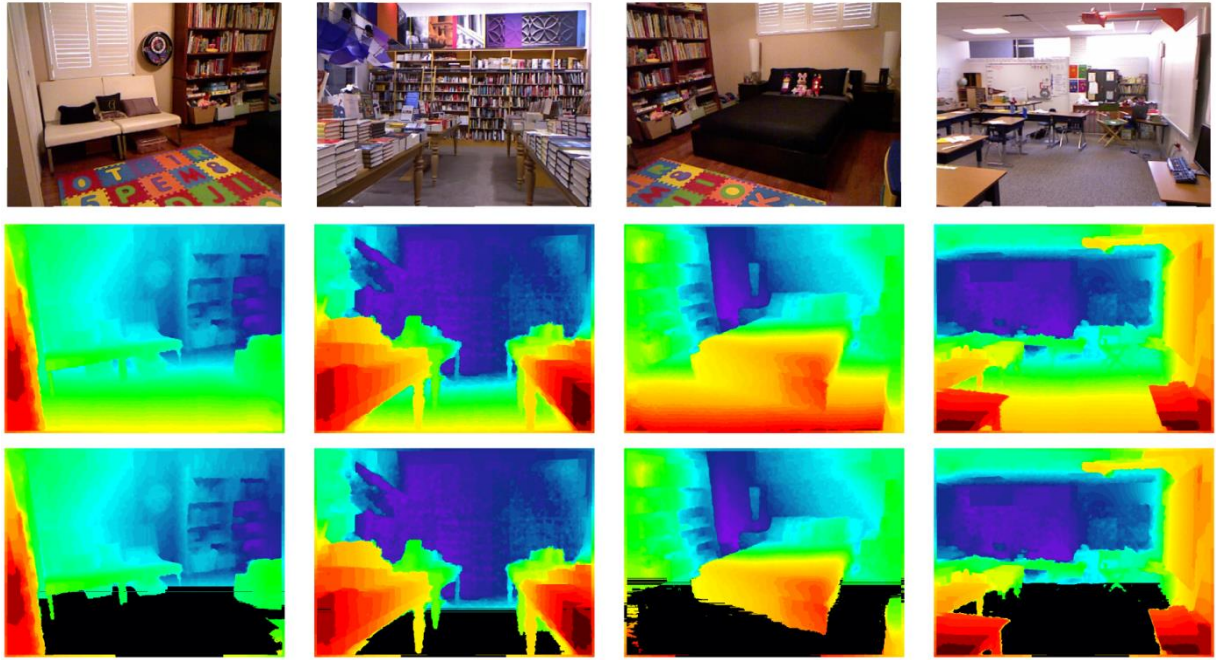

Rysunek 2.15. Obrazy ilustrujące skuteczność wydzielania podłogi metodą v-dysparycji na obrazach z bazy danych NYU Depth Dataset V2. Od góry do dołu: obraz RGB, obraz głębi, obraz głębi z zaznaczonymi pikselami należącymi do podłogi. Ze względu na mały rozmiar obrazów, głębie przedstawiono w skali kolorów (czerwony – obiekty blisko kamery, niebieski – obiekty daleko od kamery).

### 2.2.2. RANSAC

Detekcja punktów sceny, które należą do podłogi, może okazać się niewystarczająca w kontekście percepcji sceny na potrzeby wydzielenia cech charakteryzujących osobę i jej ruch. Wygodniej jest traktować podłogę jako płaszczyznę i przedstawiać ją przy pomocy równania:

$$ax + by + cz + d = 0 \qquad (2.18)$$

Celem określenia wartości współczynników $a, b, c, d$ należy zastosować metodę estymacji parametrów modelu matematycznego na podstawie zbioru obserwacji. Algorytm RANSAC (Fischler & Bolles, 1981) (ang. *Random Sample Consesus*) jest ogólną metodą estymacji parametrów modelu na podstawie danych, zaprojektowany tak, aby radzić sobie z dużą liczbą wartości odstających od poszukiwanego modelu (ang. *outliers*). Algorytm ten stosuje technikę próbkowania, która pozwala na wygenerowanie potencjalnych rozwiązań (ang. *candidate solutions*) wykorzystując jak najmniejszą liczbę obserwacji. Ze zbioru $D = \{d_1, ..., d_N\}$, do którego należą wszystkie dane wybierany jest losowo pewien podzbiór $MSS$, nazywany minimalnym zbiorem próbek (ang. *minimal sample set*). Następnie sprawdzana jest jaka część całego zbioru danych, oznaczona jako $CS$ (ang. *consensus set*), jest zgodna z modelem $\mathcal{M}$, którego wektor parametrów $\Theta$ został oszacowany na podstawie podzbioru $MSS$. Pseudokod algorytmu można zapisać następująco:

Algorytm 2 (RANSAC):

1. **do**:
2.       wybierz losowo podzbiór $MSS \epsilon D$, $|MSS| < |D|$
3.       na podstawie $MSS$ estymuj wektor parametrów $\Theta$
4.       dla każdej próbki $d \epsilon D$ wykonaj:
5.            wyznacz błąd $e_\mathcal{M}$, zdefiniowany jako odległość $d$ do modelu $\mathcal{M}(\Theta)$
6.            **if** $e_\mathcal{M}(d, \Theta) \leq \delta$ // $\delta$ jest predefiniowanym progiem
7.                 $d \epsilon CS$
8.       oblicz $|CS|$
9. **while** $\frac{|CS|}{|D|} < \tau$, gdzie $\tau$ jest predefiniowanym progiem

Oprócz warunku stopu związanego z liczbą punktów należących do *consensus set*, można ograniczyć liczbę iteracji algorytmu, zakładając, że po pewnej liczbie iteracji $h$ uzyskamy wystarczająco dobre rozwiązanie. Niech $q$ będzie prawdopodobieństwem wyboru takiego $MSS$, że w wyniku estymacji parametrów $\Theta$ modelu $\mathcal{M}$ uzyskamy idealne dopasowanie parametrów. Zatem prawdopodobieństwo wyboru $MSS$ zawierającego co najmniej jeden *outlier* wynosi $1 - q$. Prawdopodobieństwo $h$-krotnego wyboru takich $MSS$, że wszystkie z nich będą zawierały *outliers* wynosi $(1 - q)^h$. Wyboru takiego $h$, że prawdopodobieństwo to jest mniejsze niż pewien próg $\varepsilon$ można dokonać na podstawie zależności:

$$h \geq \left[ \frac{\log \varepsilon}{\log(1 - q)} \right] \tag{2.19}$$

Warto podkreślić, że przystępując do estymacji nieznanych parametrów z reguły nie mamy wiedzy *a priori* o wartości prawdopodobieństwa $q$, gdyż zależy ono od procentowego udziału *outliers* w zbiorze $D$. Można go oszacować na podstawie największej liczby *outliers* wybranych przy losowaniu $MSS$, jednak takie oszacowanie może być zbyt optymistyczne (Tordoff & Murray, 2005). Problem ten poddany jest szerzej analizie w pracach (Chum & Matas, 2005).

   Na rysunku 2.16. przedstawiono chmury punktów przykładowego procesu wydzielania podłogi (korespondują one z mapą głębi z rysunku 2.12). Kolorem czerwonym zaznaczono te elementy chmury punktów (b), które są wynikiem segmentacji podłogi metodą v-dysparycji i transformaty Hougha. Porównując ją z chmurą punktów (a) można zauważyć, że segmentacja podłogi nie przebiegła w sposób zadowalający. Jak można zauważyć na wspomnianym rysunku, dane nie są pozbawione błędnie zakwalifikowanych do płaszczyzny punktów, zaś część właściwej płaszczyzny została pominięta. Nie mniej jednak, liczba *outliers* nie jest znacząca, a więc określone tym sposobem dane są dobrą hipotezą początkową dla algorytmu RANSAC. Chmura punktów (c), która uzyskana została w oparciu o estymację parametrów równania płaszczyzny i usunięcie punktów do niej należących jest pozbawiona błędów segmentacji podłogi.

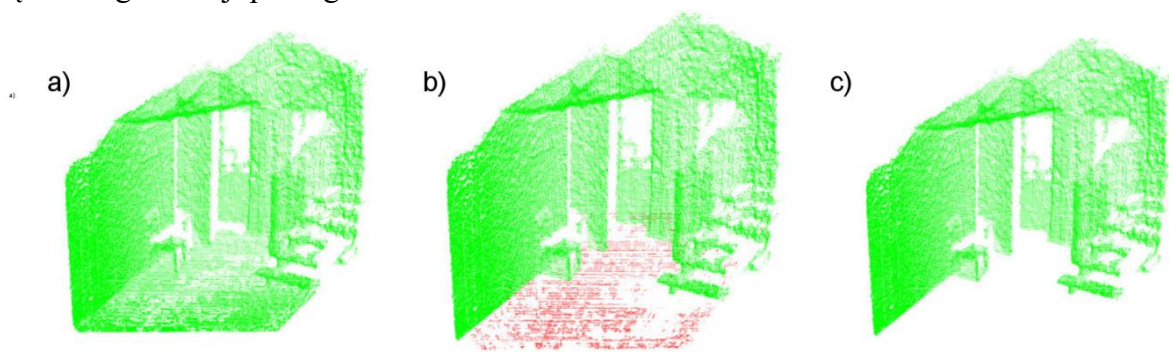

Rysunek 2.16. Wynik detekcji płaszczyzny podłogi przedstawiony za pomocą chmur punktów: a) wejściowa chmura punktów, b) wynik metody *v-disparity* i transformaty Hougha, c) wynikowa chmura punktów pozbawiona punktów należących do podłogi.

Do estymacji parametrów równania płaszczyzny można wykorzystać także inne metody (Tanahashi et al., 2001), np.: metodę najmniejszych kwadratów, metodę wartości i wektorów własnych (ang. *eigenvalues*) czy *maximum-likelihood estimation*. Metodę RANSAC wybrano ze względu na możliwość uzyskania dobrych wyników przy obecności znacznej liczby *outliers*. Wyznaczenie metodą v-disparity hipotezy początkowej dla algorytmu RANSAC pozwala na ograniczenie wpływu zaszumienia danych na dokładność procesu estymacji. Nie mniej jednak dla bardziej skomplikowanej sceny, przykładowo, jeśli znajdują się na niej przedmioty leżące na podłodze, wybór algorytmu RANSAC, jako bardziej złożonego (niż np. metoda najmniejszych kwadratów) może się okazać w pełni uzasadniony.

### 2.2.3. Budowa i aktualizacja modelu tła

Algorytmy detekcji tła i pierwszego planu (ang. *background/foreground detection*) są powszechnie wykorzystywane w systemach multimedialnych i znalazły szerokie zastosowanie m.in. w systemach monitoringu (detekcja i śledzenie obiektów), systemach interakcji człowiek-komputer, a także wykorzystywane są jako etap wstępnego przetwarzania przy rozpoznawaniu akcji i zachowań ludzi. Opracowano wiele metod detekcji tła i pierwszego planu działających w oparciu o dane z kamer RGB (Sobral & Vacavant, 2014). Do najprostszych metod zalicza się metody oparte na odjęciu tła (ang. *background substraction*). Metody te cechuje niewielki koszt obliczeniowy. Należą do najszybszych i dobrze sprawdzają się w systemach wbudowanych lub systemach czasu rzeczywistego wykorzystujących kilka kamer. Jednak metody te nie są pozbawione wad: często zawodzą przy zmianach oświetlenia, występujących cieniach itp. W praktyce najczęściej wykorzystuje sie nieco bardziej zaawansowane modele rozkładu kolorów, takie jak mieszanina Gaussianów (ang. *Mixture of Gaussians*) (Zivkovic & van der Heijden, 2006), sieci neuronowe (Maddalena & Petrosino, 2008) czy też modele oparte o inne metody statystyczne (Wren et al., 1997; Yao & Odobez, 2007; Arsic et al., 2009).

Jak już wspomniano, metody oparte na odjęciu tła należą do najprostszych. Jednak dzięki operowaniu na obrazach głębi, zamiast RGB, można uniknąć pewnych wad wspomnianych metod, w szczególności dotyczących nadmiernej czułości na zmiany oświetlenia sceny. W podstawowej wersji algorytmu, odejmowane są wartości pikseli obrazu tła od wartości pikseli obrazu z aktualnej klatki:

$$|D_t(x,y) - B_t(x,y)| > B_{thold} \qquad (2.20)$$

gdzie $D_t$ aktualna mapa głębi, $B_t$ jest referencyjnym modelem tła, a $Th$ to próg obcięcia, poniżej którego piksele są traktowane jako tło. Schemat metody budowy modelu tła przedstawiono na rys. 2.17. Model ten powinien jak najlepiej opisywać aktualne tło oraz zapewniać możliwość adaptacji w czasie. Do najpopularniejszych metod budowy modelu tego typu zaliczamy algorytmy, w których zastosowano filtry uśredniające (ang. *mean filter*) lub filtry medianowe (ang. *median filter*) w czasie. W przypadku zastosowania filtru uśredniającego dla każdego piksela tła wyznaczana jest średnia wartość na podstawie wartości pikseli z kilku obrazów. Operację tę można przedstawić w następujący sposób:

$$B_t(x,y) = \frac{1}{n}\sum_{i=1}^{n} D_{t-i}(x,y) \qquad (2.21)$$

Do budowy modelu tła można wykorzystać także operację mediany (Arsic et al., 2009) dla pikseli z kilku poprzednich obrazów:

$$B_t(x,y) = median\{D_{t-i}(x,y)\}, \qquad i \epsilon \{1,\dots,n\} \qquad (2.22)$$

Zaletą omówionych metod jest szybkość działania, wadą zaś zwiększone wymagania pamięciowe, co wynika z konieczności zapamiętania $n$ poprzednich map głębi. Metoda przedstawiona na rysunku 2.17. służy do budowy modelu tła przy inicjalizacji działania systemu oraz późniejszej jego aktualizacji na wypadek wykrycia zmian na scenie. W procesie inicjalizacji, w celu poprawnego zbudowania modelu tła, scena powinna być pusta (pozbawiona osób i innych poruszających się obiektów). Można wówczas wykorzystać, następujące po sobie obrazy. W procesie aktualizacji tła, jeśli użytkownik przebywa na scenie, obrazy zapisywane są do bufora co pewien czas, z większymi interwałami czasowymi, w ten sposób, aby uniknąć uwzględnienia pikseli należących do umieszczenia użytkownika w modelu tła. Operacja mediany pozwoli usunąć skrajne wartości pikseli, czego efektem będzie zbudowanie modelu bez osoby poruszającej się po scenie. Inercyjny czujnik ruchu noszony przez użytkownika na potrzeby detekcji upadku może także zostać wykorzystany do usprawnienia aktualizacji modelu tła. Informacja o tym czy użytkownik porusza się czy nie, pozwoli na uniknięcie włączenia pozostającej w spoczynku osoby do modelu tła (Kwolek & Kępski, 2014).

    Celem lepszego dopasowania algorytmu budowy modelu tła do specyfiki wynikającej z operowania na mapach głębi, dokonano istotnej zmiany w stosunku do klasycznej wersji algorytmu operującego na obrazach kolorowych. Jak już wcześniej wspomniano, wykorzystane urządzenie do akwizycji obrazów, dla każdego piksela zwraca wartość

odległości od kamery w milimetrach lub wartość 0 gdy nie można było dokonać pomiaru głębokości (*nmd*, oznaczone kolorem czarnym na obrazach).

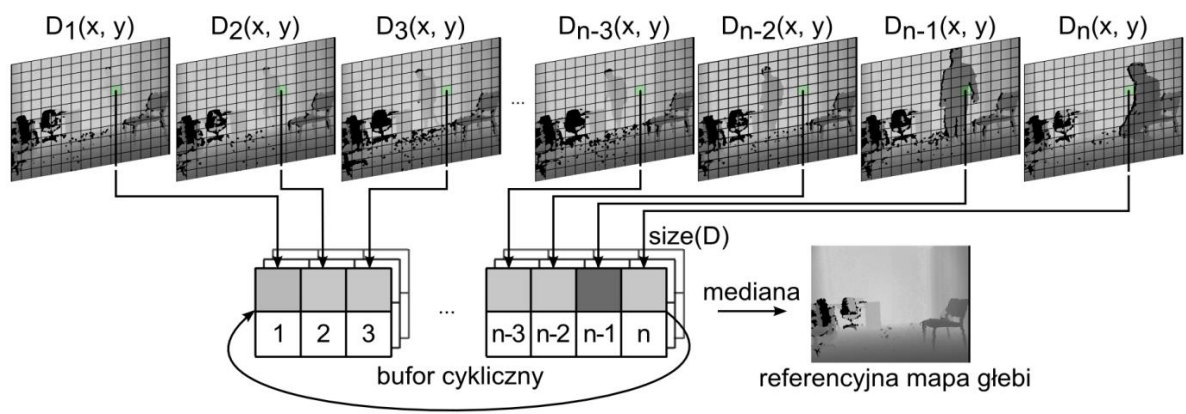

Rysunek 2.17. Schemat metody budowy modelu tła w oparciu o operację mediany obrazów w czasie.

Wynika to z niedoskonałości urządzenia. Modyfikacja algorytmu polega na pominięciu tych pikseli z $n$ obrazów, dla których wartość wynosi 0. Pozwala to uniknąć sytuacji, gdy dla współrzędnych $(x, y)$ na obrazie, w sekwencji $n$ pikseli dominować będą piksele *nmd*, skutkiem czego byłoby umieszczenie tych pikseli w referencyjnym modelu tła. Zmodyfikowaną operację mediany pikseli z $n$ poprzednich klatek można zatem przedstawić w następujący sposób:

$$B_t(x,y) = median\{D_{t-i}(x,y)\}, \qquad (2.23)$$
$$i \epsilon \{1, \dots, n\}, \ D_{t-i}(x,y) \neq 0$$

Modyfikacja ta nie spowoduje, że model tła będzie całkowicie pozbawiony pikseli o wartości zero, gdyż pewne obszary (obszary szklane, posiadające metaliczne powierzchnie, łatwo odbijające światło) będą zawsze reprezentowane na obrazie głębi jako zera. Nie mniej jednak, zastosowanie takiego podejścia pozwoli na eliminację znacznej liczby zerowych pikseli, co z kolei spowoduje powstanie gęstszej chmury punktów 3D modelu tła. Rezultaty tej modyfikacji przedstawiono na rysunku 2.18.

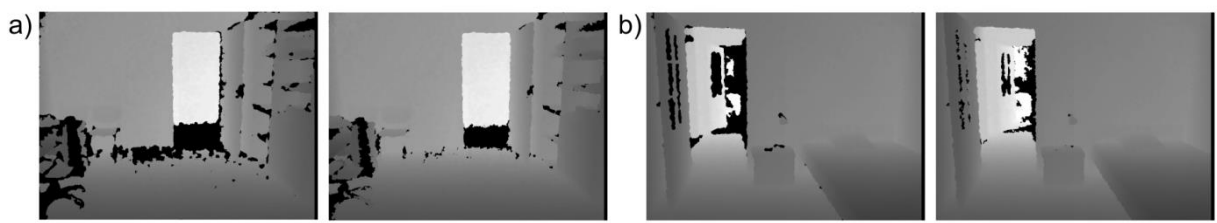

Rysunek 2.18. Pary modeli tła uzyskane klasyczną metodą mediany (po lewej stronie) oraz zmodyfikowaną metodą (obrazy po prawej stronie). Sekwencja a) zarejestrowana była przy dużym natężeniu światła słonecznego, zaś sekwencja b) przy niskim natężeniu światła.

Obraz tła uzyskany metodą zmodyfikowanej mediany jest lepszej jakości i charakteryzuje się mniejszą liczbą pikseli *nmd*. Tabela 2.3. prezentuje porównanie modeli tła zbudowanych obiema metodami, mając na względzie liczbę pikseli *nmd*.

Tabela 2.3. Porównanie wyników uzyskanych przez metody budowy modelu tła, mając na względzie udział pikseli o nieokreślonej wartości głębi.

| Sekwencja | Poziom natężenia światła słonecznego | Metoda podstawowa | | Metoda zmodyfikowana | |
|---|---|---|---|---|---|
| | | liczba pikseli *nmd* | odsetek pikseli *nmd* | liczba pikseli *nmd* | odsetek pikseli *nmd* |
| 1 | wysoki | 35696 | 11,62% | **19356** | **6,30%** |
| 2 | niski | 20077 | 6,54% | **11803** | **3,84%** |

Zastosowanie stałego interwału czasowego pomiędzy operacjami zapisu obrazów do bufora cyklicznego może prowadzić do sytuacji, w której fragment postaci użytkownika, poruszającego się ze zmienną prędkością po scenie lub przebywającego chwilowo w bezruchu, zostanie umieszczony w modelu tła. Ponadto zaobserwowano, że zazwyczaj wymagana jest modyfikacja jedynie fragmentu referencyjnej mapy głębi, gdyż zazwyczaj ingerencja użytkownika w otaczającą go scenę dotyczy jednego lub kilku obiektów tej sceny, pozostawiając pozostałe obiekty w stanie niezmienionym. Modyfikacja jedynie fragmentów modelu tła przedstawiających zmienione obszary sceny pozwoliłaby na zmniejszenie kosztu obliczeniowego poprzez znaczną redukcję liczby zbiorów pikseli, które należy poddać operacji mediany, a także pozwoliłaby zmniejszyć prawdopodobieństwo przypadkowego umieszczenia postaci w modelu tła. Na potrzeby niniejszej pracy zaproponowano modyfikację algorytmu budowy referencyjnej mapy głębi, polegającą na:

- Detekcji regionów zainteresowania algorytmu (ang. *Regions of Interest, ROI*), dzięki której regiony zainteresowania algorytmu są zdefiniowane jako obszary obrazu zawierające fragmenty sceny, które uległy modyfikacji (zob. Algorytm 3, linia 8 oraz rysunek 2.19d).
- Wykorzystaniu informacji o położeniu postaci w procesie doboru wartości głębokości umieszczanych w buforze. Informacja ta jest uzyskiwana w oparciu o segmentację metodą rozrostu obszarów (ang. *region growing*) oraz algorytmy śledzenia, opisane w kolejnym podrozdziale (zob. Algorytm 3, linia 5).

Zmodyfikowany algorytm budowy modelu tła można przedstawić w następujący sposób:

Algorytm 3 (Algorytm budowy modelu tła):

1. dany jest model tła $B(x,y)$, bufor $Q = \{D_{t-1}(x,y), D_{t-2}(x,y), \ldots, D_{t-Q_{size}}(x,y)\}$
2. pobierz nową mapę głębi $D_t(x,y)$ (zob. rysunek 2.19c.)
3. wyznacz obraz $F_t(x,y) = \begin{cases} D_t(x,y), & \text{jeśli } |D_t(x,y) - B(x,y)| \geq B_{thold} \\ 0, & \text{jeśli } |D_t(x,y) - B(x,y)| < B_{thold} \end{cases}$
4. wyznacz na $F_t(x,y)$ pola powierzchni komponentów połączonych *BLOB* i określ ich liczbę $n_{BLOB}$ // rysunek 2.19d.

5. przypisz obrazowi postaci $P_t(x,y)$ obszar *BLOB* o największym podobieństwie do $P_{t-1}(x,y)$ // wyznaczenie podobieństwa np. w oparciu o współczynniki kształtu
6. **if** $\frac{area(P_t(x,y))}{area(P_{t-1}(x,y))} > T_a$ **or** $n_{BLOB} > 1$
7.     wyznacz $F_{t-3}(x,y) = \begin{cases} D_{t-3}(x,y), & \text{jeśli } |D_{t-3}(x,y) - B(x,y)| \geq B_{thold} \\ 0, & \text{jeśli } |D_{t-3}(x,y) - B(x,y)| < B_{thold} \end{cases}$
8.     określ *ROI*, utwórz stos *S*, określ punkty startowe rozrostu obszarów (ang. *seed*) na podstawie $F_{t-3}(x,y)$ oraz utwórz tablicę logiczną *L* i zainicjuj ją wartościami *false*
9.     $P_t(x,y) = RegionGrowing(D_t(x,y), seed)$
10.     $D'_t(ROI) = D_t(ROI) - P_t(ROI)$ // rysunek 2.19f.
11.     odłóż $D'_t(ROI)$ na stos *S*
12.     $L = L\ or\ logical(D'_t(ROI))$ // rysunek 2.19g.
13.     **if** dla każdego (x,y), $L(x,y) \neq false$
14.         $B'(x,y) = median(S)$
15.         **return** $B'(x,y)$
16.     **else**
17.         pobierz mapę głębi
18.         wyznacz punkty *seed*
19.         **goto 9**
20. **else**
21.     $B'(x,y) = B(x,y)$
22.     **return** $B'(x,y)$ // rysunek 2.19h.

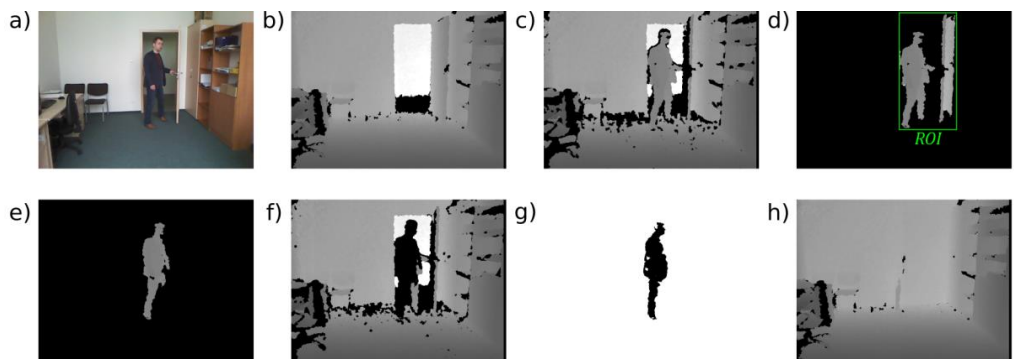

Rysunek 2.19. Ilustracja wydzielania tła w oparciu o zmodyfikowany algorytm budowy tła. a) obraz RGB b) początkowy model tła $B(x,y)$ c) mapa głębi $D(x,y)$ d) obiekty BLOB pierwszego planu e) sylwetka postaci po operacji segmentacji f) mapa głębi $D'(x,y)$ z usuniętą sylwetką człowieka g) tablica logiczna *L* h) zmodyfikowany model tła $B'(x,y)$.

Dla potrzeb analizy algorytmu załóżmy, że dany jest model tła $B(x, y)$, uzyskany podczas inicjalizacji systemu. Decyzja o aktualizacji referencyjnej mapy głębi podejmowana jest na podstawie analizy obrazu pierwszego planu $F_t(x, y)$. Celem segmentacji obiektów znajdujących się na scenie obraz ten poddawany jest operacji etykietowania elementów połączonych. W celu eliminacji szumu pomiarowego sensora zbiór komponentów połączonych jest ograniczony do obiektów posiadających określoną liczbę pikseli, większą od zadanego progu, dobranego eksperymentalnie. Potrzeba aktualizacji modelu tła może zaistnieć w następujących przypadkach:

- pojawienia się nowego komponentu na obrazie – w wyniku krótkotrwałej interakcji użytkownika z otoczeniem (np. przechodzący użytkownik przewrócił pewien obiekt sceny), interakcji innej osoby z otoczeniem (np. otwarcie drzwi przez nową osobę wchodzącą do monitorowanego pokoju użytkownika),
- zwiększenie obszaru komponentu reprezentującego użytkownika – w wyniku interakcji użytkownika z otoczeniem, podczas której osoba pozostaje w dłuższym kontakcie z przedmiotem. Skutkuje to wzrostem pola powierzchni komponentu w stosunku do poprzednich klatek sekwencji (np. otwieranie drzwi, przestawianie przedmiotów).

Pojawienie się nowego komponentu na obrazie wymaga sprawdzenia, czy nowym obiektem na scenie jest inny obiekt niż osoba. Aktualizacja modelu tła w takim wypadku byłaby niepotrzebna, co więcej, mogłaby doprowadzić do włączenia postaci do modelu tła. Weryfikacji można dokonać wykorzystując metody detekcji postaci ma mapach głębi: opracowaną dla potrzeb niniejszego rozwiązania (Kępski & Kwolek, 2014a) lub przedstawione w literaturze przedmiotu (Xia et al., 2011; Spinello & Arras, 2011). W razie konieczności aktualizacji tła, działanie algorytmu zilustrować można w następujący sposób:

- określ regiony zainteresowania (*ROI*), które zawierają zmodyfikowane elementy sceny (należące do zbioru połączonych komponentów),
- przygotuj stos *S*, na który odkładane będą mapy głębi służące do obliczenia nowego modelu tła (przy użyciu operacji mediany) oraz tablicę logiczną *L*, rozmiarem odpowiadającą regionom *ROI*,
- w każdej iteracji algorytmu dokonuj operacji segmentacji postaci (do inicjalizacji wykorzystywane jest położenie postaci przed momentem wykrycia zmiany na scenie, uzyskiwane dzięki przechowywaniu danych w buforze), mapę głębi pozbawioną postaci odłóż na stos, zaś w tablicy logicznej *L* oznacz niezerowe piksele jako *true*,
- gdy dla wszystkich pikseli z *ROI* wartości w tablicy logicznej są prawdziwe (oznacza to, że dla każdego położenia piksela istnieje w stosie taki obraz, który posiada wartość różną od 0, a więc dla wszystkich pikseli można obliczyć wartości głębokości), wykonaj operację mediany na obrazach ze stosu *S*.

Działanie przedstawionego algorytmu budowy tła zostało przebadane na sekwencjach obrazów przedstawiających osobę poruszającą się po pomieszczeniu, która dokonuje zmiany położenia dużych przedmiotów (otwiera i zamyka drzwi lub szafy, przestawia krzesła, itd.). Przykładowe rezultaty ilustrujące działanie algorytmu umieszczono w sieci web[8].

## 2.3. Algorytmy detekcji i śledzenia osób

### 2.3.1. Wydzielenie postaci

Segmentacja obrazu (ang. *image segmentation*) jest procesem podziału obrazu na obszary, które odpowiadają poszczególnym widocznym na obrazie obiektom i są jednorodne pod względem pewnych wybranych własności. Obszarami zainteresowania mogą być zbiory pikseli (punktów). Własnością, która powinna spełniać kryterium jednorodności jest wartość pikseli, która w przypadku obrazów RGB oznacza ich kolor, a w przypadku map głębi odległość od kamery. W statystyce problem ten jest szeroko opisany w literaturze i znany jako analiza skupień (ang. *cluster analysis*) (Szeliski, 2010).

Celem segmentacji jest zazwyczaj oddzielenie poszczególnych obiektów wchodzących w skład obrazu i wyodrębnienie ich od tła, na którym występują, np. wydzielenie postaci znajdującej się na scenie. Jest więc istotnym elementem percepcji sceny i etapem przygotowawczym do dalszej analizy obrazów: rozpoznawania obiektów, detekcji cech, czy też śledzenia obiektów. Większość metod segmentacji można zaliczyć do następujących kategorii (Skarbek & Koschan, 1994; Szeliski, 2010):

**Metody działające w oparciu o histogram.** Segmentacja polega na doborze progu na podstawie histogramu obrazu, zaś jej wynikiem jest obraz binarny. Histogram może być zbudowany w dziedzinie koloru, jasności lub głębi. Niskie nakłady obliczeniowe pozwalają na zastosowanie takich metod w przetwarzaniu danych video.

**Metody krawędziowe** wykorzystują fakt, że granice regionów i krawędzie w obrazie są ściśle powiązane. Krawędzie, które zostały zidentyfikowane na obrazie (przy zastosowaniu np. gradientu) muszą być połączone w celu uformowania zamkniętej krzywej otaczającej obszary jednorodne. Metody aktywnych konturów pozwalają na wykrycie krawędzi otaczających regiony o podobnych własnościach (Szeliski, 2010).

**Metody obszarowe**
- oparte o rozrost obszarów (ang. *region growing*) – polegają na poszukiwaniu elementów o podobnych właściwościach i łączeniu ich ze sobą,
- metody podziału i łączenia (ang. *split & merge*).

Istnieją także **metody hybrydowe** wykorzystujące więcej niż jedną z powyższych metod, np. łączące rozrost obszarów z informacją o przebiegu krawędzi. Osobną grupą są techniki wykorzystujące metody uczenia maszynowego (Ren & Malik, 2003; Cyganek, 2011).

Ze względu na ograniczone możliwości obliczeniowe mobilnej platformy *PandaBoard*, dla wbudowanego systemu detekcji upadku przygotowano i przebadano algorytm wydzielania sylwetki człowieka w sekwencji obrazów głębokości. Metoda ta polega na wykorzystaniu

---
[8] http://fenix.univ.rzeszow.pl/~mkepski/demo/bg/

uprzednio zbudowanego modelu tła do wydzielenia postaci. Piksel obrazu o głębokości $D_t$ i współrzędnych $(x, y)$ uznawany jest za należący do tła, jeśli wartość jego głębokości nie różni się od wartości głębokości piksela modelu tła $B_t$ o tych samych współrzędnych o więcej niż pewien predefiniowany próg $B_{thold}$ dobrany eksperymentalnie. Algorytm można przedstawić następująco:

Algorytm 4 (Algorytm wydzielania sylwetki człowieka):

1. Pobierz nową mapę głębi $D_t(x, y)$
2. Dla każdego piksela:
3.     **If** $|D_t(x,y) - B_t(x,y)| > B_{thold}$
4.         $F_t(x,y) = D_t(x,y)$
5.     **else**:
6.         $F_t(x,y) = 0$

Wynik operacji różnicy obrazów przedstawiono na rysunku 2.20. Jak można zaobserwować obraz wynikowy jest zaszumiony, lecz postać jest dość dobrze widoczna. Konieczne jest zatem przeprowadzenie operacji usunięcia szumu i wyodrębnienia z obrazu komponentów reprezentujących obiekty znajdujące się na scenie.

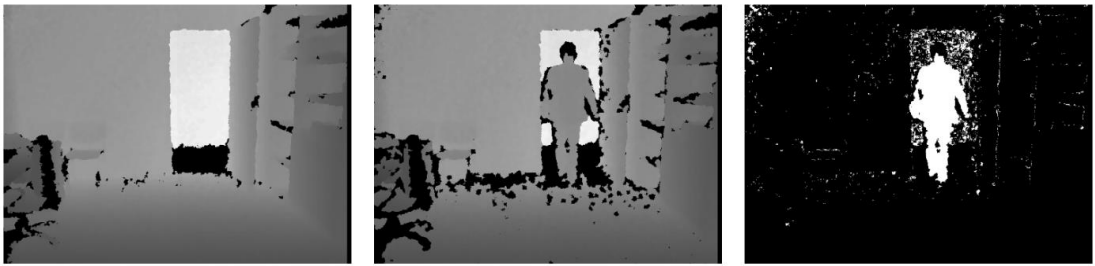

Rysunek 2.20. Operacja różnicy obrazów głębi. Od lewej: obraz tła, aktualny obraz głębi, różnica obrazów (zilustrowana w oparciu o obraz binarny).

Algorytm przetwarza wejściowy obraz binarny w ten sposób, że połączonym ze sobą obszarom pikseli nadawana jest wspólna etykieta. Można to przedstawić jako transformację obrazu binarnego $B$ w obraz symboliczny $S$, taką że:

- wszystkie elementy obrazu binarnego $B$ mające wartość tła, również będą należały do tła na obrazie $S$,
- każdy maksymalny podzbiór elementów obrazu binarnego $B$ niebędących tłem, otrzyma unikalną etykietę (reprezentowaną przez kolejne liczby naturalne) na obrazie symbolicznym $S$.

Sąsiedztwo pikseli może być 4-elementowe i dla piksela o współrzędnych $(u, v)$ może być zdefiniowane w następujący sposób:

$$\{(u-1, v), (u+1, v), (u, v-1), (u, v+1)\} \qquad (2.24)$$

lub 8-elementowe i zdefiniowane w następujący sposób:

$$\left\{\begin{array}{c}(u-1,v),(u+1,v),(u,v-1),(u,v+1),\\(u-1,v-1),(u+1,v-1),(u+1,v+1),(u-1,v+1)\end{array}\right\} \quad (2.25)$$

Większość stosowanych implementacji tej metody opiera się na algorytmie dwuprzebiegowym (ang. *two-pass*), w którym podczas pierwszego przebiegu pikselom nadawane są tymczasowe etykiety, zaś etykiety regionów sąsiadujących, które są przeznaczone do połączenia są przechowywane w tablicy. W trakcie lub po zakończeniu pierwszego przebiegu, dokonywane jest łączenie obszarów zazwyczaj za pomocą algorytmu *union-find* (Fiorio & Gustedt, 1996). Etykietowanie komponentów o 4-elementowym sąsiedztwie na przykładowym obrazie przedstawia rysunek 2.21.

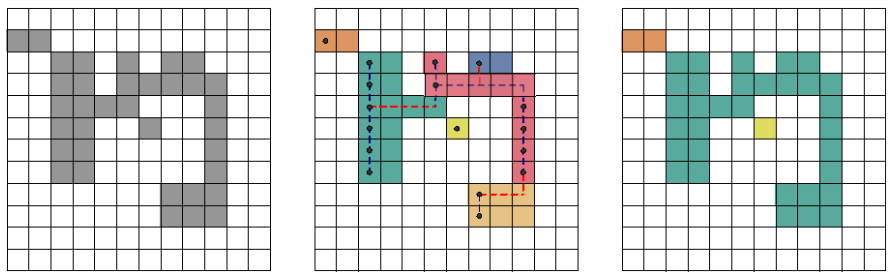

Rysunek 2.21. Etapy etykietowania komponentów połączonych, od lewej: oryginalny obraz binarny, pierwszy etap etykietowania (przebieg horyzontalny) z zaznaczonymi regionami przygotowanymi do połączenia, obraz wynikowy po operacji połączenia sąsiadujących obszarów.

Metodę etykietowania komponentów połączonych zastosowano w opracowanym systemie do usunięcia szumu i wydzielenia obiektów pierwszego planu na mapach głębi. Wprowadzono modyfikację metody, polegającą na łączeniu ze sobą we wspólny komponent tylko tych pikseli, których wartość głębi różni się nie więcej niż pewien próg $Th$, wyznaczony eksperymentalnie. Pozwala to na uniknięcie sytuacji, w której komponenty reprezentujące postać zostały obarczone szumem lub złączone z innymi obiektami, które mimo sąsiedztwa na obrazie, oddalone są w przestrzeni 3D. Połączone komponenty o odpowiednio dużej liczbie pikseli uznawane są za obiekty pierwszego planu (zob. rysunek 2.22). Gdy liczba komponentów jest większa niż 1, co zazwyczaj oznacza zmianę tła poprzez interakcję użytkownika z otoczeniem, dokonywana jest adaptacja tła z wykorzystaniem algorytmu opisanego w poprzednim podrozdziale (zob. algorytm 3).

Warto wspomnieć, że sukces metody aktualizacji tła zależy w dużej mierze od odpowiedniego doboru sekwencji obrazów, na których zostanie wykonana operacja mediany. W idealnym, pożądanym przypadku, pomimo obecności użytkownika na scenie obraz tła zostanie utworzony i uwzględniał będzie całą scenę z pominięciem użytkownika (co oznacza uniknięcie sytuacji, w której osoba lub fragment jej sylwetki zostałby "wtopiony" w tło). Obrazy muszą być dobierane z odpowiednim interwałem czasowym, co więcej interwał ten nie zawsze powinien być stały. Należy także wziąć pod uwagę kontekst, w jakim znajduje się osoba na scenie. Osoba może poruszać się z różną prędkością lub przebywać w spoczynku.

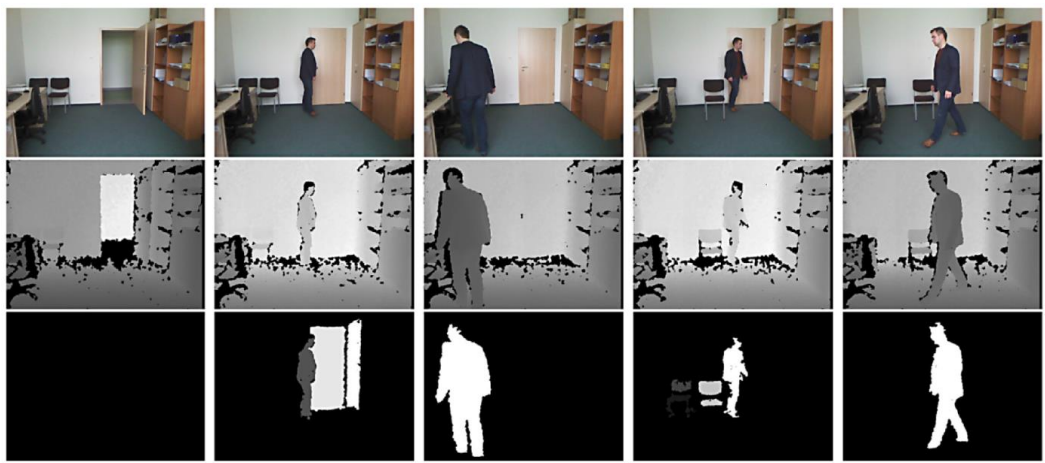

Rysunek 2.22. Działanie metody wydzielania postaci zilustrowane na kilku obrazach z sekwencji. Od góry: obraz kolorowy, mapa głębi, obraz symboliczny pierwszego planu zawierający etykiety obiektów.

Wówczas należy pominąć okresy nieaktywności osoby, a więc sytuacje w których cała sylwetka zostałaby zaliczona do nowego modelu tła. W podjęciu decyzji, które obrazy należy uwzględnić przy wyznaczaniu mediany, mogą być pomocne dane z inercyjnego czujnika ruchu, czy też śledzenie osoby na obrazach głębi, które zostanie przybliżone w następnym rozdziale.

W ramach niniejszej pracy przebadano (Kępski & Kwolek, 2012b) również możliwość powiązania metody etykietowania połączonych komponentów z algorytmem Mean-shift (Comaniciu & Meer, 2002). Jest to nieparametryczny estymator funkcji gęstości prawdopodobieństwa. Idea tego algorytmu polega na traktowaniu punktów w $d$-wymiarowej przestrzeni jak rozkład prawdopodobieństwa, której gęste regiony korespondują z maksimami funkcji prawdopodobieństwa. Algorytm ten mając do dyspozycji punkt początkowy iteracyjnie wyznacza wektor przesunięcia w kierunku maksimów funkcji rozkładu prawdopodobieństwa. Mając $n$ punktów $x_k, k = 1, \ldots, n$ w $d$-wymiarowej przestrzeni $\mathbb{R}^d$ wektor przesunięcia określany jest na podstawie następującej zależności:

$$m(x) = \frac{\sum_{k=1}^{n} x_k g\left(\left\|\frac{x - x_k}{h}\right\|^2\right)}{\sum_{k=1}^{n} g\left(\left\|\frac{x - x_k}{h}\right\|^2\right)} - x \qquad (2.26)$$

gdzie $g$ jest pewną funkcją jądra, a $h$ parametrem wygładzania. Dzięki temu Mean-shift dokonuje klasteryzacji $d$-wymiarowej przestrzeni poprzez przypisywanie każdemu punktowi wartości lokalnego maksimum korespondującej funkcji prawdopodobieństwa. Każdemu pikselowi obrazu przypisano wektor 3D określający jego położenie w przestrzeni. Następnie, po określeniu klastrów pikseli na obrazie, dokonano połączenia tych obszarów, które posiadały zbliżone wartości średniej głębokości, a więc analogicznie jak podczas drugiego przebiegu operacji etykietowania połączonych komponentów. Dzięki temu obiekty, wstępnie podzielone na kilka obszarów, mogły być połączone w jeden spójny region, korespondujący z rzeczywistą reprezentacją obiektu zainteresowania w przestrzeni 3D. Efekty tych operacji zilustrowano w sposób poglądowy na rysunku 2.23.

Algorytm Mean-shift charakteryzuje się dużym kosztem obliczeniowym w zadaniach segmentacji, w których każdy piksel jest punktem danych. W systemach wbudowanych o ograniczonej mocy obliczeniowej, złożoność ta nie pozwala na implementację zadania segmentacji w czasie rzeczywistym na obrazach o oryginalnej rozdzielczości 640x480 pikseli.

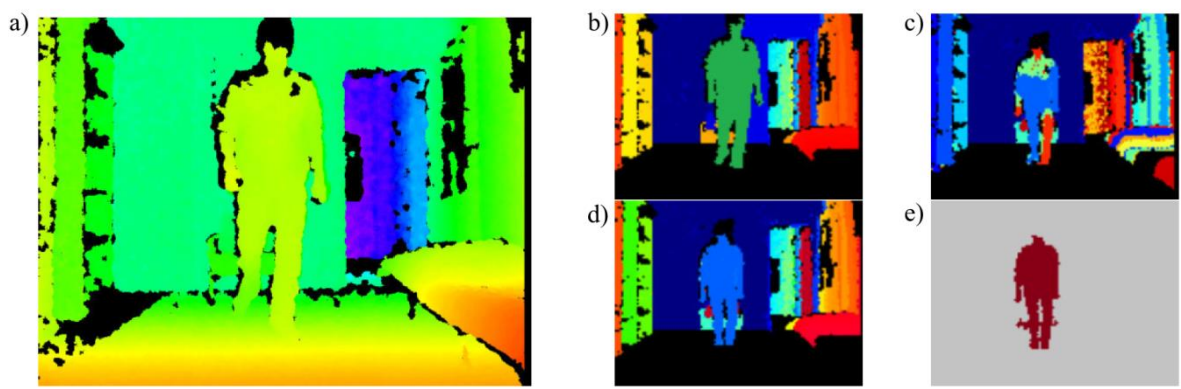

Rysunek 2.23. Wydzielenie postaci: a) mapa głębi, b), c) rezultat segmentacji Mean-shift, d) efektywniejsza segmentacja osoby metodą Mean-shift w połączeniu z metodą sąsiadujących komponentów, e) wydzielona postać.

Celem skrócenia czasu segmentacji do wartości umożliwiającej przetwarzanie w czasie rzeczywistym na platformie *PandaBoard*, dokonywano segmentacji na przeskalowanych obrazach do rozmiaru 128x96 pikseli.

Algorytm rozrostu obszarów (ang. *region growing*) jest jednym z algorytmów segmentacji polegającym na podziale obrazu na obszary, w zależności od pewnego zbioru punktów startowych (*seed points*). Algorytm ten opiera się na założeniu, że piksele w obrębie danego obszaru charakteryzują się podobną wartością intensywności. Metoda rozrostu obszarów została zaproponowana w pracy (Adams & Bischof, 1994) dla obrazów w skali szarości, lecz może być stosowana do obrazów kolorowych, wielospektralnych czy też map głębi, pod warunkiem doboru odpowiedniej metryki. Wejściem algorytmu jest obraz oraz zbiór punktów startowych pogrupowany w $n$ podzbiorów: $A_1, A_2, A_3, ..., A_n$. Podzbiory te mogą zawierać jeden lub więcej punktów. W kolejnych iteracjach algorytmu następuje rozrost początkowych obszarów, poprzez dołączanie pikseli o podobnych wartościach intensywności. Algorytm ten znajduje, taką segmentację obrazu, w której każdy element obrazu należy do dokładnie jednego obszaru powstałego z rozrostu podzbioru pikseli $A_i$, mając na względzie kryterium homogeniczności. W praktyce, obszary dołączane są do tego podzbioru, dla którego różnica δ pomiędzy wartością intensywności dla piksela a średnią intensywnością regionu jest najmniejsza. Wadą oryginalnego algorytmu jest silna zależność wyniku od kierunku przetwarzania pikseli, co szczególnie zauważalne jest, gdy obraz zawiera dużo małych obszarów o podobnej średniej intensywności pikseli. Jest to wynikiem zastosowania posortowanej listy pikseli, w której przechowywane są nieprzydzielone piksele w kolejności od najmniejszej do największej wartości δ. W kolejnych iteracjach algorytmu wartości δ pikseli znajdujących się na liście nie są aktualizowane, pomimo możliwej zmiany średnich wartości intensywności rozrastających się regionów. W pracy (Mehnert & Jackway, 1997) zaproponowano modyfikację algorytmu, polegającą na zastosowaniu kolejki priorytetowej (ang. *priority queue*, PQ) posortowanej rosnąco względem wartości δ i kilku stosów pikseli przechowujących piksele o wartości δ, zależnej od pozycji stosu w kolejce PQ. Podczas usuwania pikseli o najwyższym priorytecie (a w konsekwencji najmniejszej wartości δ) przetwarzany jest nie jeden piksel, a cały stos.

Na potrzeby wydzielania postaci w scenariuszu z ruchomą kamerą umieszczoną na obrotowej, aktywnej głowicy *pan-tilt,* przebadano możliwość zastosowania algorytmu rozrostu obszarów (Kępski & Kwolek, 2014a). Ze względu na specyfikę zagadnienia, które polega na wydzieleniu pojedynczego komponentu reprezentującego sylwetkę postaci, zaproponowano modyfikację oryginalnego algorytmu *region growing*. Polega na rezygnacji z możliwości rozrostu wielu obszarów, co pozwala zmniejszenie kosztu obliczeniowego algorytmu i łatwiejszego zastosowania w czasie rzeczywistym na mobilnej platformie obliczeniowej *PandaBoard*. Zastosowanie mapy kolejek, indeksowanej ma podstawie różnicy głębokości między dołączanymi pikselami a rozrastającym się regionem, zamiast sortowanej kolejki priorytetowej pozwala na uniknięcie konieczności sortowania w kolejnych iteracjach algorytmu. W tabeli 2.4. przedstawiono zestawienie oznaczeń wykorzystanych przy zapisie zmodyfikowanego algorytmu rozrostu obszaru. Następnie przedstawiono pseudokod zaproponowanej metody rozrostu obszaru (Algorytm 5) oraz obrazy ilustrujące rozrost obszaru co 200 iteracji algorytmu (Rysunek 2.24). Warto wspomnieć, że w scenariuszu z kamerą zamontowaną na suficie i skierowaną w dół pomieszczenia, ze względu na charakterystykę sensora głębi postać na obrazie reprezentowana jest zazwyczaj przez pewną liczbę grup pikseli o rosnących wartościach głębi (przypominających warstwice). Zastosowanie więc mapy kolejek pozwala na dołączanie wielu pikseli jednocześnie.

Tabela 2.4. Oznaczenia wykorzystane w pseudokodzie algorytmu rozrostu obszarów.

| Oznaczenia | |
| --- | --- |
| REGION_MEAN | Średnia wartość głębi rozrastającego się regionu |
| $\delta_{thold}$ | próg różnicy głębokości, wartość dobierana eksperymentalnie |
| NHQ | *Neighbours holding queue* – kolejka zawierająca wskaźniki do pikseli sąsiadujących z regionem rozrastającym się |
| QM | Mapa kolejek indeksowana na podstawie różnicy głębokości. Kluczem elementu jest δ, a wartością kolejka przechowująca piksele posiadające wartość δ. |
| FQ | Kolejka o najmniejszej wartości δ w danej iteracji |
| Etykiety | |
| NOT_LABELED | Piksel nie został odwiedzony |
| IN_NQUEUE | Piksel znajduje się w NHQ |
| IN_QUEUE | Piksel znajduje się w QM |
| LABELED | Piksel został przydzielony do rozrastającego się regionu |

Algorytm 5 (Zmodyfikowany algorytm rozrostu obszarów):

1. Nadaj etykiety pikselom regionu początkowego, oblicz REGION_MEAN, piksele sąsiadujące z regionem początkowym wstaw do NHQ, nadaj im etykietę IN_NQUEUE
2. **while** NHQ i QM są niepuste powtarzaj:
3.    **while** NHQ jest niepuste powtarzaj:
4.       usuń piksel z NHQ
5.       oblicz jego $\delta$
6.       **if** $\delta > \delta_{thold}$:
         **continue**
7.       wstaw do odpowiedniej kolejki w QM (o indeksie $\delta$)
8.       nadaj mu etykietę IN_QUEUE
9.    **if** QM jest niepuste:
10.       usuń FQ z QM
11.       **while** FQ jest niepuste:
12.          usuń piksel z FQ
13.          nadaj mu etykietę LABELED
14.          dodaj jego sąsiednie piksele NOT_LABELED do NHQ i nadaj im etykietę IN_NQUEUE
15.    zaktualizuj REGION_MEAN o wartości pikseli usuniętych z FQ

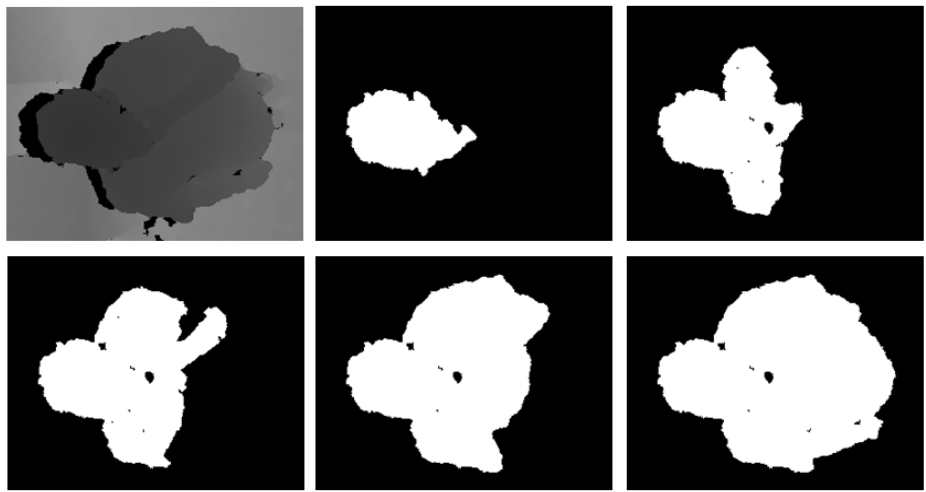

Rysunek 2.24. Obrazy ilustrujące rozrost obszaru co 200 iteracji algorytmu. Pierwszy obraz przedstawia fragment mapy głębi.

W celu uzyskania dokładniejszej segmentacji postaci oraz uniknięcia możliwości włączenia do komponentu reprezentującego postać pikseli do niej nienależących, szczególnie w przypadku, w którym osoba dokonuje interakcji z otoczeniem, można wykorzystać detektor postaci oparty o histogramy gradientów i klasyfikator SVM. W przypadku nagłego wzrostu pola powierzchni wydzielonego komponentu, podejście zaprezentowane w pracy (Kępski & Kwolek, 2014a) pozwala na dokonanie detekcji jedynie w obszarze komponentu, a nie na całej mapie głębi. Takie rozwiązanie jest korzystne obliczeniowo, szczególnie w kontekście implementacji systemu na platformie obliczeniowej z procesorem o architekturze ARM. Detekcja postaci dostarcza informacji o położeniu postaci, która może być wykorzystana do ponownej inicjalizacji algorytmu rozrostu obszarów. Przykładowe rezultaty ilustrujące działanie algorytmu rozrostu obszarów umieszczono w sieci web[9].

### 2.3.2. Śledzenie głowy osoby

W trakcie wykonywania przez użytkownika czynności życia codziennego oprócz postaci mogą pojawić się na wydzielonym obrazie pierwszego planu inne elementy sceny. Śledzenie położenia użytkownika na scenie może być w takiej sytuacji przydatne w odróżnianiu osób od innych obiektów lub jak wykazano w pracy (Rougier et al., 2006) informacja o położeniu głowy postaci może zwiększyć skuteczność procesu detekcji upadku. W algorytmach śledzenia głowy najczęściej wykorzystuje się modelowanie owalnego kształtu głowy za pomocą elipsy oraz aktywne modele wyglądu (Rymut & Kwolek, 2010). Jak wskazują wyniki badań eksperymentalnych, jakkolwiek rozpatrywanie pojedynczej hipotezy może być korzystne ze względu na wymagania obliczeniowe (Kwolek, 2005), to w praktycznych zastosowaniach najskuteczniejsze okazały się algorytmy wykorzystujące wiele hipotez. Typowymi reprezentantami tych algorytmów są filtry cząsteczkowe (algorytmy kondensacji stanu) i algorytmy optymalizacji w oparciu o rój cząstek (Kennedy & Eberhart, 1995).

Problem śledzenia można przedstawić jako zagadnienie estymacji nieznanego stanu pewnego dynamicznego systemu. Stan ten można przedstawić w uproszczeniu jako zbiór wszystkich czynników mogących mieć wpływ na dany system w przyszłych jednostkach czasu (Thrun et al., 2005). Stan oznaczamy jako $x$, natomiast stan systemu w konkretnym czasie $t$ jako $x_t$. Stan ten może być jedynie estymowany na podstawie danych pomiarowych, oznaczonych jako $z$. Klasycznym przykładem algorytmu estymacji ukrytego stanu systemu jest filtracja Bayesa.

Filtry Bayesa stosuje się zakładając, że stochastyczny proces spełnia kryteria procesu Markowa, co oznacza że stan $x_t$ zależy bezpośrednio od stanu procesu w poprzedzającej chwili czasu:

$$p(x_t| x_{1:t-1}) = p(x_t| x_{t-1}) \qquad (2.27)$$

gdzie $p$ jest warunkowym rozkładem prawdopodobieństwa. Założenia modelu Markowa dotyczą także niezależności poszczególnych pomiarów $z$:

$$p(z_t| x_{1:t}) = p(z_t| x_t) \qquad (2.28)$$

Sieć Bayesa została przedstawiona na rysunku 2.25.

---

[9] http://fenix.univ.rzeszow.pl/~mkepski/demo/rg/

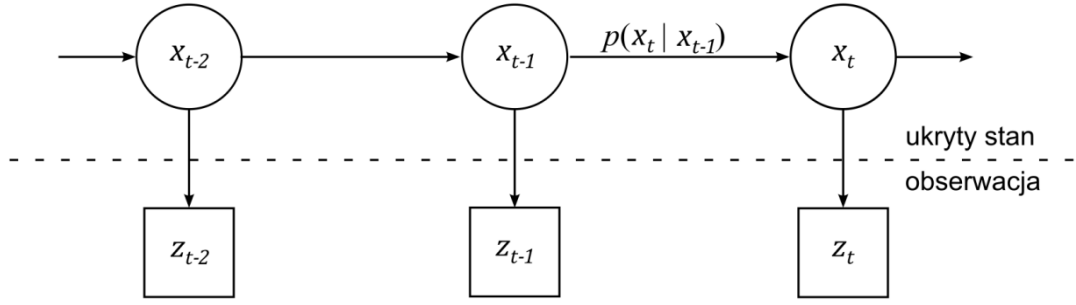

Rysunek 2.25. Sieć Bayesa

W metodach bayesowskich wiedza o aktualnym stanie systemu w chwili czasu $t$ reprezentowana jest za pomocą warunkowego rozkładu prawdopodobieństwa *a posteriori* $p(x_t| z_t)$ zmiennej stanu $x_t$. Filtry Bayesa rozwiązują problem estymacji stanu systemu poprzez zastosowanie następujących po sobie mechanizmów predykcji i korekcji. Załóżmy, że poprzedni rozkład prawdopodobieństwa $p(x_{t-1}| z_{t-1})$ jest znany w czasie $t$. Etap predykcji polega na określeniu rozkładu prawdopodobieństwa bez znajomości nadchodzącego w czasie $t$ pomiaru $z_t$:

$$p(x_{t-1}| z_{1:t-1}) \rightarrow p(x_t| z_{1:t-1}) \tag{2.29}$$

Etap ten można przedstawić za pomocą równania Chapmana-Kołmogorowa:

$$p(x_t| z_{1:t-1}) = \int p(x_t| x_{t-1}) p(x_{t-1}| z_{1:t-1}) dx_{t-1} \tag{2.30}$$

Etap korekcji polega na określeniu rozkładu prawdopodobieństwa przy znajomości pomiaru $z_t$. Etap ten następuje po wykonaniu etapu predykcji:

$$\{p(x_t| z_{1:t-1}), z_t\} \rightarrow p(x_t| z_{1:t}) \tag{2.31}$$

$$p(x_t| z_{1:t}) = p(z_t| x_t) p(x_t| z_{1:t-1}) \tag{2.32}$$

Sekwencyjną aktualizację warunkowego rozkładu prawdopodobieństwa *a posteriori* można przedstawić korzystając z zależności (2.30) i (2.32):

$$p(x_t| z_{1:t}) = p(z_t| x_t) \int p(x_t| x_{t-1}) p(x_{t-1}| z_{1:t-1}) dx_{t-1} \tag{2.33}$$

Do najczęściej spotykanych metod wyznaczania estymaty wektora stanu modelu dynamicznego systemu opartych o sieci Bayesa, można zaliczyć filtr Kalmana (ang. *Kalman Filter*) (Kalman, 1960), rozszerzony filtr Kalmana (ang. *Extended Kalman Filter*) (Ljung, 1979) oraz filtr cząsteczkowy (ang. *Particle Filter*) (Gordon et al., 1993).

Filtr Kalmana jest optymalnym estymatorem stanu systemu liniowego, wyznaczanym na podstawie pomiarów na wyjściu systemu. Rozkład *a posteriori* jest wielowymiarowym rozkładem normalnym, jeśli oprócz założeń modelu Markowa spełnione są trzy poniższe warunki:

- prawdopodobieństwo $p(x_t| x_{t-1})$ jest funkcją liniową i obarczone jest błędem o rozkładzie gaussowskim,

- prawdopodobieństwo pomiaru $p(z_t|x_t)$ także jest funkcją liniową i obarczone jest błędem o rozkładzie gaussowskim,
- początkowe prawdopodobieństwo $p(x_0)$ ma rozkład normalny.

Do oszacowania bieżącego stanu systemu filtr Kalmana wymaga jedynie znajomości stanu poprzedniego oraz wektora obserwacji. W procesie filtracji można wyróżnić dwie fazy: predykcji oraz uaktualniania. Podczas predykcji na podstawie bieżącego stanu systemu przewidywane są przyszłe wartości wektora stanu, natomiast na etapie uaktualniania wykorzystuje się bieżące obserwacje, co z kolei prowadzi do poprawy przewidywanego wektora stanu. Filtr Kalmana jest optymalnym estymatorem, ponieważ przy konkretnych założeniach może spełniać pewne kryterium, np. minimalizacji błędu średniokwadratowego estymowanych parametrów. Rozszerzony filtr Kalmana jest przykładem modyfikacji oryginalnego algorytmu, pozwalającym na zastosowanie go w systemach dyskretnych. Największą wadą rozszerzonego filtru Kalmana jest to, że poddanie zmiennej losowej o rozkładzie normalnym nieliniowym transformacjom spowoduje zmianę rozkładu na inny niż normalny. Algorytm ten jest prostym estymatorem stanu, który poprzez linearyzację przybliża bayesowskie reguły filtracji. Ponadto metoda ta nie nadaje się szczególnie dobrze do śledzenia wielu hipotez jednocześnie. Pomimo wspomnianych wad, metoda ta jest szeroko stosowana, ze względu na swoją prostotę i stosunkowo niewielką złożoność obliczeniową, która wynosi $O(k^{2,8} + n^2)$, gdzie $k$ jest długością wektora $z_t$, a $n$ liczbą wymiarów przestrzeni stanu $x_t$.

Filtr cząsteczkowy jest nieparametryczną implementacją filtracji bayesowskiej. Główną ideą jest reprezentacja rozkładu prawdopodobieństwa *a posteriori* za pomocą skończonego zbioru próbek wylosowanych na podstawie danego rozkładu. Estymatorem miary prawdopodobieństwa $p$ jest rozkład empiryczny o postaci:

$$p(x) = \frac{1}{N} \sum_{i=1}^{N} \delta(x - x^{(i)}) \tag{2.34}$$

gdzie $\delta$ jest deltą Diraca, natomiast ciąg jest $\{x^{(i)}\}_{i=1}^{N}$ ciągiem niezależnych próbek, zwanych cząstkami. Każda cząstka jest instancją stanu w czasie $t$, a więc hipotezą na temat prawdziwego, ukrytego stanu w czasie $t$. W praktyce do wygenerowania próbki stosuje się metody Monte Carlo wykorzystujące łańcuchy Markowa (MCMC) lub rozkład proponowany (funkcję istotności). Liczba cząstek jest zazwyczaj duża i zależy od wymiarów wektora stanu. Istnieją implementacje, w których liczba cząstek jest funkcją zmiennej $t$ lub innych wartości związanych z rozkładem prawdopodobieństwa $p$. Metoda oparta na funkcji ważności polega na zastąpieniu rozkładu $p$ rozkładem $q$ o podobnych właściwościach oraz późniejszym wykorzystaniu go do generacji próbki niezależnych, ważonych zmiennych losowych $\{x^{(i)}, w^{(i)}\}_{i=1}^{N}$. Wagi $w^{(i)}$ odzwierciedlają prawdopodobieństwo, że próbki pochodzą z rozkładu $p$. Wybór funkcji istotności jest istotnym krokiem, gdyż od niej zależy efektywność filtru. W najprostszej wersji filtru wykorzystywana jest sekwencyjna funkcja ważności (ang.

*Sequential Importance Sampling* – SIS), w której wagi wyznacza się z następującej zależności:

$$w_t^{(i)} = \frac{p(x_t^{(i)}|z_t)}{q(x_t^{(i)}|z_t)} \propto w_{t-1}^{(i)} \frac{p(z_t|x_t^{(i)})p(x_t^{(i)}|x_{t-1}^{(i)})}{q(x_t^{(i)}|x_{t-1}^{(i)},z_t)} \quad (2.35)$$

Aby zapobiec degeneracji próbek, czyli spadkowi wartości wag do wartości zaniedbywalnych, stosuje się mechanizm przepróbkowania (ang. *resampling*). Polega on na zastąpieniu cząstek z niewielkimi wagami nowym zbiorem cząstek $N$ z jednakowymi wagami:

$$\left\{x_t^{(i)}, w_t^{(i)}\right\} \rightarrow \left\{x_t^{(i)}, \frac{1}{N}\right\} \quad (2.36)$$

Dzięki temu algorytm skupia się na tych obszarach przestrzeni stanu, w których cząstki stosunkowo trafnie odwzorowują stan systemu. Omawiany algorytm jest rozszerzoną wersją algorytmu SIS i znany jest w literaturze pod nazwą SIR (ang. *Sampling Importance Resampling*). Schemat jego działania przedstawiono na rysunku 2.26.

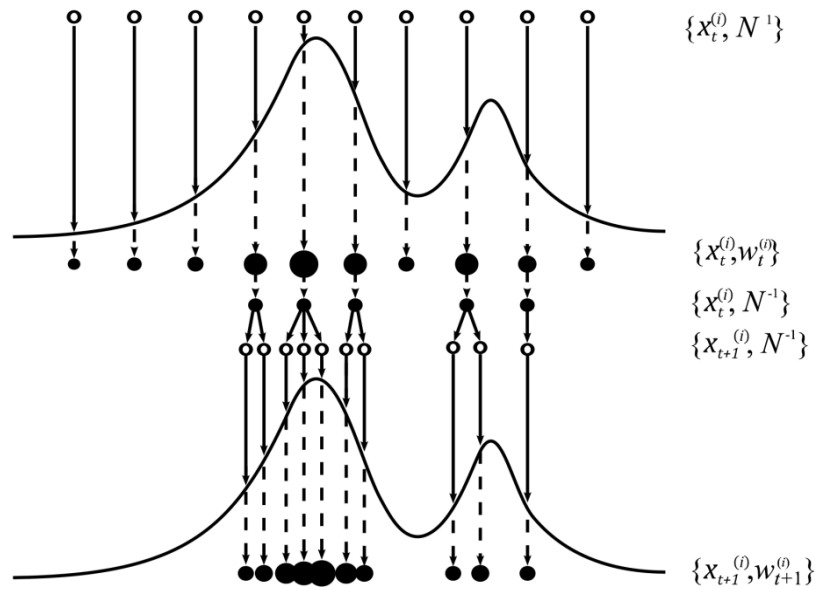

Rysunek 2.26. Ilustracja działania filtru cząsteczkowego z przepróbkowaniem.

Metoda ta polega na przeprowadzeniu następujących po sobie kroków predykcji, korekty oraz ponownego próbkowania. Dzięki temu zapobiega się propagowaniu nieistotnych cząstek do kolejnych iteracji. Algorytm filtru cząsteczkowego SIR można przedstawić w następujący sposób:

Algorytm 4:

1. $\left\{x_t^{(i)}, w_t^{(i)}\right\}_{i=1}^{N} = \text{PF}\left(\left\{x_{t-1}^{(i)}, w_{t-1}^{(i)}\right\}_{i=1}^{N}, z_t\right)$
2.     **Dla** $i = 0, 1, \ldots, N$
3.         $x_t^{(i)} \sim p\left(x_t^{(i)} \middle| x_{t-1}^{(i)}\right)$

4. $\quad w_t^{(i)} = p\left(z_t \big| x_t^{(i)}\right)$

5. $\quad W_s = \sum_{i=1}^{N} w_t^{(i)}$

6. **Dla** $i = 0, 1, \ldots, N$

7. $\quad w_t^{(i)} = \frac{w_t^{(i)}}{W_s}$

8. $\quad \left\{x_t^{(i)}, w_t^{(i)}\right\}_{i=1}^{N} = \text{Resampling}\left(\left\{x_t^{(i)}, w_t^{(i)}\right\}_{i=1}^{N}\right)$

W omawianym systemie głowa postaci jest śledzona przy wykorzystaniu filtru cząsteczkowego. Dla filtru opracowano model obserwacji $p(z_t | x_t)$, który pozwala na efektywne śledzenie obarczone niewielkim błędem. Kształt głowy postaci zamodelowany był za pomocą elipsoidy (Kępski & Kwolek, 2014c). Model obserwacji łączy w sobie wartości funkcji dopasowania pomiędzy elipsoidą a chmurą punktów oraz projekcją elipsoidy na płaszczyznę z krawędziami mapy głębokości. Motywacją do takiego sformułowania modelu była charakterystyka danych pochodząca z urządzenia. Odległość pomiędzy sąsiednimi elementami chmury punktów jest niewielka dla obiektów znajdujących się blisko kamery i rośnie wraz z nią. Obiekty znajdujące się dalej od kamery reprezentowane są w postaci warstwic, co powoduje częściową utratę informacji o ich rzeczywistym kształcie względem osi $z$ układu kamery (głębokości). Omawiane zjawisko dotyczy szczególnie wszelkiego rodzaju krzywizn. W celu określenia dopasowania modelu do chmury punktów, w pierwszej kolejności wyznaczana jest odległość punktu $(x, y, z)$ do elipsoidy za pomocą następującej zależności:

$$d = \sqrt{\frac{(x-x_0)^2}{a^2} + \frac{(y-y_0)^2}{b^2} + \frac{(z-z_0)^2}{c^2} - 1} \qquad (2.37)$$

gdzie $a, b, c$ to półosie elipsoidy. Stopień przynależności punktu do elipsoidy określany jest w następujący sposób:

$$m = 1 - \frac{d}{d_{thold}} \qquad (2.38)$$

gdzie $d_{thold}$ jest progiem ustalonym eksperymentalnie. Funkcja dopasowania modelu głowy postaci do chmury punktów określona jest więc następująco:

$$f_1 = \sum_{(x,y,z) \in S} m(x, y, z) \qquad (2.39)$$

gdzie $S(x_0, y_0, z_0)$ oznacza zbiór punktów należących do głowy postaci i jej otoczenia. Jak już wcześniej wspomniano, oprócz dopasowania modelu do chmury punktów, wyznaczane jest także dopasowanie krawędzi mapy głębi do elipsy. Elipsoida jest rzutowana na płaszczyznę obrazu wykorzystując model kamery *Kinect*. Dla każdego piksela $(u, v)$ należącego do elipsy $E$ obliczane jest dopasowanie do krawędzi na podstawie poniższej zależności:

$$p = \begin{cases} D_e(u,v) \cdot (5 - D_d(u,v)), & D_d(u,v) < 5 \\ 0, & D_d(u,v) \geq 5 \end{cases} \quad (2.40)$$

gdzie $D_e(u,v) \epsilon \{0,1\}$ jest wartością piksela na obrazie krawędzi rzutowanej elipsy, a $D_d(u,v)$ jest wartością piksela na mapie odległości od krawędzi wydzielonych z mapy głębi, a wartość 5 jest progiem dobranym eksperymentalnie. Zatem dopasowanie elipsy można przedstawić w sposób następujący:

$$f_2 = \sum_{(u,v)\epsilon E} p(u,v) \quad (2.41)$$

Model obserwacji, wykorzystujący funkcje dopasowania 2D i 3D, zdefiniowany jest za pomocą zależności:

$$p(x_t^{(i)}|z_t^{(i)}) = \frac{1}{\sqrt{2\pi\sigma^2}} \exp\left(-\frac{f_1 \cdot f_2}{2\sigma^2}\right) \quad (2.42)$$

Pomiędzy jednostką czasu $t-1$ a $t$ wszystkie cząstki są propagowane w przestrzeni stanu zgodnie z modelem lokomocji:

$$x_t^{(i)} = x_{t-1}^{(i)} + \delta, \qquad \delta = N(0, \Sigma) \quad (2.43)$$

Na rysunku 2.27. przedstawiono przykładowe obrazy ilustrujące śledzenie głowy postaci dla sekwencji z kamerą statyczną za pomocą proponowanej metody i filtru składającego się z 500 cząsteczek. Wartości δ zostały dobrane eksperymentalnie.

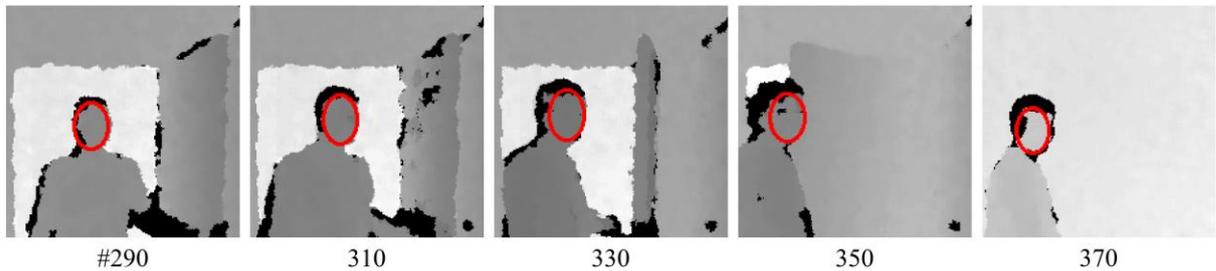

Rysunek 2.27. Przykładowe obrazy prezentujące śledzenie głowy postaci w scenariuszu ze statyczną kamerą.

Wektor stanu składa się z pięciu elementów i zawiera współrzędne 3D środka elipsoidy $(x_0, y_0, z_0)$ oraz wartości dwóch kątów obrotu modelu głowy (*pitch, roll*). Jak można zaobserwować na rysunku 2.27. ewaluacja śledzenia była przeprowadzona podczas interakcji użytkownika z obiektami sceny. W rozwiązaniu opartym jedynie o model tła doszłoby do sytuacji, w której model tła stawał się nieaktualny i na obrazie pierwszego tła pojawiłyby się inne obiekty oprócz użytkownika. Dzięki zastosowaniu śledzenia głowy osoby, uzyskano informację o położeniu istotnej części ciała człowieka. Wykorzystanie wspomnianej informacji może znacząco polepszyć skuteczność detekcji upadku.

Użyteczność śledzenia głowy w detekcji upadku została też przebadana w scenariuszu z aktywną kamerą, umieszczoną na głowicy o dwóch stopniach swobody. Śledzenie postaci w takim scenariuszu jest szczególnie istotne, gdyż informacja o położeniu użytkownika na scenie pozwala na podążanie kamery za osobą. Wymagane jest także zastąpienie metody wydzielania postaci, wykorzystywanej dla kamery statycznej, inną metodą, nie wykorzystującą odjęcia modelu tła. W niniejszym systemie wykorzystywano metodę rozrostu obszarów. Rezultaty śledzenia głowy przedstawiono na rysunku 2.28 i udostępniono w sieci web[10].

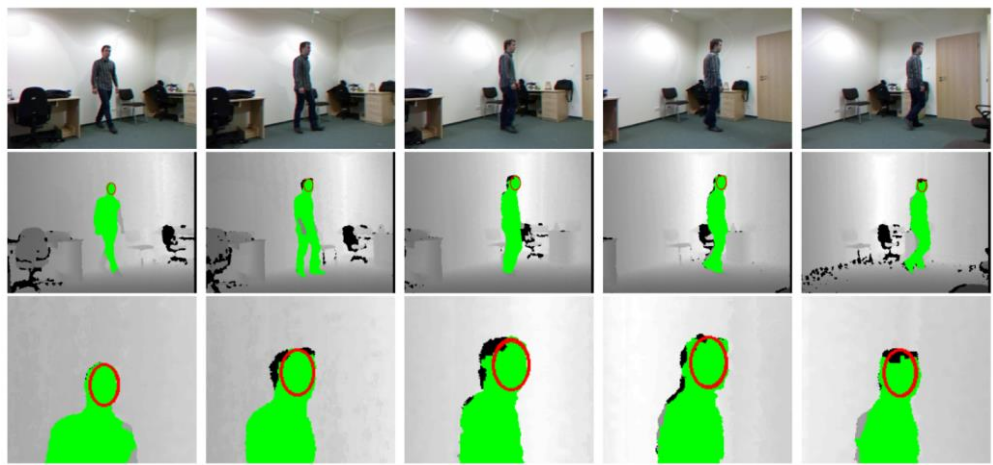

Rysunek 2.28. Wyniki śledzenia głowy postaci w scenariuszu z kamerą aktywną.

Na potrzeby scenariusza działania systemu, w którym kamera zamontowana jest na suficie i skierowana w dół pomieszczenia, opracowano algorytm śledzący oparty o filtr cząsteczkowy i deskryptor przedstawiony na rysunku 2.29. W trakcie badań eksperymentalnych przebadano szereg deskryptorów. Deskryptory budowane były w oparciu o prostokąty o różnych wielkościach i wzajemnym ułożeniu. Okazało się, że najlepsze wyniki uzyskać można w oparciu o koncentryczne ułożenie prostokątów (zob. rysunek 2.29a). Celem szybszego wyznaczenia średniej wartości głębi w każdym z prostokątów, wykorzystywano obrazy z zsumowanymi wartościami głębi (ang. *integral images*). Funkcja dopasowania modelu obserwacji filtru opiera się o histogramy różnic średnich głębi wyznaczonych w poszczególnych oknach deskryptora względem elementu środkowego (Kępski & Kwolek, 2016b; Kępski & Kwolek, 2016a).

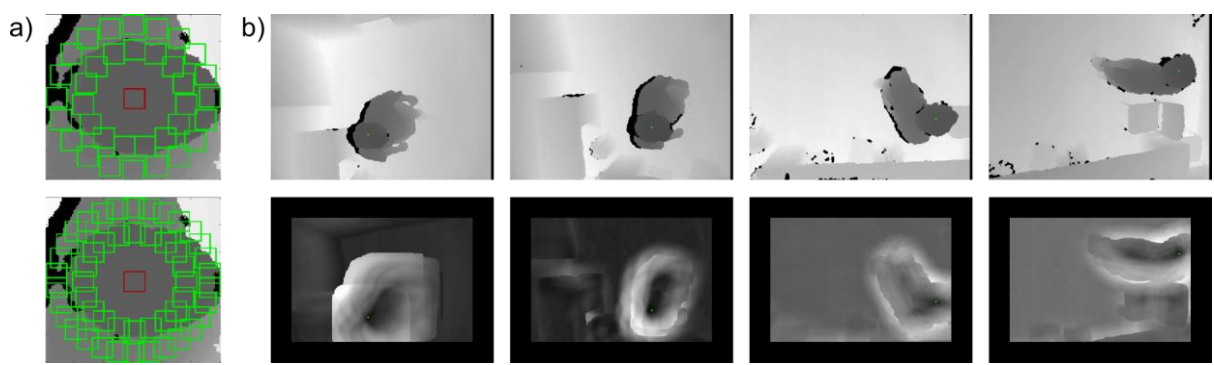

Rysunek 2.29. Ilustracja śledzenia postaci w scenariuszu z kamerą "u góry". a) przykładowe przebadane deskryptory b) sekwencja obrazów głębi i odpowiadających im map prawdopodobieństwa.

---

[10] http://fenix.univ.rzeszow.pl/~mKępski/demo/tracking

## 2.4. Podsumowanie

W rozdziale omówiono algorytmy przetwarzania obrazów i percepcji sceny. Zaprezentowano modyfikacje algorytmów, które zostały opracowane dla potrzeb skutecznej detekcji i ekstrakcji postaci na mapach głębi. Celem zwiększenia obszaru obserwacji, a w szczególności zapewnienia obserwacji całego pokoju o powierzchni 15-20 m$^2$ przez kamerę zamocowaną na wysokości około 3 m, zaproponowano wykorzystanie kamery aktywnej. Śledzenie osoby realizowane jest w oparciu o zbudowaną głowicę *pan/tilt*. W celu zwiększenia niezawodności ekstrakcji osoby zaproponowano modyfikacje istniejących algorytmów, a także zaproponowano nowy algorytm śledzenia osoby. Zaproponowane usprawnienia dotyczyły algorytmów ekstrakcji tła oraz algorytmu rozrostu obszarów na obrazach głębi. Dzięki zaproponowanym udoskonaleniom uzyskano konkurencyjne rozwiązania do wydzielania postaci na mapach głębi, charakteryzujące się nie tylko możliwością działania w czasie rzeczywistym na platformie obliczeniowej z procesorem o architekturze ARM, ale także wysoką skutecznością ekstrakcji osoby.

# 3. Detekcja upadku – wyniki badań eksperymentalnych

## 3.1. Charakterystyka danych akcelerometrycznych na potrzeby detekcji upadku

Autorzy prac (Bourke et al., 2007; Noury et al., 2008; Kangas et al., 2008; Li et al., 2009) zwracają uwagę na wysoką skuteczność metod detekcji upadku w oparciu o dane akcelerometryczne. Większość badań, dotyczących wykorzystania inercyjnych czujników ruchu w detekcji upadku, przeprowadzono w oparciu o zbiór symulowanych akcji wykonywanych na niewielkich grupach osób, z czego jedynie czynności dnia codziennego wykonywane były przez osoby starsze. Biorąc pod uwagę statystyki (Bagalà et al., 2012), zgromadzenie odpowiedniej liczby upadków osób starszych nie jest łatwym zadaniem. Przy założeniu, że 30% osób powyżej 65. roku życia upada raz w roku, do zarejestrowania stu zdarzeń upadku wymagana jest akwizycja w przybliżeniu 100 000 dni aktywności monitorowanej osoby. Z tego powodu większość prac, wykorzystuje dane symulowane przez zdrowe, młode osoby, niejednokrotnie profesjonalnie lub zawodowo uprawiające sport (Chen et al., 2005). Co więcej, ze względu na liczbę fałszywych alarmów, algorytmy często badane są na podstawie przygotowanego scenariusza wykonywanych akcji, a nie w środowisku docelowego użytkownika podczas jego normalnego funkcjonowania. Skuteczność metody zaproponowanej w (Bourke et al., 2007) została przebadana na symulowanych upadkach oraz ADLs wykonywanych przez 10 starszych osób. Osoby starsze wykonywały akcje należące do dwóch typów: siadanie (na fotelu, krześle, łóżku, itp.) oraz chodzenie. Każda osoba powtarzała ściśle określony scenariusz trzykrotnie.

W literaturze (Noury et al., 2008) zaproponowano podział aktywności człowieka związanej z typowym upadkiem, na następujące fazy:

1. normalna czynność dnia codziennego, taka jak chodzenie, wstawanie z krzesła czy łóżka, itp.,
2. faza krytyczna, w której człowiek znajduje się w niekontrolowanym ruchu przypominającym spadek swobodny, po którym następuje uderzenie (ang. *impact*),
3. faza po upadku, zazwyczaj związana z leżeniem na podłodze,

4. faza próby powrotu do normalnej postawy, która może zakończyć się niepowodzeniem lub w ostateczności może do niej nie dojść, np. w wyniku urazu lub utraty przytomności.

Większość algorytmów wykorzystujących predefiniowany próg definiuje akcję jako upadek, jeśli wartość wektora przyspieszenia określonego wzorem:

$$SV_{total} = \sqrt{A_x^2 + A_y^2 + A_z^2} \tag{3.1}$$

jest większa od przyjętej wartości progu *UFT* (ang. *lower fall threshold*). Podejście to opiera się na założeniu, że wartość wektora przyspieszenia w momencie uderzenia znacznie przekracza wartości typowe dla normalnych, codziennie wykonywanych akcji. Część metod definiuje inny próg, określony jako *LFT* (ang. *lower fall threshold*). Wykorzystuje on fakt, że podczas upadku, tuż przed uderzeniem, przyspieszenie ciała typowo rośnie, a więc wartość mierzona przez akcelerometr maleje (w stanie spoczynku wartość zmierzona przez akcelerometr wynosi 1 [g], a w trakcie spadku swobodnego 0 [g], zob. rozdział 2.1.2). Zatem metody te klasyfikują akcję jako upadek, gdy wartość $SV_{total} < LFT$. Ilustracji tych metod dostarcza rysunek 3.1.

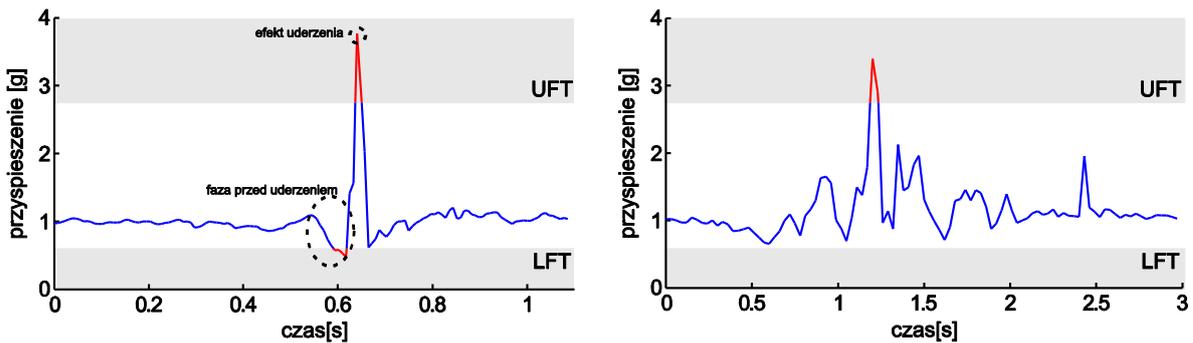

Rysunek 3.1. Dane akcelerometryczne zarejestrowane podczas upadku. Po lewej: modelowy zbiór wartości powstały w wyniku filtracji danych zarejestrowanych. Po prawej: rzeczywiste dane, pochodzące z bazy UR Fall Detection Dataset. LFT = 0.6 [g], UFT = 2.75 [g] wg. pracy (Bourke et al., 2007)

Przedstawione na rysunku dane modelowe powstały w wyniku filtracji danych zarejestrowanych podczas upadku, celem zobrazowania elementów wspólnych dla większości upadków. Charakterystyczna faza poprzedzająca uderzenie występuje, gdy osoba wykonuje bezwładny ruch w kierunku podłoża. Metody wykorzystujące próg *LFT* działają w oparciu o analizę wartości przyspieszeń zarejestrowanych w tej fazie upadku. W wyniku badań na potrzeby niniejszej pracy zauważono, że niektóre upadki (np. z krzesła lub takie, podczas których osoba próbuje się w jakiś sposób asekurować) nie charakteryzują się wyraźnym spadkiem przyspieszenia podczas fazy poprzedzającej uderzenie o podłoże, tak jak upadki z pozycji stojącej. Powoduje to spadek skuteczności metody detekcji opartej na progu *LFT*. Przykładem może być przedstawiony na rysunku 3.1. przebieg akcelerometryczny, będący częścią bazy UR Fall Detection Dataset.

W ramach niniejszej pracy dokonano kilkugodzinnej akwizycji danych akcelerometrycznych dla pięciu starszych osób (kobiety i mężczyźni w wieku powyżej 65 lat, z czego jedna osoba w wieku powyżej 80 lat). Celem badań było określenie skuteczności prostych metod opartych o predefiniowany próg oraz poznanie typowych wartości przyspieszenia i prędkości kątowej dla czynności dnia codziennego. Rysunek 3.2. prezentuje przykładowe dane dla trzech czynności: schodzenia po schodach, podnoszenia przedmiotów oraz siadania. Celem porównania danych akcelerometrycznych, przedstawiono także upadek symulowany przez młodą osobę w wieku 25 lat. Jak można zauważyć, wartości przyspieszenia powyżej definiowanego w literaturze progu *UFT* występują nie tylko dla upadku, ale też dla czynności takich jak siadanie, chodzenie po schodach, podnoszenie przedmiotów i podobnych. Co więcej, niektóre czynności wykonywane energicznie (np. siadanie) charakteryzują się podobnym przebiegiem wartości przyspieszenia jak upadek, w szczególności dla obu tych akcji występuje faza przed uderzeniem ciała o pewną bryłę sztywną. Wartości te zależą od wielu czynników, często niezależnych od użytkownika, jak wysokość krzesła czy fotela, twardość podłoża itp.

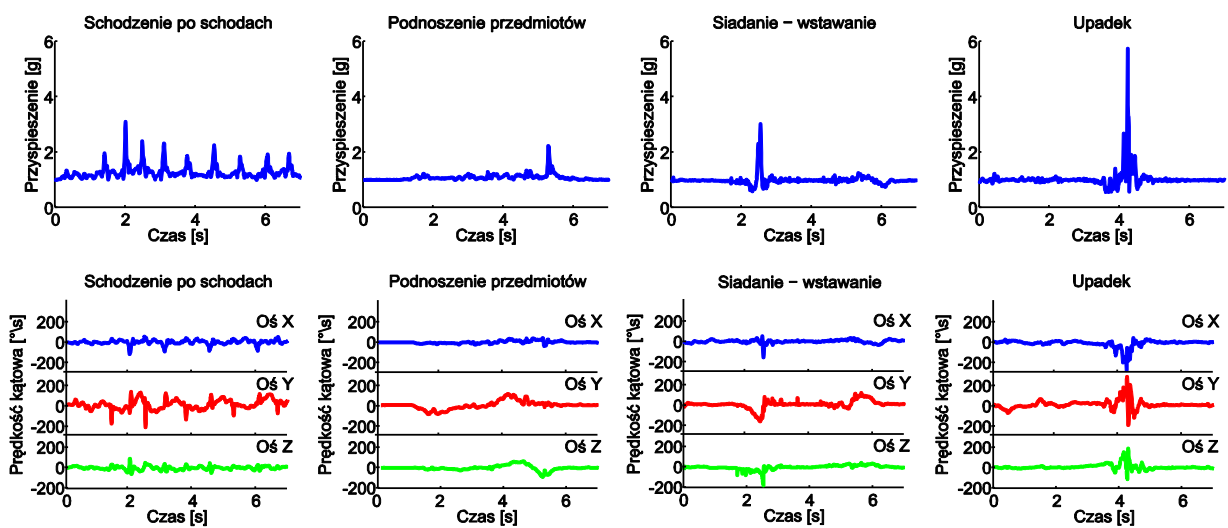

Rysunek 3.2. Przykładowe dane przyspieszenia i prędkości kątowej dla czynności życia codziennego i upadku.

Analizując zgromadzone w dłuższym horyzoncie czasowym dane, można zauważyć, że duża liczba wartości przekracza progi *UFT* i *LFT*. Maksymalna liczba fałszywych alarmów uzyskana dla godzinnego okna czasowego wynosi odpowiednio 15 oraz 12. Przykładowe dane dla ponad godzinnego eksperymentu przedstawiono na rysunku 3.3. Rezultaty przeprowadzonych badań potwierdzają przedstawioną w pracy (Bagalà et al., 2012) tezę o niskiej specyficzności metod detekcji upadku działających w oparciu o próg wartości wektora przyspieszenia, określanego na podstawie zależności (3.1).

Część prac badawczych wykorzystuje założenie, że upadek kończy się w większości przypadków w pozycji leżącej (lub rzadziej siedzącej) i dokonuje detekcji w oparciu o orientację ciała człowieka (ang. *posture*). Niektórzy autorzy (Karantonis et al., 2006) dokonują wydzielenia statycznego komponentu grawitacyjnego z danych przyspieszenia lub wykorzystują żyroskop do określenia orientacji (Hwang et al., 2004).

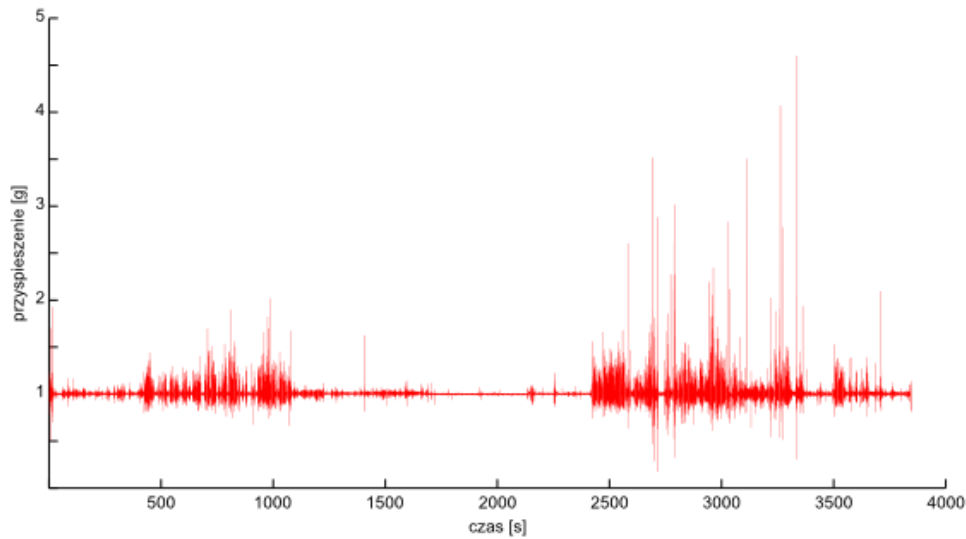

Rysunek 3.3. Dane akcelerometryczne zarejestrowane podczas monitoringu osoby w wieku powyżej 80 lat.
Maksymalna wartość $SV_{total}$ wynosiła 4,96 [g], minimalna 0.17 [g].

Zmiana pozy człowieka związana z upadkiem jest definiowana jako zmiana orientacji ciała przed i po uderzeniu (Boissy et al., 2007) lub jako orientacja ciała przyjmowana kilka sekund po uderzeniu (Karantonis et al., 2006; Tamura, 2005). Podejście to pozwala na likwidację części fałszywych alarmów, ale zdarza się, że zawodzi przy akcjach mających podobną charakterystykę przyspieszenia jak upadki (np. energiczne siadanie). Często stosuje się metody polegające na połączeniu predefiniowanego progu, orientacji i innych cech, takich jak prędkość. Przykładem takiego podejścia może być algorytm VELOCITY + IMPACT + POSTURE przedstawiony w pracy (Bourke et al., 2010). Metoda ta osiąga najlepszą czułość i specyficzność detekcji upadku spośród algorytmów przebadanych w pracy (Bagalà et al., 2012), odpowiednio 83 i 97 procent. Wadą takiego podejścia jest, jak zauważyli autorzy, pominięcie znacznej liczby upadków kończących się na miękkich amortyzujących powierzchniach czy upadków z krzesła. Warto podkreślić, że tak naprawdę opisywana w pracach orientacja nie odnosi się stricte do ciała osoby, a jedynie położenia sensora inercyjnego. Ten zaś może w niekontrolowanym środowisku poza laboratorium ulec przesunięciu, w wyniku niepoprawnego czy też niedokładnego przymocowania do ciała lub ubioru. Dlatego też w niniejszej pracy jako jedno z założeń przyjęto wykorzystanie inercyjnych sensorów wraz z kamerą *Kinect*.

Na podstawie własnych badań oraz przeglądu literatury dokonano oceny zalet i wad technologii detekcji upadku w oparciu o czujniki inercyjne. Do zalet niewątpliwie można zaliczyć:

- łatwiejszą niż w przypadku systemów wizyjnych akwizycję danych, ze względu na rozmiar danych i urządzenia,
- możliwość segmentacji w czasie akcji wykonywanych przez człowieka (dokładnie wiemy kiedy wystąpiło uderzenie osoby o podłoże podczas upadku, co nie jest tak oczywiste w przypadku systemów wizyjnych),

- możliwość wykrywania okresów nieaktywności użytkownika, co szczególnie wpływa na wykorzystanie mocy obliczeniowej w systemach wbudowanych działających w systemie 24/7/365.

Metody oparte o sensory inercyjne nie są pozbawione wad, do których można zaliczyć:

- niesatysfakcjonującą czułość i specyficzność metod detekcji upadku,
- brak możliwości monitorowania użytkownika w niektórych okresach jego aktywności (np. gdy się przebiera, lub kładzie spać).

## 3.2. Założenia i zasada działania systemu

Wraz z rozwojem badań nad metodami detekcji upadku ukształtowały się wymagania stawiane tej technologii (Igual et al., 2013; Yu, 2008). W pracy (Igual et al., 2013) i pokrewnych określono największe wyzwania i potencjalne ograniczenia tej technologii. Według autorów do skutecznego wdrożenia metod detekcji upadku potrzebna jest:

- Wysoka skuteczność detekcji upadków w środowisku użytkowników z grupy docelowej – detekcja powinna charakteryzować się wysoką specyficznością i czułością. Mimo osiągnięcia wyników zbliżonych do pożądanych w warunkach laboratoryjnych, dokładność algorytmów w scenariuszach życia codziennego (ang. *performance under real-life conditions*) pozostaje kwestią otwartą, czego dowodzą prace (Bagalà et al., 2012; Igual et al., 2013).
- Użyteczność technologii – łatwość w obsłudze i maksymalna automatyzacja procesu inicjalizacji działania oraz minimalizacja czasu potrzebnego na obsługę systemu przez użytkownika są ważnym celem w procesie projektowania systemów telemedycznych. Należy uwzględniać potrzeby potencjalnych użytkowników, oraz ocenić ich potencjalną wiedzę na temat obsługi urządzeń elektronicznych. Opracowywaną technologię, mimo swojego potencjalnego wyrafinowania pod względem stosowanych metod, powinna charakteryzować prostota obsługi.
- Akceptacja użytkowników – jest powiązana ze skutecznością i użytecznością technologii (Kurniawan, 2008). Zbyt wysoka liczba fałszywych alarmów, bądź skomplikowana obsługa mogą być przeszkodą w skutecznym zastosowaniu metod w praktyce, ze względu na opór osób starszych. Ponadto zagadnienie prywatności jest kluczowe, gdyż osoby mogą czuć się monitorowane, co z kolei może mieć negatywny wpływ na ich jakość życia.

Mając na względzie powyższe wymagania, zalety i wady systemów noszonych przez użytkownika oraz systemów wizyjnych, a także potencjał aplikacyjny przygotowywanego rozwiązania, opracowano założenia systemu detekcji upadku. Do najważniejszych z nich należą:

- system działa w oparciu o kamerę *Kinect* i inercyjny czujnik ruchu,
- celem zachowania prywatności użytkownika analizie poddane są tylko mapy głębi,

- możliwe jest dowolne umieszczenie kamery (na wprost lub na suficie pomieszczenia), w każdym z przypadków kamera może być statyczna lub zamontowana na głowicy aktywnej,
- system działa na platformie deweloperskiej dla urządzeń mobilnych *PandaBoard ES* z procesorem o architekturze ARM,
- możliwe jest działanie w oparciu o dane z obu urządzeń, bądź jednego z nich.

Wybór kamery głębi, jako źródła strumienia wizyjnego, umotywowany był kilkoma czynnikami. Przede wszystkim, celem maksymalizacji czasu podczas którego użytkownik jest monitorowany wykluczono stosowanie kamer RGB, gdyż takowe mogą działać tylko w obecności światła dziennego lub sztucznego o odpowiednim natężeniu. Jak wykazano eksperymentalnie w rozdziale 2., kamera głębi z powodzeniem może znaleźć zastosowanie w pomieszczeniach o dowolnym natężeniu światła lub jego całkowitego braku. Prywatność osoby, a więc kolejna z cech dobrego systemu detekcji upadku, jest bardziej respektowana, gdy analizie poddawane są jedynie mapy głębi. W razie wyposażenia systemu o moduł zdalnej weryfikacji zdarzenia, przykładowo na podstawie krótkiego filmu wideo, zdalny operator będący opiekunem czy członkiem rodziny użytkownika będzie w stanie stwierdzić upadek jedynie na podstawie danych o położeniu sylwetki w przestrzeni, co nie będzie rażącym naruszeniem prywatności.

Dowolność w umiejscowieniu kamery zwiększa potencjał aplikacyjny technologii, gdyż pozwala na większą elastyczność przy instalacji systemu w otoczeniu użytkownika. Dzięki temu system będzie nadawał się do wykorzystania w większej liczbie pomieszczeń o różnej wysokości, kształcie i powierzchni. Związana jest z tym jednak konieczność opracowania metod uniwersalnych, bądź dopasowanych stricte do danego ustawienia. Ze względu na troskę o wysoką skuteczność detekcji opracowano osobne metody detekcji dla kamer położonych "na wprost" i "u góry". Przez kamerę umieszczoną "na wprost" rozumiemy urządzenie położone w pobliżu jednej ze ścian i odpowiednio nachylone, w zależności od wysokości pokoju i położenia. Kamera "u góry" oznacza urządzenie zamontowane na suficie, którego oś optyczna jest prostopadła do podłogi. Ze względu na konieczność zastosowania systemu w pomieszczeniach o większej powierzchni, lub takich których proporcje są nietypowe (jedna ściana wyraźnie dłuższa) oraz ograniczenia w kącie widzenia kamery, opracowano rozwiązania programowe i sprzętowe pozwalające na montaż kamery na aktywnej głowicy *pan/tilt*. Większa elastyczność położenia kamery może pozwolić na uniknięcie potrzeby większej ingerencji podczas instalacji systemu w środowisku osoby starszej (konieczność przeprowadzenia przewodów zasilających i sygnałowych), co z kolei może mieć pozytywny wpływ na akceptowalność technologii przez użytkowników.

Wysoka skuteczność systemu wiąże się nie tylko z opracowaniem odpowiednio dokładnych metod detekcji, ale też z takim zaprojektowaniem systemu, aby pozwalał na możliwie ciągły monitoring osoby (Mathie et al., 2004). Zapewnienie nieprzerwanego monitoringu osoby w oparciu tylko o czujnik ruchu jest nierealne, gdyż osoba musi się przebierać, kąpać, itp. Urządzenie takie po pewnym czasie użytkowania wymagało będzie naładowania baterii. Z kolei zastosowanie samej kamery ogranicza możliwość monitoringu do obszarów będących w jej polu widzenia.

W dostępnych w literaturze opisach systemów detekcji upadku, kwestia platformy sprzętowej jest najczęściej pomijana (Cucchiara et al., 2007; Zhang et al., 2012) lub sprowadza się do informacji, że system został uruchomiony na komputerze klasy PC (Miaou et al., 2006). Zastosowanie typowego komputera w systemie detekcji upadku zainstalowanym w środowisku użytkownika może być źródłem niedogodności: zwiększonego zużycia energii elektrycznej, ciągłego hałasu pochodzącego od systemu chłodzenia procesora, itp. Aby zwiększyć potencjał aplikacyjny takiego systemu, zdecydowano, że platformą obliczeniową będzie jednopłytowy komputer z procesorem w architekturze ARM. Docelowe zastosowanie takiego urządzenia, jakim są systemy wbudowane (ang. *embedded systems*) niesie za sobą szereg korzyści. Platforma charakteryzująca się niewielkimi rozmiarami, małym zużyciem energii oraz niebędąca źródłem hałasu (dzięki zastosowaniu pasywnych układów chłodzących) powinna korzystnie wpłynąć na użyteczność systemu i akceptowalność użytkowników. Ponadto, niewielkie rozmiary i zużycie energii pozwalają na większą skalowalność systemu, która polegałaby na budowie rozproszonego układu z wieloma urządzeniami śledzącymi i dokonującymi detekcji akcji osoby w dużych budynkach (np. szpitalach czy domach opieki społecznej).

Architektura systemu detekcji upadku została przedstawiona na rysunku 3.4. Na poziomie konceptualnym można ją przedstawić jako 4 warstwy: akwizycji, przetwarzania, komunikacji i klasyfikacji. Taki podział wprowadzono celem przedstawienia uproszczonego modelu systemu i zaprezentowanie etapów jego działania.

**Warstwa akwizycji**

Warstwa akwizycji danych prezentuje sposób pozyskania danych wejściowych z urządzenia *Kinect* oraz inercyjnego czujnika ruchu. Mapy głębi są pobierane z wykorzystaniem protokołu USB, natomiast dane akcelerometryczne pobierane są bezprzewodowo z akcelerometru przy wykorzystaniu protokołu Bluetooth. Do akwizycji danych wykorzystano bibliotekę OpenNI, skompilowaną dla architektury ARM na podstawie kodów źródłowych, natomiast dla czujnika IMU przygotowano oprogramowanie w oparciu o kody źródłowe dostarczone przez producenta.

**Warstwa przetwarzania**

Przed wydzieleniem postaci i jej cech, mapy głębi zostają poddane etapowi wstępnego przetwarzania (ang. *preprocessing*) by usunąć ich niedoskonałości, za które można uważać szum i obszary o nieznanej wartości pikseli. Z obserwacji przedstawionych w rozdziale 2. wynika, że mapy głębi charakteryzują się dość wiarygodnymi pomiarami dla płaskich powierzchni umiarkowanie obijających światło oraz obszarami pikseli *nmd* na krawędziach przedmiotów, refleksyjnych powierzchniach i cieniach obiektów. Ponadto występują błędy pomiaru, które charakteryzują się określeniem wartości głębi, odbiegającej jednak od wartości rzeczywistych. Częściowe lub całkowite usunięcie takiego szumu może być dokonane metodą filtracji medianowej, dwuliniowej (ang. *bilateral filter*) (Tomasi & Manduchi, 1998) lub technikami wykorzystującymi metody probabilistyczne (Smolka et al., 1999). W tym celu zastosowano filtr medianowy o rozmiarze 5 na 5 pikseli. Nieciągłość w dziedzinie głębokości, wynikającą z obszarów *nmd*, nie zostaje usunięta pomimo zastosowania filtracji. Istniejące metody filtracji wykorzystują pary obrazów RGB-D (J. Yang et al., 2012; Chen & Shi, 2013; Richardt et al., 2012) do usuwania tych obszarów z map głębi. Wykorzystują one informacje o kolorze obiektów, zakładając że obszary o podobnym kolorze z dużym prawdopodobieństwem mają zbliżone wartości głębokości. Proponowane podejście zakłada wykorzystanie jedynie informacji o głębi, więc metody te w systemie detekcji upadku nie znajdują zastosowania. Usunięcie obszarów *nmd* jest dokonywane metodą filtracji medianowej w czasie, przy wykorzystaniu 3 ostatnich obrazów.

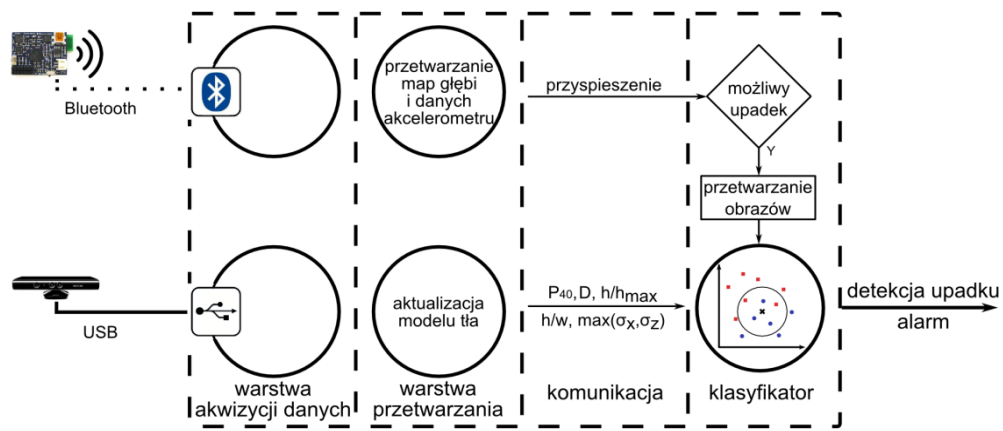

Rysunek 3.4. Schemat architektury systemu detekcji upadku

Inne metody, które można zaliczyć do warstwy przetwarzania także dotyczą przetwarzania map głębi i są to:

- budowa i aktualizacja modelu tła,
- wydzielenie pierwszego planu,
- detekcja postaci,
- śledzenie postaci,
- wydzielenie cech postaci.

Metody te zaprezentowane zostały w rozdziale 2., za wyjątkiem wydzielania cech postaci, które omówiono bardziej szczegółowo w rozdziale 3.3.

## 3.3. Algorytm detekcji upadku

Zgodnie z założeniami przyjętymi podczas projektowania systemu, opracowano algorytmy detekcji upadku dla dwóch różnych ustawień kamery: "na wprost" i "u góry". Pomimo cech wspólnych, obydwa algorytmy dostosowano do specyfiki ustawienia kamery oraz danych wejściowych. Obraz sylwetki człowieka, czy też dowolnego przedmiotu o złożonym kształcie będzie zmieniał się w zależności od ustawienia kamery na scenie. W niniejszym podrozdziale przedstawiono i omówiono obydwa algorytmy oraz cechy postaci wydzielane na potrzeby klasyfikacji. Do prezentacji algorytmów wykorzystano diagramy aktywności UML.

### 3.3.1. Algorytm dla kamery umieszczonej "na wprost"

Diagram aktywności UML algorytmu detekcji upadku dla kamery umieszczonej "na wprost" przedstawiono na rysunku 3.5. Pierwszym etapem jest akwizycja danych z kamery głębi i bezprzewodowego czujnika ruchu. Mapy głębi zostają poddane operacjom wstępnego przetwarzania (filtracji) i zostają zapisane do bufora cyklicznego. Przechowywanie map w buforze jest niezbędne do wykonania operacji budowy i aktualizacji modelu tła. W następnym kroku algorytmu wykonywana jest detekcja ruchu w oparciu o akcelerometr, co pozwala na dokonanie tej operacji niskim kosztem obliczeniowym. Gdy osoba pozostaje w bezruchu przez pewien zadany okres czasu, system przestaje dokonywać wydzielania postaci na obrazach i zapisując nowe dane do bufora czeka na ruch postaci. Pozwala to uniknąć przetwarzania każdej klatki obrazu dzięki wykryciu okresów, w których monitorowana osoba jest nieaktywna (przykładowo śpi lub odpoczywa). W przypadku braku danych akcelerometrycznych (osoba nie nosi akcelerometru lub utracono połączenie z urządzeniem) konieczne jest wydzielenie postaci.

Jeżeli osoba znajduje się w ruchu, zostaje wykonana operacja wydzielenia pierwszego planu i etykietowania połączonych komponentów. Przypadek pojawienia się więcej niż jednego połączonego komponentu na obrazie różnicy aktualnej mapy głębi i modelu tła traktowany jest jako zmiana elementów sceny (np. poprzez interakcję użytkownika z otoczeniem). System rozpoczyna procedurę aktualizacji modelu tła. W sytuacji braku zmian otoczenia użytkownika realizowana jest procedura detekcji upadku, to znaczy w sytuacji gdy występuje podejrzenie o jego wystąpieniu. Wstępna hipoteza o upadku jest uzyskiwana w wyniku przekroczenia wartości przyspieszenia uzyskanego z czujnika ruchu. Potwierdzenie lub odrzucenie tej hipotezy jest realizowane poprzez wydzielenie cech sylwetki postaci na obrazach głębi oraz poddaniu ich klasyfikacji. Gdy wykryto upadek, system zgłasza alarm, w przeciwnym razie kontynuowane jest działanie algorytmu dla nowych danych. W przypadku gdy system działa na podstawie tylko obrazów, konieczna jest detekcja upadku dla każdego obrazu, gdyż nie ma możliwości określenia wstępnej hipotezy w oparciu o przyspieszenie.

Detekcję upadku można rozpatrywać jako problem klasyfikacji dwuklasowej. Przygotowanie odpowiedniego klasyfikatora wymaga opracowania zbioru cech postaci wydzielanych z map głębi oraz wyboru cech istotnych z punktu widzenia klasyfikacji. Wybór cech jest kluczowym problemem w dziedzinie widzenia i uczenia maszynowego i w dużym stopniu wpływa na skuteczność klasyfikacji oraz wynik działania systemu wizyjnego (Dollar et al., 2007). Cecha o dużej sile dyskryminacyjnej powinna:

- nieść dużą ilość informacji (ang. *informative*),
- być niewrażliwa na szum lub określony zbiór transformacji,

- wyznaczanie powinno być zrealizowane niskim kosztem obliczeniowym,
- charakteryzować się niewielką wariancją wewnątrz klasy decyzyjnej oraz dużą wariancją pomiędzy klasami.

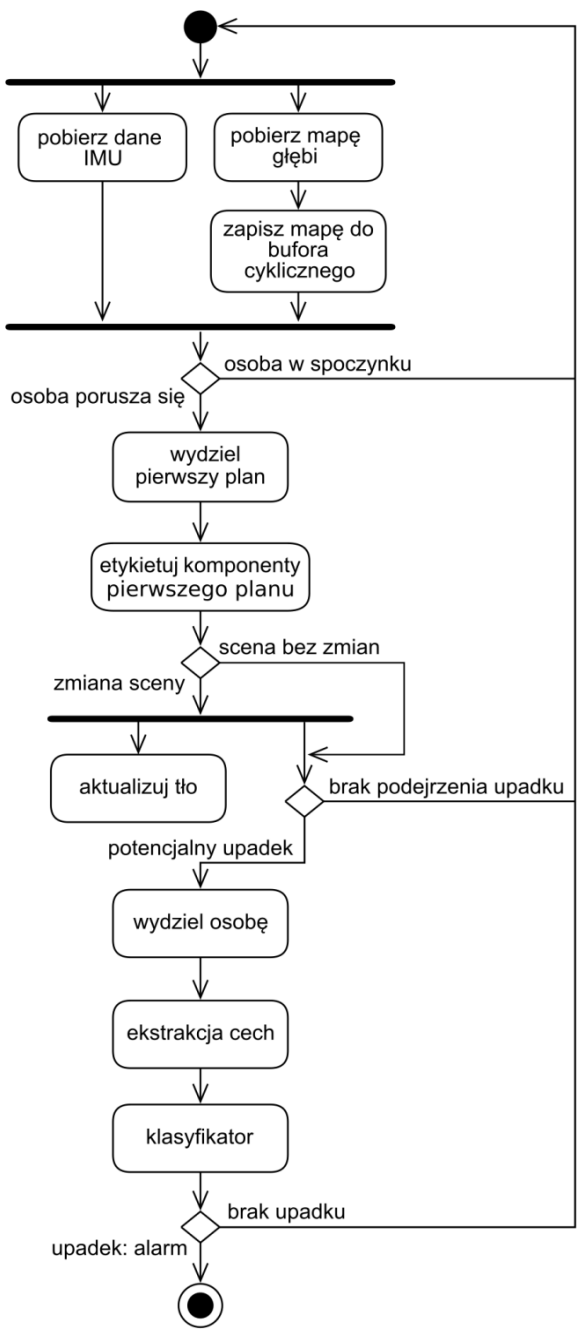

Rysunek 3.5. Diagram aktywności UML ilustrujący algorytm detekcji upadku dla kamery umieszczonej na wprost.

W kontekście systemu detekcji upadku opartego o obrazy głębi można wyróżnić także inne pożądane cechy:

- odporność na szum charakterystyczny dla sensora głębi,

- odporność na przesłonięcia sylwetki czy złączenia sylwetki z elementami sceny,
- czas wyznaczania pozwalający na działanie systemu w czasie rzeczywistym.

Do opracowania cech wykorzystano zbiór danych URFD oraz sekwencje zarejestrowane na potrzeby prac (Kępski & Kwolek, 2013; Kwolek & Kępski, 2013). Sekwencje te prezentują sylwetki osób w różnych pozach, występujących w czynnościach życia codziennego, oraz osoby leżące na podłodze i symulujące pozycję po upadku. Na sekwencjach zarejestrowano 33 młode osoby (studentów Uniwersytetu Rzeszowskiego). Ze zbioru reprezentatywnych obrazów głębi wydzielono cechy opisujące sylwetkę człowieka w danej chwili. Ostatecznie ze zbioru cech wybrano następujące deskryptory:

- $H/W$ – stosunek wysokości do szerokości wydzielonej postaci, wyznaczony na podstawie map głębi (rysunek 3.6),
- $H/H_{max}$ – stosunek wysokości wydzielonej postaci w danej klatce obrazu do jej rzeczywistej wysokości w postawie wyprostowanej, wyznaczony na podstawie chmury punktów,
- $D$ – odległość środka ciężkości postaci od płaszczyzny podłogi wyrażona w milimetrach,
- $max(\sigma_x, \sigma_z)$ – maksymalne odchylenie standardowe wartości punktów należących do postaci od jej środka ciężkości, wzdłuż osi X i Z układu współrzędnych kamery *Kinect*,
- $P_{40}$ – stosunek liczby punktów należących do postaci, leżących w prostopadłościanie o wysokości 40 cm, umieszczonym nad podłogą, do liczby wszystkich punktów należących do postaci (rysunek 3.7).

Oprócz wymienionych wyżej deskryptorów zostały wydzielone inne cechy, m.in.: współczynniki wypełnienia poszczególnych segmentów obrysu sylwetki postaci, osie duże i małe obszaru reprezentującego postać na obrazie, współczynniki kształtu. Jakość cech była weryfikowana na podstawie miary przyrostu informacji (ang. *information gain*) (Cover & Thomas, 1991). Do ewaluacji wykorzystano algorytm InfoGainAttributeEval pochodzący z biblioteki Weka (Hall et al., 2009).

Cecha $H/W$ jest często spotykana w literaturze w kontekście detekcji upadku (Mastorakis & Makris, 2012; Anderson et al., 2006; Liu et al., 2010). Gdy osoba stoi, stosunek wysokości do szerokości prostokąta jest duży, zazwyczaj większy od 1. Gdy osoba znajduje się w pozycji leżącej, wartość ta jest znacznie mniejsza. Jednak detekcja upadku jedynie w oparciu o tą cechę charakteryzuje się skutecznością około 80%, jak przedstawiono w pracy (Stone & Skubic, 2014). Cecha ta może być podatna na przysłonięcia i prowadzić do błędów klasyfikacji drugiego rodzaju, gdy część osoby po upadku znajduje się poza polem widzenia kamery. Ponadto, ten sam błąd może wystąpić, gdy osoba zostanie niezbyt precyzyjnie wydzielona, np. w wyniku zmiany położenia obiektów na scenie w momencie upadku.

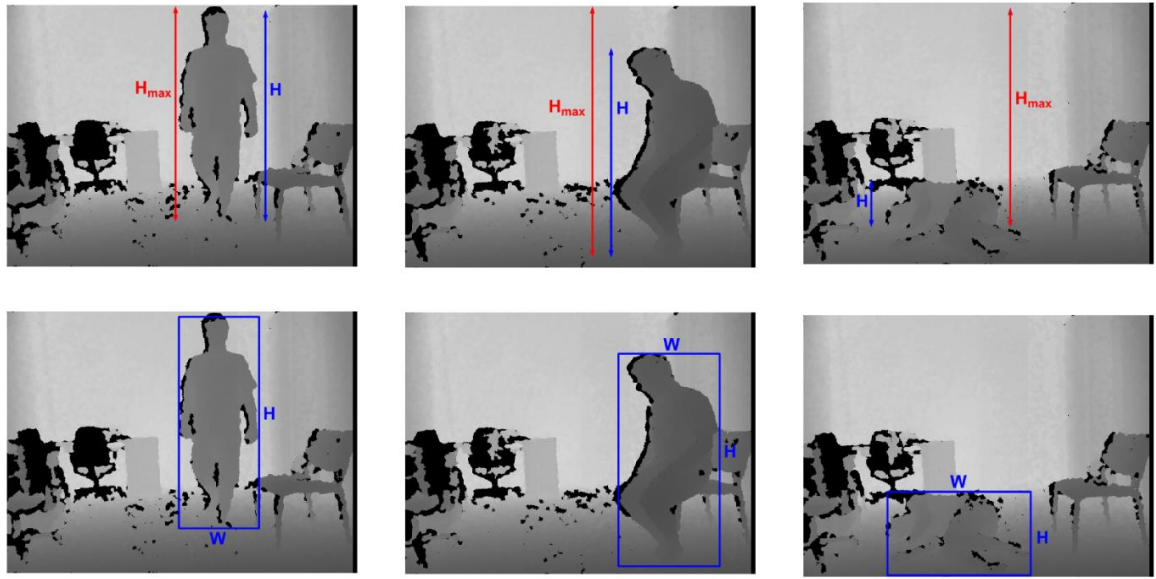

Rysunek 3.6. Cechy $H/W$ oraz $H/H_{max}$ przedstawione dla różnych sylwetek postaci: stojącej, siedzącej i leżącej.

Miary $H/H_{max}$ oraz $D$ uzależnione są przede wszystkim od wysokości sylwetki postaci w danej chwili. Cecha $D$ została już przedstawiona w literaturze (Rougier et al. 2011), jednak jej niewielka skuteczność wynikała z zastosowania jej samodzielnie oraz wyborze prostego sposobu wnioskowania o upadku w oparciu o próg wartości. Cechy $max(\sigma_x, \sigma_z)$ oraz $P_{40}$, nie były prezentowane uprzednio w żadnych pracach naukowych innych zespołów badawczych. Zostały one zaproponowane celem odróżnienia od upadku akcji, w których osoba schyla się po przedmiot, kuca lub wykonuje podobne czynności. Duże wartości odchylenia standardowego punktów należących do postaci względem osi $x$ lub $y$, są charakterystyczne dla leżącej sylwetki. Istotne jest to, że odchylenie standardowe wyznaczane jest na podstawie chmury punktów zamiast mapy głębi. Dzięki temu można uniezależnić wartość omawianego atrybutu od orientacji leżącej osoby względem kamery. Cecha $P_{40}$ ma za zadanie odróżniać sytuację, w których postać znajduje się blisko podłogi, wydzielonej wcześniej algorytmami v-dysparycji i RANSAC (rysunek 3.7).

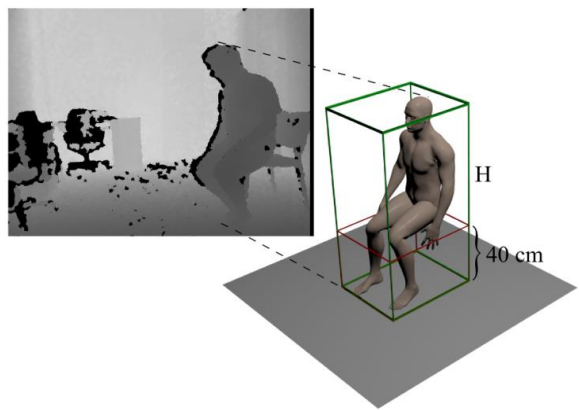

Rysunek 3.7. Cecha $P_{40}$ przedstawiona na mapie głębi i odpowiadającym mu modelu postaci 3D.

## 3.3.2. Algorytm dla kamery umieszczonej "u góry"

Umieszczenie kamery na suficie skierowanej tak, że oś kamery jest w przybliżeniu prostopadła do podłogi, pozwala na zmniejszenie przesłonięć oraz trudności praktycznych ze wspomnianym wcześniej minimalnym zasięgiem urządzenia. Jednak biorąc pod uwagę kąt widzenia kamery oraz typowe wysokości pomieszczeń mieszkalnych lub użytkowych (sale szpitalne, korytarze, domy opieki społecznej) to dla zakresu wysokości pomieszczenia od 2,5 do 3,3 metra wielkość monitorowanego obszaru waha się od 6 m$^2$ do 10,5 m$^2$. Obszar ten można zwiększyć stosując nakładkę *Nyko Zoom*[11]. Urządzenie to składa się z zestawu soczewek o szerokim kącie widzenia (ang. *fisheye lens*) i służy do zwiększenia pola widzenia urządzenia, kosztem zmniejszenia jego zasięgu. Efektem zastosowania tej nakładki jest obraz o dużej dystorsji. W efekcie zastosowania urządzenia *Nyko Zoom* można uzyskać obraz obejmujący większy fragment sceny (rysunek 3.8). Obraz ten charakteryzuje się jednak dużymi skupiskami pikseli *nmd*, obejmującymi swoim zasięgiem nie tylko miejsca typowe dla urządzenia (cienie, powierzchnie refleksyjne), ale też duże fragmenty sceny znajdujące się na obrazie w pobliżu jego rogów. Ze względu na te niedoskonałości, wpływające bezpośrednio na jakość danych wejściowych, zaniechano zastosowania tego typu soczewek dla potrzeb detekcji upadku.

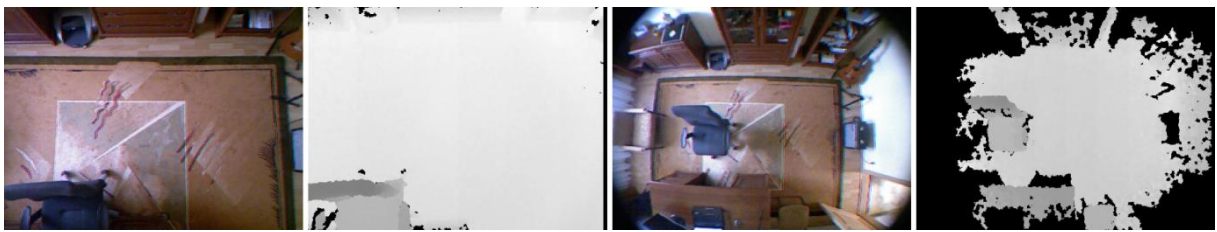

Rysunek 3.8. Porównanie obrazów z kamery *Kinect* bez nakładki *Nyko* (dwa pierwsze obrazy) oraz z nakładką (obraz trzeci i czwarty).

Zbyt mała powierzchnia monitorowania, a także brak możliwości zadawalającego jej zwiększenia stosując nakładki na obiektyw, doprowadziły do umieszczenia kamery na obrotowej głowicy *pan-tilt* (konstrukcja głowicy opisana została w rozdziale 2.). Algorytm działania systemu w konfiguracji z kamerą aktywną powinien uwzględniać rozwiązania następujących zagadnień (nie występujących, czy też mniej znaczących w systemie z kamerą statyczną):

- brak możliwości zastosowania algorytmu wydzielenia pierwszego planu w oparciu o referencyjny model tła,
- potrzeba dokładnego określenia pozycji śledzonej postaci wynikająca z konieczności wyznaczenia parametrów sterowania głowicy.

Brak możliwości użycia referencyjnego modelu do detekcji osoby wynika z tego, że dla każdego położenia kamery istnieje inny obraz tła. Wystąpienie zmiany położenia elementów sceny prowadziłoby do konieczności wyznaczania referencyjnego modelu tła dla każdego położenia kamery. Ponadto, algorytm tworzenia modelu tła wymaga akwizycji kilkuset obrazów oraz zapamiętania w buforze części z nich. Przy dynamicznie poruszającej się osobie po scenie taka operacja byłaby kosztowna i trudna do wykonania. Z tego powodu metoda wydzielania postaci, zastosowana w przypadku kamery statycznej, została zastąpiona algorytmem segmentacji w oparciu o rozrost obszarów. Dodatkowo, obiekt śledzony jest poddany weryfikacji detektorem działającym na podstawie histogramów gradientów skierowanych (HOG) w celu sprawdzenia czy obiekt reprezentuje faktycznie sylwetkę osoby.

---

[11] http://nyko.com/products/product-detail/?name=Zoom

Diagram aktywności UML algorytmu detekcji upadku dla kamery umieszczonej "u góry" przedstawiono na rysunku 3.9. Pierwszym etapem, podobnie jak w ustawieniu kamery "na wprost", jest realizowana akwizycja danych z kamery głębi i bezprzewodowego sensora ruchu oraz filtracja map głębi. Następnie za pomocą algorytmu rozrostu obszarów wydzielana jest osoba. Punkty startowe, od których rozpoczynana jest segmentacja są znane dzięki informacji o położeniu postaci na wcześniejszej klatce obrazu. Po segmentacji postaci i weryfikacji jej poprawności przy użyciu HOG dalsza analiza wydzielanego obszaru jest przeprowadzana w dwojakim celu: wyznaczenie środka postaci w celu obliczenia parametrów sterowania głowicy oraz klasyfikacja pozy leżącej postaci na potrzeby wyznaczenia wstępnej hipotezy o wystąpieniu upadku. W razie braku podejrzenia wystąpienia upadku, algorytm rozpoczyna pracę dla nowych danych, w przeciwnym wypadku konieczna jest dalsza analiza akcji wykonanej przez człowieka. Sam fakt stwierdzenia, że osoba znajduje się w pozycji leżącej jest niewystarczający do detekcji upadku charakteryzującej się wysoką swoistością, gdyż intencją osoby mogło być znalezienie się w pozycji leżącej. Odróżnienie takiego zachowania od upadku jest możliwe dzięki analizie wartości przyspieszeń uzyskanych z akcelerometru oraz analizie zmian sylwetki postaci w czasie. Dla niniejszego systemu zaproponowano cechy opisujące akcje postaci pod względem dynamiki zmian sylwetki, nazwane *Dynamic Transitions* (Kępski & Kwolek, 2014b; Kępski & Kwolek, 2014a). Metoda ta pozwala odróżnić akcje wykonywane dynamicznie, charakteryzujące się szybką zmianą sylwetki (np. upadek) od typowych ADLs. Decyzja po analizie cech dynamicznych prowadzi do alarmu (w razie wystąpienia upadku) lub do kontynuacji pracy systemu dla nowych danych.

Detekcja pozy leżącej postaci została zrealizowana na podstawie zbioru następujących cech:

- $H/H_{max}$ – stosunek wysokości postaci w danej klatce obrazu (odległości głowy od podłogi) do jej rzeczywistej wysokości,
- $area$ – wielkość obszaru reprezentującego postać na obrazie przeskalowana dla stałej odległości,
- $l/w$ – stosunek długości osi wielkiej do długości osi małej obszaru reprezentującego postać na obrazie binarnym $I(x,y)$, powstałym w wyniku wydzielenia osoby na mapie głębi.

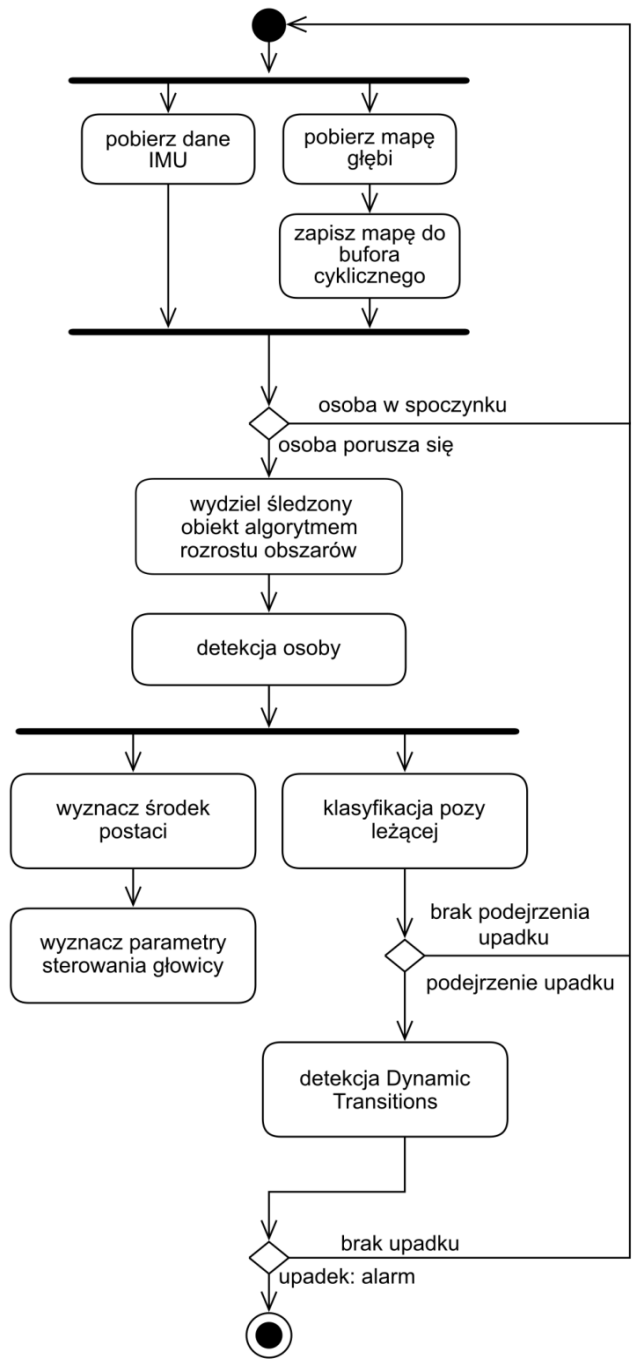

Rysunek 3.9. Diagram aktywności UML ilustrujący algorytm detekcji upadku dla kamery umieszczonej u góry pomieszczenia, wyposażonej w obrotową głowicę aktywną.

Długości osi wielkiej i małej obszaru odpowiadającego postaci zostały wyznaczone w następujący sposób:

$$l = 0.707\sqrt{(a+c) + \sqrt{b^2 + (a-c)^2}} \tag{3.2}$$

$$w = 0.707\sqrt{(a+c) - \sqrt{b^2 + (a-c)^2}} \tag{3.3}$$

gdzie $a, b, c$ to współczynniki, zdefiniowane jako:

$$a = \frac{M_{20}}{M_{00}} - x_c^2 \tag{3.4}$$

$$b = 2\left(\frac{M_{11}}{M_{00}} - x_c y_c\right) \tag{3.5}$$

$$c = \frac{M_{02}}{M_{00}} - y_c^2 \tag{3.6}$$

zaś $M_{00}, M_{02}, M_{11}, M_{20}$ to momenty bezwładności obrazu binarnego $I(x, y)$, zdefiniowane jako:

$$M_{pq} = \sum_x \sum_y x^p y^q I(x, y) \tag{3.7}$$

W celu wyznaczenia wartości cechy *area* niezbędne jest przeskalowanie obszaru komponentu reprezentującego postać. Wynika to z tego, że w momencie akwizycji obrazu dokonywana jest projekcja perspektywiczna, w której obiekty znajdujące się bliżej na obrazie zajmują większy obszar. Dzięki znajomości odległości obiektów od kamery i parametrów kalibracji możliwe jest dokonanie skalowania tak, aby ułatwić porównanie rzeczywistych rozmiarów obiektów znajdujących się w różnych odległościach od kamery.

Dotychczas zaprezentowano w jaki sposób można dokonać detekcji upadku na podstawie klasyfikacji pozy leżącej, analizując pojedynczą mapę głębi. Niemniej jednak upadek jest dynamicznym procesem, który charakteryzuje się krótkim czasem trwania, często w literaturze definiowanym jako mieszczący się w przedziale 0,4 do 0,8 sekundy. Podczas upadku następuje nagła zmiana odległości głowy osoby od podłogi oraz zmiana orientacji ciała na horyzontalną lub zbliżoną do horyzontalnej (gdy upadek kończy się w nietypowej pozycji, np. półsiedzącej). Aby odróżnić upadki od innych akcji, takich jak kładzenie się do łóżka, zaproponowano analizę zmian cech opisujących sylwetkę w czasie (Kępski & Kwolek, 2015). Ponieważ upadki, mogą różnić się między sobą ułożeniem ciała, kierunkiem upadku, prędkością ciała (różnica występuje szczególnie pomiędzy upadkiem z pozycji nieruchomej, a podczas ruchu, np. chodzenia), a także ustawieniem postaci względem kamery, konieczne jest zaproponowanie takich cech, które będą wspólne dla jak największej liczby możliwych wzorców kinematycznych upadku. Jedną z takich cech wspólnych jest gwałtowna zmiana orientacji ciała z pionowej na horyzontalną, połączona z dużymi wartościami przyspieszenia w momencie uderzenia o podłoże, a także zwiększenie obszaru zajmowanego przez postać na mapie głębi.

## 3.4. Wyniki badań eksperymentalnych

### 3.4.1. Wykorzystane klasyfikatory

Z przeglądu literatury wynika, że w pracach dotyczących detekcji upadku najczęściej wykorzystywane są drzewa decyzyjne, algorytm SVM oraz metoda *k*-najbliższych sąsiadów (ang. *k-Nearest Neighbor*, *k*-NN). Stosunkowo nieliczne są prace, w których wykorzystywane są podejścia oparte o logikę rozmytą czy też inne metody. Tym niemniej, potencjał aplikacyjny wielu z nich, a w szczególności metod opartych o *soft-computing,* czy też zespoły klasyfikatorów (Woźniak et al., 2014) może być znaczący. W niniejszej pracy, detekcja upadku realizowana była w oparciu o klasyfikator *k*-NN, SVM oraz wnioskowanie rozmyte. Wyniki badań nad detekcją upadku w oparciu o wnioskowanie rozmyte przedstawiono w rozdziale 4., natomiast w niniejszym rozdziale przedstawiono wyniki uzyskane w oparciu o klasyfikatory, które są najczęściej wykorzystywane w pokrewnych badaniach.

Metoda *k*-najbliższych sąsiadów (Cover & Hart, 1967) jest nieparametryczną metodą wykorzystywaną często w zadaniach klasyfikacji. Metoda ta opiera się na określaniu odległości między próbką testową, a zbiorem próbek uczących. Odległość euklidesowa próbki $x_i$ zawierającej $p$-elementowy wektor cech $(x_{i1}, x_{i2}, \ldots, x_{ip})$ od próbki $x_l$ określona jest następująco:

$$d(x_i, x_l) = \sqrt{(x_{i1} - x_{l1})^2 + (x_{i2} - x_{l2})^2 + \cdots + (x_{ip} - x_{lp})^2} \qquad (3.8)$$

Klasyfikacja dla $k = 1$, przebiega na przypisaniu próbce testowej $x_i$ etykiety klasy jej najbliższego sąsiada. Gdy $k > 1$, próbka testowa otrzymuje etykietę klasy, która najczęściej występuje wśród jej *k*-sąsiadów. Metoda ta należy do tzw. leniwych technik klasyfikacji (ang. *lazy learning*), które nie generują modelu (hipotezy) dla funkcji docelowej, lecz w momencie klasyfikacji próbek testowych odpowiedź klasyfikatora jest uzyskiwana na podstawie próbek trenujących. Do zalet tego typu technik można zaliczyć zdolność do klasyfikacji problemów o zmieniającym się zbiorze treningowym bez konieczności ponownej budowy modelu. Jednak strategia wyznaczania odległości pomiędzy próbkami na etapie klasyfikacji obarczona jest pewnym kosztem obliczeniowym. Najprostsze metody *brute force* polegają na obliczaniu odległości między wszystkimi próbkami, co prowadzi do złożoności obliczeniowej $O[DN^2]$ dla $N$ próbek o liczbie wymiarów $D$. W celu ograniczenia kosztu obliczeniowego algorytmu klasyfikacji podjęto próby implementacji algorytmu *k*-NN w oparciu o różne struktury danych, m. in. *kd*-drzewa czy inne rodzaje drzew binarnych (Bentley, 1975). Wcześniejsza budowa drzewa pozwala na uniknięcie potrzeby wyznaczania wszystkich odległości potrzebnych do wybrania $k$ sąsiadów, co w konsekwencji pozwala na zmniejszenie złożoności obliczeniowej wyszukiwania sąsiadów do $O[DN \log N]$ lub mniejszej (Arya et al., 1998).

Podstawową ideą algorytmu SVM jest znalezienie hiperpłaszczyzny pozwalającej na separowanie danych w możliwie optymalny sposób (Cortes & Vapnik, 1995). Jeśli w zadaniu klasyfikacji binarnej zbiór danych jest liniowo separowalny, wówczas istnieje co najmniej jedna hiperpłaszczyzna:

$$\hat{g}(x) = w^T x + w_0 = 0 \qquad (3.9)$$

pozwalająca na rozgraniczenie wektorów danych należących do różnych klas. Margines hiperpłaszczyzny rozdzielającej można zdefiniować w następujący sposób:

$$m = (d^+ + d^-) \qquad (3.10)$$

gdzie $d^+$ i $d^-$ to odległości między hiperpłaszczyzną rozdzielającą a najbliższym punktem z klasy reprezentującej dodatnie i ujemne przykłady, odpowiednio. Intuicyjnie, za lepsze można uznać te płaszczyzny, które przebiegają możliwie daleko od obiektów obydwu klas, zapewniające jak największy margines separacji. Hiperpłaszczyzna taka zapewnia najlepszą zdolność klasyfikatora do uogólniania i jest nazywana optymalną hiperpłaszczyzną rozdzielającą (ang. *optimal separating hyperplane*, OSH). Klasyfikator liniowy, w którym hiperpłaszczyzną rozdzielającą jest OSH nazywany jest liniowym klasyfikatorem SVM. Problem poszukiwania hiperpłaszczyzny z maksymalnym marginesem, a więc wyznaczenia wektora wag $w$ i stałej $w_0$, jest problemem optymalizacji kwadratowej, dla którego przedstawiono efektywne rozwiązania w pracy (Cristianini & Shawe-Taylor, 1999). Dla problemu nieseparowalnego liniowo znalezienie optymalnej hiperpłaszczyzny polega na wprowadzeniu zmiennych osłabiających $\{\xi_i\}_{i=1}^n$, które są miarą odchylenia danego wektora uczącego od przypadku liniowej separowalności. Wprowadzenie w procesie optymalizacji do funkcji celu sumy składnika, uwzględniającego wartości zmiennych osłabiających oraz parametru $C$, w postaci:

$$c_o = C \left( \sum_{i=1}^n \xi_i \right) \qquad (3.11)$$

pozwala na uwzględnienie kosztu $c_o$ związanego z błędną klasyfikacją. Dzięki dobraniu odpowiedniej wartości parametru $C$ możliwe jest uzyskanie kompromisu pomiędzy liczbą błędnie sklasyfikowanych próbek, a szerokością marginesu separacji. Należy wspomnieć, że oprócz liniowego klasyfikatora SVM istnieją jego nieliniowe odmiany wykorzystujące transformację przestrzeni wejściowej za pomocą nieliniowego przekształcenia (tzw. funkcji jądra).

### Wyniki detekcji upadku dla kamery umieszczonej "na wprost"

Badania zrealizowano na sekwencjach obrazów głębi ze zbioru danych UR Fall Dataset (zob. rozdz. 1.4). Jak już wspomniano, przygotowane sekwencje zawierają upadki oraz typowe czynności dnia codziennego, które wykonane były przez 6 osób, zarówno w środowisku laboratoryjnym jak i domowym. W tabeli 3.1 zestawiono wykonane czynności z podziałem na upadki i czynności dnia codziennego. Zrealizowano upadki rozpoczynające się zarówno od pozycji stojącej jak i siedzącej. Warto podkreślić, że w większości prac dotyczących detekcji upadku rozpatruje się jedynie upadki z pozycji stojącej. Tym niemniej upadki z pozycji siedzącej mogą stanowić duży odsetek upadków, w szczególności w środowisku domowym. Ogólna liczba sekwencji wynosi 70, spośród których 30 prezentuje upadki.

Tabela 3.1. Zestawienie sekwencji wykorzystanych przy badaniach skuteczności opracowanych algorytmów.

|  | typ akcji | liczba akcji |
|---|---|---|
| upadek | z pozycji stojącej | 15 |
| upadek | z pozycji siedzącej | 15 |
| ADLs | siadanie | 10 |
| ADLs | leżenie | 10 |
| ADLs | kucanie | 10 |
| ADLs | schylanie | 10 |

Z przeglądu literatury wynika (Igual et al., 2013), że wiele systemów do detekcji upadku wykorzystuje jedynie informację o pozie leżącej. Celem przebadania użyteczności tego podejścia, w oparciu o przygotowany zestaw deskryptorów dokonano klasyfikacji pozy osoby. Czułość, swoistość, dokładność i precyzję klasyfikatora określono w oparciu o zestaw cech omówiony w podrozdziale 3.3.1. Wspomniane parametry określono w oparciu o następujące cechy: $H/W$, $H/H_{max}$, $D$, $max(\sigma_x, \sigma_z)$, $P_{40}$. W tabeli 3.2 zebrano uzyskane wyniki klasyfikacji pozy osoby. Jak zaobserwować można, osiągnięto wysoką czułość i swoistość systemu, jednak wyniki klasyfikacji nie są pozbawione błędów pierwszego i drugiego rodzaju. Co więcej, w praktycznym zastosowaniu, wspomniane błędy przełożyłyby się na wystąpienie fałszywych alarmów oraz pominięcie niektórych upadków.

Tabela 3.2. Macierz pomyłek dla klasyfikacji pozy osoby.

|  |  |  | Rzeczywiste pozy osoby | | |
|---|---|---|---|---|---|
|  |  |  | osoba w pozie leżącej | osoba w pozie nieleżącej | |
| Przewidywane pozy osoby | k-NN(3) | osoba w pozie leżącej | 898 | 6 | Dokładność = 99,55% |
| Przewidywane pozy osoby | k-NN(3) | osoba w pozie nieleżącej | 5 | 1516 | Dokładność = 99,55% |
| Przewidywane pozy osoby | k-NN(3) |  | Czułość = 99,45% | Swoistość = 99,61% | Precyzja = 99,34% |

Skuteczność detekcji upadku dla algorytmu zaprezentowanego w podrozdziale 3.3.1 przebadano na wspomnianych sekwencjach danych z bazy URFD. W algorytmie zaprezentowanym na rysunku 3.5, w bloku odpowiedzialnym za klasyfikację (zob. blok "klasyfikator") wykorzystano klasyfikatory k-NN oraz SVM. Celem oceny wskaźników jakościowych zrealizowano badania eksperymentalne, których wyniki zestawiono w tabeli 3.3. W omawianej tabeli zamieszczono także wyniki uzyskane przez reprezentatywne i powszechnie przywoływane w literaturze metody. Jak można zauważyć, wykorzystanie danych pochodzących z noszonego przez osobę inercyjnego sensora umożliwia znaczne polepszenie jakości detekcji.

Tabela 3.3. Wyniki uzyskane w oparciu o opracowane metody detekcji upadku na zbiorze danych UR Fall Dataset dla kamery umieszczonej na wprost.

|  |  | *k*-NN + akcelerometr | SVM + akcelerometr | *k*-NN | UFT (Bourke et al., 2007) | LFT (Bourke et al., 2007) |
|---|---|---|---|---|---|---|
| **Wyniki** | **Dokładność** | 95,71% | 94,28% | 90,00% | 88,57% | 78,57% |
|  | **Precyzja** | 90,90% | 88,24% | 81,08% | 78,95% | 68,29% |
|  | **Czułość** | 100,00% | 100,00% | 100% | 100,00% | 93,33% |
|  | **Swoistość** | 92,50% | 90,00% | 82,5% | 80,00% | 67,50% |

Mając na względzie brak akceptacji istniejących rozwiązań przez seniorów, głównie ze względu na liczbę fałszywych alarmów przy wysokiej czułości urządzenia, które występują podczas ciągłego monitoringu osoby, można stwierdzić, że jednoczesne wykorzystanie danych wizyjnych i akcelerometru jest racjonalne. W szczególności, dzięki użyciu deskryptorów opisujących ruch osoby i obrazów głębi możliwe jest zmniejszenie liczby fałszywych alarmów w porównaniu do systemów operujących na sekwencjach obrazów lub pomiarach z akcelerometru czy żyroskopu. Jak można zaobserwować, metody działające jedynie w oparciu o akcelerometr charakteryzują się dużą liczbą błędów pierwszego rodzaju, co przekłada się na niższą swoistość wspomnianych rozwiązań. Z kolei detekcja upadku jedynie w oparciu o klasyfikację pozy w jakiej znajduje się osoba (zob. wyniki uzyskiwane przez *k*-NN), prowadzi do niezadowalającej swoistości i precyzji metody, mając na względzie praktyczne zastosowania w systemach nieprzerwanego monitoringu osoby.

Jedną z przyczyn dla których klasyfikator oparty jedynie o cechy obrazów nie uzyskuje wyższych wskaźników jakości detekcji upadku jest to, że cechy, które są powszechnie wykorzystywane w systemach do detekcji upadku charakteryzują się dużą wariancją wewnątrzklasową przy małej wariancji międzyklasowej. Jak można zaobserwować na rysunku 3.10, na którym zilustrowano grupowanie się cech dla obrazów wykorzystywanych w badaniach nad klasyfikacją pozy, zob. tabela 3.2, cechy zaproponowane w niniejszej pracy mają znaczącą siłę dyskryminującą.

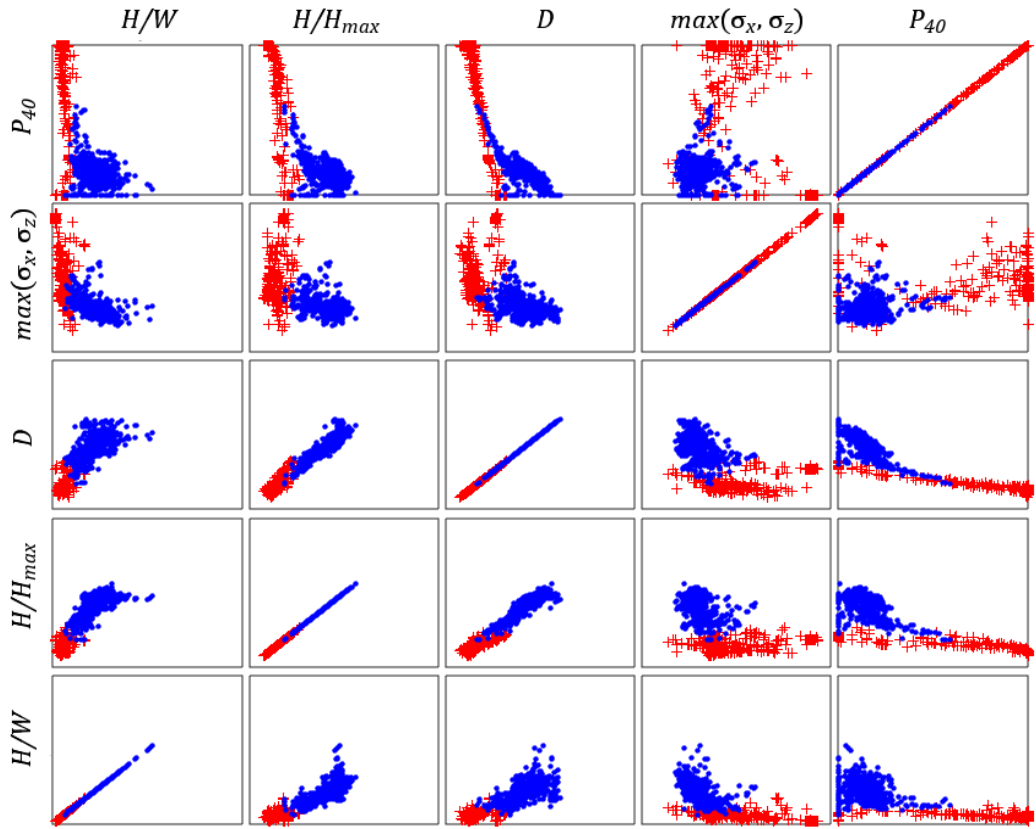

Rysunek 3.10. Ilustracja grupowania się cech dla obrazów wykorzystywanych w badaniach nad klasyfikacją pozy.

### 3.4.2. Wyniki detekcji upadku dla kamery umieszczonej "u góry"

Z przeglądu literatury dotyczącej detekcji upadku wynika, że większość zespołów badawczych nie rozpatruje scenariuszy, w których kamera umieszczona jest u góry pomieszczenia. Tym niemniej, w typowych systemach monitoringu w oparciu o kamery RGB, dość powszechnie wykorzystuje się kamery skierowane w dół i umieszczone na obrotowych głowicach. Wyniki zamieszczone w niniejszym podrozdziale dotyczą rozwiązania zaprezentowanego w podrozdziale 3.3.2, zakładającego umiejscowienie kamery na suficie pomieszczenia. Liczba obrazów, na których realizowano badania eksperymentalne jest równa 525. Na wymieniony zbiór składa się 248 obrazów przedstawiających osoby w pozie leżącej oraz 277 obrazów przedstawiających osoby w pozostałych pozach. Jak można zaobserwować w tabeli 3.4, w oparciu o obrazy zarejestrowane przez kamerę umieszczoną u góry uzyskać można wyniki, które nie ustępują wynikom zaprezentowanym w tabeli 3.2. Omawiane wyniki uzyskane zostały w oparciu o zestaw trzech deskryptorów, podczas gdy wyniki przedstawione w tabeli 3.2 uzyskano w oparciu o zestaw pięciu cech. Warto także podkreślić, że w scenariuszu z kamerą umieszczoną u góry, liczba przysłonięć występujących na obrazach jest mniejsza.

Tabela 3.4. Wyniki detekcji pozy leżącej.

|  |  |  | Rzeczywiste pozy osoby | | |
|---|---|---|---|---|---|
|  |  |  | osoba w pozie leżącej | osoba w pozie nieleżącej |  |
| Przewidywane pozy osoby | *k*-NN | osoba w pozie leżącej | 244 | 9 | Dokładność = 97,52% |
|  |  | osoba w pozie nieleżącej | 4 | 268 |  |
|  |  |  | Czułość = 98,39% | Swoistość = 96,75% | Precyzja = 96,44% |
|  | SVM | osoba w pozie leżącej | 244 | 10 | Dokładność = 97,33% |
|  |  | osoba w pozie nieleżącej | 4 | 267 |  |
|  |  |  | Czułość = 98,39% | Swoistość = 96,39% | Precyzja = 96,06% |

Jak już wspomniano w rozdziale 2.3.1, w trakcie wydzielania osoby jedynie w oparciu o rozrost obszarów może dojść do dołączenia do osoby obszarów do niej nienależących. W tabeli 3.5 znajdują się wyniki uzyskane w oparciu o algorytm, który może być wykorzystany do usprawnienia wydzielania osoby w razie zajścia wspomnianej sytuacji (Kępski & Kwolek, 2016).

Tabela 3.5. Wyniki detekcji postaci metodą HOG-SVM.

|  | Dokładność | Precyzja | Czułość | Swoistość |
|---|---|---|---|---|
| rotacja komponentu | 99,45 | 98,21 | 100 | 99,22 |
| brak rotacji komponentu | 98,91 | 98,18 | 98,18 | 99,22 |

Omawiana metoda detekcji postaci została przebadana na zbiorze liczącym 892 obrazy, z których 60% wykorzystano do uczenia klasyfikatora. Przedstawione wyniki uzyskano na zbiorach rozłącznych. Jak można zaobserwować, dzięki transformacji komponentu reprezentującego sylwetkę postaci do orientacji kanonicznej, wyniki detekcji są lepsze mając na względzie czułość i dokładność.

Celem przebadania przydatności zaproponowanych cech *dynamic transitions*, zob. rozdział 3.3.2, zrealizowano eksperyment, w którym określono czułość i specyficzność dla $\Delta t \in (400ms, 800ms)$. W oparciu o uzyskane wyniki sporządzono wykres obrazujący przebieg krzywej ROC, który zaprezentowany jest na rysunku 3.11. Jak można zaobserwować, uzyskane pole powierzchni pod krzywą ROC jest znaczące.

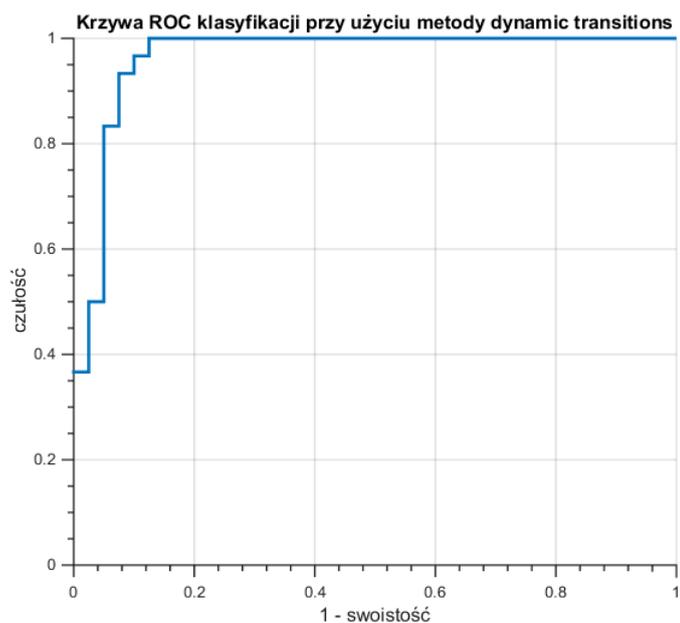

Rysunek 3.11. Krzywa ROC klasyfikacji upadku przy użyciu metody *dynamic transitions*.

## 3.5. Podsumowanie

W rozdziale zaprezentowano wyniki badań eksperymentalnych, które uzyskano w oparciu o zaproponowane metody detekcji upadku. Przebadano scenariusz zakładający ustawienie kamery "na wprost" oraz kamery zamocowanej pod sufitem i skierowanej w dół. Uzyskane wyniki badań pokazały, że jest to obiecujące podejście, pomimo tego, że w literaturze przedmiotu scenariusz z kamerą udostępniającą mapy głębi i skierowaną w dół nie był do tej pory rozpatrywany. Na podstawie zrealizowanych badań empirycznych pokazano, że w oparciu o zaproponowany zestaw cech do detekcji osoby w pozie leżącej uzyskać można wyniki o wysokiej czułości i specyficzności. Uzyskana krzywa ROC dla cech dynamicznych wskazuje na ich znaczącą siłę dyskryminacyjną.

# 4. Detekcja upadku w oparciu o rozmyty układ wnioskujący

Wcześniejsze badania, przedstawione w rozdziale 3, wykazały użyteczność cech opisujących pozę oraz dynamikę ruchu postaci na potrzeby detekcji upadku. W niniejszym rozdziale przedstawiono koncepcję wykorzystania rozmytego systemu wnioskującego do detekcji upadku, celem poprawienia uzyskiwanych wyników, a przede wszystkim specyficzności detekcji systemu. Rozdział rozpoczyna się przedstawieniem pojęcia logiki rozmytej i wnioskowania rozmytego. W dalszej części rozdziału przedstawiony został system wnioskujący oraz uzyskane wyniki. Rezultaty badań porównano z uprzednio uzyskanymi wynikami.

## 4.1. Logika rozmyta

Klasyczna logika dwuwartościowa, może być niewystarczająca do formalnego opisu zjawisk otaczającego nas świata, które mają charakter wieloznaczny i nieprecyzyjny. W praktyce dwuwartościowe określenie przynależności do zbioru nie zawsze opisuje stan faktyczny (Ross, 2010). W wyniku zapotrzebowania na metody opisujące zjawiska o wspomnianym charakterze, powstały pewne alternatywne do logiki klasycznej systemy logiczne. Jednym z takich systemów jest tzw. logika rozmyta. Jej twórcą jest profesor Lofti A. Zadeh. W 1965 roku opublikował on teorię zbiorów rozmytych, a w 1973 roku stworzył system logiki rozmytej. Zapoczątkowało to dynamiczny rozwój nowych teorii i metod w informatyce, matematyce, sterowaniu i automatyce oraz w innych dziedzinach nauki i techniki. Warto wspomnieć, że logika rozmyta nie jest pierwszą próbą odejścia od klasycznej logiki dwuwartościowej. Rozwój tej dziedziny badań zapoczątkowały prace profesora Jana Łukasiewicza (Łukasiewicz, 1920) i E.L. Posta (Post, 1921) z początku lat dwudziestych XX wieku czy też Kurta Gödla (Gödel, 1932).

W celu precyzyjnego przedstawienia zagadnień logiki wielowartościowej przedstawione zostanie pojęcie normy triangularnej (trójkątnej). Powstanie norm triangularnych związane jest z zagadnieniem probabilistycznych przestrzeni metrycznych. Normy triangularne posłużyły w rozważaniach nad tymi przestrzeniami jako narzędzie do uogólnienia nierówności trójkąta. W późniejszym okresie zauważono, iż normy te nadają się do interpretacji spójników logicznych w logikach wielowartościowych (Gottwald, 1999).

Normą triangularną (w skrócie t-normą) nazywamy operację binarną $t: [0,1] \times [0,1] \to [0,1]$ spełniającą następujące warunki:

$$\forall a, b \in [0,1]: a \, t \, b = b \, t \, a \tag{4.1}$$

$$\forall a, b, c \in [0,1]: (a \, t \, b) \, t \, c = a \, t \, (b \, t \, c) \tag{4.2}$$

$$\forall a, b, c, d \in [0,1]: (a \leq b \, \& \, c \leq d) \Rightarrow a \, t \, c \leq b \, t \, d \tag{4.3}$$

oraz

$$\forall a \in [0,1]: a \, t \, 1 = a \tag{4.4}$$

czyli odpowiednio warunki: przemienności, łączności, monotoniczności oraz elementu neutralnego. Konormą triangularną (w skrócie t-konormą) nazywamy operację binarną $s: [0,1] \times [0,1] \to [0,1]$ spełniającą warunki (4.1), (4.2) i (4.3) oraz warunek:

$$\forall a \in [0,1]: a \, s \, 0 = a \tag{4.5}$$

Do podstawowych norm i konorm triantularnych można zaliczyć:

1. t-norma minimum ∧:

$$a \wedge b = \min(a, b) \tag{4.6}$$

2. t-konorma maximum ∨:

$$a \vee b = \max(a, b) \tag{4.7}$$

3. t-norma algebraiczna $t_a$:

$$a \, t_a \, b = ab \tag{4.8}$$

4. t-konorma algebraiczna $s_a$:

$$a \, s_a \, b = a + b - ab \tag{4.9}$$

5. t-norma Łukasiewicza $t_L$:

$$a \, t_L \, b = 0 \vee (a + b - 1) \tag{4.10}$$

6. t-konorma Łukasiewicza $s_L$:

$$a \, s_L \, b = 1 \wedge (a + b) \tag{4.11}$$

7. t-norma graniczna $t_D$:

$$a \, t_D \, b = \begin{cases} 0, & \text{jeżeli } a \vee b < 1 \\ a \wedge b, & \text{jeżeli } a \vee b = 1 \end{cases} \tag{4.12}$$

8. t-norma graniczna $s_D$:

$$a \, s_D \, b = \begin{cases} 1, & \text{jeżeli } a \wedge b > 0 \\ a \vee b, & \text{jeżeli } a \wedge b = 0 \end{cases} \tag{4.13}$$

Teoria zbiorów rozmytych zaproponowana w (Zadeh, 1965) rozszerza klasyczną teorię zbiorów, o możliwość częściowej przynależności do zbioru, która jest określana wartością z przedziału [0,1].Teoria dostarcza definicji zbioru rozmytego $A$ w przestrzeni $X$ jako zbioru par:

$$A = \{(x, \mu_A(x)); x \in X\} \tag{4.14}$$

w którym:

$$A = \mu_A : X \to [0, 1] \tag{4.15}$$

jest funkcją przynależności do zbioru rozmytego. Funkcja ta przypisuje każdemu elementowi jego stopień przynależności, przy czym:

1) $\mu_A(x) = 1$, oznacza całkowitą przynależność elementu $x$ do zbioru rozmytego,
2) $\mu_A(x) = 0$, oznacza brak przynależności elementu $x$ do zbioru rozmytego,
3) $0 < \mu_A(x) < 1$, oznacza częściową przynależność elementu $x$ do zbioru rozmytego.

Zbiór o skończonej liczbie elementów, zdefiniowany jak powyżej, można przedstawić w sposób symboliczny:

$$A = \frac{\mu_A(x_1)}{x_1} + \frac{\mu_A(x_2)}{x_2} + \cdots + \frac{\mu_A(x_n)}{x_n} = \sum_{i=1}^{n} \frac{\mu_A(x_i)}{x_i} \tag{4.16}$$

gdzie zapis $\frac{\mu_A(x_n)}{x_n}$ oznacza parę $A = \{(x, \mu_A(x)); x \in X\}$, a znak + sumę mnogościową elementów. Dla przestrzeni $X$ o nieskończonej liczbie elementów, zbiór rozmyty symbolicznie zapisujemy jako:

$$A = \int_X \frac{\mu_A(x)}{x} \tag{4.17}$$

Funkcje przynależności dla zbiorów rozmytych mogą przyjmować różne kształty: singleton, funkcja gaussowska, typu dzwonowego, klasy *s*, klasy π, klas *y* i *L* (kształtem przypominająca trapez), oraz klasy *t* (trójkątna). Przykładowe funkcje przynależności przedstawiono na rysunku 3.6.

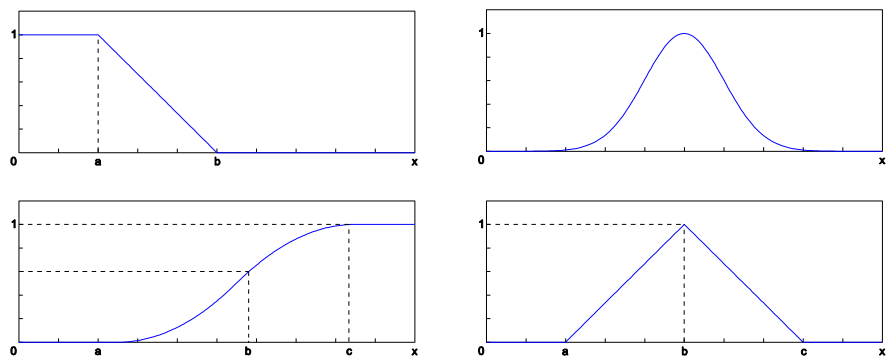

Rysunek 4.1. Przykładowe funkcje przynależności: a) funkcja klasy L, b) funkcja gaussowska, c) funkcja klasy s, d) funkcja klasy t.

Na zbiorach rozmytych, podobnie jak klasycznych można zdefiniować operacje mnogościowe oraz algebraiczne. Normy triangularne i negacje znajdują zastosowanie przy definiowaniu operacji na zbiorach rozmytych.

**Przecięciem** zbiorów rozmytych $A, B \subseteq X$ jest zbiór rozmyty $A \cap B$ o funkcji przynależności:

$$\mu_{A \cap B}(x) = t(\mu_A(x), \mu_B(x)) \tag{4.18}$$

dla każdego x ∈ X.

**Sumą** zbiorów rozmytych $A, B \subseteq X$ jest zbiór rozmyty $A \cup B$ określony funkcją przynależności:

$$\mu_{A \cup B}(x) = s(\mu_A(x), \mu_B(x)) \tag{4.19}$$

dla każdego x ∈ X.

**Dopełnieniem** zbioru rozmytego A ⊆ X jest zbiór rozmyty $\hat{A}$, którego funkcja przynależności ma postać:

$$\mu_{\hat{A}}(x) = 1 - \mu_A(x) \tag{4.20}$$

Należy wprowadzić jeszcze pojęcia **nośnika** i **wysokości** zbioru rozmytego. Nośnikiem zbioru rozmytego *A* jest taki zbiór elementów, których wartość funkcji przynależności jest większa od zera. Wysokością zbioru nazywamy największą wartość funkcji przynależności.

**Relacją rozmytą** *R* pomiędzy dwoma zbiorami (niepustymi i nierozmytymi) *X*, *Y* jest zbiór rozmyty określony na iloczynie kartezjańskim $X \times Y$, tj.:

$$R \subseteq X \times Y = \{(x, y) : x \in X, y \in Y\} \tag{4.21}$$

Relacja ta jest zbiorem par $R = \{((x, y), \mu_R(x, y))\}$ gdzie $\mu_R$ jest nową funkcją przynależności $\mu_R : X \times Y \to [0, 1]$.

Nierosnącą funkcję $N : [0, 1] \to [0, 1]$ nazywamy negacją jeżeli spełniony jest następujący warunek:

$$N(0) = 1 \land N(1) = 0 \tag{4.22}$$

Negację *N* nazywamy negacją typu *strict*, jeżeli jest ciągła i malejąca. Negację *N* nazywamy negacją typu *strong* jeżeli jest typu *strict* oraz jest inwolucją, tzn.:

$$N(N(a)) = a \tag{4.23}$$

Przykładowymi negacjami są:

1. Negacja Zadeha (*strong*):

$$N(a) = 1 - a \tag{4.24}$$

2. Negacja Yadera (*strict*):

$$N(a) = (1 - a^p)^{\frac{1}{p}}, \ p > 0 \tag{4.25}$$

3. Negacja Sugeno (*strong*):

$$N(a) = 1 - a \tag{4.26}$$

## 4.2. Wnioskowanie rozmyte

Wnioskowanie rozmyte (wnioskowanie przybliżone lub przybliżone rozumowanie, ang. *fuzzy inference*) realizowane w oparciu o bazę wiedzy jest sposobem określania rozmytego zbioru wyjściowego $B'$ na podstawie odpowiednich procedur transformacji oraz wejściowego zbioru rozmytego $A'$. Teoria wnioskowania rozmytego rozszerza dotychczas znane modele wnioskowania o możliwość wykorzystania nieprecyzyjnych stwierdzeń, charakterystycznych dla logiki rozmytej.

W tradycyjnej logice dwuwartościowej wnioskujemy o prawdziwości zdań – wniosków, na podstawie prawdziwości innych zdań nazywanych przesłankami. Schemat wnioskowania ma tą właściwość, że jeżeli prawdziwe są wszystkie przesłanki, to prawdziwy jest wniosek. Ogólny schemat wnioskowania zapisywany jest w następujący sposób:

$$\frac{\begin{array}{c} A_1 \\ \ldots \\ A_n \end{array}}{B} \tag{4.27}$$

gdzie $A_1 \ldots A_n$ przesłanki, a $B$ wniosek. Regułami wnioskowania nazywamy wzorce opisujące dozwolone sposoby bezpośredniego wyprowadzania nowych formuł ze znanych formuł. Dwie podstawowe reguły wnioskowania to *modus ponens*:

$$\frac{\begin{array}{c} A \\ A \rightarrow B \end{array}}{B} \tag{4.28}$$

oraz *modus tollens*:

$$\frac{\begin{array}{c} \overline{B} \\ A \rightarrow B \end{array}}{\overline{A}} \tag{4.29}$$

Reguła (4.23) mówi, że jeżeli uznajemy prawdziwość poprzednika prawdziwej implikacji, to musimy uznać też prawdziwość jej następnika. Reguła (4.24) może być zastosowana do dowolnych dwóch formuł, z których jedna jest implikacją, a druga – zanegowanym następnikiem tej implikacji. Wówczas reguła generuje formułę będącą zanegowanym poprzednikiem implikacji.

Reguły wnioskowania w logice dwuwartościowej można rozszerzyć na przypadek rozmyty. Niech $A$, $A' \subseteq X$ oraz $B$, $B' \subseteq Y$ są zbiorami rozmytymi, a $x$ i $y$ zmiennymi lingwistycznymi. Przesłanka przyjmuje postać:

$$x \text{ jest } A' \qquad (4.30)$$

Gdzie $A'$ jest zbiorem rozmytym, zdefiniowanym w niepustej przestrzeni $X$ o funkcji przynależności (4.15). Implikacja postaci:

$$\text{Jeżeli } x \text{ jest } A \text{ to } y \text{ jest } B \qquad (4.31)$$

Jest równoważna relacji rozmytej $R \in X \times Y$. Wniosek reguły rozmytej odnosi się do pewnego zbioru rozmytego $B'$, który jest określony przez złożenie zbioru rozmytego $A'$ i rozmytej implikacji $A \rightarrow B$:

$$B' = A' \circ (A \rightarrow B) \qquad (4.32)$$

Zatem funkcja przynależności zbioru rozmytego $B'$ przyjmuje postać:

$$\mu_{B'}(y) = \sup\{\mu_{A'}(x) \overset{T}{*} \mu_{A \rightarrow B}(x, y)\} \qquad (4.33)$$

Uogólnioną rozmytą regułę wnioskowania *modus ponens* można przedstawić na schemacie wnioskowania:

$$\frac{\begin{array}{c} x \text{ jest } A' \\ \text{Jeżeli } x \text{ jest } A \text{ to } y \text{ jest } B \end{array}}{y \text{ jest } B'} \qquad (4.34)$$

Analogicznie można wyprowadzić uogólnioną rozmytą regułę wnioskowania *modus tollens* korzystając z reguły (4.29). Można przedstawić ją w następujący sposób:

$$\frac{\begin{array}{c} y \text{ jest } B' \\ \text{Jeżeli } x \text{ jest } A \text{ to } y \text{ jest } B \end{array}}{x \text{ jest } A'} \qquad (4.35)$$

Zbiór rozmyty $A'$ reprezentujący wniosek reguły *modus tollens* można określić jako:

$$A' = (A \rightarrow B) \circ B' \qquad (4.36)$$

a funkcję przynależności:

$$\mu_{A'}(x) = \sup\{\mu_{A \rightarrow B}(x, y) \overset{T}{*} \mu_{B'}(y)\} \qquad (4.37)$$

Funkcję $I : [0, 1]^2 \rightarrow [0,1]$ nazywamy implikacją rozmytą, jeżeli spełnia poniższe warunki:

a) jeżeli $a_1 \leq a_3$, to $I(a_1, a_2) \geq I(a_3, a_2)$ dla wszystkich $a_1, a_2, a_3 \in [0,1]$
b) jeżeli $a_2 \leq a_3$, to $I(a_1, a_2) \leq I(a_1, a_3)$ dla wszystkich $a_1, a_2, a_3 \in [0,1]$
c) $I(0, a_2) = 1$ dla wszystkich $a_2 \in [0,1]$
d) $I(a_1, 1) = 1$ dla wszystkich $a_1 \in [0,1]$
e) $I(1, 0) = 0$

Przykładowymi implikacjami są:

1. Implikacja Kleene-Dienesa (binarna):

$$I(a, b) = \max\{1 - a, b\} \qquad (4.38)$$

2. Implikacja Łukasiewicza:

$$I(a,b) = \min\{1, 1 - a + b\} \quad (4.39)$$

3. Implikacja Reichenbacha:

$$I(a,b) = 1 - a + ab \quad (4.40)$$

4. Implikacja Fodora:

$$I(a,b) = \begin{cases} 1, & \text{jeżeli } a \leq b \\ \max\{1 - a, b\}, & \text{jeżeli } a > b \end{cases} \quad (4.41)$$

5. Implikacja Gödla:

$$I(a,b) = \begin{cases} 1, & \text{jeżeli } a \leq b \\ b, & \text{jeżeli } a > b \end{cases} \quad (4.42)$$

Rozmyte systemy wnioskujące są jednym z najbardziej rozpowszechnionych zastosowań logiki rozmytej (Guillaume, 2001). Są one szczególnie przydatne w rozwiązywaniu wielu zagadnień dotyczących sterowania, klasyfikacji czy wspomagania decyzji. Ich szerokie zastosowanie wynika z dwóch własności: z jednej strony potrafią radzić sobie z pojęciami lingwistycznymi, a z drugiej są uniwersalnymi aproksymatorami wejścia na wyjście.

Schemat typowego systemu wnioskującego przedstawiono na rysunku 4.2. Składa się on z następujących elementów (Rutkowski, 2006):

- bazy reguł (ang. *knowledge base*),
- bloku rozmywania (ang. *fuzzification*),
- bloku wnioskowania (ang. *inference*),
- bloku wyostrzania (ang. *defuzzification*).

**Blok rozmywania** odpowiedzialny jest za odwzorowanie wartości sygnału wejściowego w zbiór rozmyty. Operacja ta jest nazywana rozmywaniem. Powstały zbiór rozmyty jest wejściem bloku wnioskowania.

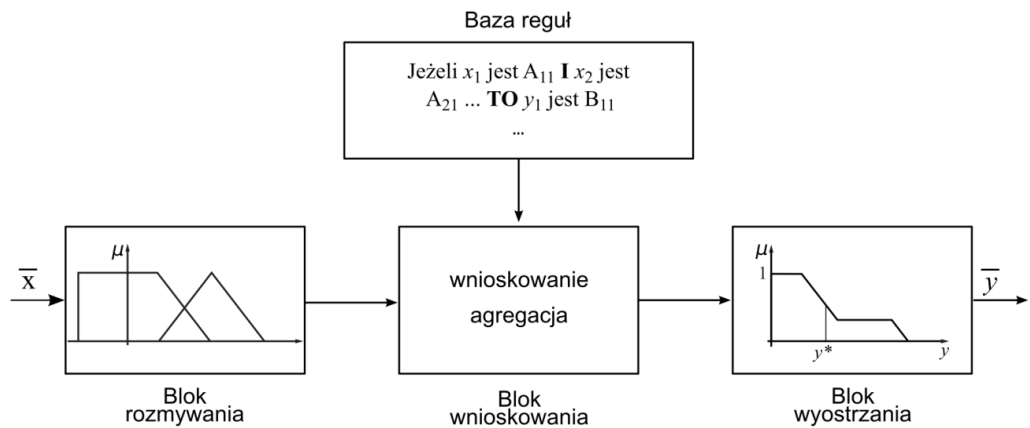

Rysunek 4.2. Schemat rozmytego systemu wnioskującego.

**Baza reguł**, nazywana niekiedy modelem lingwistycznym, jest zbiorem rozmytych reguł postaci:

Jeżeli $x_1$ jest $A_1$ I $x_2$ jest $A_2$ TO $y_1$ jest $B_3$

gdzie *A*, *B* to zbiory rozmyte, $x_1$, $x_2$ to zmienne wejściowe modelu lingwistycznego a $y_1$ zmienna wyjściowa modelu lingwistycznego. Baza reguł zawiera informacje definiowane przez eksperta z danej dziedziny, do których należą wartości lingwistyczne zmiennych oraz definicje zbiorów rozmytych. Zaczerpnięta z dziedziny logiki matematycznej, ogólna postać regułowej reprezentacji wiedzy składa się z części nazywanej poprzednikiem (ang. *antecendent*) oraz części nazywanej następnikiem (ang. *consequent*).

**Blok wnioskowania** dla każdej reguły z bazy reguł określa stopień prawdziwości części przesłankowej.

**Blok wyostrzania** realizuje zagadnienie odwzorowania zbiorów rozmytych w jedną wartość $\bar{y} \in Y$, będącą wyznaczonym sterowaniem na wyjściu systemu wnioskującego. Jeżeli wielkością wyjściową bloku wnioskowania jest *N* zbiorów rozmytych, wówczas wartość wyjścia można wyznaczyć korzystając z następujących metod:

1. Metoda maksimum funkcji przynależności (ang. *height method*, *max membership principle*). Wartość $\bar{y}$ jest równa wysokości zbioru i opisana równaniem:

$$\bar{y} = \sup_{y \in Y} \mu_{B'}(y) \tag{4.43}$$

2. Metoda środka ciężkości (ang. *centroid method*), w której wartość wyjścia zdefiniowana jest jako środek ciężkości funkcji $\mu_{B'}(y)$:

$$\bar{y} = \frac{\int_Y y \mu_{B'}(y) dy}{\int_Y \mu_{B'}(y) dy} \tag{4.44}$$

3. Metoda średniej maksimów (ang. *mean max membership*), która jest ściśle związana z metodą opartą o maksimum funkcji przynależności. Jednak zamiast maksimum globalnego, wartość wyjścia definiowana jest jako średnia arytmetyczna dwóch lub więcej maksimów lokalnych (Lee, 1990).

4. Metoda średniej ważonej (ang. *weighted average method*), w której wartość wyjścia wyznaczana jest za pomocą wzoru:

$$\bar{y} = \frac{\sum_{k=1}^{N} \mu_{\bar{B}^k}(\bar{y}^k) \bar{y}^k}{\sum_{k=1}^{N} \mu_{\bar{B}^k}(\bar{y}^k)} \tag{4.45}$$

gdzie $\bar{y}^k$ jest punktem, w którym funkcja $\mu_{\bar{B}^k}(y)$ przyjmuje wartość maksymalną. Punkt ten jest nazywany środkiem zbioru rozmytego.

Dokonując przeglądu literatury przedmiotu (Leekwijck & Kerre, 1999) spotkać można również inne metody wyostrzania: metodę najmniejszego maksimum (SOM, ang. *smallest value of maximum*), metodę środka maksimum (MOM, ang. *middle value of maximum*), metodę największego maksimum (LOM, ang. *largest value of maximum*), metodę środka sum (COS, ang. *center value of sums*) czy też metodę podstawowego rozkładu wyostrzania (BADD, ang. *BAsic Defuzzification Distribution*).

## 4.3. Metody projektowania rozmytych systemów wnioskujących

Budowa rozmytego układu wnioskującego wymaga odpowiedniego zaprojektowania jego elementów składowych. W procesie projektowania układu możemy wyróżnić następujące zagadnienia:

- wybór zmiennych wejściowych,
- podział przestrzeni wejścia na grupę odpowiednich zbiorów rozmytych, tj. dobór funkcji przynależności,
- budowa bazy reguł,
- wybór operatorów (t-norm i t-konorm) oraz metod wyostrzania.

Projektowanie systemu wnioskującego może być zrealizowane przez człowieka – eksperta w danej dziedzinie, której dotyczy problem sterowania lub podejmowania decyzji, lub automatycznie na podstawie pewnego zbioru danych uczących.

### 4.3.1. Projektowanie w oparciu o wiedzę ekspercką

Możliwość projektowania układów w oparciu o wiedzę ekspercką wynika z charakteru logiki rozmytej, w szczególności jej przystosowania do opisu pojęć lingwistycznych, a także z naturalnych zdolności człowieka do obserwacji otaczającego go świata i umiejętności opisu, czy też modelowania zachodzących w nim zjawisk (Ross, 2010). Jako przykład może posłużyć zagadnienie doboru funkcji przynależności dla zmiennej określającej temperaturę powietrza (Rys. 4.3). W niniejszym zagadnieniu, każda z funkcji odpowiada przynależności do jednego ze zbiorów rozmytych reprezentujących wartości temperatury: Zimno, Chłodno, Ciepło, Gorąco.

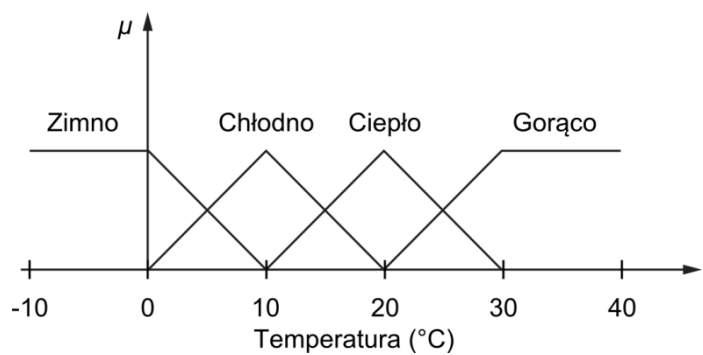

Rysunek 4.3 Funkcje przynależności dla zmiennej lingwistycznej temperatura.

Warto podkreślić, że definicja przynależności do zbiorów rozmytych powinna być dostosowana do rozpatrywanego problemu i jego kontekstu, a także osoby definiującej (eksperta). W rozpatrywanym przykładzie, definicja pojęć lingwistycznych odnosi się do temperatury otoczenia człowieka, w innym przypadku (np. określenia temperatury pracy pewnego urządzenia przemysłowego) dziedzina i przeciwdziedzina funkcji mogłyby być inne.

Odrębnym etapem projektowania rozmytego systemu wnioskującego jest budowa bazy reguł. Różni eksperci mogą formułować wiedzę w różny, indywidualnie arbitralny sposób, natomiast na etapie projektowania bazy wiedzy zagwarantowany musi zostać brak anomalii w regułach. W literaturze przedmiotu (Ligęza, 2006) znaleźć można kategoryzację anomalii, które mogą wystąpić w procesie projektowania bazy reguł:

- **Redundancja**:
  - reguły równoważne,
  - reguły pochłaniające się,
  - reguły identyczne,
  - reguły nieużywane.
- **Niespójność**:
  - reguły niejednoznaczne,
  - reguły sprzeczne,
  - reguły niespójne logicznie.
- **Redukcja**:
  - niewłaściwa redukcja reguł,
  - eliminacja niezbędnych zmiennych.
- **Niekompletność**:
  - logiczna niekompletność,
  - fizyczna niekompletność.

W literaturze dotyczącej systemów rozmytych wymienia się następujące, wymagane, własności reguł rozmytego modelu wiedzy:

- kompletność modelu wiedzy,
- niesprzeczność bazy reguł,
- ciągłość bazy reguł,
- brak redundancji w bazie reguł.

Rozmyty model wiedzy jest kompletny (ang. *completeness*) jeżeli każdemu wektorowi wejść $\bar{\mathbf{x}} = [x_1, x_2, ..., x_n]$ można przyporządkować wartość ostrą $\bar{y}$ na wyjściu modelu. Oznacza to, że niezależnie od wejścia przesłanka przynajmniej jednej z reguł będzie spełniona, a reguła będzie wykorzystana w procesie wnioskowania. Rozmyty model wiedzy posiada sprzeczną (ang. *conflict rules*) bazę wiedzy, jeżeli istnieją w niej dwie reguły o identycznych przesłankach, ale o różnych konkluzjach. Rozmyty model wiedzy posiada ciągłą (ang. *continuous*) bazę wiedzy, jeżeli nie ma w niej sąsiednich reguł $R(j), R(k)$ ze zbiorami rozmytymi w konkluzji $B^j, B^k$, o pustym iloczynie: $B^j \cap B^k \neq \emptyset$.

Brak nadmiarowości bazy reguł oznacza unikanie sytuacji, w których w bazie reguł znajduje się więcej niż jedna reguła o tej samej przesłance i identycznej konkluzji. Najczęściej pożądaną sytuacją jest nie tylko unikanie nadmiarowości, ale dążenie do jak najprostszego modelu wiedzy. Mniejsza liczba reguł pozwala na łatwiejszą interpretację działania systemu przez człowieka. Szerszy opis metod projektowania i weryfikacji poprawności rozmytych systemów wnioskujących zamieszczono w pracach (Ross, 2010; Ligęza, 2006; Lunardhi & Passino, 1995).

### 4.3.2. Projektowanie na podstawie danych

Pomimo połączenia zdolności logiki rozmytej do modelowania pojęć lingwistycznych i naturalnych umiejętności człowieka do opisu otaczających go zjawisk, metody projektowania systemów wnioskujących w oparciu o wiedzę ekspercką mogą okazać się niewystarczające (Guillaume, 2001). Trudności w projektowaniu systemów przez eksperta mogą przysparzać np.: braki w wiedzy o charakterze modelowanego zjawiska lub jego znaczna złożoność. Wraz z rozwojem technik uczenia maszynowego wiele zespołów badawczych podjęło wysiłki w celu opracowania metod automatycznego budowania rozmytych modeli wnioskujących na podstawie danych. Metody te można podzielić na dwie grupy: metody pozyskiwania bazy wiedzy oraz optymalizacji parametrów modelu. Zadaniem metod automatycznego pozyskiwania baz wiedzy jest uzyskanie możliwie najmniejszego zbioru reguł, który pozwala na jak najdokładniejsze odwzorowanie modelowanego zjawiska. Proces budowy bazy wiedzy można podzielić na dwa etapy: etap indukcji reguł (ang. *rule induction*) oraz etap optymalizacji bazy reguł (ang. *rule-base optimization*) (Guillaume, 2001). Etap optymalizacji bazy reguł jest szczególnie ważny w procesie projektowania systemów wnioskujących do modelowania zachowania złożonych obiektów czy zjawisk. Liczba wygenerowanych reguł może być znacząca, co w połączeniu z mnogością zmiennych sprawia, że baza wiedzy takiego systemu może być trudna do interpretacji przez człowieka. Istotnymi zagadnieniami w takim przypadku stają się zadania selekcji reguł i wyboru zmiennych, określane jako optymalizacja struktury systemu wnioskującego (ang. *structure optimization*). Do najczęściej stosowanych metod indukcji reguł można zaliczyć: metodę Wanga-Mendela, metody oparte o klasteryzację fuzzy c-means, metody wykorzystujące drzewa decyzyjne czy też algorytmy genetyczne. Zadanie wyboru zmiennych może być realizowane w sposób lokalny oraz globalny. Podczas globalnego doboru zmiennych, niektóre z nich są usuwane i nie są uwzględniane w żadnej z reguł. Lokalny dobór zmiennych pozwala na pominięcie danej zmiennej w niektórych regułach, prowadząc do powstania reguł niepełnych. W procesie doboru zmiennych często stosuje się metody oparte o minimalizację kryterium regularności (ang. *regularity criterion*), algorytmy genetyczne, czy też drzewa decyzyjne.

Proces optymalizacji parametrów modelu rozmytego polega głównie na wyznaczaniu parametrów funkcji przynależności dla zmiennych wejściowych i wyjściowych systemu, przy jednoczesnej minimalizacji błędu układu wnioskującego szacowanego względem modelowanego zjawiska czy procesu. W zadaniu tym wykorzystywana jest jedna z miar błędu, np. błąd średniokwadratowy (ang. *Mean Squared Error*, MSE) czy średni błąd bezwzględny (ang. *Mean Absolute Error*) (Piegat, 2001).

## 4.4. Rozmyty system wnioskujący

Wyniki badań zaprezentowane w rozdziale 3. wykazały użyteczność cech opisujących pozę oraz dynamikę ruchu postaci na potrzeby detekcji upadku. Zaproponowany system podejmował decyzję w oparciu o układ hierarchiczny, w którym hipoteza o wystąpieniu upadku była stawiana na podstawie analizy danych akcelerometrycznych, a następnie potwierdzana bądź odrzucana na podstawie analizy obrazu (zob. rysunek 4.4). Wnioskowanie na podstawie obrazów przebiegało dwuetapowo:

1. detekcja pozy leżącej na podstawie pojedynczej mapy głębi,
2. analiza ruchu postaci przy użyciu *dyamic transitions*.

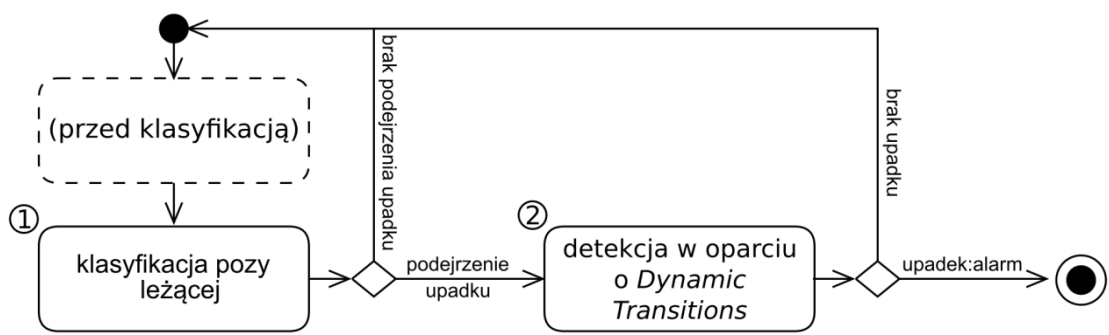

Rysunek 4.4. Schemat blokowy systemu detekcji upadku działającego w oparciu o logikę rozmytą.

Omawiany model systemu detekcji upadku wykazał się wysoką czułością oraz specyficznością, co zostało zaprezentowane w pracach (Kwolek & Kępski, 2014; Kwolek & Kępski, 2015). Ponadto do zalet takiego podejścia należą:

- Możliwość doboru czułości systemu – w zależności od zapotrzebowania system może informować o tym, że użytkownik leży na podłodze, bądź dokonywać pełnej detekcji upadku.
- Modułowość – pozwala udoskonalać poszczególne etapy detekcji bez konieczności przeprojektowywania całego systemu.
- Możliwość uzupełniania danych uczących *on-line* – zastosowania metod *lazy learning* pozwala na dodawanie do bazy danych trenujących próbek danych (sprawdzonych przez człowieka, np. operatora systemu lub opiekuna osoby starszej) bez potrzeby ponownego uczenia modelu,
- Prostota implementacji – system działa w oparciu o sprawdzone metody klasyfikacji, które zostały uwzględnione w wielu bibliotekach *OpenSource*, także dla urządzeń wbudowanych.
- Niski koszt obliczeniowy – zastosowanie układu hierarchicznego, w którym hipoteza opiera się na przetwarzaniu danych akcelerometrycznych pozwala uniknąć części lub całości operacji na obrazach czy też sekwencjach obrazów. Zwiększa to potencjał aplikacyjny systemu, szczególnie w kontekście docelowej platformy obliczeniowej z procesorem o architekturze ARM.

W dotychczasowych badaniach nad detekcją upadku zaobserwowano, że akcja ta posiada zazwyczaj dwie charakterystyczne własności:

- nagłą zmianę sylwetki charakteryzującą się dużym przyspieszeniem, rzadko spotykanym przy czynnościach dnia codziennego, szczególnie u osób starszych,
- zmiana sylwetki kończy się zazwyczaj pozycją leżącą.

Dokonując przeglądu literatury przedmiotu, zauważyć można, że niektóre zespoły badawcze (Miaou et al., 2006) skupiają się na samej detekcji pozy leżącej, pomijając ruch postaci. Takie podejście wynika z założenia, że jeśli osoba starsza znajduje się w pozycji leżącej w miejscu do tego nieprzeznaczonym (na podłodze), to jest to bezpośrednim efektem upadku. Oparcie reguł decyzyjnych o takie założenie może być właściwe dla osób o ograniczonej zdolności ruchowej, lecz w przypadku osób bardziej sprawnych może prowadzić do fałszywych alarmów.

W toku badań zauważono, że zastosowanie hierarchicznego systemu detekcji może w sporadycznych przypadkach prowadzić do obniżenia czułości lub specyficzności. Spadek czułości może być skutkiem wystąpienia błędów drugiego rodzaju, spowodowanych:

- Błędami klasyfikacji pozy osoby – upadek osoby monitorowanej może zakończyć się w pozycji półleżącej, czy też siedzącej. Biorąc pod uwagę podobieństwo takiej konfiguracji sylwetki do niektórych czynności ADLs oraz mnogość możliwych położeń ciała względem kamery, poza ta może zostać błędnie sklasyfikowana jako próbka negatywna.
- Błędami detekcji *dynamic transitions* – największe zmiany sylwetki postaci zachodzą przy upadku z pozycji stojącej, przy czym przypadek ten nie wyczerpuje całego uniwersum możliwych akcji. W niektórych upadkach z pozycji siedzącej, czy klęczącej zmiany cech w czasie mogą być niewystarczające do klasyfikacji akcji jako upadku. Ponadto, osoba może się asekurować przed upadkiem (poprzez próbę przytrzymania się obiektów z jej otoczenia), czy też mogą wystąpić przesłonięcia.

Spadek wartości swoistości może być skutkiem wystąpienia błędów pierwszego rodzaju, spowodowanych:

- Błędami klasyfikacji pozy osoby – możliwe w przypadku błędów segmentacji postaci, lub wystąpienia przysłonięć, przez co poza ta może zostać błędnie sklasyfikowana jako próbka pozytywna.
- Błędami detekcji *dynamic transitions* – zbyt gwałtowne wykonanie czynności ADLs, które kończą się w pozycji leżącej, może prowadzić do niepoprawnej klasyfikacji tych akcji jako upadek, co zaobserwowano w (Kwolek & Kępski, 2014) dla sekwencji ze zbioru UR Fall Dataset.

Oparcie systemu decyzyjnego o metody klasyfikacji stwarza możliwość porównania cech klasyfikowanego przypadku z szerszym lub węższym spektrum próbek uczących (zebranych podczas upadków symulowanych bądź rzeczywistych), co jednak prowadzi do konieczności operowania pojęciami przynależności binarnej, które w nomenklaturze logiki rozmytej zostałyby określone jako wartości ostre. Gdy w takim podejściu wystąpią wspomniane wcześniej błędy, rośnie prawdopodobieństwo niepoprawnej klasyfikacji akcji.

W celu uniknięcia ryzyka błędnej decyzji wynikającej ze wspomnianych własności systemu hierarchicznego i działającego w oparciu o klasyfikatory, zaproponowano zastosowanie logiki rozmytej do detekcji upadku. Na etapie projektowania przyjęto następujące założenia:

- **System uzyskuje hipotezę o wystąpieniu upadku w oparciu o dane akcelerometryczne.** Podobnie jak omawiane w rozdziale 3. algorytmy, docelową platformą dla niniejszego systemu jest urządzenie *PandaBoard* bądź pokrewne, co w konsekwencji wymaga ograniczenia nakładów obliczeniowych. Jak już wcześniej wspomniano, analiza danych akcelerometrycznych pozwala uniknąć konieczności ciągłego przetwarzania sekwencji obrazów.
- **Wejściem systemu są zmienne odpowiadające cechom statycznym i dynamicznym wykorzystywanym we wcześniejszych algorytmach.** Dotychczasowe badania wykazały użyteczność tych cech w problemie detekcji upadku. Wykorzystane atrybuty to cechy statyczne: $H/W$, $H/H_{max}$, $max(\sigma_x, \sigma_z)$, $P_{40}$ oraz cechy dynamiczne: $H(t)/H(t - \Delta t)$, $D(t)/D(t - \Delta t)$, $SV_{total}$.
- **Baza reguł zaprojektowana w oparciu o wiedzę ekspercką.** Czytelność reguł zaprojektowanych przez człowieka pozwoli na analizę i modyfikację systemu w dużo większym stopniu niż jest to możliwe w przypadku systemów generowanych automatycznie.
- **Porównanie z systemami zaprojektowanymi w oparciu o dane.** Analiza porównawcza wyników pozwoli na ocenę przydatności zaproponowanego podejścia.

Schemat ideowy proponowanego rozwiązania przedstawiono na rysunku 4.5.

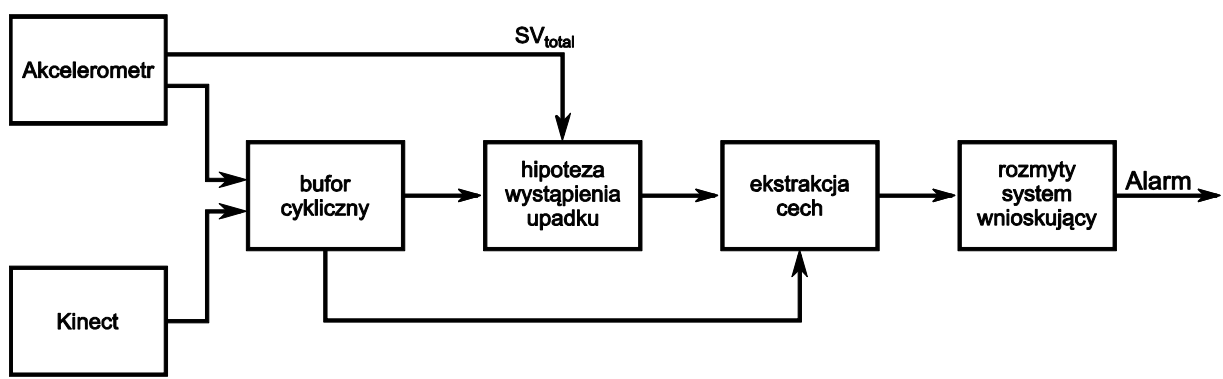

Rysunek 4.5. Schemat blokowy systemu detekcji upadku działającego w oparciu o logikę rozmytą.

Oznaczając liczbę wejść modelu jako $N$ i przyjmując, że każda wielkość wejściowa jest scharakteryzowana identyczną ilością $J$ zbiorów rozmytych, to maksymalna liczba $R$ reguł określona jest wzorem: $R = J^N$. Zakładając, że w rozpatrywanym problemie każdy atrybut scharakteryzowany zostanie trzema zbiorami rozmytymi, przy siedmiu zmiennych wejściowych baza wiedzy powinna składać się z 2187 reguł, co jest wartością wykraczającą poza możliwości bezbłędnego zaprojektowania przez eksperta.

Celem przezwyciężenia tych trudności praktycznych, zaproponowano rozwiązanie dla systemu wnioskującego, które wprowadza dwie syntetyczne zmienne lingwistyczne:

- **Poza postaci** (w skrócie: Pose) – będąca odwzorowaniem wartości cech statycznych na pewien zbiór rozmyty,
- **Dynamika ruchu postaci** (w skrócie Transition) – będąca odwzorowaniem wartości atrybutów dynamicznych na pewien zbiór rozmyty.

Do uzyskania obu zmiennych wykorzystywane są rozmyte układy wnioskujące Mamdaniego (nazwane odpowiednio Static oraz Transition), posiadające odpowiednio 4 i 3 wejścia. Za ostateczną decyzję o wystąpieniu upadku odpowiedzialny jest układ wnioskujący Takagi-Sugeno, posiadający dwie zmienne wejściowe: Pose i Transition oraz jedną zmienną wyjściową, będącą funkcją wejścia, decydującą o detekcji upadku (nazwany Decision). Schemat systemu przedstawiono na rysunku 4.6.

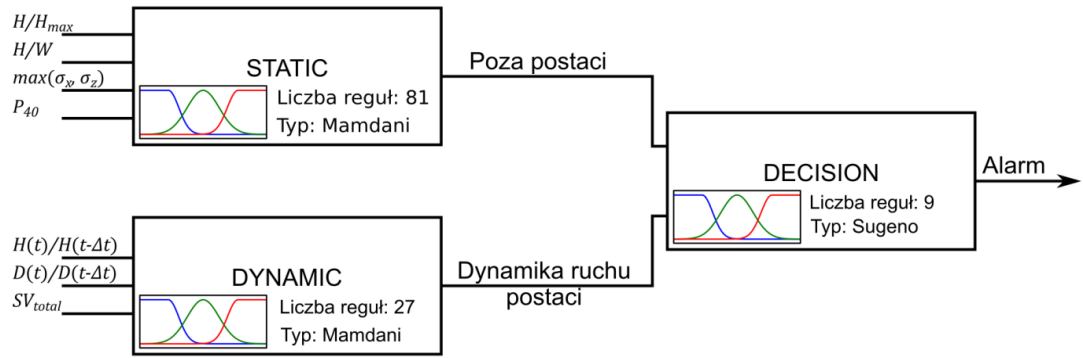

Rysunek 4.6. Schemat rozmytego systemu wnioskującego.

Porównując zaproponowane podejście z klasycznym systemem rozmytym o 7 wejściach można zaobserwować następujące zalety rozmytego systemu hierarchicznego:

- **redukcję liczby reguł** – połączenie trzech systemów o liczbie reguł: 81, 12 i 9 odpowiada 2187 regułom jednego systemu wnioskującego,
- **zrównoważenie mocy decyzyjnej cech statycznych i dynamicznych** – wprowadzenie syntetycznych zmiennych lingwistycznych pozwala na wyważenie decyzji pomiędzy kategoriami atrybutów w przypadku ich różnej liczebności,
- **łatwość modyfikacji** – wprowadzenie zmian w jednym z komponentów systemu nie powoduje konieczności ingerencji w inne komponenty, czy też przeprojektowania całego układu wnioskującego,
- **czytelność i łatwość analizy bazy wiedzy.**

### 4.4.1. Układ Static

Zadaniem rozmytego układu wnioskującego Static jest określenie pozy w jakiej znajduje się człowiek. Układ ten posiada 4 zmienne wejściowe, zaś każda z nich posiada trzy funkcje przynależności:

- wysoka wartość cechy (ang. *high*) – charakterystyczna dla sylwetki stojącego człowieka,
- średnia wartość cechy (ang. *medium*) – charakterystyczna dla sylwetki człowieka siedzącego, schylonego bądź kucającego,
- niska wartość cechy (ang. *low*) – charakterystyczna dla sylwetki człowieka w pozycji leżącej.

Parametry funkcji przynależności dobrano automatyczne posługując się algorytmem Fuzzy c-means. Parametry te określono w wyniku sklasteryzowania 600 obrazów przedstawiających postacie w różnych pozach, w tym także podczas upadku i w trakcie wykonywania czynności ADL. Wspomniane zmienne lingwistyczne reprezentują następujące deskryptory:

- $H/W$ – stosunek wysokości do szerokości wydzielonej postaci, wyznaczony na podstawie map głębi (zob. rysunek 4.7),
- $H/H_{max}$ – stosunek wysokości wydzielonej postaci w danej klatce obrazu do jej rzeczywistej wysokości w postawie wyprostowanej, wyznaczony na podstawie chmury punktów (zob. rysunek 4.8),
- $max(\sigma_x, \sigma_z)$ – maksymalne odchylenie standardowe wartości punktów należących do postaci od jej środka geometrycznego, wzdłuż osi X i Z układu współrzędnych kamery *Kinect* (zob. rysunek 4.9),
- $P_{40}$ – stosunek liczby punktów należących do postaci, leżących w prostopadłościanie o wysokości 40 cm umieszczonym nad podłogą, do liczby wszystkich punktów należących do postaci (zob. rysunek 4.10).

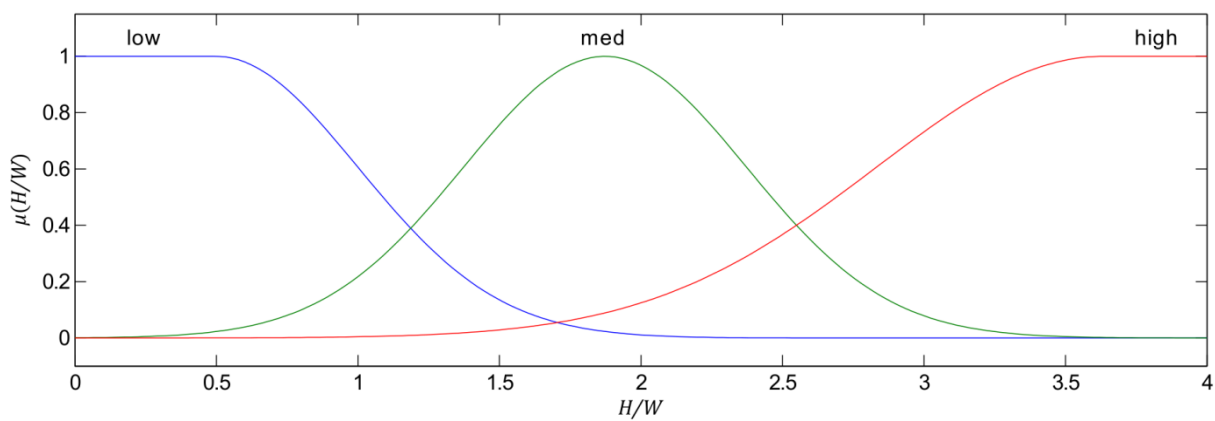

Rysunek 4.7. Funkcje przynależności dla zmiennej $H/W$.

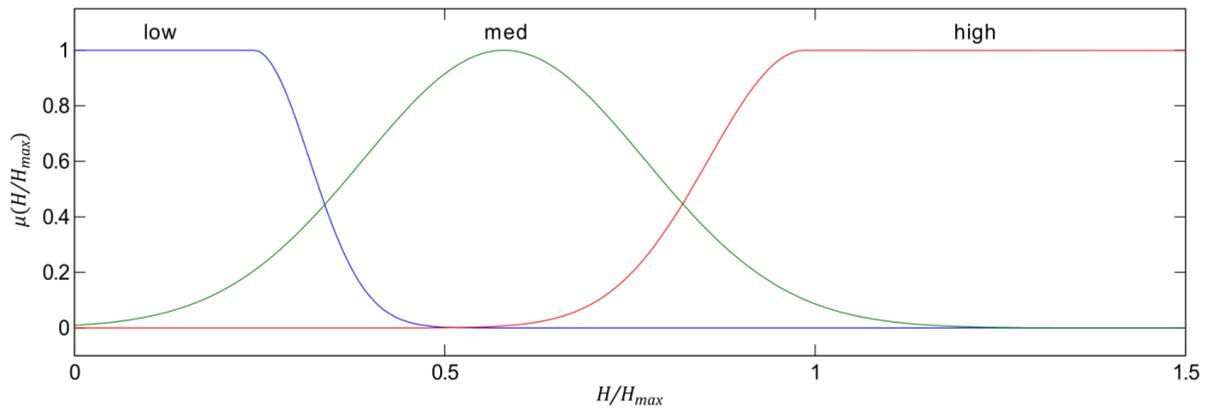

Rysunek 4.8. Funkcje przynależności dla zmiennej $H/H_{max}$.

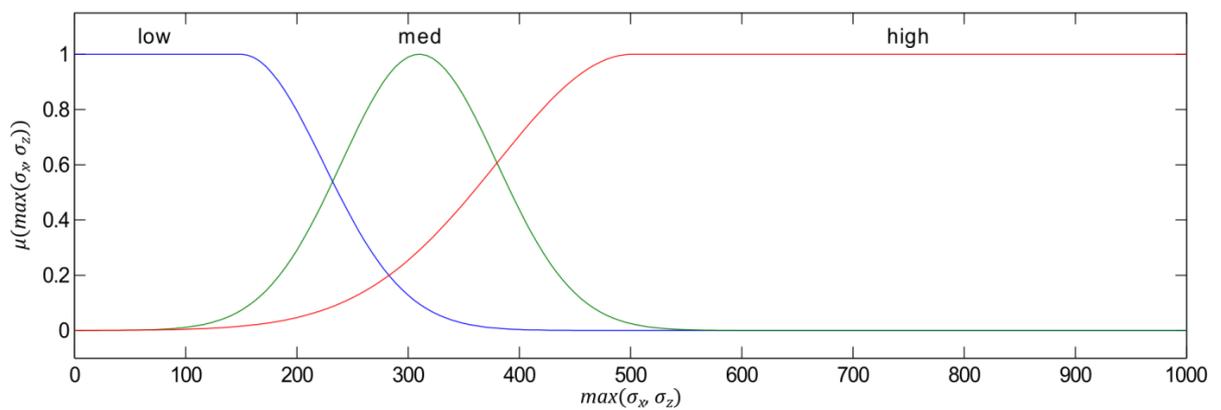

Rysunek 4.9. Funkcje przynależności dla zmiennej $max(\sigma_x, \sigma_z)$.

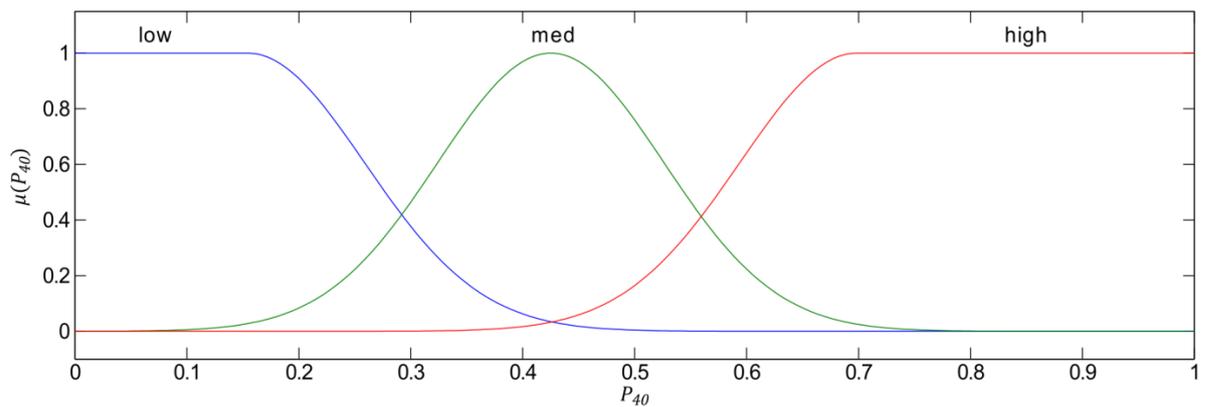

Rysunek 4.10. Funkcje przynależności dla zmiennej $P_{40}$.

### 4.4.2. Układ Transition

Zadaniem rozmytego układu wnioskującego Transition jest zamodelowanie charakterystyki ruchu postaci, towarzyszącego zmianie pozy sylwetki człowieka. Układ ten posiada 3 zmienne wejściowe, wśród których dwie zmienne reprezentują deskryptory uzyskane na podstawie map głębi, zaś jedna z nich reprezentuje dane z sensora inercyjnego. Omawiany układ operuje na następujących deskryptorach:

- $H(t)/H(t-\Delta t)$ – stosunek wysokości sylwetki postaci w czasie $t$ do wysokości w czasie $t - \Delta t$, gdzie $\Delta$ jest parametrem wyznaczonym eksperymentalnie (zob. rysunek 4.11),
- $D(t)/D(t-\Delta t)$ – stosunek odległości punktu ciężkości sylwetki postaci do płaszczyzny podłogi w czasie $t$ do odległości w czasie $t - \Delta t$, gdzie $\Delta$ jest parametrem wyznaczonym eksperymentalnie (zob. rysunek 4.12),
- $SV_{total}$ – wartość przyspieszenia uzyskana z sensora inercyjnego w czasie $t$, wyznaczona na podstawie zależności (1.7) (zob. rysunek 4.13).

Dla zmiennych wejściowych $H(t)/H(t-\Delta t)$ oraz $D(t)/D(t-\Delta t)$ zaprojektowano następujące funkcje przynależności:

- wysoka wartość cechy – charakterystyczna dla postaci wykonującej wolny ruch w kierunku płaszczyzny podłogi,
- niska wartość cechy – charakterystyczna dla postaci wykonującej szybki ruch w kierunku płaszczyzny podłogi.

Z kolei wartości zmiennej wejściowej $SV_{total}$ mogą przynależeć do trzech zbiorów rozmytych:

- wysoka wartość przyspieszenia,
- średnia wartość przyspieszenia,
- niska wartość przyspieszenia.

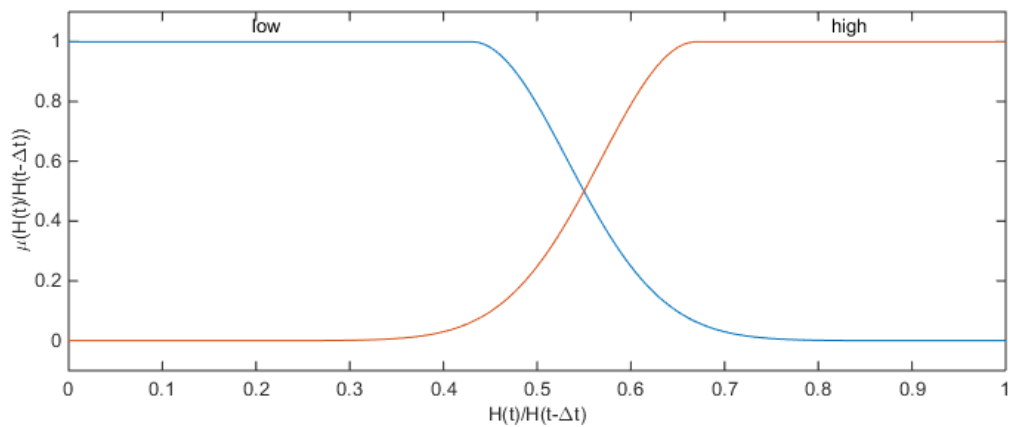

Rysunek 4.11. Funkcje przynależności dla zmiennej $H(t)/H(t-\Delta t)$.

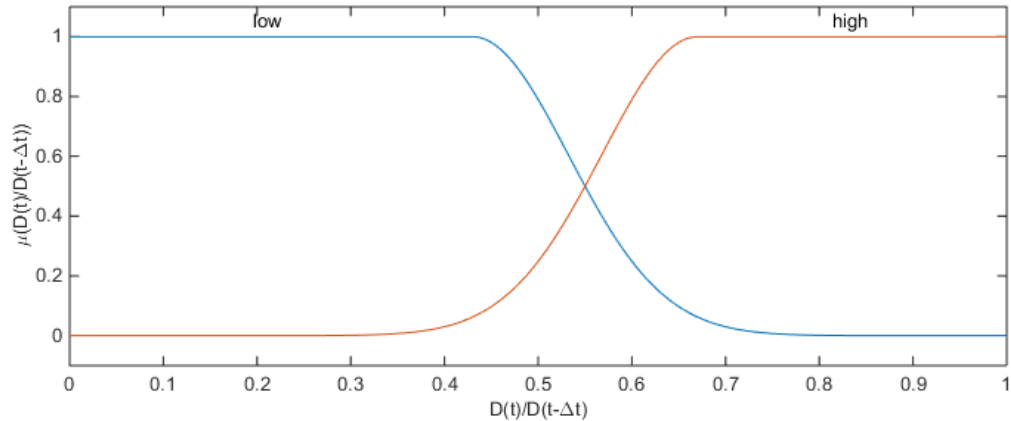

Rysunek 4.12. Funkcje przynależności dla zmiennej $D(t)/D(t-\Delta t)$.

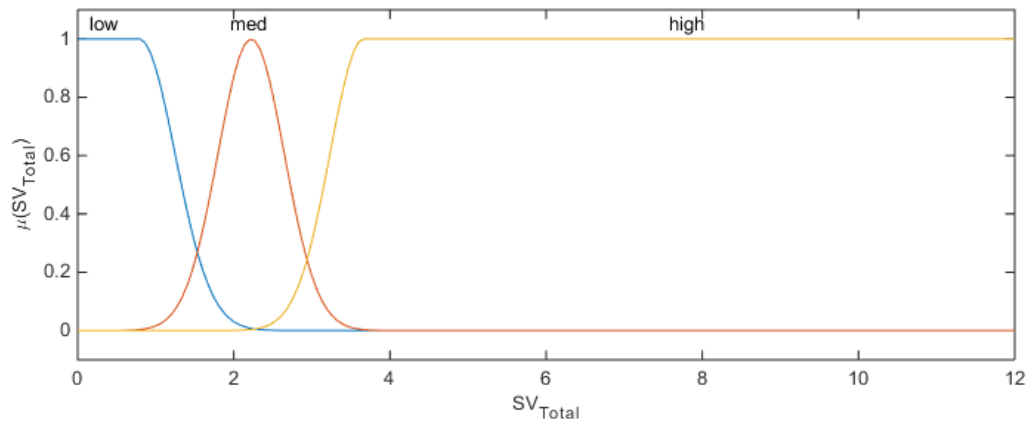

Rysunek 4.13. Funkcje przynależności dla zmiennej $SV_{total}$.

### 4.4.3. Układ Decision

Układ wnioskujący Decision jest typu Takagi-Sugeno i odpowiada za podejmowanie decyzji o tym czy dana akcja kończy się upadkiem czy też jest czynnością dnia codziennego. Operuje on na dwóch zmiennych wejściowych, które reprezentują wyjścia układów Pose i Transition. Zmienna Pose posiada następujące funkcje przynależności:

- isLy – człowiek znajduje się w pozie leżącej,
- mayLy – możliwe, że człowiek znajduje się w pozie leżącej lub pozie tymczasowej, charakterystycznej dla sylwetki znajdującej się w trakcie zmiany pozycji,
- notLy – poza charakterystyczna dla osoby stojącej, chodzącej lub siedzącej.

Opracowane funkcje przynależności dla zmiennych wejściowych przedstawiono na rysunkach 4.14 i 4.15.

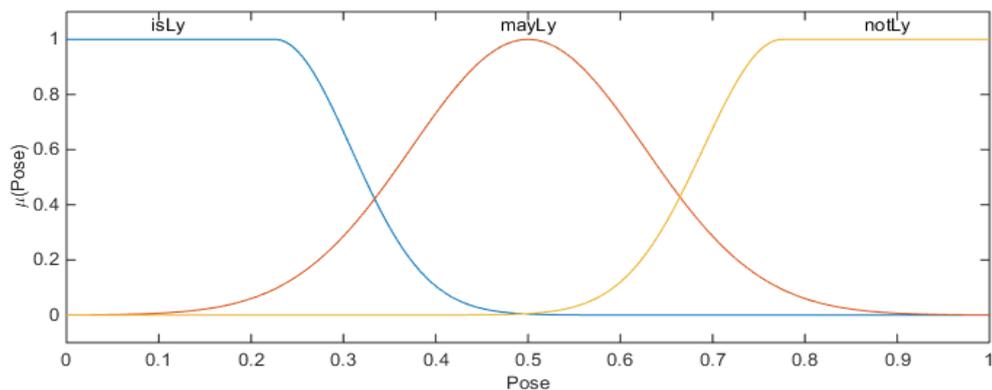

Rysunek 4.14. Funkcje przynależności dla zmiennej Pose.

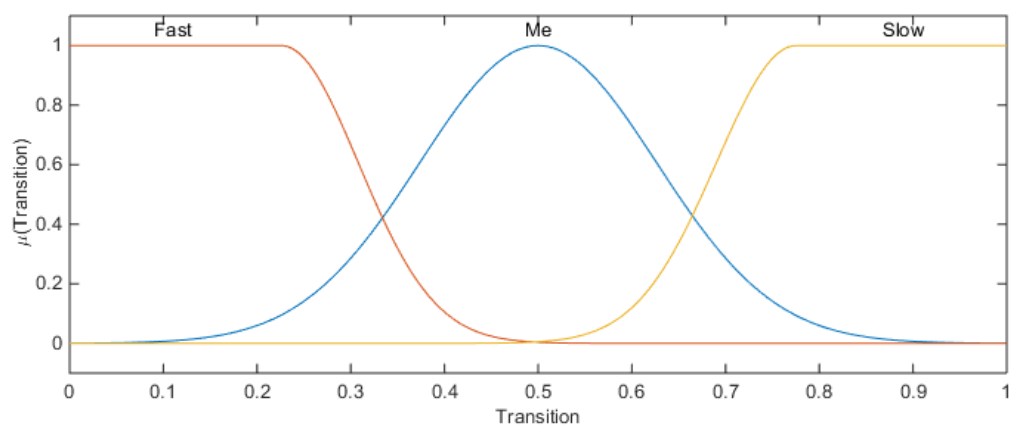

Rysunek 4.15. Funkcje przynależności dla zmiennej Transition.

## 4.5. Wyniki badań

W celu określenia skuteczności działania rozmytego systemu wnioskującego dla detekcji upadku, przeprowadzono badania eksperymentalne na danych ze zbioru UR Fall Detection Dataset. W trakcie projektowania układu wnioskującego celem oceny jakości użytych deskryptorów jak i skuteczności wnioskowania w oparciu o przygotowaną przez eksperta bazę reguł, wpierw przebadano układ Static, wnioskujący o pozie sylwetki człowieka. Dzięki dodaniu na wyjściu układu bloku wyostrzania, możliwe było uzyskanie binarnych wyników klasyfikacji pozy oraz porównanie ich z wynikami uzyskanymi za pomocą klasyfikatora SVM (Kwolek & Kępski, 2014). Do ewaluacji skuteczności detekcji wspomnianych metod wybrano 2425 obrazów z bazy URFD i innych sekwencji zarejestrowanych w typowych pomieszczeniach domowych oraz biurowych. Wybrany zbiór składał się z 1522 obrazów przedstawiających sylwetkę człowieka podczas wykonywania czynności ADL, takich jak chodzenie, siadanie, schylanie się oraz kucanie. We wspomnianym zbiorze danych liczba obrazów przedstawiających osobę leżącą na podłodze była równa 903. Macierz pomyłek dla klasyfikacji pozy sylwetki człowieka przy użyciu układu Static przedstawiono w tabeli 4.1.

Tabela 4.1. Macierz pomyłek dla klasyfikacji pozy sylwetki człowieka przy użyciu rozmytego systemu wnioskującego.

|  |  | **Rzeczywiste pozy osoby** | | |
|---|---|---|---|---|
|  |  | **osoba w pozie leżącej** | **osoba w pozie nieleżącej** |  |
| **Uzyskane wyniki** | **osoba w pozie leżącej** | 903 | 19 | Dokładność = 99,22% |
|  | **osoba w pozie nieleżącej** | 0 | 1503 | Precyzja = 97,94% |
|  |  | Czułość = 100% | Swoistość = 98,75% |  |

Jak można zauważyć w tabeli 4.1, klasyfikacja pozy sylwetki za pomocą rozmytego układu wnioskującego Static charakteryzuje się wysoką czułością, lecz nie jest pozbawiona błędów pierwszego rodzaju. Celem porównania układu rozmytego i układu klasycznego zbudowanego w oparciu o SVM, zrealizowano dodatkowe badania porównawcze, których wyniki zebrano w tabeli 4.2.

Tabela 4.2. Porównanie wyników klasyfikacji pozy sylwetki człowieka uzyskanych przy użyciu rozmytego systemu wnioskującego i klasyfikatora SVM.

|  |  | **Rozmyty system wnioskujący** | **SVM** (Kwolek & Kępski, 2014) |
|---|---|---|---|
| **Uzyskane wyniki** | **Dokładność** | 99,22% | 99,67% |
|  | **Precyzja** | 97,94% | 100% |
|  | **Czułość** | 100,00% | 99,05% |
|  | **Swoistość** | 98,75% | 100% |

Jak już wspomniano wcześniej, oprócz wnioskowania o pozie sylwetki człowieka, system wykorzystuje cechy opisujące ruch postaci i modeluje charakter tego ruchu w oparciu o trzy zbiory rozmyte. Mając na względzie to, że upadek jest akcją dynamiczną, której towarzyszy nagła zmiana orientacji ciała ludzkiego, takie deskryptory mogą być przydatne w procesie wnioskującym. Rysunek 4.16 przedstawia przebieg wartości cechy $H/H_{max}$ w czasie dla przykładowej sekwencji ze zbioru URFD.

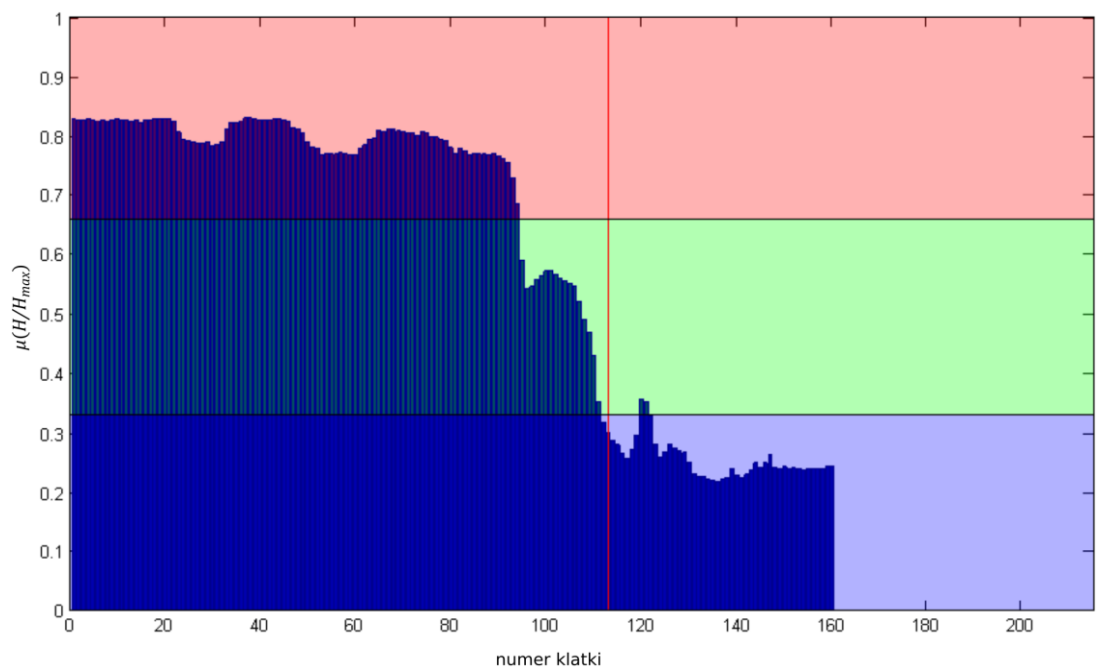

Rysunek 4.16. Wartość cechy $H/H_{max}$ dla przykładowej sekwencji ze zbioru URFD. Czerwoną pionową linią zaznaczono moment uderzenia ciała osoby o podłoże.

Jak można zauważyć na rysunku 4.16 wartość cechy $H/H_{max}$ gwałtownie spada tuż przed momentem uderzenia ciała osoby o podłoże. W badaniach eksperymentalnych wykorzystywano dwie wspomniane wcześniej cechy: $H(t)/H(t-\Delta t)$ oraz $D(t)/D(t-\Delta t)$. Dobrano także eksperymentalnie próg $\Delta t$. Po przeanalizowaniu przedziału wartości (450, 1000), okazało się, że wartość $\Delta t$ równa 700 ms zapewnia kompromis między czułością a swoistością. Aby zwiększyć moc dyskryminacyjną układu wnioskującego dla czynności ADL, które są wykonywane szybko i charakteryzują się nagłym spadkiem wartości cechy $H/H_{max}$, przyjęto, że trzecią zmienną wejściową będzie wartość przyspieszenia zarejestrowana przez noszony przez użytkownika sensor inercyjny.

W dalszej fazie badań eksperymentalnych oceniano skuteczność detekcji upadku w oparciu jedynie o cechy dynamiczne (wykorzystywane przez układ Transition) oraz skuteczność hierarchicznego rozmytego systemu wnioskującego. Uzyskane wyniki zestawiono w tabeli 4.3. Jak można zauważyć, oba układy charakteryzują się wysoką skutecznością, jednak połączenie układów Static i Transition w układ hierarchiczny, a w konsekwencji wprowadzenie nowych zmiennych lingwistycznych doprowadziło do polepszenia pozostałych parametrów: dokładności, precyzji i swoistości.

W tabeli 4.4. zestawiono wyniki kilku metod detekcji upadku, które uzyskano na danych z bazy UR Fall Detection Dataset. Oprócz wyników uzyskanych przy użyciu rozmytego systemu wnioskującego, zaprezentowano w niej także wyniki uzyskane w oparciu o klasyfikator SVM i wygenerowany na podstawie danych uczących układ wnioskujący ANFIS.

Tabela 4.3. Porównanie wyników detekcji upadku przy użyciu rozmytego układu wnioskującego Transition i hierarchicznego rozmytego systemu wnioskującego Static + Transition + Decision. Do wyznaczenia skuteczności wykorzystano bazę UR Fall Detection Dataset (sekwencje 1-70).

|  |  | **Static + Transition + Decision** | **Transition** |
|---|---|---|---|
| **Uzyskane wyniki** | **Dokładność** | 97,14% | 92,86% |
|  | **Precyzja** | 93,75% | 85,71% |
|  | **Czułość** | 100,00% | 100,00% |
|  | **Swoistość** | 95,00% | 87,50% |

Metody SVM i ANFIS operowały na takim samym zestawie cech jak projektowany system wnioskujący: $H/H_{max}$, $H/W$, $max(\sigma_x, \sigma_z)$, $P_{40}$, $H(t)/H(t - \Delta t)$, $D(t)/D(t - \Delta t)$ oraz $SV_{total}$. W celach porównawczych w tabeli ujęto także wyniki dla metod pokrewnych, które można uznać za reprezentatywne.

Tabela 4.4. Porównanie wyników kilku metod detekcji upadku dla bazy UR Fall Detection Dataset (sekwencje 1-70).

|  |  | **Metoda** | | | | |
|---|---|---|---|---|---|---|
|  |  | **Rozmyty system wnioskujący** | **SVM (Kwolek & Kępski, 2015)** | **ANFIS** | ***Treshold UFT* (Bourke et al., 2007)** | ***Treshold LFT* (Bourke et al., 2007)** |
| **Uzyskane wyniki** | **Dokładność** | 97,14% | 95,71% | 95,71% | 88,57% | 78,57% |
|  | **Precyzja** | 93,75% | 90,90% | 90,90% | 78,95% | 68,29% |
|  | **Czułość** | 100,00% | 100,00% | 100,00% | 100,00% | 93,33% |
|  | **Swoistość** | 95,00% | 92,50% | 92,50% | 80,00% | 67,50% |

Jak można zauważyć, zarówno metody SVM jak i ANFIS charakteryzują się wysoką czułością detekcji upadku. Jednak ich swoistość w porównaniu do zaprojektowanego hierarchicznego rozmytego systemu wnioskującego jest niższa, co w konsekwencji prowadzi do większej liczby fałszywych alarmów. Warto wspomnieć, że liczba reguł systemu ANFIS wynosi 256, co z kolei przekłada się na większą trudność analizy bazy wiedzy przez eksperta w porównaniu do bazy wiedzy projektowanego systemu, która zawiera jedynie 102 reguły.

## 4.6. Podsumowanie

Istotnym ograniczeniem, z którym spotykają się projektanci systemów do detekcji upadku jest brak wystarczającej liczby danych uczących, w szczególności rejestrowanych w rzeczywistych, niesymulowanych warunkach. Co więcej, zbudowanie bazy danych, która uwzględniałaby wszystkie możliwe scenariusze upadku jest zadaniem trudnym i kosztownym. Mając na względzie powyższe ograniczenia, zbudowanie dobrze generalizującego detektora nie jest zadaniem trywialnym. W wyniku podjętych badań opracowano rozwiązanie, które uzyskało wysoką czułość i swoistość dla bazy wiedzy o niewielkiej liczbie reguł. Liczba reguł jest na tyle mała, że możliwe jest przeanalizowanie tych reguł przez osoby zajmujące się detekcją upadku.

# Podsumowanie

W pracy opracowano i przebadano metody detekcji upadku człowieka na podstawie sekwencji map głębi oraz danych pochodzących z sensora inercyjnego. Zagadnienie to nie należy do łatwych, lecz jest często podejmowane przez różne grupy badawcze ze względu na zapotrzebowanie na automatyczne systemy detekcji upadku. Mając na względzie znaczący wzrost długości życia w krajach rozwiniętych oraz to, że rozwiązania komercyjne nie są pozbawione wad można stwierdzić, że zagadnienie detekcji upadku jest problemem aktualnym. Jak wskazują badania, czynnikami najbardziej frustrującymi seniorów, wykorzystujących tradycyjne systemy detekcji upadku, jest liczba fałszywych alarmów. Rozwiązania takie wymagają ciągłej interwencji użytkownika przez co nie zostały w pełni zaakceptowane przez swoją grupę docelową. Jednocześnie ograniczenie liczby fałszywych alarmów może niekiedy wiązać się ze spadkiem czułości takiego systemu, co mogłoby doprowadzić do sytuacji, w której część upadków zostałby przez system niewykryta.

W pracy szczególną uwagę poświęcono przebadaniu algorytmów, które umożliwiłyby efektywną detekcję upadku, przy zachowaniu wysokiej swoistości. Zaprezentowano i porównano dwie metody: metodę opartą o uczenie z nadzorem oraz metodę wykorzystującą wnioskowanie rozmyte. Metody te pozwoliły uzyskać wskaźniki jakościowe świadczące o wysokiej czułości i swoistości detekcji upadku, co zwiększa ich potencjał aplikacyjny. Zaproponowane metody zostały zaprojektowane tak, aby możliwe było uruchomienie ich na platformie obliczeniowej z procesorem w architekturze ARM. Oparcie systemu o dane z powszechnie dostępnego sensora głębi *Microsoft Kinect* pozwala na uniknięcie konieczności kalibracji i zmniejszenie kosztów instalacji. Dzięki zaproponowaniu szeregu modyfikacji algorytmów wstępnego przetwarzania obrazów, uzyskano rozwiązania pozwalające na detekcję i śledzenie osoby oraz detekcję upadku w czasie rzeczywistym. Dzięki inercyjnemu sensorowi ruchu noszonemu przez użytkownika uzyskano możliwość segmentacji w czasie wykonywanych akcji. Segmentacja ta jest szczególnie przydatna w procesie detekcji upadku, gdyż umożliwia określenie momentu uderzenia postaci o podłoże. Zaproponowano także algorytmy aktualizacji modelu tła, wykorzystujące informacje o kontekście w jakim znajduje się postać, pozyskane na podstawie analizy danych z sensora ruchu. Pozwalają one na wykrycie okresów nieaktywności użytkownika, co z kolei ułatwia budowę reprezentatywnego modelu tła. Skuteczność opracowanych algorytmów określono na podstawie przygotowanej bazy danych, zawierającej dane zarejestrowane przy użyciu dwóch kamer oraz inercyjnego sensora ruchu. Opracowane algorytmy zostały zaimplementowane w językach Matlab oraz C++. Do najważniejszych osiągnięć można zaliczyć:

- opracowanie oraz zaimplementowanie algorytmów detekcji upadku oparte o uczenie z nadzorem; metody te charakteryzują się wysoką czułością i małą liczbą fałszywych alarmów, a także pozwalają na implementację systemu na platformie obliczeniowej z procesorem ARM; w ramach prac przygotowano zestaw cech, zbiory uczące i testujące, a także przedstawiono wyniki klasyfikacji;
- opracowanie rozmytego systemu detekcji upadku będącego hierarchią trzech układów wnioskujących, charakteryzującego się wysoką czułością i specyficznością, a także możliwością łatwej analizy bazy wiedzy przez eksperta; w ramach prac zaprojektowano zmienne lingwistyczne, bazę reguł, przebadano system i porównano wyniki z metodami opartymi o uczenie z nadzorem, a także inne pokrewne algorytmy detekcji upadku;
- opracowanie architektury systemu uwzględniającej możliwości obliczeniowe platformy z procesorem ARM; zaproponowanie wykorzystania bufora cyklicznego oraz wstępnej analizy danych, pozwalające na uniknięcie przetwarzania wszystkich obrazów *on-line*;
- przygotowanie zestawu algorytmów i ich modyfikacji na potrzeby sytemu detekcji upadku: algorytmów detekcji podłogi (metodami *v-disparity* i RANSAC), śledzenia postaci w oparciu o filtr cząsteczkowy, zmodyfikowanej metody rozrostu obszarów oraz metody budowy modelu tła;
- przygotowanie i udostępnienie w sieci web zbioru sekwencji UR Fall Detection Dataset, który posłużył ocenie skuteczności detekcji metod opracowanych w niniejszej pracy; zbiór ten jest jedynym publicznie dostępnym zbiorem łączącym sekwencje obrazów z danymi pochodzącymi z inercyjnego sensora ruchu;
- przygotowanie aktywnej głowicy obrotowej *pan/tilt* dedykowanej dla kamery *Microsoft Kinect* i opracowanie oprogramowania sterownika działającego na mikrokontrolerze Arduino.

Przeprowadzone eksperymenty oraz zaprezentowane wyniki detekcji upadku, a także ich analiza dowodzą słuszności tezy pracy, co oznacza, że algorytmy wykorzystujące obrazy głębi pozwalają na uzyskanie wskaźników jakościowych świadczących o wysokiej czułości i swoistości detekcji upadku. Wykorzystanie informacji o kontekście poprawia skuteczność detekcji upadku.

Dalsze badania skupią się na rozwoju rozmytego systemu detekcji upadku oraz nad poprawą skuteczności i efektywności algorytmów opartych o uczenie nadzorowane. Część uwagi zostanie poświęcona opracowaniu nowych deskryptorów opisujących dane z inercyjnego sensora ruchu. Planowane jest także opracowanie efektywnych rozwiązań dla detekcji upadku w scenariuszach z przysłonięciami.

# Bibliografia

A. Sixsmith, N. Johnson & R. Whatmore (2005) Pyroelectric IR sensor arrays for fall detection in the older population. *J. Phys. IV France*, 128, pp. 153–160.

Adams, R. & Bischof, L. (1994) Seeded region growing. *IEEE Transactions on Pattern Analysis and Machine Intelligence*, 16 (6), pp. 641–647.

Aggarwal, J.K. & Ryoo, M.S. (2011) Human Activity Analysis: A Review. *ACM Comput. Surv.*, 43 (3), pp. 16:1–16:43.

Aggarwal, J.K. & Xia, L. (2014) Human activity recognition from 3D data: A review. *Pattern Recognition Letters*, 48, pp. 70–80.

Anderson, D., Keller, J.M., Skubic, M., Chen, X. & He, Z. (2006) Recognizing falls from silhouettes. In: *Annual International Conference of the IEEE Engineering in Medicine and Biology Society*. IEEE, pp. 6388–91.

Anderson, D., Luke, R.H., Keller, J.M., Skubic, M., Rantz, M. & Aud, M. (2009) Linguistic Summarization of Video for Fall Detection Using Voxel Person and Fuzzy Logic. *Computer Vision and Image Understanding*, 113 (1), pp. 80–89.

Arsic, D., Lyutskanov, A., Rigoll, G. & Kwolek, B. (2009) Multi camera person tracking applying a graph-cuts based foreground segmentation in a homography framework. In: *IEEE International Workshop on Performance Evaluation of Tracking and Surveillance*. IEEE, pp. 1–8.

Arya, S., Mount, D.M., Netanyahu, N.S., Silverman, R. & Wu, A.Y. (1998) An optimal algorithm for approximate nearest neighbor searching fixed dimensions. *Journal of the ACM*, 45 (6), pp. 891–923.

Auvinet, E., Multon, F., Saint-Arnaud, A., Rousseau, J. & Meunier, J. (2011) Fall detection with multiple cameras: an occlusion-resistant method based on 3-D silhouette vertical distribution. *IEEE Transactions on Information Technology in Biomedicine*, 15 (2), pp. 290–300.

Auvinet, E., Rougier, C., Meunier, J., St-Arnaud, A. & Rousseau, J. (2010) *Multiple cameras fall data set*. Montréal (Québec).

Bagalà, F., Becker, C., Cappello, A., Chiari, L., Aminian, K., Hausdorff, J.M., Zijlstra, W. & Klenk, J. (2012) Evaluation of accelerometer-based fall detection algorithms on real-world falls. *PloS ONE*, 7 (5), p.e37062.

Bentley, J.L. (1975) Multidimensional binary search trees used for associative searching. *Communications of the ACM*, 18 (9), pp. 509–517.

Besl, P.J. & McKay, N.D. (1992) Method for registration of 3-D shapes. In: P. S. Schenker ed. *Sensor Fusion IV: Control Paradigms and Data Structures*. International Society for Optics and Photonics, pp. 586–606.

Black, M.J., Yacoob, Y., Jepson, A.D. & Fleet, D.J. (1997) Learning parameterized models of image motion. In: *IEEE Conference on Computer Vision and Pattern Recognition*. IEEE, pp. 561–567.

Bobick, A.F. & Davis, J.W. (2001) The recognition of human movement using temporal templates. *IEEE Transactions on Pattern Analysis and Machine Intelligence*, 23 (3), pp. 257–267.

Boissy, P., Choquette, S., Hamel, M. & Noury, N. (2007) User-based motion sensing and fuzzy logic for automated fall detection in older adults. *Telemedicine Journal and e-Health*, 13 (6), pp. 683–93.

Bourke, A.K. & Lyons, G.M. (2008) A threshold-based fall-detection algorithm using a bi-axial gyroscope sensor. *Medical Engineering & Physics*, 30 (1), pp. 84–90.

Bourke, A.K., O'Brien, J. V & Lyons, G.M. (2007) Evaluation of a threshold-based tri-axial accelerometer fall detection algorithm. *Gait & Posture*, 26 (2), pp. 194–9.

Bourke, A.K., van de Ven, P., Gamble, M., O'Connor, R., Murphy, K., Bogan, E., McQuade, E., Finucane, P., Olaighin, G. & Nelson, J. (2010) Evaluation of waist-mounted tri-axial accelerometer based fall-detection algorithms during scripted and continuous unscripted activities. *Journal of Biomechanics*, 43 (15), pp. 3051–7.

Brownsell, S.J., Bradley, D.A., Bragg, R., Catlin, P. & Carlier, J. (2000) Do community alarm users want telecare? *Journal of Telemedicine and Telecare*, 6 (4), pp. 199–204.

Brzoza-Woch, R., Ruta, A. & Zieliński, K. (2013) Remotely reconfigurable hardware–software platform with web service interface for automated video surveillance. *Journal of Systems Architecture*, 59 (7), pp. 376–388.

Chen, C. & Shi, G. (2013) A color-guided, region-adaptive and depth-selective unified framework for Kinect depth recovery. In: *IEEE International Workshop on Multimedia Signal Processing*. IEEE, pp. 007–012.

Chen, J., Kwong, K., Chang, D., Luk, J. & Bajcsy, R. (2005) Wearable sensors for reliable fall detection. In: *Annual International Conference of the IEEE Engineering in Medicine and Biology Society*. IEEE, pp. 3551–4.

Christensen, K., Doblhammer, G., Rau, R. & Vaupel, J.W. (2009) Ageing populations: the challenges ahead. *Lancet*, 374 (9696), pp. 1196–208.

Chum, O. & Matas, J. (2005) Matching with PROSAC — Progressive Sample Consensus. In: *IEEE Conference on Computer Vision and Pattern Recognition*. IEEE, pp. 220–226.

Comaniciu, D. & Meer, P. (2002) Mean shift: a robust approach toward feature space analysis. *IEEE Transactions on Pattern Analysis and Machine Intelligence*, 24 (5), pp. 603–619.

Cootes, T.F., Taylor, C.J., Cooper, D.H. & Graham, J. (1995) Active Shape Models-Their Training and Application. *Computer Vision and Image Understanding*, 61 (1), pp. 38–59.

Cortes, C. & Vapnik, V. (1995) Support-Vector Networks. *Machine Learning*, 20 (3), pp. 273–297.

Cover, T. & Hart, P. (1967) Nearest neighbor pattern classification. *IEEE Transactions on Information Theory*, 13 (1), pp. 21–27.

Cover, T.M. & Thomas, J.A. (1991) *Elements of information theory*. Wiley-Interscience.

Cristianini, N. & Shawe-Taylor, J. (1999) An introduction to support Vector Machines: and other kernel-based learning methods.

Cucchiara, R., Prati, A. & Vezzani, R. (2007) A multi-camera vision system for fall detection and alarm generation. *Expert Systems*, 24 (5), pp. 334–345.

Cupillard, F., Bremond, F. & Thonnat, M. (2002) Group behavior recognition with multiple cameras. In: *IEEE Workshop on Applications of Computer Vision*. IEEE, pp. 177–183.

Cyganek, B. (2011) One-Class Support Vector Ensembles for Image Segmentation and Classification. *Journal of Mathematical Imaging and Vision*, 42 (2-3), pp. 103–117.

Dai, P., Di, H., Dong, L., Tao, L. & Xu, G. (2008) Group interaction analysis in dynamic context. *IEEE Transactions on Systems, Man, and Cybernetics. Part B, Cybernetics*, 39 (1), pp. 34 – 42.

Dalal, N. & Triggs, B. (2005) Histograms of Oriented Gradients for Human Detection. In: *IEEE Conference on Computer Vision and Pattern Recognition*. IEEE, pp. 886–893.

Dollar, P., Belongie, S. & Perona, P. (2010) The Fastest Pedestrian Detector in the West. In: *BMVC*. BMVA Press, pp. 68.1–68.11.

Dollar, P., Tu, Z., Tao, H. & Belongie, S. (2007) Feature Mining for Image Classification. In: *IEEE Conference on Computer Vision and Pattern Recognition*. IEEE, pp. 1–8.

Duda, R.O. & Hart, P.E. (1972) Use of the Hough transformation to detect lines and curves in pictures. *Communications of the ACM*, 15 (1), pp. 11–15.

Fahrenberg, J., Foerster, F., Smeja, M. & Müller, W. (1997) Assessment of posture and motion by multichannel piezoresistive accelerometer recordings. *Psychophysiology*, 34 (5), pp. 607–12.

Fiorio, C. & Gustedt, J. (1996) Two linear time Union-Find strategies for image processing. *Theoretical Computer Science*, 154 (2), pp. 165–181.

Fischler, M.A. & Bolles, R.C. (1981) Random sample consensus: a paradigm for model fitting with applications to image analysis and automated cartography. *Communications of the ACM*, 24 (6), pp. 381–395.

Fuller, G.F. (2000) Falls in the elderly. *American Family Physician*, 61 (7), pp. 2159–68, 2173–4.

Gödel, K. (1932) Zum Intuitionistischen Aussagenkalkül. *Anzieger Akademie der Wissenschaften Wien, Math. - naturwissensch*, 69, pp. 65–66.

Gordon, N.J., Salmond, D.J. & Smith, A.F.M. (1993) Novel approach to nonlinear/non-Gaussian Bayesian state estimation. *Radar and Signal Processing, IEE Proceedings F*, 140 (2), pp. 107–113.

Gottwald, S. (1999) Many-Valued Logic And Fuzzy Set Theory. In: U. Höhle & S. E. Rodabaugh eds. *Mathematics of Fuzzy Sets*. The Handbooks of Fuzzy Sets Series. Boston, MA, Springer US, pp. 5–89.

Guillaume, S. (2001) Designing fuzzy inference systems from data: An interpretability-oriented review. *IEEE Transactions on Fuzzy Systems*, 9 (3), pp. 426–443.

Hall, M., Frank, E., Holmes, G., Pfahringer, B., Reutemann, P. & Witten, I.H. (2009) The WEKA data mining software. *ACM SIGKDD Explorations Newsletter*, 11 (1), p.10.

Heinrich, S., Rapp, K., Rissmann, U., Becker, C. & König, H.-H. (2010) Cost of falls in old age: a systematic review. *Osteoporosis International*, 21 (6), pp. 891–902.

Hwang, J.Y., Kang, J.M., Jang, Y.W. & Kim, H. (2004) Development of novel algorithm and real-time monitoring ambulatory system using Bluetooth module for fall detection in the elderly. *Annual International Conference of the IEEE Engineering in Medicine and Biology Society*, 3, pp. 2204–7.

Igual, R., Medrano, C. & Plaza, I. (2013) Challenges, issues and trends in fall detection systems. *BioMedical Engineering OnLine*, 12, p.66.

Jingen, L., Jiebo, L. & Shah, M. (2009) Recognizing realistic actions from videos 'in the wild'. In: *IEEE Conference on Computer Vision and Pattern Recognition*. IEEE, pp. 1996–2003.

Johansson, G. (1975) Visual motion perception. *Scientific American*, 232(6), pp. 76–88.

Kalman, R.E. (1960) A New Approach to Linear Filtering and Prediction Problems. *Journal of Basic Engineering*, 82 (1), p.35.

Kangas, M., Konttila, A., Lindgren, P., Winblad, I. & Jämsä, T. (2008) Comparison of low-complexity fall detection algorithms for body attached accelerometers. *Gait & Posture*, 28 (2), pp. 285–91.

Karantonis, D.M., Narayanan, M.R., Mathie, M., Lovell, N.H. & Celler, B.G. (2006) Implementation of a Real-Time Human Movement Classifier Using a Triaxial Accelerometer for Ambulatory Monitoring. *IEEE Transactions on Information Technology in Biomedicine*, 10 (1), pp. 156–167.

Kasprzak, W., Wilkowski, A. & Czapnik, K. (2012) Hand gesture recognition based on free-form contours and probabilistic inference. *International Journal of Applied Mathematics and Computer Science*, 22 (2), pp. 437–448.

Keijsers, N.L.W., Horstink, M.W.I.M. & Gielen, S.C.A.M. (2003) Movement parameters that distinguish between voluntary movements and levodopa-induced dyskinesia in Parkinson's disease. *Human Movement Science*, 22 (1), pp. 67–89.

Kellokumpu, V., Zhao, G. & Pietikäinen, M. (2008) Human activity recognition using a dynamic texture based method. In: *BMVC*.

Kennedy, J. & Eberhart, R. (1995) Particle swarm optimization. In: *International Conference on Neural Networks*. IEEE, pp. 1942–1948.

Kępski, M. & Kwolek, B. (2014a) Detecting human falls with 3-axis accelerometer and depth sensor. In: *Annual International Conference of the IEEE Engineering in Medicine and Biology Society*. IEEE, pp. 770–773.

Kępski, M. & Kwolek, B. (2015) Embedded system for fall detection using body-worn accelerometer and depth sensor. In: *IEEE International Conference on Intelligent Data Acquisition and Advanced Computing Systems: Technology and Applications*. IEEE, pp. 755–759.

Kępski, M. & Kwolek, B. (2016a) Event-driven system for fall detection using body-worn accelerometer and depth sensor. *IEEE Transactions on Instrumentation and Measurement*, submitted – selected papers from *IDAACS2015 Conf*.

Kępski, M. & Kwolek, B. (2012a) Fall detection on embedded platform using kinect and wireless accelerometer. In: *Computers Helping People with Special Needs*. Lecture Notes in Computer Science. Springer Berlin Heidelberg, pp. 407–414.

Kępski, M. & Kwolek, B. (2016b) Fall Detection Using Body-worn Accelerometer and Depth Maps Acquired by Active Camera. In: *Hybrid Artificial Intelligence Systems*. Lecture Notes in Computer Science

Kępski, M. & Kwolek, B. (2014b) Fall Detection Using Ceiling-Mounted 3D Depth Camera. In: *International Conference on Computer Vision Theory and Applications*. pp. 640–647.

Kępski, M. & Kwolek, B. (2012b) Human fall detection by mean shift combined with depth connected components. In: *Computer Vision and Graphics*. Lecture Notes in Computer Science. Springer Berlin Heidelberg, pp. 457–464.

Kępski, M. & Kwolek, B. (2014c) Person Detection and Head Tracking to Detect Falls in Depth Maps. In: *Computer Vision and Graphics*. Lecture Notes in Computer Science. Springer International Publishing, pp. 324–331.

Kępski, M. & Kwolek, B. (2013) Unobtrusive Fall Detection at Home Using Kinect Sensor. In: *Computer Analysis of Images and Patterns*. Lecture Notes in Computer Science. Springer Berlin Heidelberg, pp. 457–464.

Kępski, M., Kwolek, B. & Austvoll, I. (2012) Fuzzy inference-based reliable fall detection using kinect and accelerometer. In: *Artificial Intelligence and Soft Computing*. Lecture Notes in Computer Science. Springer Berlin Heidelberg, pp. 266–273.

Khan, S.M. & Shah, M. (2005) Detecting group activities using rigidity of formation. In: *Proceedings of the ACM International Conference on Multimedia*. New York, New York, USA, ACM Press, p.403.

Khoshelham, K. & Elberink, S.O. (2012) Accuracy and resolution of Kinect depth data for indoor mapping applications. *Sensors*, 12 (2), pp. 1437–54.

Konstantinidis, E.I. & Bamidis, P.D. (2015) Density based clustering on indoor kinect location tracking: A new way to exploit active and healthy aging living lab datasets. In: *IEEE International Conference on Bioinformatics and Bioengineering*. IEEE, pp. 1–6.

Krzeszowski, T., Michalczuk, A., Kwolek, B., Switonski, A. & Josinski, H. (2013) Gait recognition based on marker-less 3D motion capture. In: *IEEE International Conference on Advanced Video and Signal Based Surveillance*. IEEE, pp. 232–237.

Kurniawan, S. (2008) Older people and mobile phones: A multi-method investigation. *International Journal of Human-Computer Studies*, 66 (12), pp. 889–901.

Kwolek, B. (2005) CamShift-Based Tracking in Joint Color-Spatial Spaces. In: *Computer Analysis of Images and Patterns*. Lecture Notes in Computer Science. Springer Berlin Heidelberg, pp. 693–700.

Kwolek, B. (2003) Visual system for tracking and interpreting selected human actions. *Journal of WSCG*, 11 (1-3).

Kwolek, B. & Kępski, M. (2013) Fall Detection Using Kinect Sensor and Fall Energy Image. In: *Hybrid Artificial Intelligent Systems*. Lecture Notes in Computer Science. Springer Berlin Heidelberg, pp. 294–303.

Kwolek, B. & Kępski, M. (2016) Fuzzy inference-based fall detection using kinect and body-worn accelerometer. *Applied Soft Computing*, 40, pp. 305–318.

Kwolek, B. & Kępski, M. (2014) Human fall detection on embedded platform using depth maps and wireless accelerometer. *Computer Methods and Programs in Biomedicine*, 117 (3), pp. 489–501.

Kwolek, B. & Kępski, M. (2015) Improving fall detection by the use of depth sensor and accelerometer. *Neurocomputing*, 168, pp. 637–645.

Labayrade, R., Aubert, D. & Tarel, J.-P. (2002) Real time obstacle detection in stereovision on non flat road geometry through 'v-disparity' representation. In: *IEEE Intelligent Vehicle Symposium*. IEEE, pp. 646–651.

Lan, T., Wang, Y., Yang, W., Robinovitch, S.N. & Mori, G. (2012) Discriminative latent models for recognizing contextual group activities. *IEEE Transactions on Pattern Analysis and Machine Intelligence*, 34 (8), pp. 1549–62.

Laptev, I. (2005) On Space-Time Interest Points. *International Journal of Computer Vision*, 64 (2-3), pp. 107–123.

Laptev, I., Marszalek, M., Schmid, C. & Rozenfeld, B. (2008) Learning realistic human actions from movies. In: *IEEE Conference on Computer Vision and Pattern Recognition*. IEEE, pp. 1–8.

Lee, C.C. (1990) Fuzzy logic in control systems: fuzzy logic controller. II. *IEEE Transactions on Systems, Man, and Cybernetics*, 20 (2), pp. 419–435.

Lee, T. & Mihailidis, A. (2005) An intelligent emergency response system: preliminary development and testing of automated fall detection. *Journal of Telemedicine and Telecare*, 11 (4), pp. 194–8.

Leekwijck, W. Van & Kerre, E.E. (1999) Defuzzification: criteria and classification. *Fuzzy Sets and Systems*, 108 (2), pp. 159–178.

Li, Q., Stankovic, J.A., Hanson, M.A., Barth, A.T., Lach, J. & Zhou, G. (2009) Accurate, Fast Fall Detection Using Gyroscopes and Accelerometer-Derived Posture Information. In: *International Workshop on Wearable and Implantable Body Sensor Networks*. IEEE, pp. 138–143.

Li, W., Zhang, Z. & Liu, Z. (2010) Action recognition based on a bag of 3D points. In: *IEEE Conference on Computer Vision and Pattern Recognition Workshops*. IEEE, pp. 9–14.

Li, Y., Zeng, Z., Popescu, M. & Ho, K.C. (2010) Acoustic fall detection using a circular microphone array. *Annual International Conference of the IEEE Engineering in Medicine and Biology Society*, 2010, pp. 2242–5.

Ligęza, A. (2006) *Logical Foundations for Rule-Based Systems*. Springer Berlin Heidelberg.

Liu, C.-L., Lee, C.-H. & Lin, P.-M. (2010) A fall detection system using k-nearest neighbor classifier. *Expert Systems with Applications*, 37 (10), pp. 7174–7181.

Ljung, L. (1979) Asymptotic behavior of the extended Kalman filter as a parameter estimator for linear systems. *IEEE Transactions on Automatic Control*, 24 (1), pp. 36–50.

Lunardhi, A.D. & Passino, K.M. (1995) Verification Of Qualitative Properties Of Rule-Based Expert Systems. *Applied Artificial Intelligence*, 9 (6), pp. 587–621.

Łukasiewicz, J. (1920) O logice trójwartościowej. *Ruch Filozoficzny*, 5, pp. 170–171.

Maddalena, L. & Petrosino, A. (2008) A self-organizing approach to background subtraction for visual surveillance applications. *IEEE Transactions on Image Processing*, 17 (7), pp. 1168–77.

Mastorakis, G. & Makris, D. (2012) Fall detection system using Kinect's infrared sensor. *Journal of Real-Time Image Processing*, 9 (4), pp. 635–646.

Mathie, M.J., Coster, A.C.F., Lovell, N.H. & Celler, B.G. (2004) Accelerometry: providing an integrated, practical method for long-term, ambulatory monitoring of human movement. *Physiological Measurement*, 25 (2), pp. R1–20.

Mazurek, P., Wagner, J. & Morawski, R.Z. (2015) Acquisition and preprocessing of data from infrared depth sensors to be applied for patients monitoring. In: *IEEE International Conference on Intelligent Data Acquisition and Advanced Computing Systems: Technology and Applications (IDAACS)*. IEEE, pp. 705–710.

Mehnert, A. & Jackway, P. (1997) An improved seeded region growing algorithm. *Pattern Recognition Letters*, 18 (10), pp. 1065–1071.

Mehran, R., Oyama, A. & Shah, M. (2009) Abnormal crowd behavior detection using social force model. In: *IEEE Conference on Computer Vision and Pattern Recognition*. IEEE, pp. 935–942.

Miaou, S.-G., Sung, P.-H. & Huang, C.-Y. (2006) A Customized Human Fall Detection System Using Omni-Camera Images and Personal Information. In: *Transdisciplinary Conference on Distributed Diagnosis and Home Healthcare*. IEEE, pp. 39–42.

Moeslund, T.B., Hilton, A. & Krüger, V. (2006) A survey of advances in vision-based human motion capture and analysis. *Computer Vision and Image Understanding*, 104 (2-3), pp. 90–126.

Nevitt, M.C., Cummings, S.R. & Hudes, E.S. (1991) Risk factors for injurious falls: a prospective study. *Journal of Gerontology*, 46 (5), pp. M164–70.

Nijsen, T.M.E., Aarts, R.M., Cluitmans, P.J.M. & Griep, P.A.M. (2010) Time-frequency analysis of accelerometry data for detection of myoclonic seizures. *IEEE Transactions on Information Technology in Biomedicine*, 14 (5), pp. 1197–203.

Nijsen, T.M.E., Arends, J.B.A.M., Griep, P.A.M. & Cluitmans, P.J.M. (2005) The potential value of three-dimensional accelerometry for detection of motor seizures in severe epilepsy. *Epilepsy & Behavior : E&B*, 7 (1), pp. 74–84.

Niu, W., Long, J., Han, D. & Wang, Y.-F. (2004) Human activity detection and recognition for video surveillance. In: *IEEE International Conference on Multimedia and Expo*. IEEE, pp. 719–722.

Nixon, M.S. & Carter, J.N. (2006) Automatic Recognition by Gait. *Proceedings of the IEEE*, 94 (11), pp. 2013–2024.

Noury, N., Fleury, A., Rumeau, P., Bourke, A.K., Laighin, G.O., Rialle, V. & Lundy, J.E. (2007) Fall detection - Principles and Methods. In: *Annual International Conference of the IEEE Engineering in Medicine and Biology Society*. IEEE, pp. 1663–6.

Noury, N., Rumeau, P., Bourke, A.K., ÓLaighin, G. & Lundy, J.E. (2008) A proposal for the classification and evaluation of fall detectors. *IRBM*, 29 (6), pp. 340–349.

Ohn-Bar, E. & Trivedi, M.M. (2013) Joint Angles Similarities and HOG2 for Action Recognition. In: *IEEE Conference on Computer Vision and Pattern Recognition Workshops*. IEEE, pp. 465–470.

Parameswaran, V. & Chellappa, R. (2006) View Invariance for Human Action Recognition. *International Journal of Computer Vision*, 66 (1), pp. 83–101.

Parra-Dominguez, G.S., Taati, B. & Mihailidis, A. (2012) 3D Human Motion Analysis to Detect Abnormal Events on Stairs. In: *International Conference on 3D Imaging, Modeling, Processing, Visualization & Transmission*. IEEE, pp. 97–103.

Perry, J.T., Kellog, S., Vaidya, S.M., Youn, J.-H., Ali, H. & Sharif, H. (2009) Survey and evaluation of real-time fall detection approaches. In: *International Symposium on High Capacity Optical Networks and Enabling Technologies*. IEEE, pp. 158–164.

Piegat, A. (2001) *Fuzzy Modeling and Control*. Physica-Verlag HD.

Piorek, M. & Winiecki, W. (2015) On calibration and parametrization of low power ultrawideband radar for close range detection of human body and bodily functions. In: *IEEE International Conference on Intelligent Data Acquisition and Advanced Computing Systems: Technology and Applications*. IEEE, pp. 639–645.

Planinc, R. & Kampel, M. (2012) Introducing the use of depth data for fall detection. *Personal and Ubiquitous Computing*, 17 (6), pp. 1063–1072.

Poppe, R. (2010) A survey on vision-based human action recognition. *Image and Vision Computing*, 28 (6), pp. 976–990.

Porikli, F., Ivanov, Y. & Haga, T. (2008) Robust Abandoned Object Detection Using Dual Foregrounds. *EURASIP Journal on Advances in Signal Processing*, 2008, pp. 1–12.

Post, E.L. (1921) Introduction to a General Theory of Elementary Propositions. *American Journal of Mathematics*, 43 (3), pp. 163–185.

Ren, X. & Malik, J. (2003) Learning a classification model for segmentation. In: *IEEE International Conference on Computer Vision*. IEEE, pp. 10–17 vol.1.

Richardt, C., Stoll, C., Dodgson, N.A., Seidel, H.-P. & Theobalt, C. (2012) Coherent Spatiotemporal Filtering, Upsampling and Rendering of RGBZ Videos. *Computer Graphics Forum*, 31 (2), pp. 247–256.

Rojas, I., Joya, G. & Cabestany, J. eds. (2013) Indoor Activity Recognition by Combining One-vs.-All Neural Network Classifiers Exploiting Wearable and Depth Sensors. In: *Advances in Computational Intelligence*. Lecture Notes in Computer Science. Berlin, Heidelberg, Springer Berlin Heidelberg.

Ross, T.J. (2010) *Fuzzy Logic with Engineering Applications*. Chichester, UK, John Wiley & Sons, Ltd.

Rougier, C., Auvinet, E., Rousseau, J., Mignotte, M. & Meunier, J. (2011) Fall detection from depth map video sequences. In: *Int. Conf. on Toward Useful Services for Elderly and People with Disabilities*. Springer-Verlag, pp. 121–128.

Rougier, C., Meunier, J., St-Arnaud, A. & Rousseau, J. (2006) Monocular 3D head tracking to detect falls of elderly people. In: *Annual International Conference of the IEEE Engineering in Medicine and Biology Society*. IEEE, pp. 6384–6387.

Rougier, C., Meunier, J., St-Arnaud, A. & Rousseau, J. (2011) Robust Video Surveillance for Fall Detection Based on Human Shape Deformation. *IEEE Transactions on Circuits and Systems for Video Technology*, 21 (5), pp. 611–622.

Rubenstein, L.Z. & Josephson, K.R. (2002) The epidemiology of falls and syncope. *Clinics in Geriatric Medicine*, 18 (2), pp. 141–58.

Rusu, R.B. & Cousins, S. (2011) 3D is here: Point Cloud Library (PCL). In: *IEEE International Conference on Robotics and Automation*. IEEE, pp. 1–4.

Rutkowski, L. (2006) *Metody i techniki sztucznej inteligencji: inteligencja obliczeniowa*. Warszawa, Wydawnictwo Naukowe PWN.

Rymut, B. & Kwolek, B. (2010) GPU-Supported Object Tracking Using Adaptive Appearance Models and Particle Swarm Optimization. In: L. Bolc, R. Tadeusiewicz, L. J. Chmielewski, & K. Wojciechowski eds. *Computer Vision and Graphics*. Lecture Notes in Computer Science. Berlin, Heidelberg, Springer Berlin Heidelberg, pp. 227–234.

Shoaib, M., Dragon, R. & Ostermann, J. (2011) Context-aware visual analysis of elderly activity in a cluttered home environment. *EURASIP Journal on Advances in Signal Processing*, 2011 (1), p.129.

Shotton, J., Fitzgibbon, A., Cook, M., Sharp, T., Finocchio, M., Moore, R., Kipman, A. & Blake, A. (2011) Real-time human pose recognition in parts from single depth images.

In: *IEEE Conference on Computer Vision and Pattern Recognition*. IEEE, pp. 1297–1304.

Skarbek, W. & Koschan, A. (1994) *Colour Image Segmentation - A Survey*.

Smolka, B., Wojciechowski, K.W. & Szczepanski, M. (1999) Random walk approach to image enhancement. In: *Proceedings International Conference on Image Analysis and Processing*. IEEE Comput. Soc, pp. 174–179.

Sobral, A. & Vacavant, A. (2014) A comprehensive review of background subtraction algorithms evaluated with synthetic and real videos. *Computer Vision and Image Understanding*, 122, pp. 4–21.

Sokolova, M. & Lapalme, G. (2009) A systematic analysis of performance measures for classification tasks. *Information Processing & Management*, 45 (4), pp. 427–437.

Spinello, L. & Arras, K.O. (2011) People detection in RGB-D data. In: *IEEE/RSJ International Conference on Intelligent Robots and Systems*. IEEE, pp. 3838–3843.

Stone, E.E. & Skubic, M. (2011) Evaluation of an inexpensive depth camera for passive in-home fall risk assessment. *Journal of Ambient Intelligence and Smart Environments*, 3 (4), pp. 349–361.

Stone, E.E. & Skubic, M. (2014) Fall Detection in Homes of Older Adults Using the Microsoft Kinect. *IEEE Journal of Biomedical and Health Informatics*.

Stone, E.E. & Skubic, M. (2013) Unobtrusive, continuous, in-home gait measurement using the Microsoft Kinect. *IEEE Transactions on Biomedical Engineering*, 60 (10), pp. 2925 – 2932.

Switonski, A., Polanski, A. & Wojciechowski, K.W. (2011) Human Identification Based on Gait Paths. In: J. Blanc-Talon, R. Kleihorst, W. Philips, D. Popescu, & P. Scheunders eds. *Advanced Concepts for Intelligent Vision Systems*. Lecture Notes in Computer Science. Berlin, Heidelberg, Springer Berlin Heidelberg, pp. 531–542.

Szeliski, R. (2010) *Computer Vision: Algorithms and Applications*. Springer-Verlag New York, Inc.

Tamura, T. (2005) Wearable accelerometer in clinical use. In: *Annual International Conference of the IEEE Engineering in Medicine and Biology Society*. pp. 7165–7166.

Tanahashi, H., Hirayu, H., Niwa, Y. & Yamamoto, K. (2001) Comparison of local plane fitting methods for range data. In: *IEEE Conference on Computer Vision and Pattern Recognition*. IEEE, pp. I–663–I–669.

Thrun, S., Burgard, W. & Fox, D. (2005) *Probabilistic Robotics (Intelligent Robotics and Autonomous Agents)*. The MIT Press.

Thurau, C. & Hlavac, V. (2008) Pose primitive based human action recognition in videos or still images. In: *IEEE Conference on Computer Vision and Pattern Recognition*. IEEE, pp. 1–8.

Tian, Y.-L. & Hampapur, A. (2005) Robust Salient Motion Detection with Complex Background for Real-Time Video Surveillance. In: *IEEE Workshops on Applications of Computer Vision*. IEEE, pp. 30–35.

Tomasi, C. & Manduchi, R. (1998) Bilateral filtering for gray and color images. In: *IEEE International Conference on Computer Vision*. IEEE, pp. 839–846.

Tordoff, B.J. & Murray, D.W. (2005) Guided-MLESAC: faster image transform estimation by using matching priors. *IEEE Transactions on Pattern Analysis and Machine Intelligence*, 27 (10), pp. 1523–35.

Vernon, D. (1991) *Machine Vision: Automated Visual Inspection and Robot Vision*. Prentice Hall.

Wang, J., Liu, Z., Chorowski, J., Chen, Z. & Wu, Y. (2012) Robust 3d action recognition with random occupancy patterns. In: A. Fitzgibbon, S. Lazebnik, P. Perona, Y. Sato, & C. Schmid eds. *Computer Vision – ECCV 2012*. Lecture Notes in Computer Science. Berlin, Heidelberg, Springer Berlin Heidelberg, pp. 872–885.

Wang, M., Huang, C. & Lin, H. (2006) An Intelligent Surveillance System Based on an Omnidirectional Vision Sensor. In: *IEEE Conference on Cybernetics and Intelligent Systems*. IEEE, pp. 1–6.

Webster, D. & Celik, O. (2014) Systematic review of Kinect applications in elderly care and stroke rehabilitation. *Journal of Neuroengineering and Rehabilitation*, 11, p.108.

Weinland, D., Özuysal, M. & Fua, P. (2010) Making action recognition robust to occlusions and viewpoint changes. In: *Computer Vision – ECCV 2010*. Springer-Verlag, pp. 635–648.

Woźniak, M., Graña, M. & Corchado, E. (2014) A survey of multiple classifier systems as hybrid systems. *Information Fusion*, 16, pp. 3–17.

Wren, C.R., Azarbayejani, A., Darrell, T. & Pentland, A.P. (1997) Pfinder: real-time tracking of the human body. *IEEE Transactions on Pattern Analysis and Machine Intelligence*, 19 (7), pp. 780–785.

Xia, L., Chen, C.-C. & Aggarwal, J.K. (2011) Human detection using depth information by Kinect. In: *IEEE Conference on Computer Vision and Pattern Recognition Workshops*. IEEE, pp. 15–22.

Yang, H.-D., Park, A.-Y. & Lee, S.-W. (2007) Gesture Spotting and Recognition for Human–Robot Interaction. *IEEE Transactions on Robotics*, 23 (2), pp. 256–270.

Yang, J., Ye, X., Li, K. & Hou, C. (2012) Depth recovery using an adaptive color-guided auto-regressive model. In: A. Fitzgibbon, S. Lazebnik, P. Perona, Y. Sato, & C. Schmid eds. *Computer Vision – ECCV 2012*. Lecture Notes in Computer Science. Berlin, Heidelberg, Springer Berlin Heidelberg, pp. 158–171.

Yang, X., Zhang, C. & Tian, Y. (2012) Recognizing actions using depth motion maps-based histograms of oriented gradients. In: *Proceedings of the ACM International Conference on Multimedia*. New York, New York, USA, ACM Press, p.1057.

Yao, A., Gall, J., Fanelli, G. & Gool, L. Van (2011) Does Human Action Recognition Benefit from Pose Estimation? In: *BMVC*. pp. 67.1–67.11.

Yao, J. & Odobez, J.-M. (2007) Multi-Layer Background Subtraction Based on Color and Texture. In: *IEEE Conference on Computer Vision and Pattern Recognition*. IEEE, pp. 1–8.

Yu, X. (2008) Approaches and principles of fall detection for elderly and patient. In: *International Conference on e-Health Networking, Applications and Services*. IEEE, pp. 42–47.

Zadeh, L.A. (1965) Fuzzy sets. *Information and Control*, 8 (3), pp. 338–353.

Zhang, C., Tian, Y. & Capezuti, E. (2012) Privacy preserving automatic fall detection for elderly using RGBD cameras. In: K. Miesenberger, A. Karshmer, P. Penaz, & W. Zagler eds. *Computers Helping People with Special Needs*. Lecture Notes in Computer Science. Berlin, Heidelberg, Springer Berlin Heidelberg, pp. 625–633.

Zhang, Z., Conly, C. & Athitsos, V. (2015) A survey on vision-based fall detection. , p.46.

Zigel, Y., Litvak, D. & Gannot, I. (2009) A method for automatic fall detection of elderly people using floor vibrations and sound - proof of concept on human mimicking doll falls. *IEEE Transactions on Biomedical Engineering*, 56 (12), pp. 2858–67.

Zivkovic, Z. & van der Heijden, F. (2006) Efficient adaptive density estimation per image pixel for the task of background subtraction. *Pattern Recognition Letters*, 27 (7), pp. 773–780.

# DODATEK

Baza reguł dla rozmytego układu wnioskującego Static:

**Baza wiedzy** zawiera 81 reguł, określonych na 4 zmiennych wejściowych i 3 zbiorach rozmytych dla każdej ze zmiennych. Oznaczenia wykorzystanie w opisie bazy wiedzy to:

- Hi - wysoka wartość cechy (ang. *high*),
- Me - średnia wartość cechy (ang. *medium*),
- Lo - niska wartość cechy (ang. *low*),
- notLy - poza nieleżąca (ang. *not lying*),
- mayLy - możliwa poza leżąca (ang. *maybe lying*),
- isLy - poza leżąca (ang. *lying*).

Zbiór reguł:

R1. Jeśli ($P_{40}$ jest Lo) i ($H/W$ jest Hi) i ($max(\bar{\sigma})$ jest Lo) i ($H/H_{max}$ jest Hi) to (*Pose* jest notLy)

R2. Jeśli ($P_{40}$ jest Lo) i ($H/W$ jest Hi) i ($max(\bar{\sigma})$ jest Lo) i ($H/H_{max}$ jest Me) to (*Pose* jest notLy)

R3. Jeśli ($P_{40}$ jest Lo) i ($H/W$ jest Hi) i ($max(\bar{\sigma})$ jest Lo) i ($H/H_{max}$ jest Lo) to (*Pose* jest notLy)

R4. Jeśli ($P_{40}$ jest Lo) i ($H/W$ jest Hi) i ($max(\bar{\sigma})$ jest Me) i ($H/H_{max}$ jest Hi) to (*Pose* jest notLy)

R5. Jeśli ($P_{40}$ jest Lo) i ($H/W$ jest Hi) i ($max(\bar{\sigma})$ jest Me) i ($H/H_{max}$ jest Me) to (*Pose* jest notLy)

R6. Jeśli ($P_{40}$ jest Lo) i ($H/W$ jest Hi) i ($max(\bar{\sigma})$ jest Me) i ($H/H_{max}$ jest Lo) to (*Pose* jest mayLy)

R7. Jeśli ($P_{40}$ jest Lo) i ($H/W$ jest Hi) i ($max(\bar{\sigma})$ jest Hi) i ($H/H_{max}$ jest Hi) to (*Pose* jest notLy)

R8. Jeśli ($P_{40}$ jest Lo) i ($H/W$ jest Hi) i ($max(\bar{\sigma})$ jest Hi) i ($H/H_{max}$ jest Me) to (*Pose* jest mayLy)

R9. Jeśli ($P_{40}$ jest Lo) i ($H/W$ jest Hi) i ($max(\bar{\sigma})$ jest Hi) i ($H/H_{max}$ jest Lo) to (*Pose* jest mayLy)

R10. Jeśli ($P_{40}$ jest Lo) i ($H/W$ jest Me) i ($max(\bar{\sigma})$ jest Lo) i ($H/H_{max}$ jest Hi) to (*Pose* jest notLy)

R11. Jeśli ($P_{40}$ jest Lo) i ($H/W$ jest Me) i ($max(\bar{\sigma})$ jest Lo) i ($H/H_{max}$ jest Me) to (*Pose* jest notLy)

R12. Jeśli ($P_{40}$ jest Lo) i ($H/W$ jest Me) i ($max(\bar{\sigma})$ jest Lo) i ($H/H_{max}$ jest low) to (*Pose* jest mayLy)

R13. Jeśli ($P_{40}$ jest Lo) i ($H/W$ jest Me) i ($max(\bar{\sigma})$ jest Me) i ($H/H_{max}$ jest Hi) to (*Pose* jest notLy)

R14. Jeśli ($P_{40}$ jest Lo) i ($H/W$ jest Me) i ($max(\bar{\sigma})$ jest Me) i ($H/H_{max}$ jest Me) to (*Pose* jest mayLy)

R15. Jeśli ($P_{40}$ jest Lo) i ($H/W$ jest Me) i ($max(\bar{\sigma})$ jest Me) i ($H/H_{max}$ jest Lo) to (*Pose* jest mayLy)

R16. Jeśli ($P_{40}$ jest Lo) i ($H/W$ jest Me) i ($max(\bar{\sigma})$ jest Hi) i ($H/H_{max}$ jest Hi) to (*Pose* jest mayLy)

R17. Jeśli ($P_{40}$ jest Lo) i ($H/W$ jest Me) i ($max(\bar{\sigma})$ jest Hi) i ($H/H_{max}$ jest Me) to (*Pose* jest mayLy)

R18. Jeśli ($P_{40}$ jest Lo) i ($H/W$ jest Me) i ($max(\bar{\sigma})$ jest Hi) i ($H/H_{max}$ jest Lo) to (*Pose* jest mayLy)

R19. Jeśli ($P_{40}$ jest Lo) i ($H/W$ jest Lo) i ($max(\bar{\sigma})$ jest Lo) i ($H/H_{max}$ jest Hi) to (*Pose* jest notLy)

R20. Jeśli ($P_{40}$ jest Me) i ($H/W$ jest Lo) i ($max(\bar{\sigma})$ jest Lo) i ($H/H_{max}$ jest Me) to (*Pose* jest mayLy)

R21. Jeśli ($P_{40}$ jest Lo) i ($H/W$ jest Lo) i ($max(\bar{\sigma})$ jest Lo) i ($H/H_{max}$ jest Lo) to (*Pose* jest mayLy)

R22. Jeśli ($P_{40}$ jest Lo) i ($H/W$ jest Lo) i ($max(\bar{\sigma})$ jest Me) i ($H/H_{max}$ jest Hi) to (*Pose* jest mayLy)

R23. Jeśli ($P_{40}$ jest Lo) i ($H/W$ jest Lo) i ($max(\bar{\sigma})$ jest Me) i ($H/H_{max}$ jest Me) to (*Pose* jest mayLy)

R24. Jeśli ($P_{40}$ jest Lo) i ($H/W$ jest Lo) i ($max(\bar{\sigma})$ jest Me) i ($H/H_{max}$ jest Lo) to (*Pose* jest mayLy)

R25. Jeśli ($P_{40}$ jest Lo) i ($H/W$ jest Lo) i ($max(\bar{\sigma})$ jest Hi) i ($H/H_{max}$ jest Hi) to (*Pose* jest mayLy)

R26. Jeśli ($P_{40}$ jest Lo) i ($H/W$ jest Lo) i ($max(\bar{\sigma})$ jest Hi) i ($H/H_{max}$ jest Me) to (*Pose* jest mayLy)

R27. Jeśli ($P_{40}$ jest Lo) i ($H/W$ jest Lo) i ($max(\bar{\sigma})$ jest Hi) i ($H/H_{max}$ jest Lo) to (*Pose* jest isLy)

R28. Jeśli ($P_{40}$ jest Me) i ($H/W$ jest Hi) i ($max(\bar{\sigma})$ jest Lo) i ($H/H_{max}$ jest Hi) to (*Pose* jest notLy)

R29. Jeśli ($P_{40}$ jest Me) i ($H/W$ jest Hi) i ($max(\bar{\sigma})$ jest Lo) i ($H/H_{max}$ jest Me) to (*Pose* jest notLy)

R30. Jeśli ($P_{40}$ jest Me) i ($H/W$ jest Hi) i ($max(\bar{\sigma})$ jest Lo) i ($H/H_{max}$ jest Lo) to (*Pose* jest mayLy)

R31. Jeśli ($P_{40}$ jest Me) i ($H/W$ jest Hi) i ($max(\bar{\sigma})$ jest Me) i ($H/H_{max}$ jest Hi) to (*Pose* jest notLy)

R32. Jeśli ($P_{40}$ jest Me) i ($H/W$ jest Hi) i ($max(\bar{\sigma})$ jest Me) i ($H/H_{max}$ jest Me) to (*Pose* jest mayLy)

R33. Jeśli ($P_{40}$ jest Me) i ($H/W$ jest Hi) i ($max(\bar{\sigma})$ jest Me) i ($H/H_{max}$ jest Lo) to (*Pose* jest mayLy)

R34. Jeśli ($P_{40}$ jest Me) i ($H/W$ jest Hi) i ($max(\bar{\sigma})$ jest Hi) i ($H/H_{max}$ jest Hi) to (*Pose* jest mayLy)

R35. Jeśli ($P_{40}$ jest Me) i ($H/W$ jest Hi) i ($max(\bar{\sigma})$ jest Hi) i ($H/H_{max}$ jest Me) to (*Pose* jest mayLy)

R36. Jeśli ($P_{40}$ jest Me) i ($H/W$ jest Hi) i ($max(\bar{\sigma})$ jest Hi) i ($H/H_{max}$ jest Lo) to (*Pose* jest mayLy)

R37. Jeśli ($P_{40}$ jest Me) i ($H/W$ jest Me) i ($max(\bar{\sigma})$ jest Lo) i ($H/H_{max}$ jest Hi) to (*Pose* jest notLy)

R38. Jeśli ($P_{40}$ jest Me) i ($H/W$ jest Me) i ($max(\bar{\sigma})$ jest Lo) i ($H/H_{max}$ jest Me) to (*Pose* jest mayLy)

R39. Jeśli ($P_{40}$ jest Me) i ($H/W$ jest Me) i ($max(\bar{\sigma})$ jest Lo) i ($H/H_{max}$ jest Lo) to (*Pose* jest mayLy)

R40. Jeśli ($P_{40}$ jest Me) i ($H/W$ jest Me) i ($max(\bar{\sigma})$ jest Me) i ($H/H_{max}$ jest Hi) to (*Pose* jest mayLy)

R41. Jeśli ($P_{40}$ jest Me) i ($H/W$ jest Me) i ($max(\bar{\sigma})$ jest Me) i ($H/H_{max}$ jest Me) to (*Pose* jest mayLy)

R42. Jeśli ($P_{40}$ jest Me) i ($H/W$ jest Me) i ($max(\bar{\sigma})$ jest Me) i ($H/H_{max}$ jest Lo) to (*Pose* jest mayLy)

R43. Jeśli ($P_{40}$ jest Me) i ($H/W$ jest Me) i ($max(\bar{\sigma})$ jest Hi) i ($H/H_{max}$ jest Hi) to (*Pose* jest mayLy)

R44. Jeśli ($P_{40}$ jest Me) i ($H/W$ jest Me) i ($max(\bar{\sigma})$ jest Hi) i ($H/H_{max}$ jest Me) to (*Pose* jest mayLy)

R45. Jeśli ($P_{40}$ jest Me) i ($H/W$ jest Me) i ($max(\bar{\sigma})$ jest Hi) i ($H/H_{max}$ jest Lo) to (*Pose* jest isLy)

R46. Jeśli ($P_{40}$ jest Me) i ($H/W$ jest Lo) i ($max(\bar{\sigma})$ jest Lo) i ($H/H_{max}$ jest Hi) to (*Pose* jest mayLy)

R47. Jeśli ($P_{40}$ jest Me) i ($H/W$ jest Lo) i ($max(\bar{\sigma})$ jest Lo) i ($H/H_{max}$ jest Me) to (*Pose* jest mayLy)

R48. Jeśli ($P_{40}$ jest Me) i ($H/W$ jest Lo) i ($max(\bar{\sigma})$ jest Lo) i ($H/H_{max}$ jest Lo) to (*Pose* jest mayLy)

R49. Jeśli ($P_{40}$ jest Me) i ($H/W$ jest Lo) i ($max(\bar{\sigma})$ jest Me) i ($H/H_{max}$ jest Hi) to (*Pose* jest mayLy)

R50. Jeśli ($P_{40}$ jest Me) i ($H/W$ jest Lo) i ($max(\bar{\sigma})$ jest Me) i ($H/H_{max}$ jest Me) to (*Pose* jest mayLy)

R51. Jeśli ($P_{40}$ jest Me) i ($H/W$ jest Lo) i ($max(\bar{\sigma})$ jest Me) i ($H/H_{max}$ jest Lo) to (*Pose* jest isLy)

R52. Jeśli ($P_{40}$ jest Me) i ($H/W$ jest Lo) i ($max(\bar{\sigma})$ jest Hi) i ($H/H_{max}$ jest Hi) to (*Pose* jest mayLy)

R53. Jeśli ($P_{40}$ jest Me) i ($H/W$ jest Lo) i ($max(\bar{\sigma})$ jest Hi) i ($H/H_{max}$ jest Me) to (*Pose* jest isLy)

R54. Jeśli ($P_{40}$ jest Me) i ($H/W$ jest Lo) i ($max(\bar{\sigma})$ jest Hi) i ($H/H_{max}$ jest Lo) to (*Pose* jest isLy)

R55. Jeśli ($P_{40}$ jest Hi) i ($H/W$ jest Hi) i ($max(\bar{\sigma})$ jest Lo) i ($H/H_{max}$ jest Hi) to (*Pose* jest notLy)

R56. Jeśli ($P_{40}$ jest Hi) i ($H/W$ jest Hi) i ($max(\bar{\sigma})$ jest Lo) i ($H/H_{max}$ jest Me) to (*Pose* jest mayLy)

R57. Jeśli ($P_{40}$ jest Hi) i ($H/W$ jest Hi) i ($max(\bar{\sigma})$ jest Lo) i ($H/H_{max}$ jest Lo) to (*Pose* jest mayLy)

R58. Jeśli ($P_{40}$ jest Hi) i ($H/W$ jest Hi) i ($max(\bar{\sigma})$ jest Me) i ($H/H_{max}$ jest Hi) to (*Pose* jest mayLy)

R59. Jeśli ($P_{40}$ jest Hi) i ($H/W$ jest Hi) i ($max(\bar{\sigma})$ jest Me) i ($H/H_{max}$ jest Me) to (*Pose* jest mayLy)

R60. Jeśli ($P_{40}$ jest Hi) i ($H/W$ jest Hi) i ($max(\bar{\sigma})$ jest Me) i ($H/H_{max}$ jest Lo) to (*Pose* jest mayLy)

R61. Jeśli ($P_{40}$ jest Hi) i ($H/W$ jest Hi) i ($max(\bar{\sigma})$ jest Hi) i ($H/H_{max}$ jest Hi) to (*Pose* jest mayLy)

R62. Jeśli ($P_{40}$ jest Hi) i ($H/W$ jest Hi) i ($max(\bar{\sigma})$ jest Hi) i ($H/H_{max}$ jest Me) to (*Pose* jest mayLy)

R63. Jeśli ($P_{40}$ jest Hi) i ($H/W$ jest Hi) i ($max(\bar{\sigma})$ jest Hi) i ($H/H_{max}$ jest Lo) to (*Pose* jest isLy)

R64. Jeśli ($P_{40}$ jest Hi) i ($H/W$ jest Me) i ($max(\bar{\sigma})$ jest Lo) i ($H/H_{max}$ jest Hi) to (*Pose* jest mayLy)

R65. Jeśli ($P_{40}$ jest Hi) i ($H/W$ jest Me) i ($max(\bar{\sigma})$ jest Lo) i ($H/H_{max}$ jest Me) to (*Pose* jest mayLy)

R66. Jeśli ($P_{40}$ jest Hi) i ($H/W$ jest Me) i ($max(\bar{\sigma})$ jest Lo) i ($H/H_{max}$ jest Lo) to (*Pose* jest mayLy)

R67. Jeśli ($P_{40}$ jest Hi) i ($H/W$ jest Me) i ($max(\bar{\sigma})$ jest Me) i ($H/H_{max}$ jest Hi) to (*Pose* jest mayLy)

R68. Jeśli ($P_{40}$ jest Hi) i ($H/W$ jest Me) i ($max(\bar{\sigma})$ jest Me) i ($H/H_{max}$ jest Me) to (*Pose* jest mayLy)

R69. Jeśli ($P_{40}$ jest Hi) i ($H/W$ jest Me) i ($max(\bar{\sigma})$ jest Me) i ($H/H_{max}$ jest Lo) to (*Pose* jest isLy)

R70. Jeśli ($P_{40}$ jest Hi) i ($H/W$ jest Me) i ($max(\bar{\sigma})$ jest Hi) i ($H/H_{max}$ jest Hi) to (*Pose* jest mayLy)

R71. Jeśli ($P_{40}$ jest Hi) i ($H/W$ jest Me) i ($max(\bar{\sigma})$ jest Hi) i ($H/H_{max}$ jest Me) to (*Pose* jest isLy)

R72. Jeśli ($P_{40}$ jest Hi) i ($H/W$ jest Me) i ($max(\bar{\sigma})$ jest Hi) i ($H/H_{max}$ jest Lo) to (*Pose* jest isLy)

R73. Jeśli ($P_{40}$ jest Hi) i ($H/W$ jest Lo) i ($max(\bar{\sigma})$ jest Lo) i ($H/H_{max}$ jest Hi) to (*Pose* jest mayLy)

R74. Jeśli ($P_{40}$ jest Hi) i ($H/W$ jest Lo) i ($max(\bar{\sigma})$ jest Lo) i ($H/H_{max}$ jest Me) to (*Pose* jest mayLy)

R75. Jeśli ($P_{40}$ jest Hi) i ($H/W$ jest Lo) i ($max(\bar{\sigma})$ jest Lo) i ($H/H_{max}$ jest Lo) to (*Pose* jest isLy)

R76. Jeśli ($P_{40}$ jest Hi) i ($H/W$ jest Lo) i ($max(\bar{\sigma})$ jest Me) i ($H/H_{max}$ jest Hi) to (*Pose* jest mayLy)

R77. Jeśli ($P_{40}$ jest Hi) i ($H/W$ jest Lo) i ($max(\bar{\sigma})$ jest Me) i ($H/H_{max}$ jest Me) to (*Pose* jest isLy)

R78. Jeśli ($P_{40}$ jest Hi) i ($H/W$ jest Lo) i ($max(\bar{\sigma})$ jest Me) i ($H/H_{max}$ jest Lo) to (*Pose* jest isLy)

R79. Jeśli ($P_{40}$ jest Hi) i ($H/W$ jest Lo) i ($max(\bar{\sigma})$ jest Hi) i ($H/H_{max}$ jest Hi) to (*Pose* jest isLy)

R80. Jeśli ($P_{40}$ jest Hi) i ($H/W$ jest Lo) i ($max(\bar{\sigma})$ jest Hi) i ($H/H_{max}$ jest Me) to (*Pose* jest isLy)

R81. Jeśli ($P_{40}$ jest Hi) i ($H/W$ jest Lo) i ($max(\bar{\sigma})$ jest Hi) i ($H/H_{max}$ jest Lo) to (*Pose* jest isLy)

Baza reguł dla rozmytego układu wnioskującego Transition:

**Baza wiedzy** zawiera 12 reguł, określonych na 3 zmiennych wejściowych, dwie z nich posiadają 2 zbiory rozmyte, natomiast jedna 3 zbiory. Oznaczenia wykorzystanie w opisie bazy wiedzy to:

- Hi - wysoka wartość cechy (ang. *high*),
- Me - średnia wartość cechy (ang. *medium*),
- Lo - niska wartość cechy (ang. *low*),
- Fast - duża szybkość zmiany położenia ciała człowieka,
- Me - średnia szybkość zmiany położenia ciała człowieka,
- Slow - mała szybkość zmiany położenia ciała człowieka.

Zbiór reguł:

R1. Jeśli ($H(t)/H(t-\Delta t)$ jest Lo) i ($D(t)/D(t-\Delta t)$ jest Lo) i ($SV_{total}$ jest Hi) to (Transition jest Fast)

R2. Jeśli ($H(t)/H(t-\Delta t)$ jest Lo) i ($D(t)/D(t-\Delta t)$ jest Lo) i ($SV_{total}$ jest Me) to (Transition jest Fast)

R3. Jeśli ($H(t)/H(t-\Delta t)$ jest Lo) i ($D(t)/D(t-\Delta t)$ jest Lo) i ($SV_{total}$ jest Lo) to (Transition jest Me)

R4. Jeśli ($H(t)/H(t-\Delta t)$ jest Lo) i ($D(t)/D(t-\Delta t)$ jest Hi) i ($SV_{total}$ jest Hi) to (Transition jest Fast)

R5. Jeśli ($H(t)/H(t-\Delta t)$ jest Lo) i ($D(t)/D(t-\Delta t)$ jest Hi) i ($SV_{total}$ jest Me) to (Transition jest Me)

R6. Jeśli ($H(t)/H(t-\Delta t)$ jest Lo) i ($D(t)/D(t-\Delta t)$ jest Hi) i ($SV_{total}$ jest Lo) to (Transition jest Slow)

R7. Jeśli ($H(t)/H(t-\Delta t)$ jest Hi) i ($D(t)/D(t-\Delta t)$ jest Lo) i ($SV_{total}$ jest Hi) to (Transition jest Fast)

R8. Jeśli ($H(t)/H(t-\Delta t)$ jest Hi) i ($D(t)/D(t-\Delta t)$ jest Lo) i ($SV_{total}$ jest Me) to (Transition jest Me)

R9. Jeśli ($H(t)/H(t-\Delta t)$ jest Hi) i ($D(t)/D(t-\Delta t)$ jest Lo) i ($SV_{total}$ jest Lo) to (Transition jest Slow)

R10. Jeśli ($H(t)/H(t-\Delta t)$ jest Hi) i ($D(t)/D(t-\Delta t)$ jest Hi) i ($SV_{total}$ jest Hi) to (Transition jest Me)

R11. Jeśli ($H(t)/H(t-\Delta t)$ jest Hi) i ($D(t)/D(t-\Delta t)$ jest Hi) i ($SV_{total}$ jest Me) to (Transition jest Slow)

R12. Jeśli ($H(t)/H(t-\Delta t)$ jest Hi) i ($D(t)/D(t-\Delta t$ jest Hi) i ($SV_{total}$ jest Lo) to (Transition jest Slow)

Baza reguł dla rozmytego układu wnioskującego Decision:

**Baza wiedzy** zawiera 9 reguł, określonych na 2 zmiennych wejściowych, które posiadają 3 zbiory rozmyte. Oznaczenia wykorzystanie w opisie bazy wiedzy to:

- notLy - poza nieleżąca (ang. *not lying*),
- mayLy - możliwa poza leżąca (ang. *maybe lying*),
- isLy - poza leżąca (ang. *lying*).
- Fast - duża szybkość zmiany położenia ciała człowieka,
- Me - średnia szybkość zmiany położenia ciała człowieka,
- Slow - mała szybkość zmiany położenia ciała człowieka
- Fall - decyzja o klasyfikacji akcji jako upadek,
- No-fall - decyzja o klasyfikacji akcji jako ADL.

Zbiór reguł

R1. Jeśli (Static is isLy) i (Transition is Me) then (Decision is Fall)

R2. Jeśli (Static is mayLy) i (Transition is Me) then (Decision is Fall)

R3. Jeśli (Static is isLy) i (Transition is Fast) then (Decision is Fall)

R4. Jeśli (Static is mayLy) i (Transition is Fast) then (Decision is Fall)

R5. Jeśli (Static is notLy) i (Transition is Fast) then (Decision is No-fall)

R6. Jeśli (Static is mayLy) i (Transition is Slow) then (Decision is No-fall)

R7. Jeśli (Static is notLy) i (Transition is Slow) then (Decision is No-fall)

R8. Jeśli (Static is notLy) i (Transition is Me) then (Decision is No-fall)

R9. Jeśli (Static is isLy) i (Transition is Slow) then (Decision is No-fall)

# DETEKCJA UPADKU I WYBRANYCH AKCJI NA SEKWENCJACH OBRAZÓW CYFROWYCH

**mgr inż. Michał Kępski, Uniwersytet Rzeszowski**

**Streszczenie:** W ostatnich latach obserwuje się duży wzrost zainteresowania zagadnieniem rozpoznawania akcji, a w szczególności jedną z jego dziedzin jaką jest detekcja upadku. Trudności związane z popularyzacją obecnych komercyjnych systemów detekcji upadku w środowisku seniorów wiążą się z niedoskonałością technologii, brakiem wystarczającej dokładności detekcji, dużą liczbą fałszywych alarmów oraz niewystarczającym poszanowaniem prywatności osoby podczas akwizycji i przetwarzania danych. Niniejsza praca wychodzi naprzeciw tym oczekiwaniom. Praca ma charakter empiryczny i dotyczy komputerowych systemów wizyjnych. Zasadnicza część pracy sytuuje się w obszarze rozpoznawania akcji i zachowań osoby. W pracy opracowano, przebadano oraz zaimplementowano algorytmy umożliwiające detekcję upadku na podstawie sekwencji obrazów oraz bezprzewodowego sensora inercyjnego noszonego przez monitorowaną osobę. Przebadano i dobrano zestaw deskryptorów dla obrazów głębi pozwalający na klasyfikację pozy w jakiej znajduje się osoba, a także akcji, która jest przez nią wykonywana. Badania eksperymentalne zrealizowano w oparciu o przygotowane repozytorium danych składające się z synchronizowanych obrazów oraz danych z akcelerometru. Badania zrealizowano w scenariuszu ze statyczną kamerą umieszczoną na wprost oraz aktywną kamerą obserwującą scenę z góry. Opracowane metody detekcji upadku pozwoliły uzyskać wskaźniki jakościowe świadczące o wysokiej czułości i swoistości zaproponowanych algorytmów. Algorytmy projektowano pod kątem małego zapotrzebowania na moc obliczeniową, a w szczególności możliwości ich uruchomienia na platformie ARM. W oparciu o opracowane rozwiązania zrealizowano badania eksperymentalne, które polegały na detekcji oraz śledzeniu osoby w czasie rzeczywistym, a także detekcji upadku.

# FALL DETECTION AND SELECTED ACTION RECOGNITION USING IMAGE SEQUENCES

**Michał Kępski, MSc, University of Rzeszów**

**Abstract:** In recent years a growing interest on action recognition is observed, including detection of fall accident for the elderly. However, despite many efforts undertaken, the existing technology is not widely used by elderly, mainly because of its flaws like low precision, large number of false alarms, inadequate privacy preserving during data acquisition and processing. This research work meets these expectations. The work is empirical and it is situated in the field of computer vision systems. The main part of the work situates itself in the area of action and behaviour recognition. Efficient algorithms for fall detection were developed, tested and implemented using image sequences and wireless inertial sensor worn by a monitored person. A set of descriptors for depth maps has been elaborated to permit classification of pose as well as the action of a person. Experimental research was carried out based on the prepared data repository consisting of synchronized depth and accelerometric data. The study was carried out in the scenario with a static camera facing the scene and an active camera observing the scene from above. The experimental results showed that the developed algorithms for fall detection have high sensitivity and specificity. The algorithm were designed with regard to low computational demands and possibility to run on ARM platforms. Several experiments including person detection, tracking and fall detection in real-time were carried out to show efficiency and reliability of the proposed solutions.